\documentclass[10pt,twocolumn,letterpaper]{article}

\usepackage[pagenumbers]{cvpr} 

\usepackage{graphicx}
\usepackage{amsmath}
\usepackage{amssymb}
\usepackage{booktabs}
\usepackage{tabularx}

\usepackage[export]{adjustbox}
\usepackage{multicol}
\usepackage{multirow}

\usepackage{colortbl}
\usepackage[ruled]{algorithm2e} 

\SetAlFnt{\small}
\SetAlCapFnt{\small}
\SetAlCapNameFnt{\small}
\SetAlCapHSkip{0pt}

\definecolor{cYellow}{rgb}{1,1, 0.6}
\definecolor{cOrange}{rgb}{1, 0.8, 0.6}
\definecolor{cRed}{rgb}{1, 0.6, 0.6}

\newcommand{\kd}{\boldsymbol{k}_{\hspace*{-0.1mm}d}}
\newcommand{\ks}{\boldsymbol{k}_{\hspace*{-0.1mm}s}}
\newcommand{\korm}{\boldsymbol{k}_{\hspace*{-0.1mm}\mathrm{orm}}}

\newcommand\blfootnote[1]{%
  \begingroup
  \renewcommand\thefootnote{}\footnote{#1}%
  \addtocounter{footnote}{-1}%
  \endgroup
}

\usepackage[pagebackref,breaklinks,colorlinks]{hyperref}
\usepackage[accsupp]{axessibility}  

\usepackage[capitalize]{cleveref}
\crefname{section}{Sec.}{Secs.}
\Crefname{section}{Section}{Sections}
\Crefname{table}{Table}{Tables}
\crefname{table}{Tab.}{Tabs.}

\def\cvprPaperID{} 
\def\confName{}
\def\confYear{}

\begin{document}

\title{Extracting Triangular 3D Models, Materials, and Lighting From Images}

\author{Jacob Munkberg$^1$
	\and
	Jon Hasselgren$^1$
	\and
	Tianchang Shen$^{1,2,3}$
	\and 
	Jun Gao$^{1,2,3}$
	\and 
	\vspace{2mm}
	Wenzheng Chen$^{1,2,3}$
	\and
	Alex Evans$^1$
	\and
	Thomas M\"uller$^1$
	\and
	Sanja Fidler$^{1,2,3}$
	\and
	$^1$NVIDIA \ \ \ $^2$University of Toronto \ \ \ $^3$Vector Institute
}

\newcolumntype{Y}{>{\centering\arraybackslash}X} 


\newcommand{\figTeaser}{
\begin{figure}[t]
	\centering
	\setlength{\tabcolsep}{1pt}
	{\small
	\begin{tabular}{cc}
		\includegraphics[width=0.49\columnwidth]{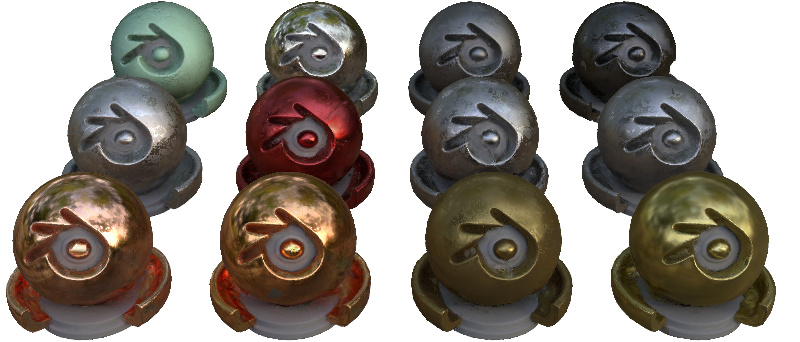} &
		\includegraphics[width=0.49\columnwidth]{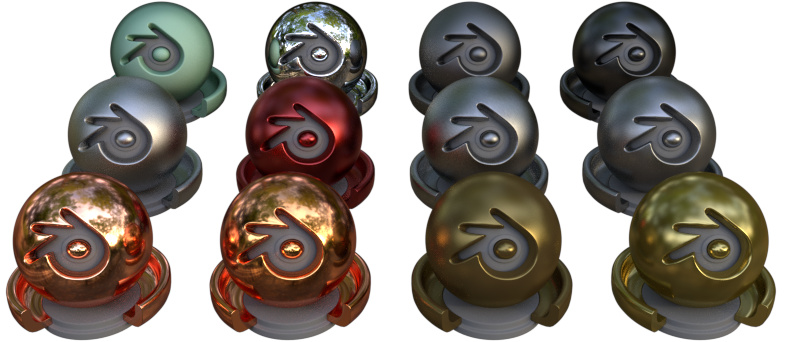} \\
		Our shaded model & Reference image \\
		\includegraphics[width=0.49\columnwidth]{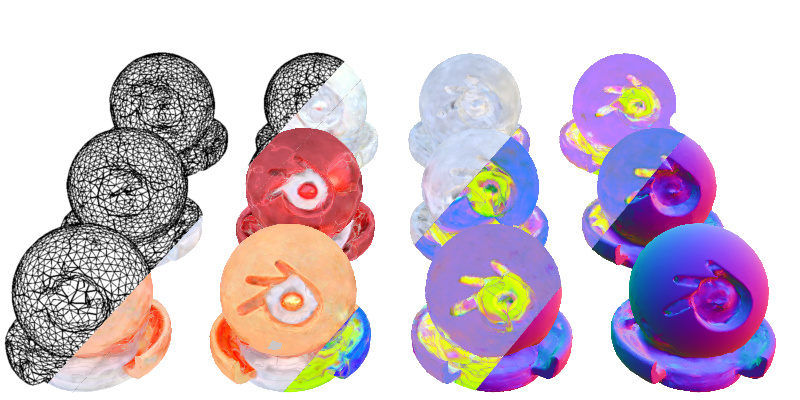} &
		\includegraphics[width=0.49\columnwidth]{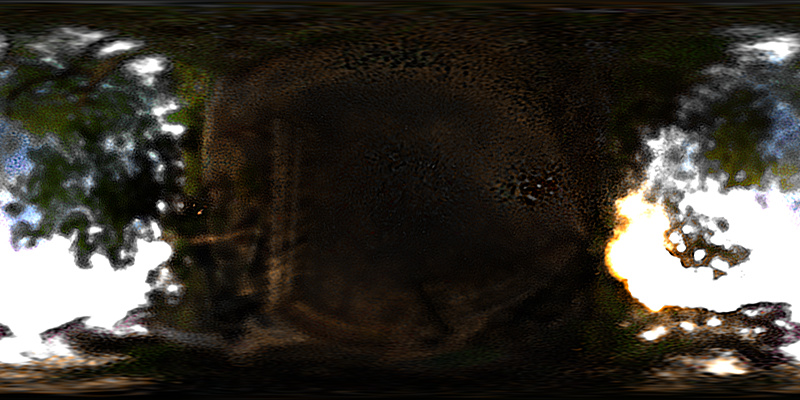} \\
        Mesh~/$\kd$/$\korm$/$\mathbf{n}$ & Extracted probe \\
    \end{tabular}
	}
	\vspace{-2mm}
     \caption{Our result on the \textsc{Materials} scene, reconstructed from 100 images from the NeRF synthetic dataset.}
	\vspace{-1mm}
  \label{fig:teaser}
\end{figure}
}


\newcommand{\figSystem}{
\begin{figure*}[t]
	\centering
		\includegraphics[width=0.99\textwidth]{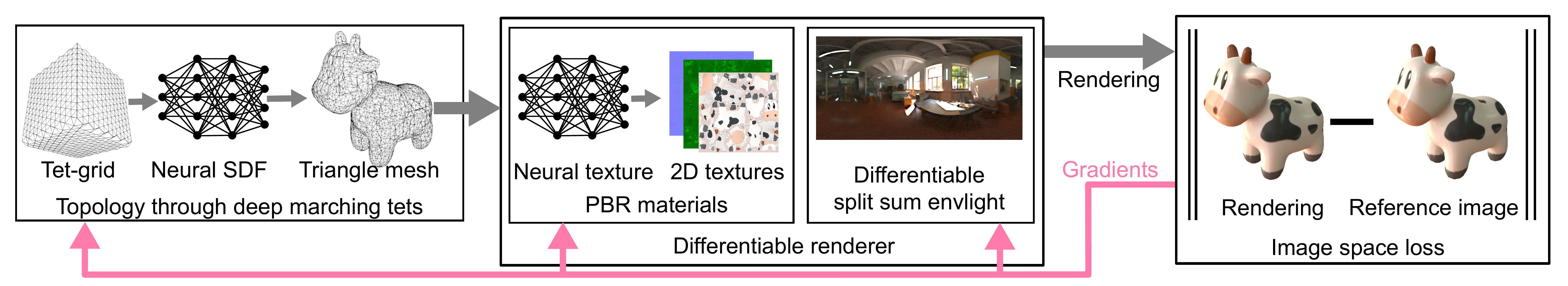}
	\vspace{-3mm}
	\caption{
		{\bf Overview of our approach.} We learn topology, materials, and environment map lighting 
		jointly from 2D supervision. We leverage differentiable marching tetrahedrons
		to directly optimize topology of a triangle mesh. While the topology is drastically changing, 
		we learn materials through volumetric texturing, efficiently encoded using an MLP with
		positional encoding. Finally, we introduce a differentiable
		version of the split sum approximation for environment lighting. Our output representation 
		is a triangle mesh with spatially varying 2D textures and a high dynamic range environment map,
		which can be used unmodified in standard game engines. The system is trained end-to-end, supervised
		by loss in image space, with gradient-based optimization of all stages. Spot model by Keenan Crane.
	}
	\vspace{-2.5mm}
	\label{fig:system}
\end{figure*}
}

\newcommand{\sign}{\text{sign}}
\newcommand{\figDMTetFormulation}{
\begin{figure}
\begin{minipage}[c]{0.50\columnwidth}
\includegraphics[width=\textwidth]{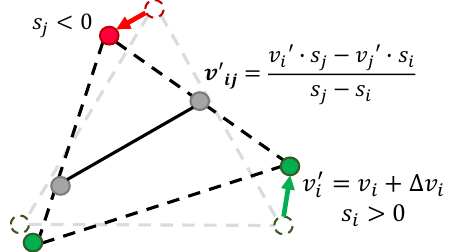}
\end{minipage}\hfill
\begin{minipage}[c]{0.50\columnwidth}
\caption{Marching Tetrahedra extracts faces from a tetrahedral grid with 
grid vertices, $v_i' = v_i + \Delta v_i$ and scalar SDF values, $s_i$. 
For tets with $\sign(s_i) \neq \sign(s_j)$, 
faces are extracted, and the face vertices, $v_{ij}$, are determined by 
by linear interpolation.} \label{fig:DMTetFormulation}
\end{minipage}
\vspace{-10mm}
\end{figure}
}


\newcommand{\figRegAblation}{
\begin{figure}[t]
	\centering
        \setlength{\tabcolsep}{1pt}
        \begin{tabular}{ccc}
            \includegraphics[width=0.32\columnwidth]{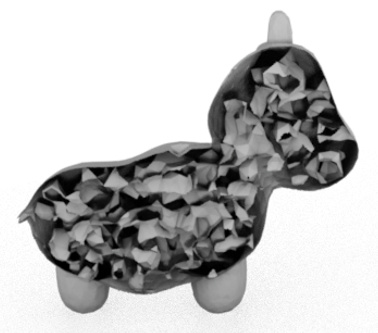} &
            \includegraphics[width=0.32\columnwidth]{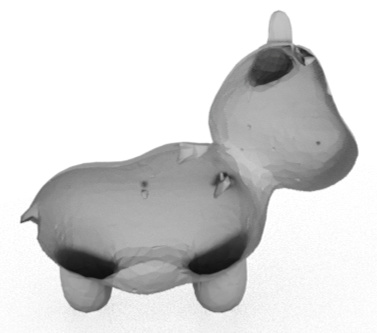} & 
            \includegraphics[width=0.32\columnwidth]{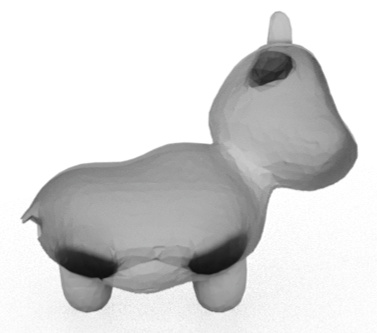}
        \end{tabular}
	\caption{Cross sections of shapes optimized without regularization loss on SDF (left), with smoothness loss 
	used by Liao et~al.~\cite{Liao2018} (middle) and with our regularization loss (right). The random faces inside the object are 
	removed by the regularization loss on SDF.
	}
	\label{fig:RegAblation}
\end{figure}
}


\newcommand{\figRegVisibility}{
\begin{figure}[t]
	\centering
        \setlength{\tabcolsep}{1pt}
        \begin{tabular}{cc}
            \includegraphics[trim={0cm 5cm 0cm 0cm}, width=0.48\columnwidth]{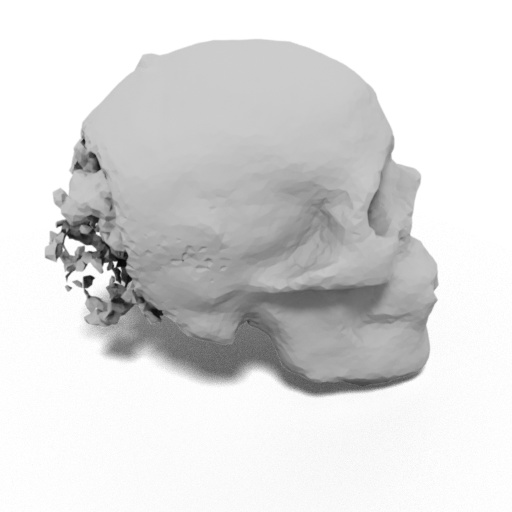} &
            \includegraphics[trim={0cm 5cm 0cm 0cm}, width=0.48\columnwidth]{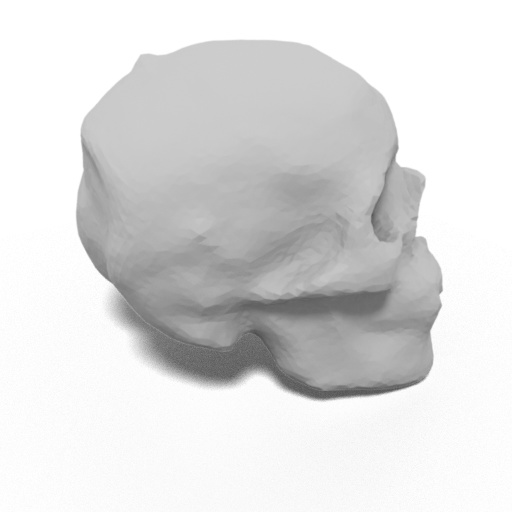}
        \end{tabular}
	\caption{Reconstruction of scan 65 from the DTU MVS dataset~\protect\cite{Jensen2014} without (left) and with (right) the regularization loss based on visibility of faces. The regularization loss removes floaters behind the object that are not visible from the training views.}
	\label{fig:RegVisibility}
\end{figure}
}


\newcommand{\figDamicornisAblation}{
\begin{figure}[t]
	\centering
    \setlength{\tabcolsep}{1pt}
    {\small
    \begin{tabular}{cc}
        \includegraphics[width=0.48\columnwidth]{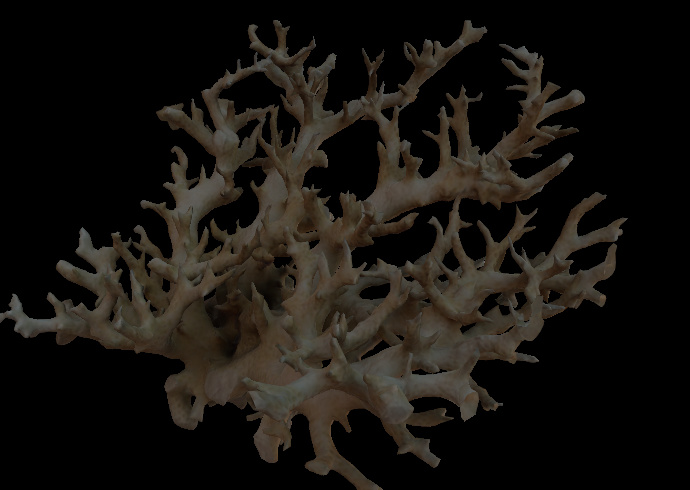} &
        \includegraphics[width=0.48\columnwidth]{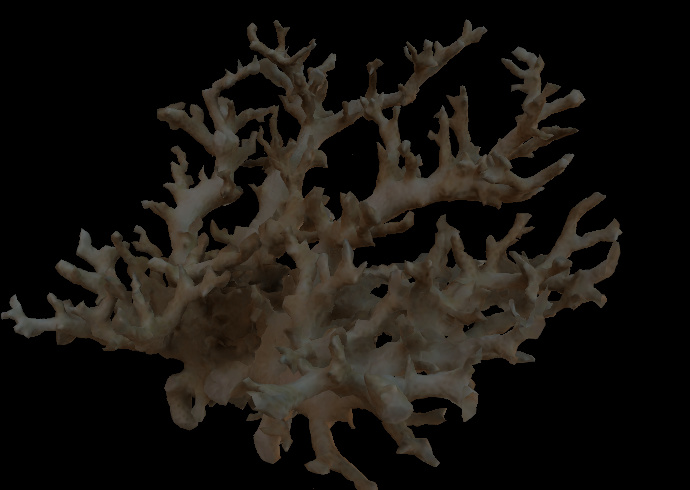} \\
		\includegraphics[width=0.48\columnwidth]{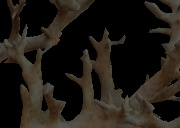} &
        \includegraphics[width=0.48\columnwidth]{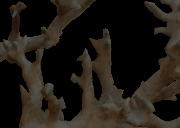} \\
        Grid & MLP 6 \\
    \end{tabular}
	}
	\caption{Comparing grid vs. MLP parametrization of DMTet on the synthetic \textsc{Damicornis} dataset. 
	Directly optimizing per-vertex SDF and offsets on a grid is faster to train, and better captures
	high-frequency geometric details than parametrizing DMTet with a network.}
	\label{fig:DamicornisAblation}
\end{figure}
}


\newcommand{\figDTUAblation}{
\begin{figure}[t]
	\centering
    \setlength{\tabcolsep}{1pt}
    {\small
    \begin{tabular}{ccc}
        \includegraphics[trim={2cm 1cm 2cm 0cm},width=0.34\columnwidth]{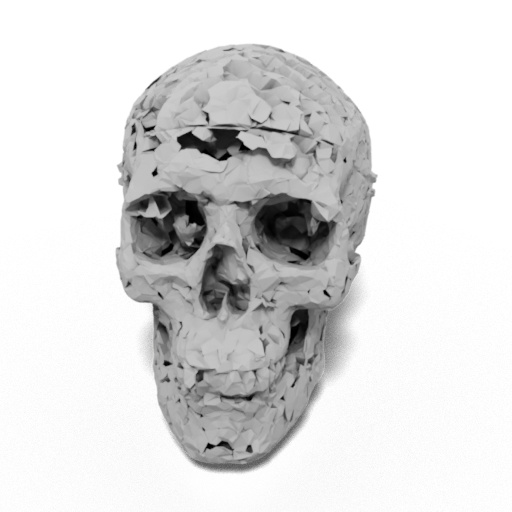} &
        \includegraphics[trim={2cm 0cm 2cm 0cm},width=0.32\columnwidth]{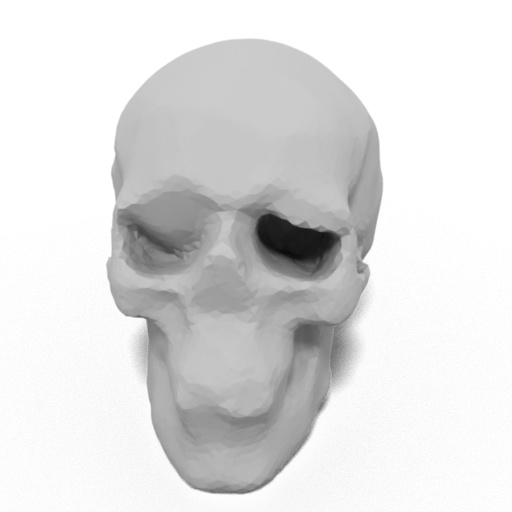} & 
        \includegraphics[trim={2cm 0cm 2cm 0cm},width=0.32\columnwidth]{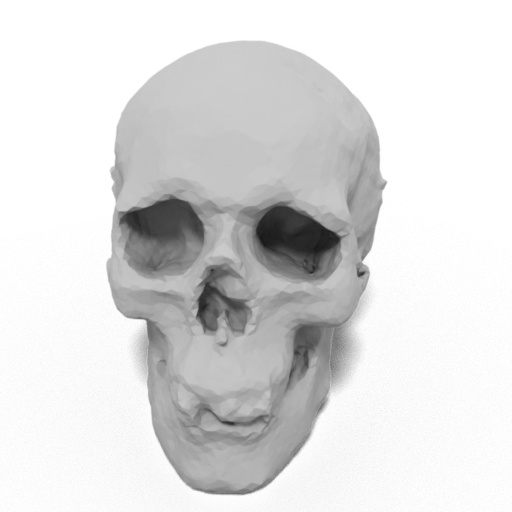}\\
        Grid & MLP f:4 & MLP f:6 \\
    \end{tabular}
	}
	\caption{Comparing grid vs. MLP parametrizations of DMTet on scan 65 from the DTU MVS dataset~\protect\cite{Jensen2014}. Directly optimizing SDF values at grid vertices leads to a surface
	 with high-frequency noise (left). In contrast, if we use an MLP to parametrize the SDF values, we can regularize the geometry, 
	 with smoothness controlled by the frequency of positional encoding. We use the positional encoding in NeRF~\cite{Mildenhall2020} with frequency set to 4 (middle) and 6 (right) 
	 respectively.}
	\label{fig:DTUAblation}
\end{figure}
}


\newlength{\mydtusize}
\setlength{\mydtusize}{0.16\textwidth}

\newcommand{\figDTUNeural}{
	\begin{figure*}
		\centering
		\setlength{\tabcolsep}{1pt}
		{\small
		\begin{tabular}{lcccccc}
			& Reference & Our & $\kd$ & $\korm$ & normals & HDR probe \\

			\rotatebox[origin=c]{90}{Scan 65} &
			\raisebox{-0.5\height}{\includegraphics[width=\mydtusize]{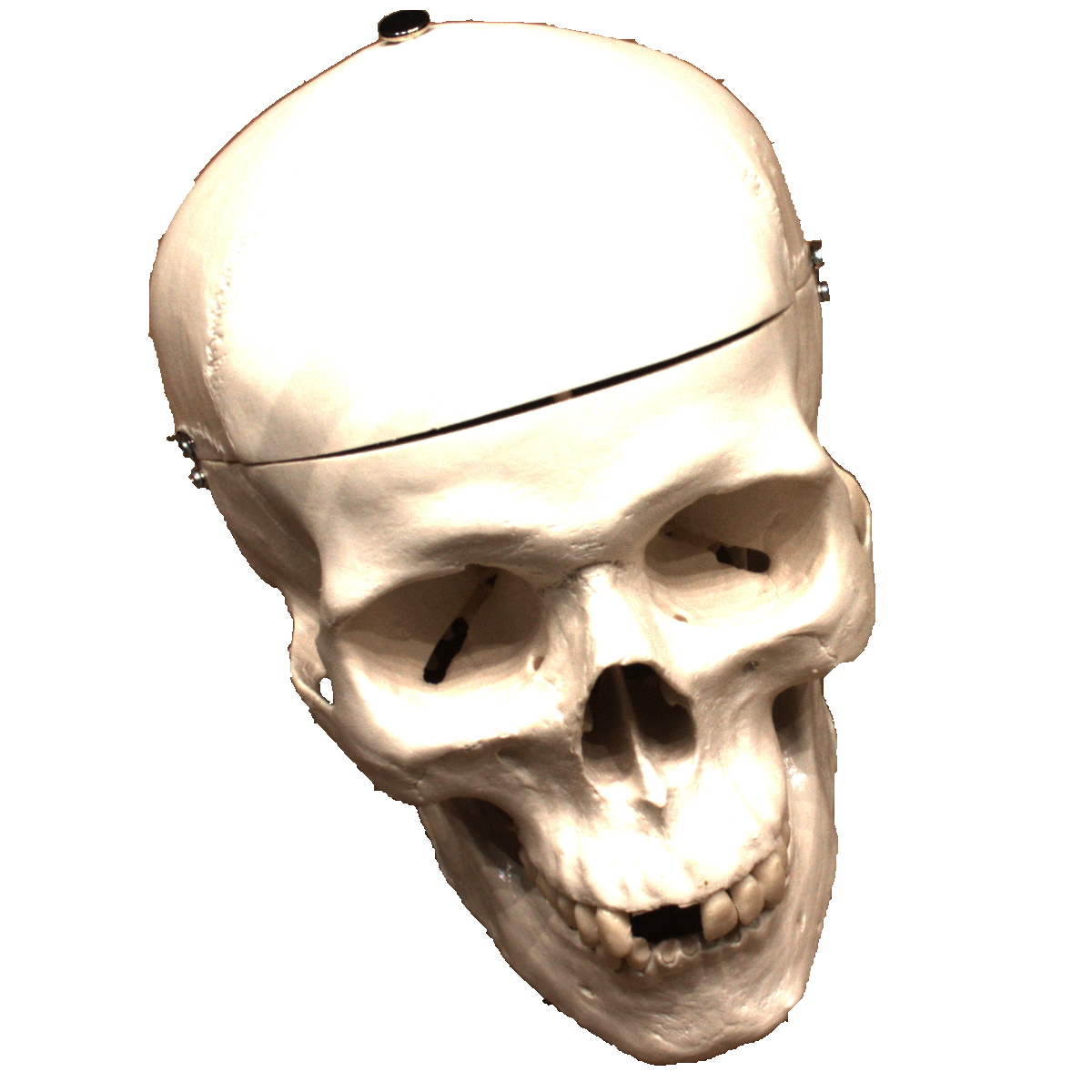}} &
			\raisebox{-0.5\height}{\includegraphics[width=\mydtusize]{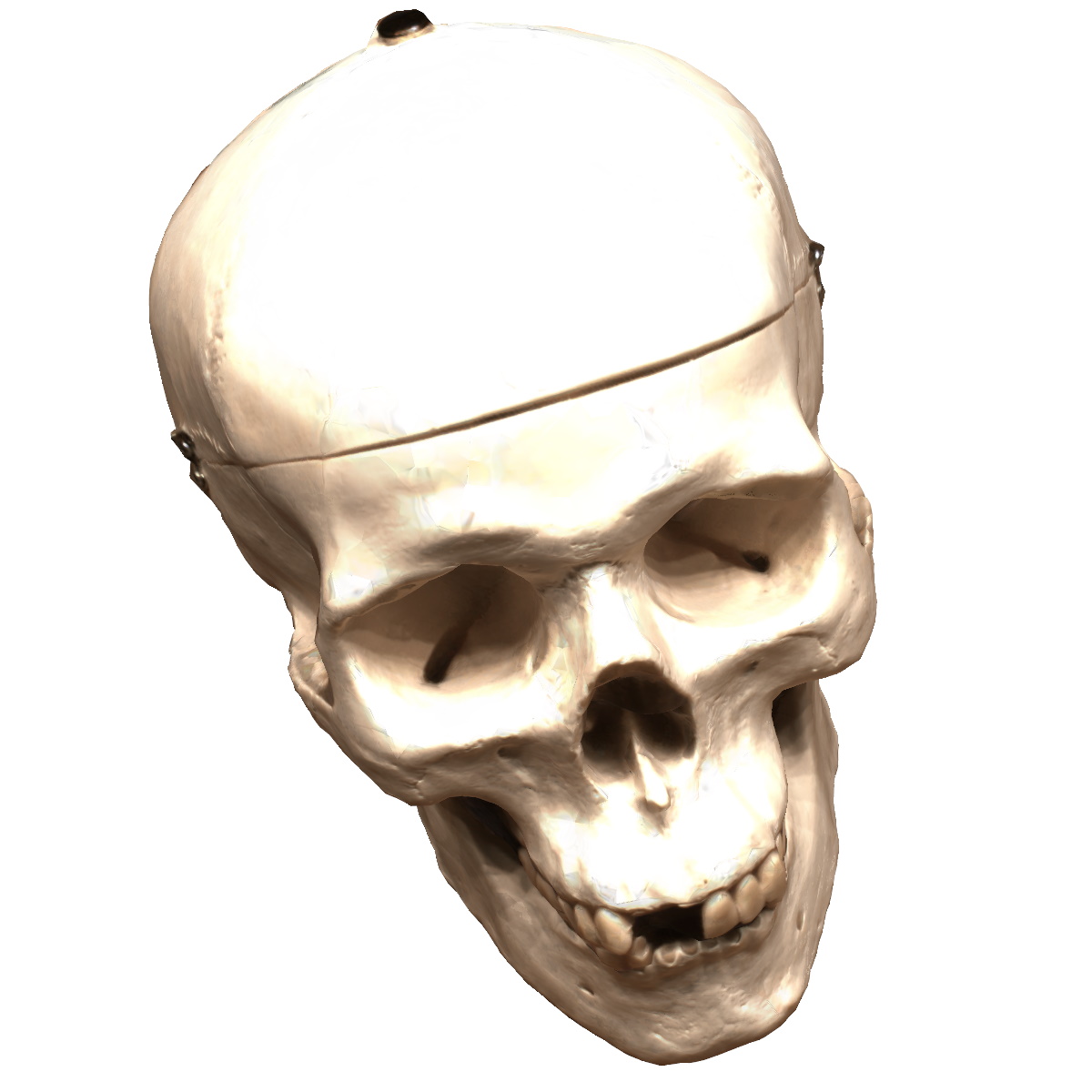}} &
			\raisebox{-0.5\height}{\includegraphics[width=\mydtusize]{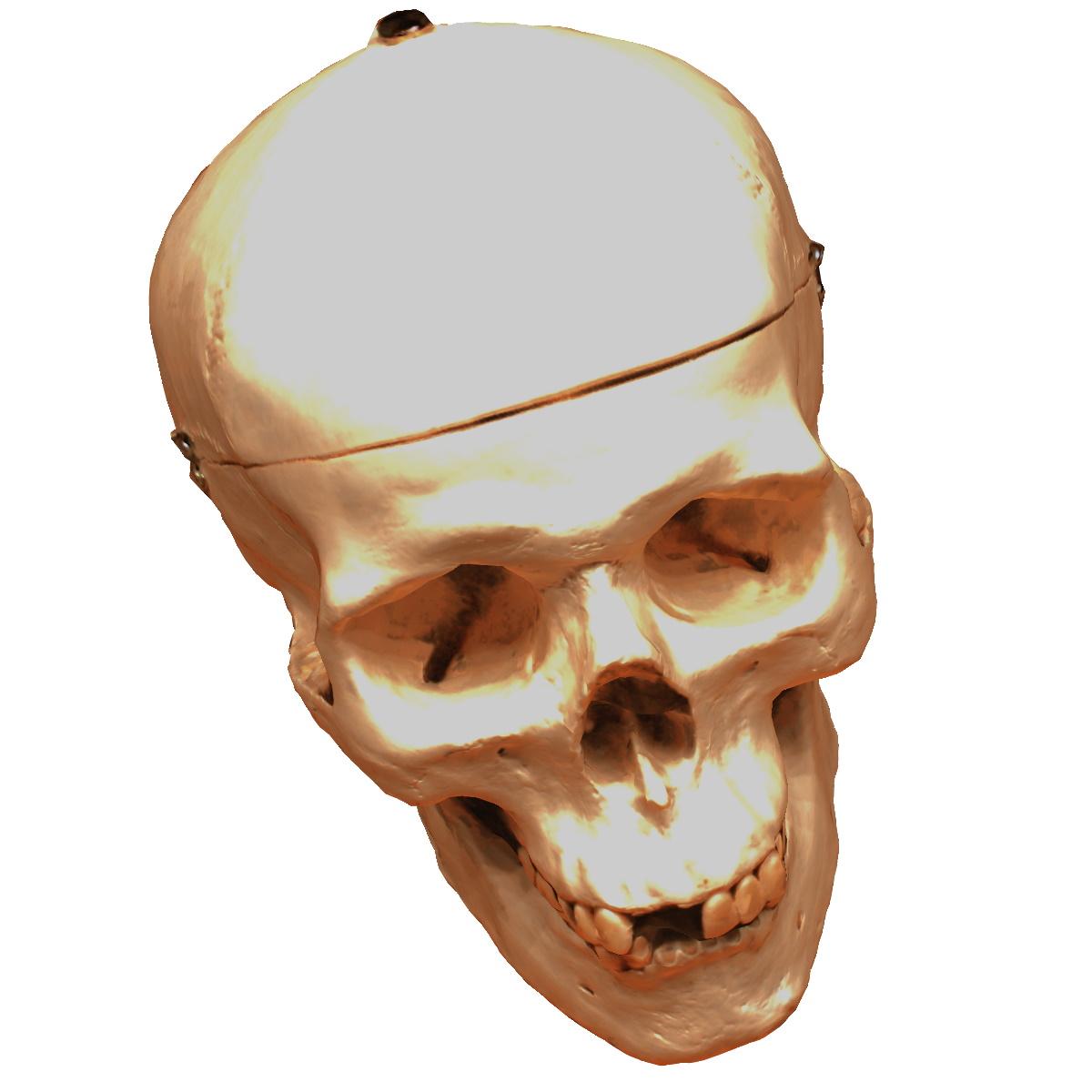}} &
			\raisebox{-0.5\height}{\includegraphics[width=\mydtusize]{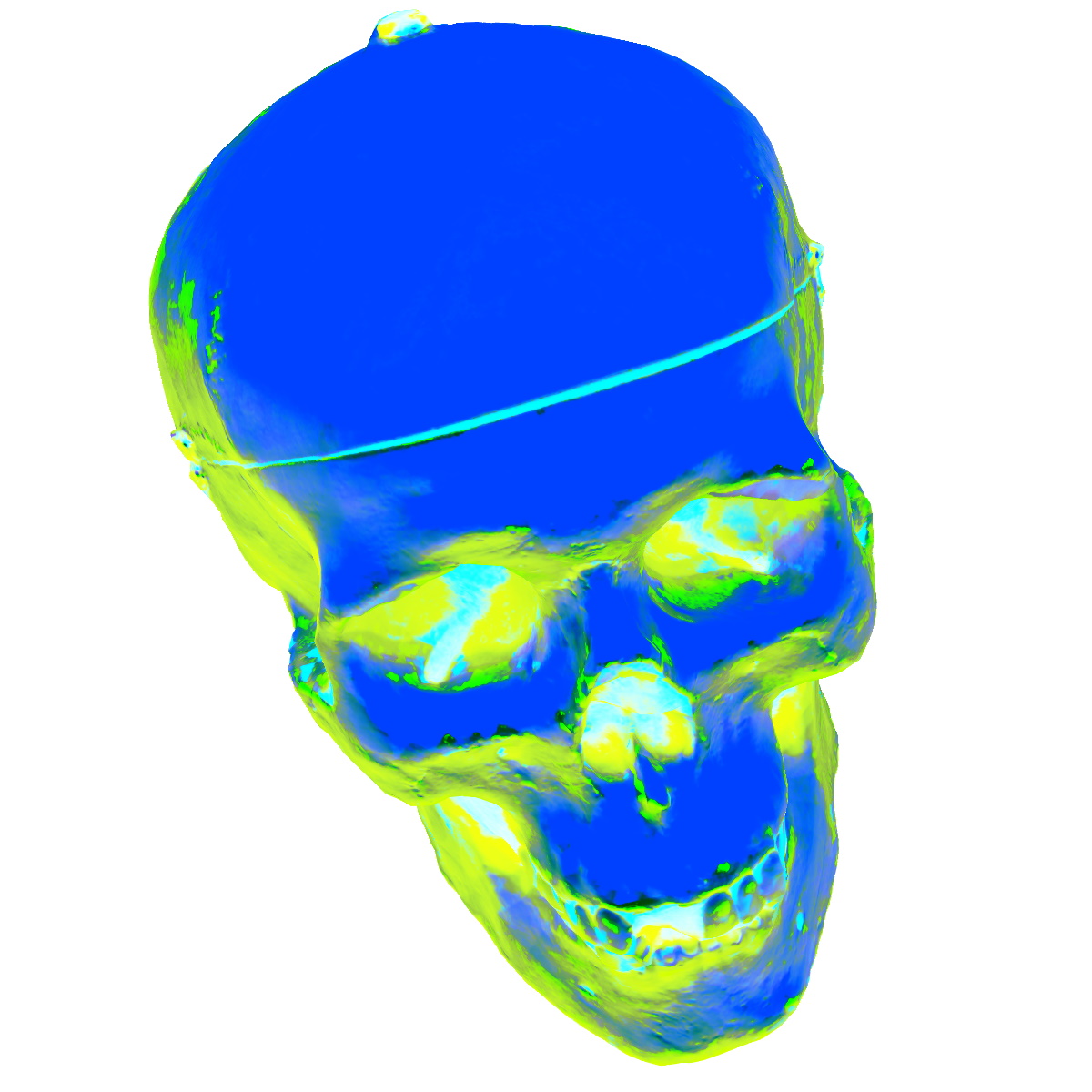}} &
			\raisebox{-0.5\height}{\includegraphics[width=\mydtusize]{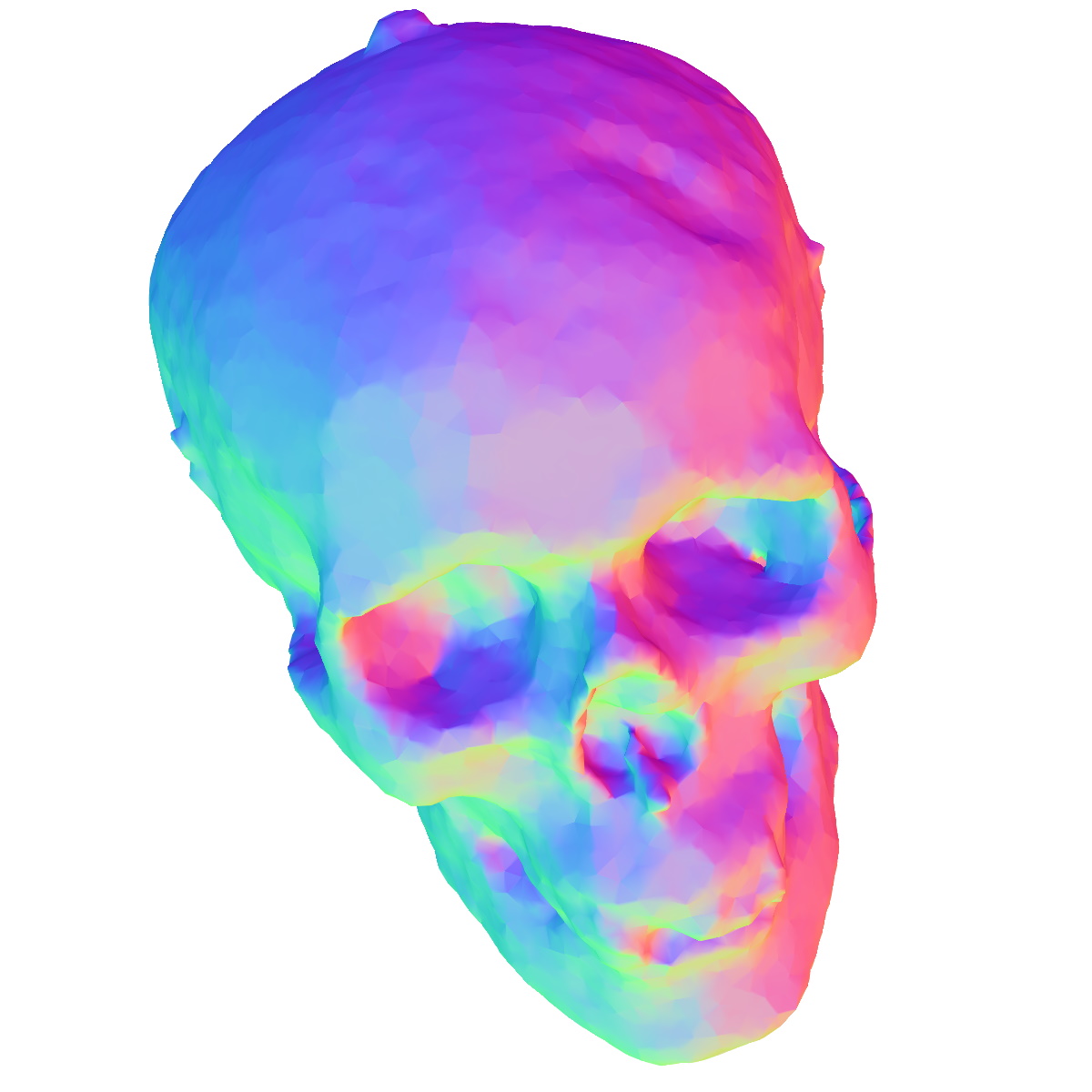}} &
			\raisebox{-0.5\height}{\includegraphics[width=\mydtusize]{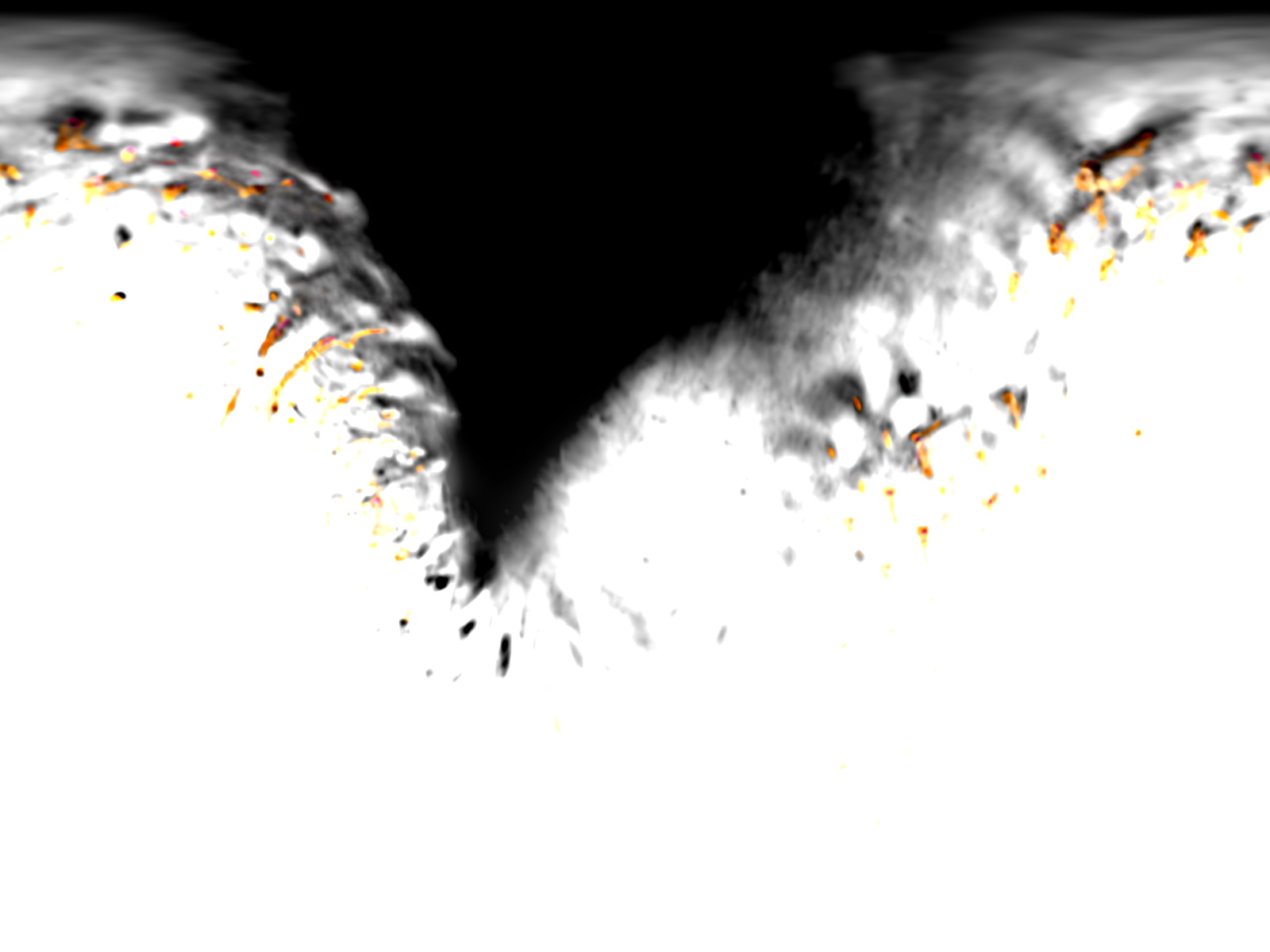}} \\

			\rotatebox[origin=c]{90}{Scan 106} &
			\raisebox{-0.5\height}{\includegraphics[width=\mydtusize]{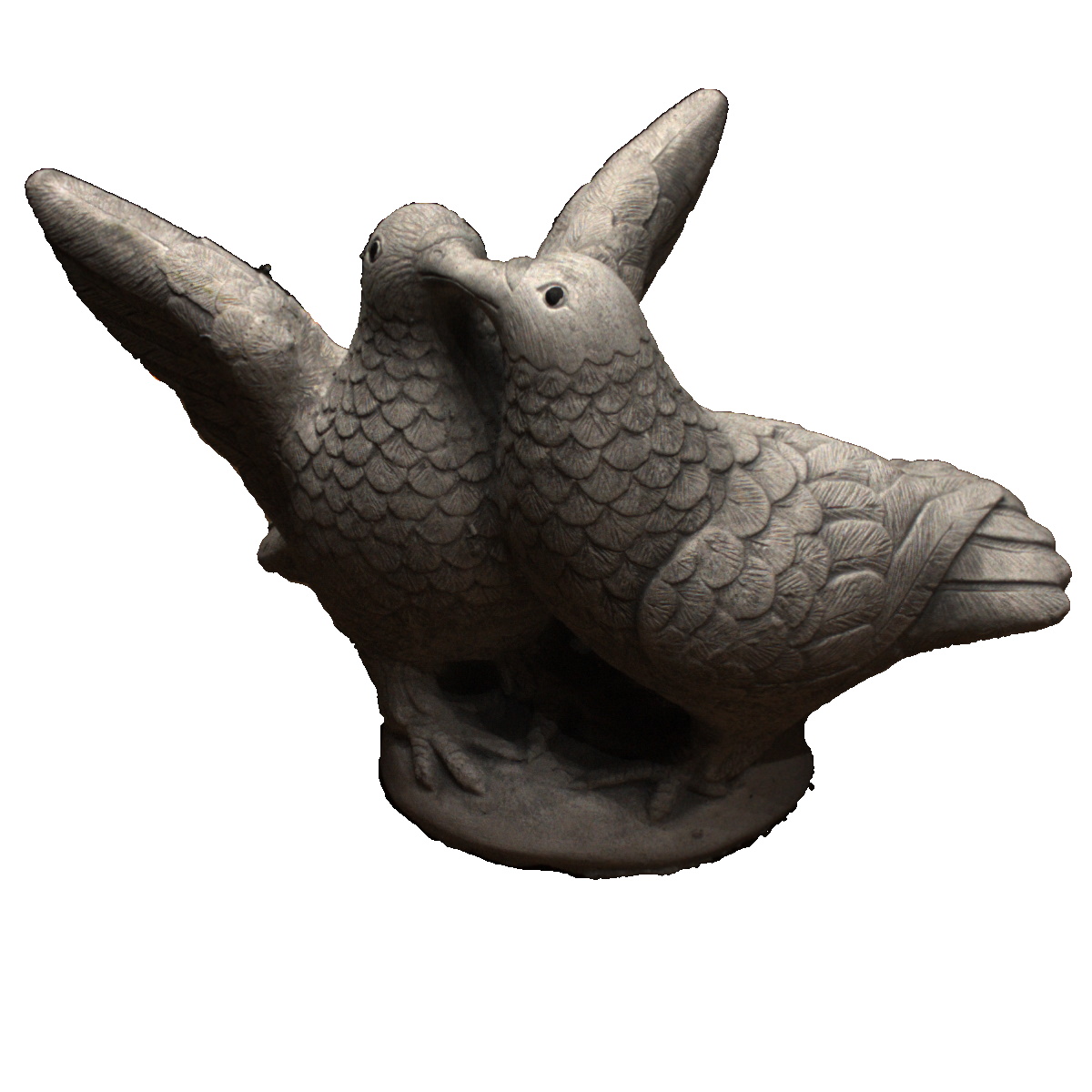}} &
			\raisebox{-0.5\height}{\includegraphics[width=\mydtusize]{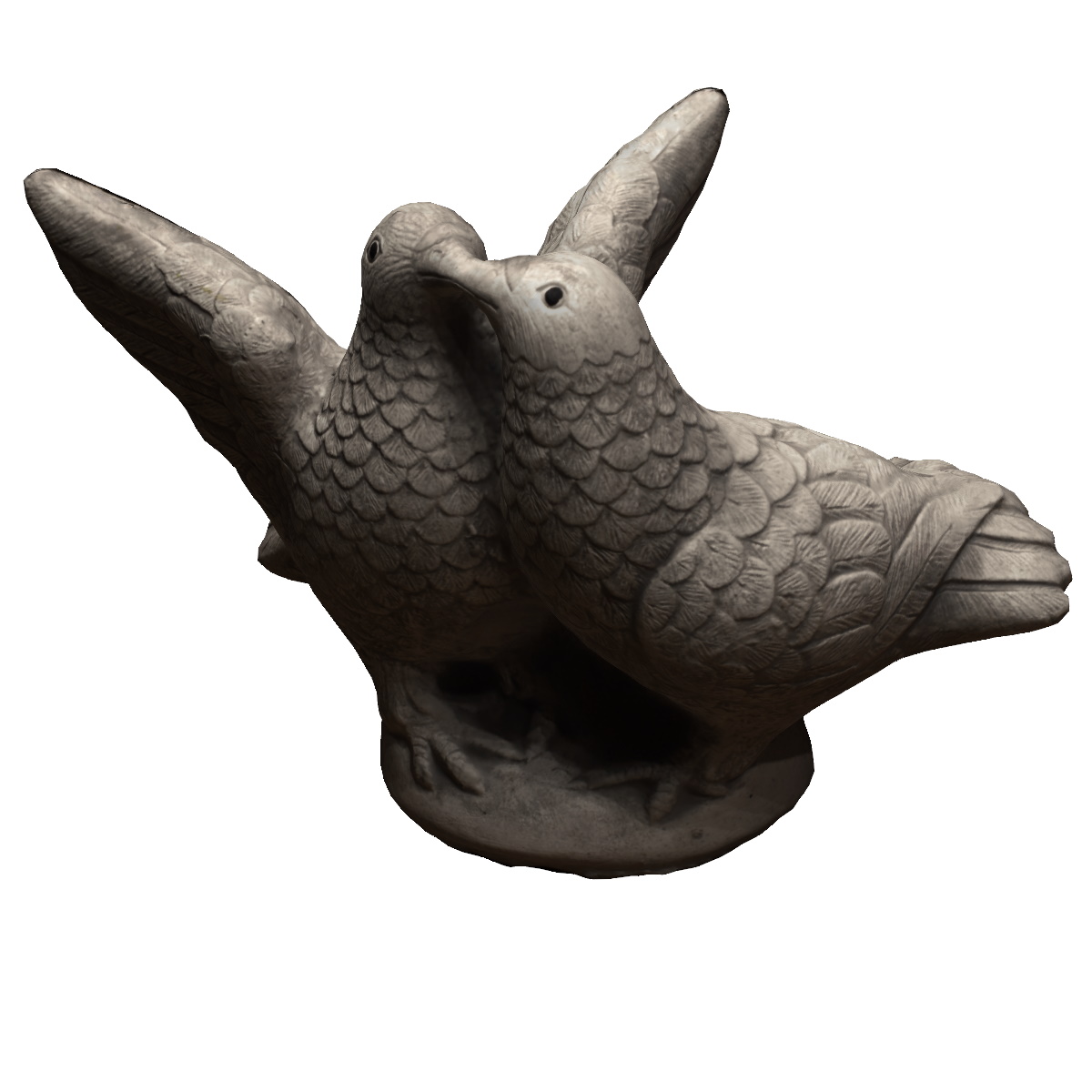}} &
			\raisebox{-0.5\height}{\includegraphics[width=\mydtusize]{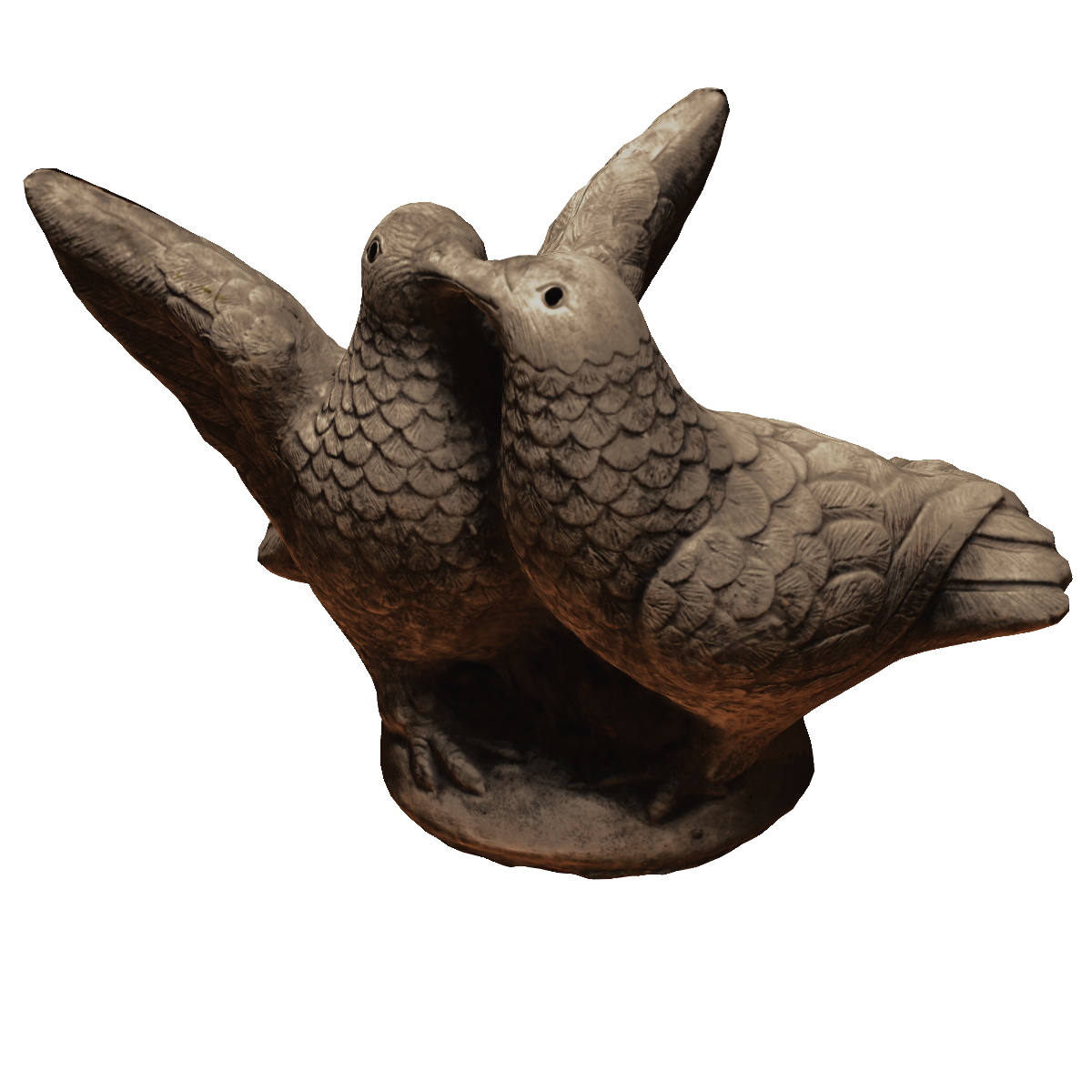}} &
			\raisebox{-0.5\height}{\includegraphics[width=\mydtusize]{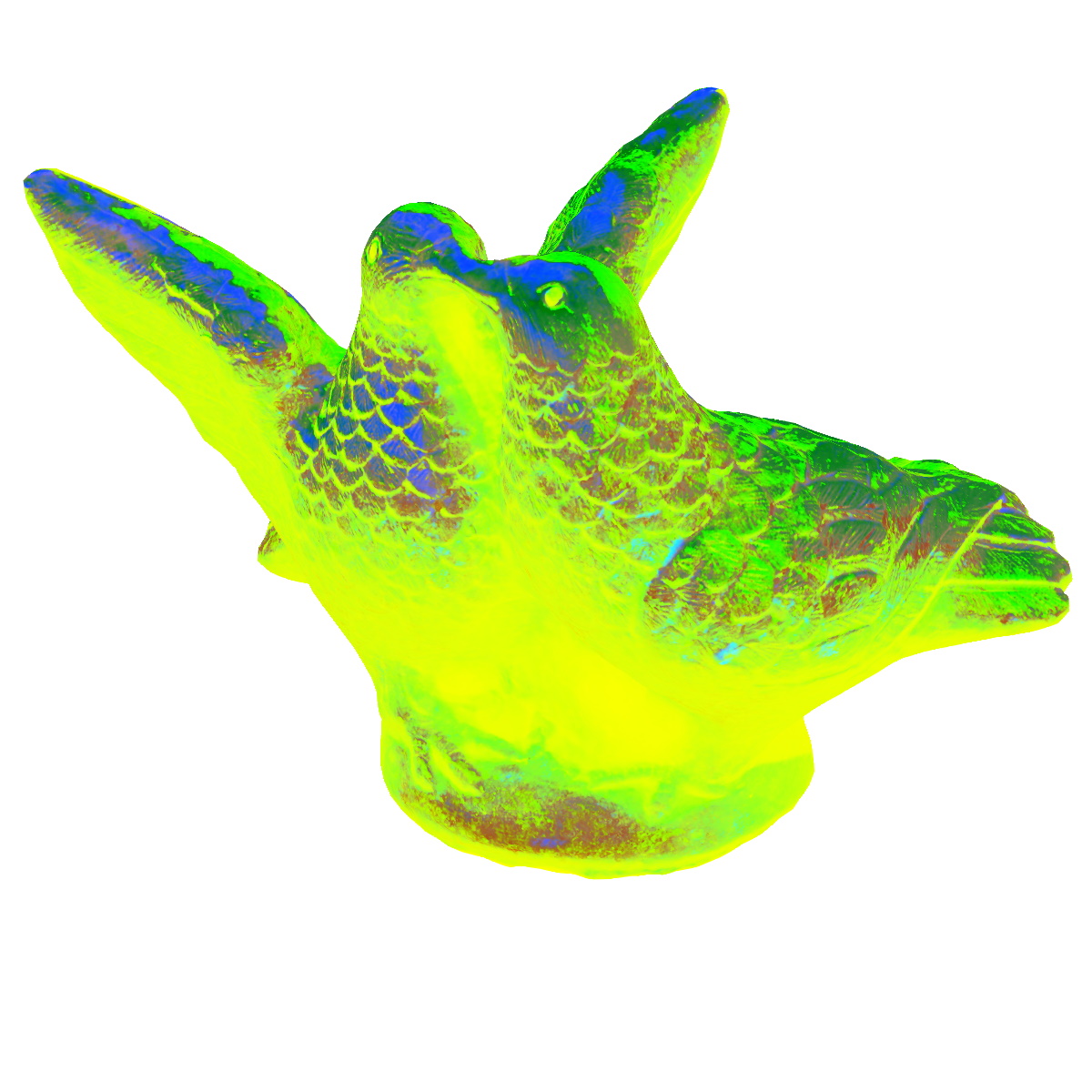}} &
			\raisebox{-0.5\height}{\includegraphics[width=\mydtusize]{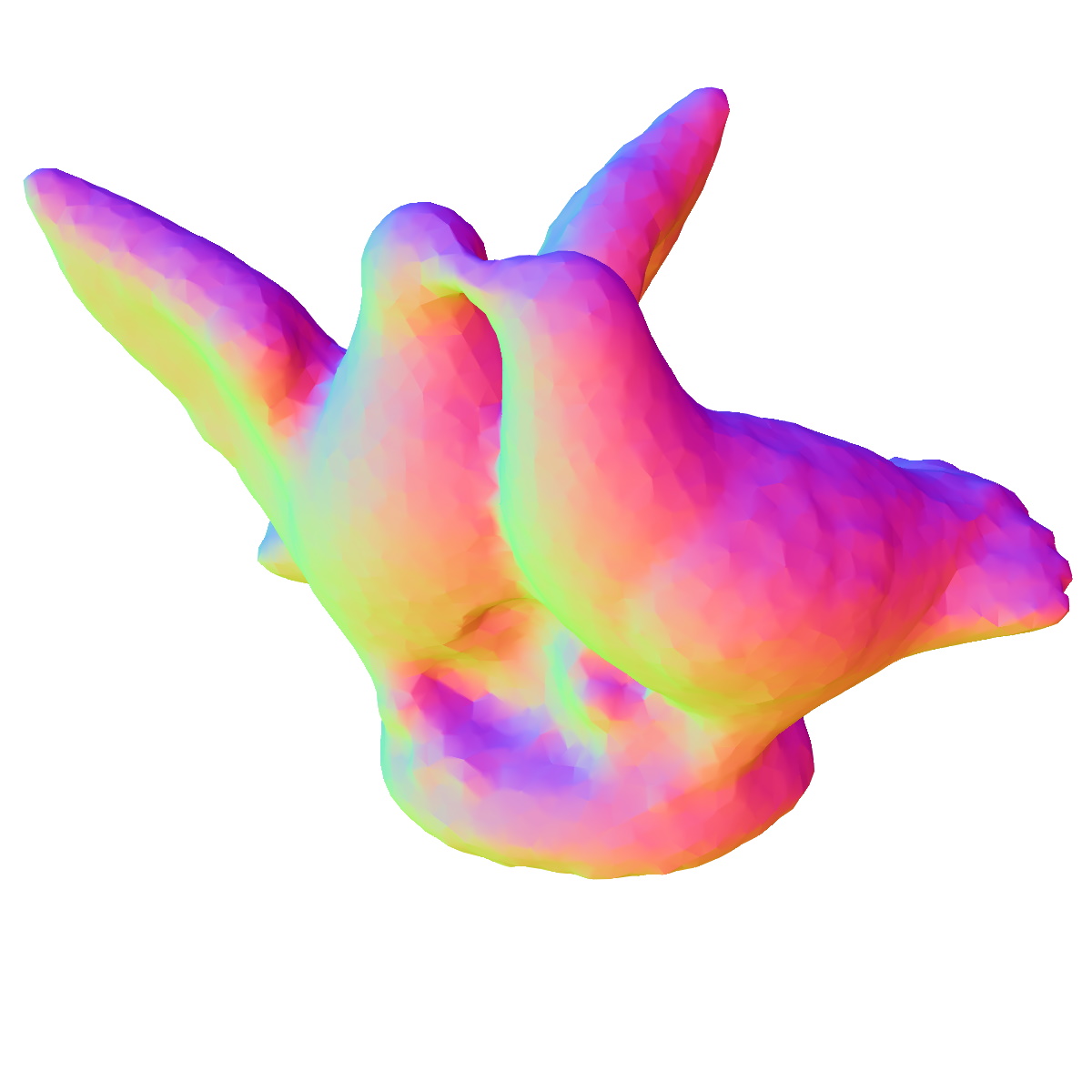}} &
			\raisebox{-0.5\height}{\includegraphics[width=\mydtusize]{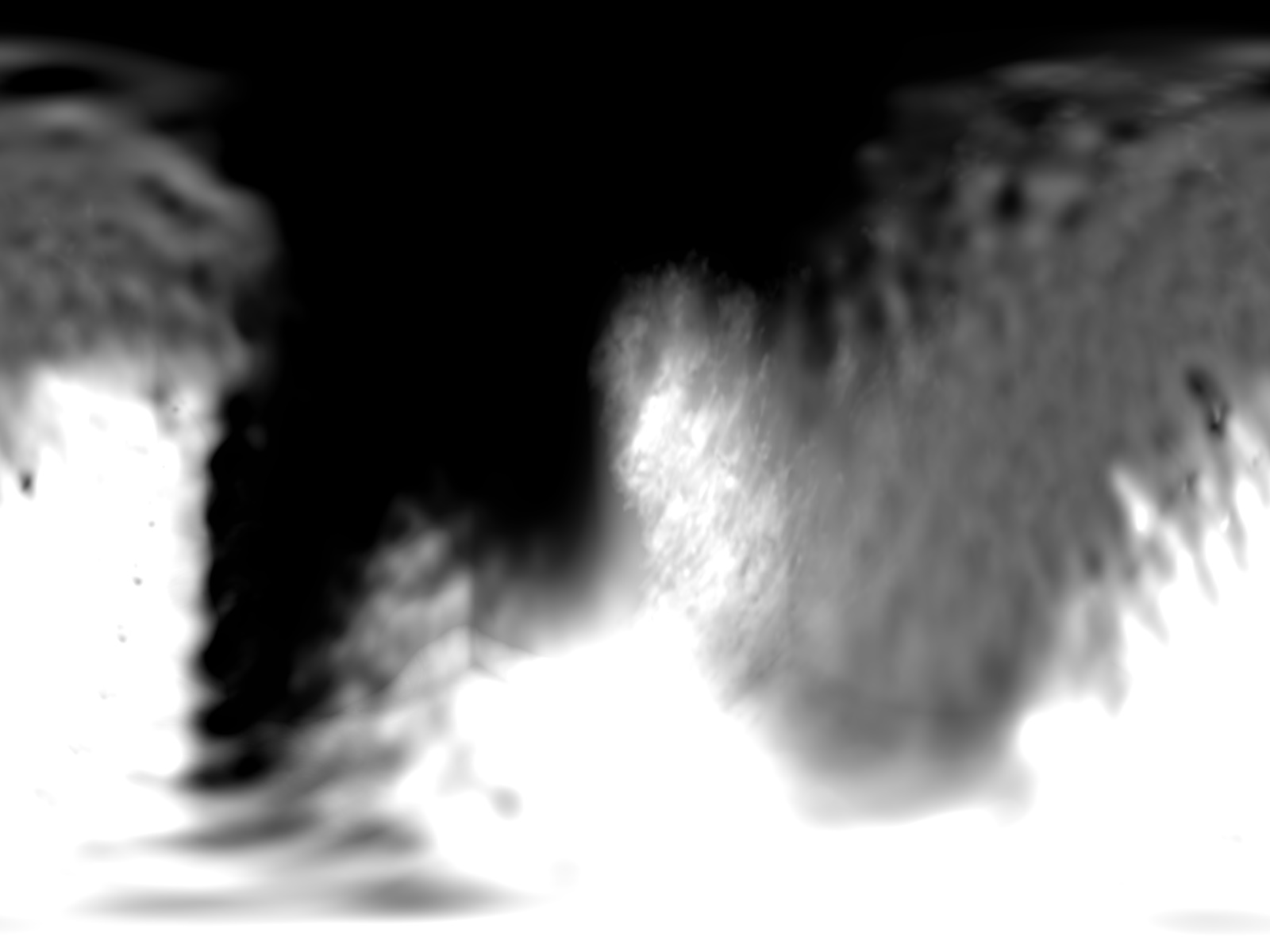}} \\

			\rotatebox[origin=c]{90}{Scan 118} &
			\raisebox{-0.5\height}{\includegraphics[width=\mydtusize]{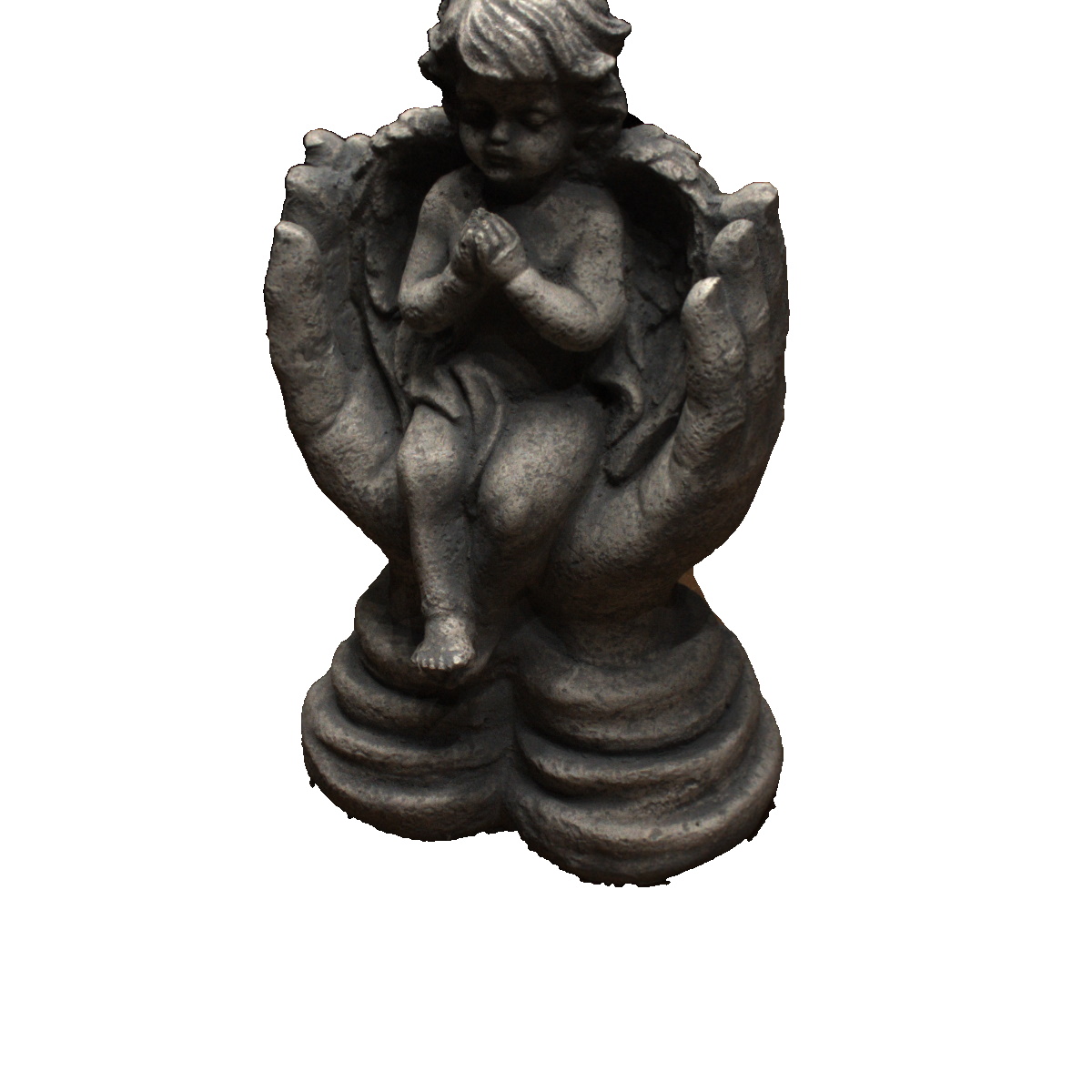}} &
			\raisebox{-0.5\height}{\includegraphics[width=\mydtusize]{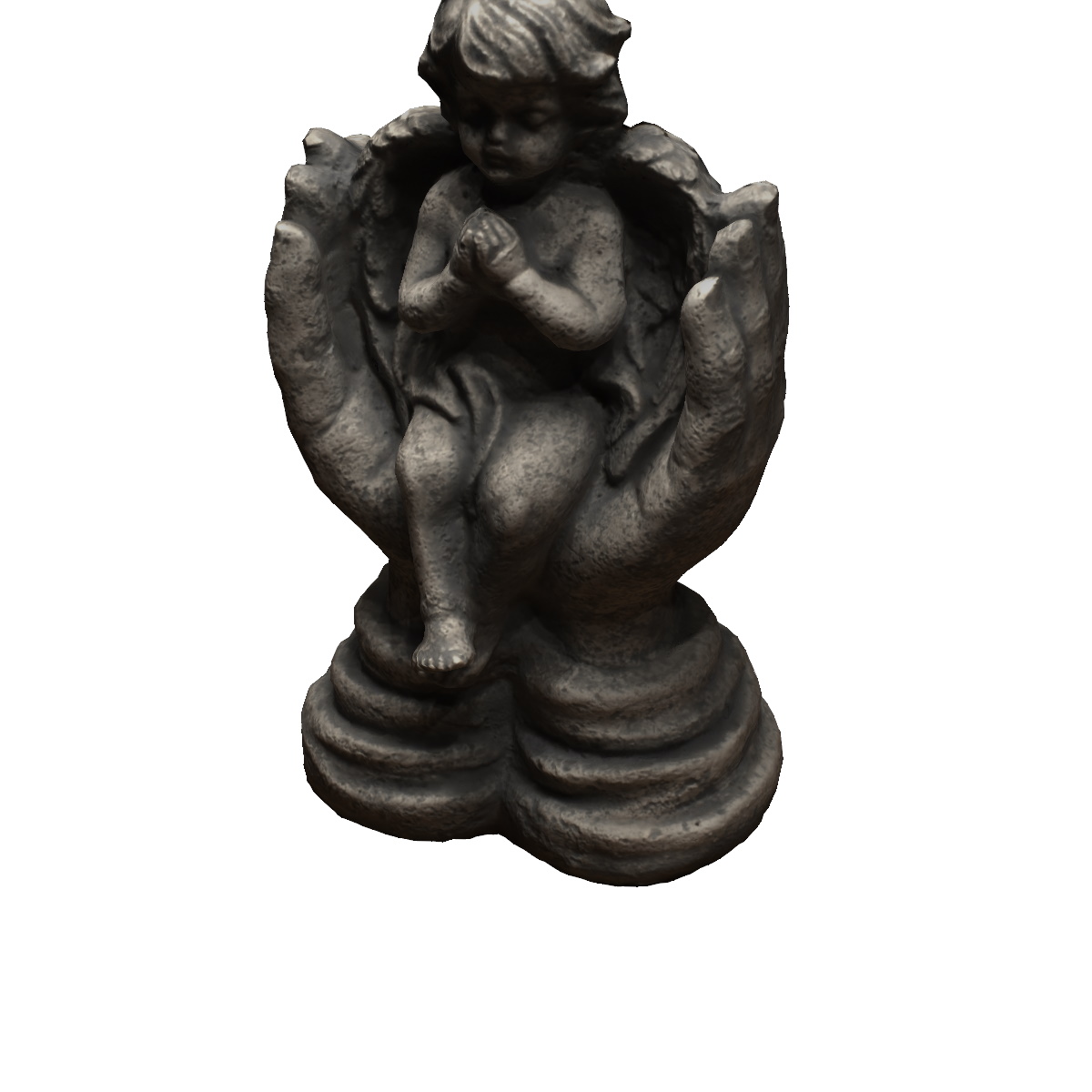}} &
			\raisebox{-0.5\height}{\includegraphics[width=\mydtusize]{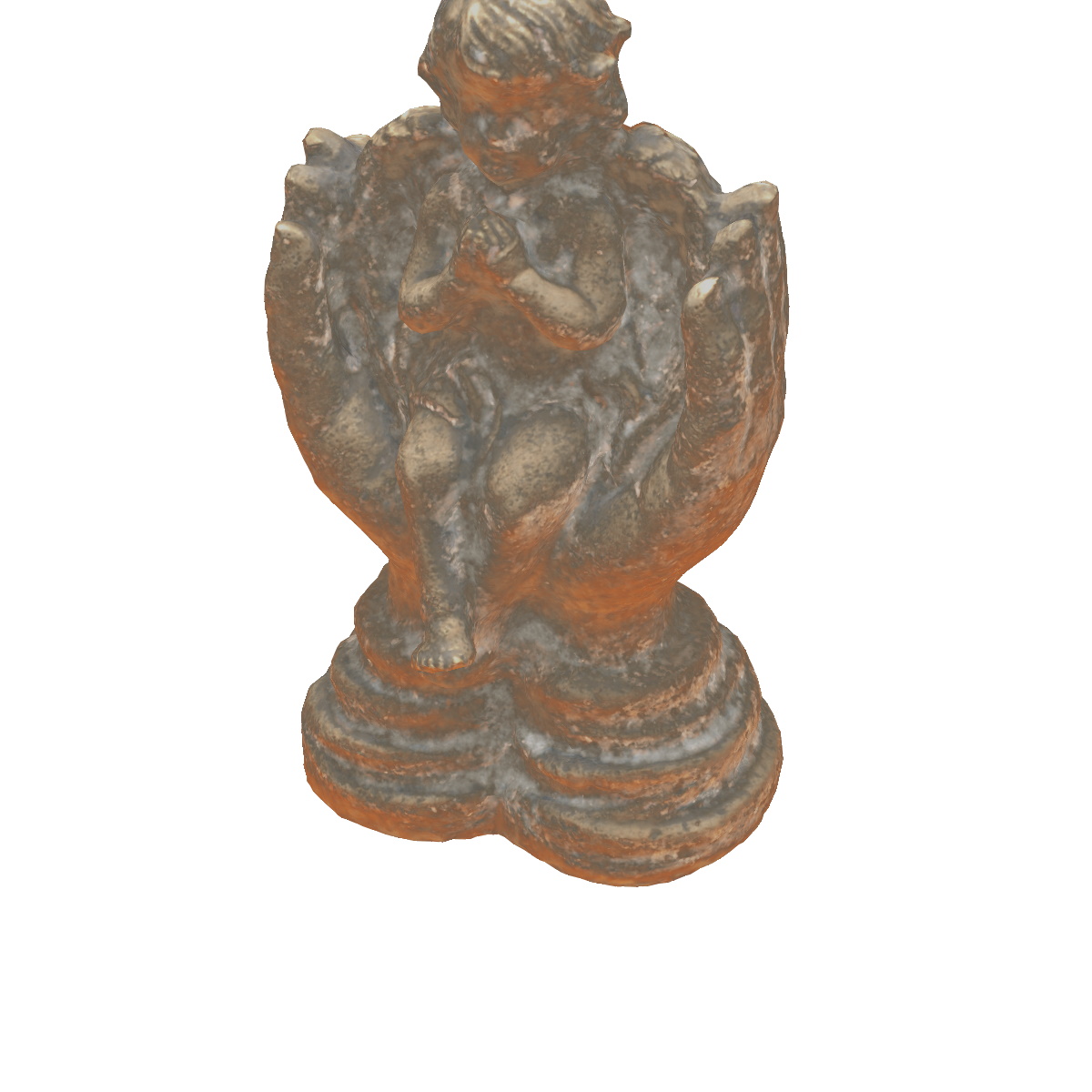}} &
			\raisebox{-0.5\height}{\includegraphics[width=\mydtusize]{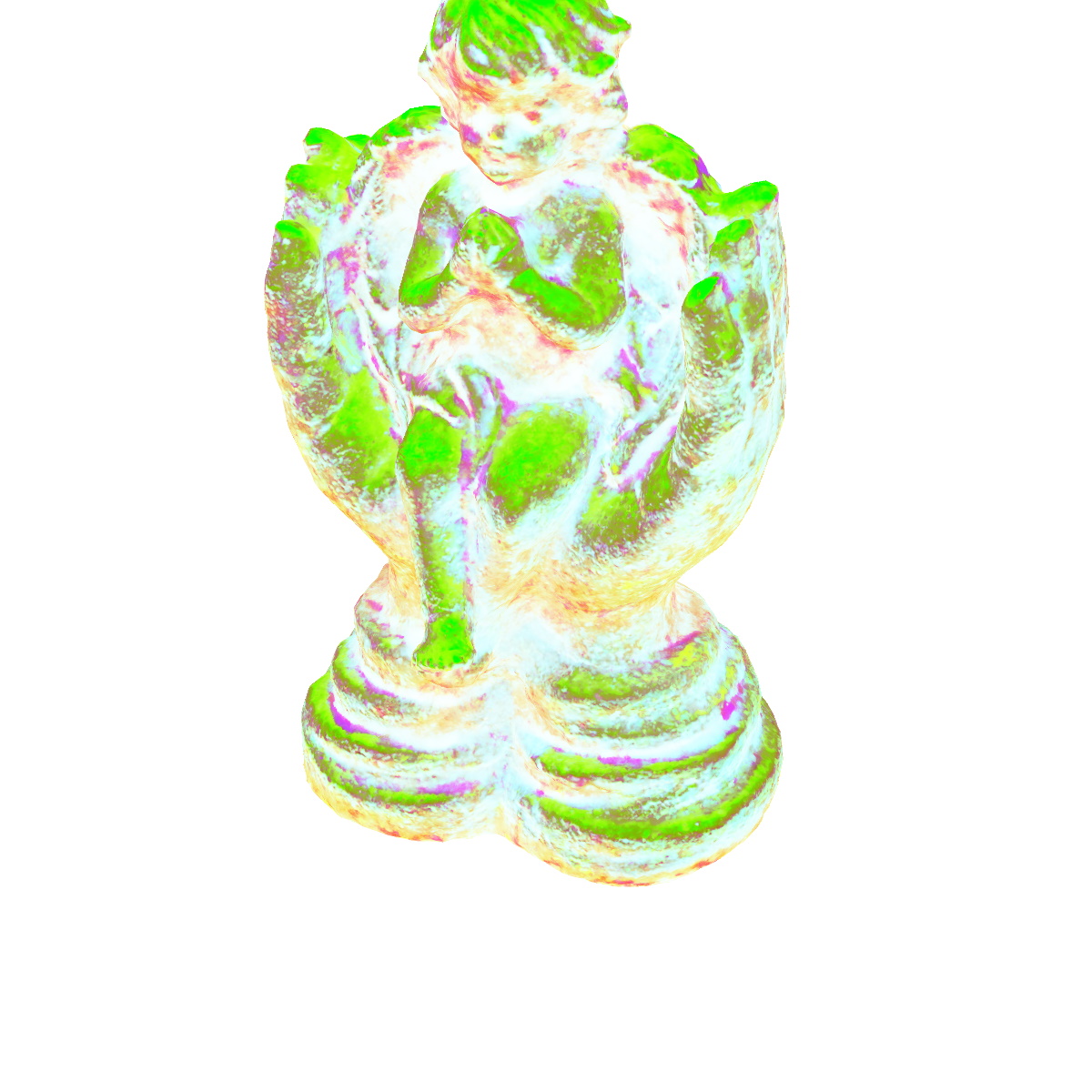}} &
			\raisebox{-0.5\height}{\includegraphics[width=\mydtusize]{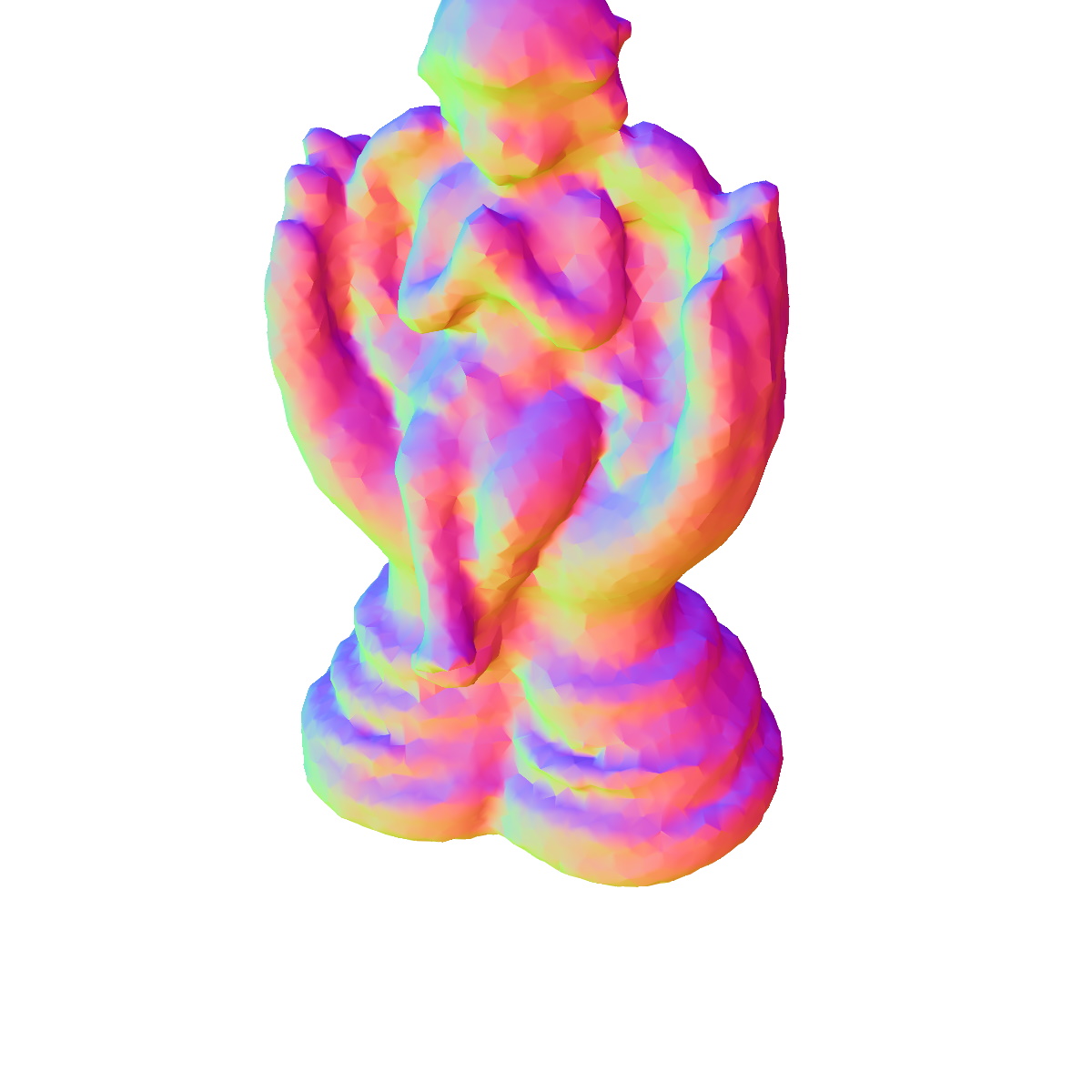}} &
			\raisebox{-0.5\height}{\includegraphics[width=\mydtusize]{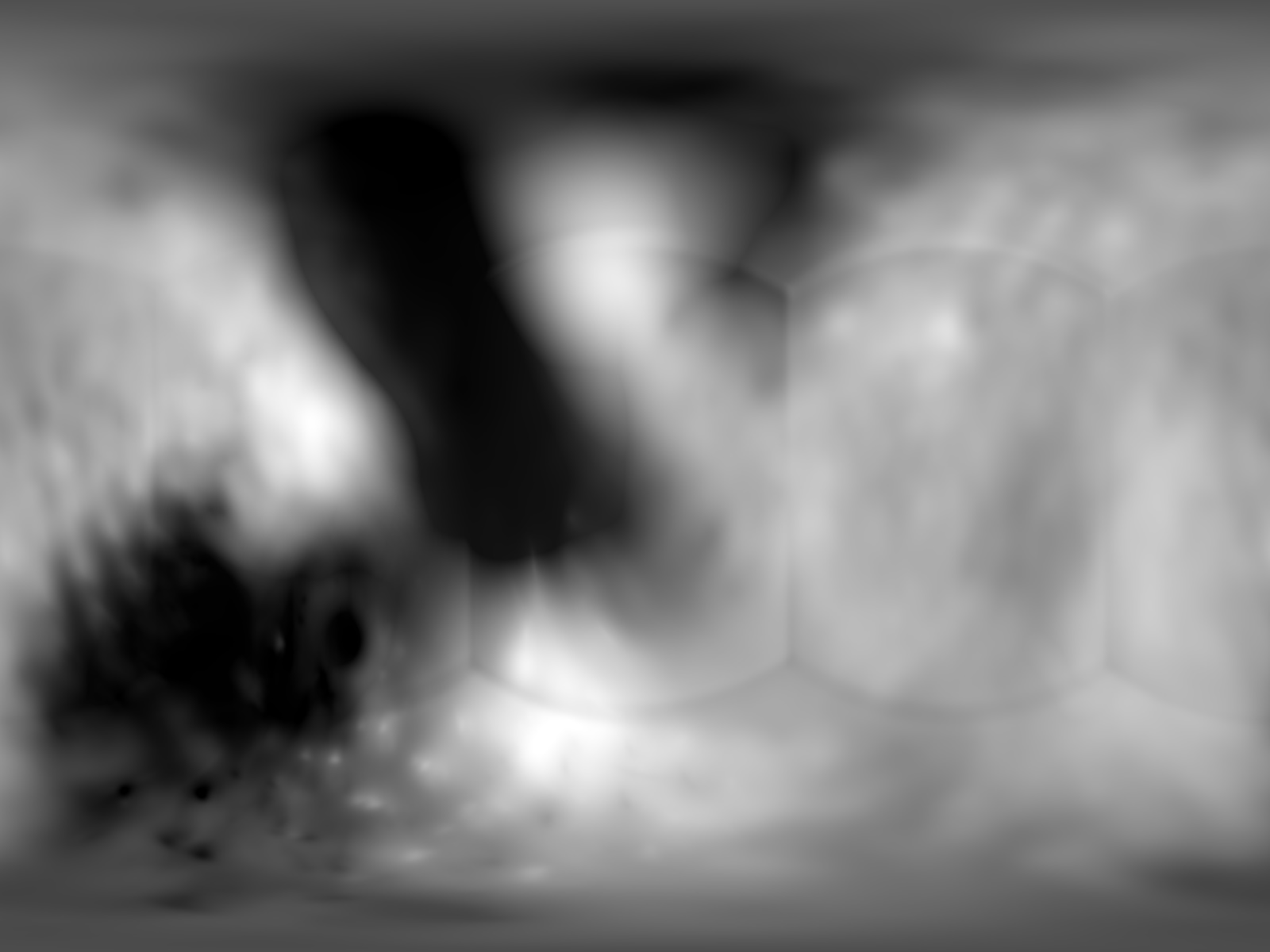}} \\

			& Reference & Our & $\kd$ & $\korm$ & normals & HDR probe \\
		\end{tabular}
		}
		\caption{
			Our decomposition results on scan 65, 106, and 118 of the DTU MVS dataset~\protect\cite{Jensen2014}. 
			Our model is trained on a reduced subset (49 of the 64 views) which has more consistent lighting across views, 
			labelled by DVR~\cite{Niemeyer2020CVPR}. However, we still penalize mask loss on the excluded views.
		}
		\label{fig:DTUNeural}
	\end{figure*}
}


\newcommand{\figMaterialModel}{
\begin{figure}
	\centering
	\setlength{\tabcolsep}{1pt}
	\begin{tabular}{cccc}
		\includegraphics[height=0.23\columnwidth]{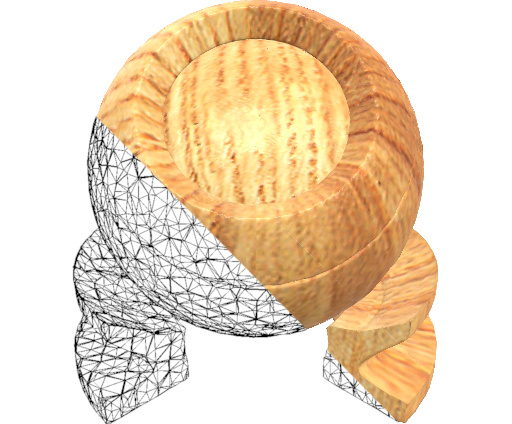} &
        \includegraphics[height=0.23\columnwidth]{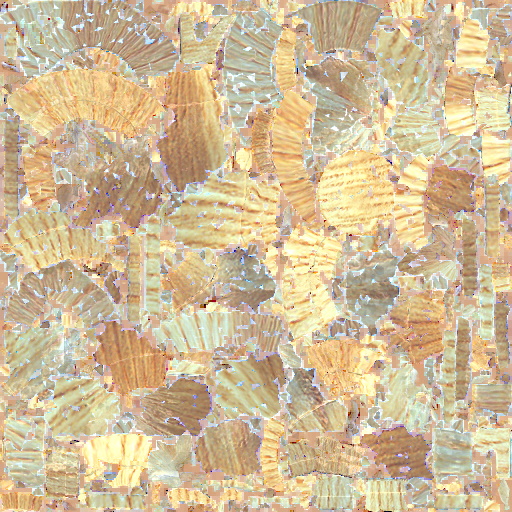} &
		\includegraphics[height=0.23\columnwidth]{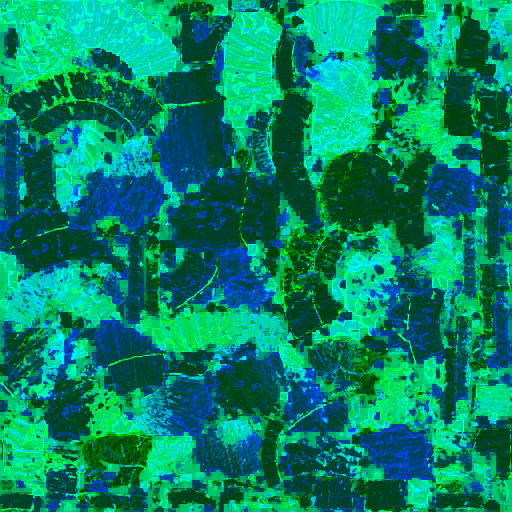} &
		\includegraphics[height=0.23\columnwidth]{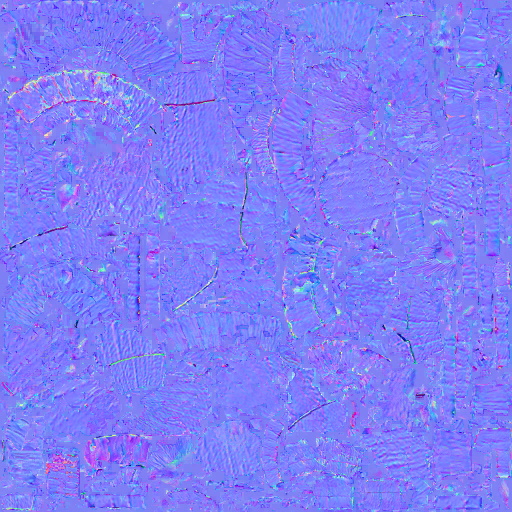} \\
		\small{wire/shaded} & \small{$\kd$} & \small{$\korm$} & \small{normals} \\
	\end{tabular}
	\vspace{-3mm}
	\caption{      
		We represent 3D models as a triangular mesh and 
        a set of spatially varying materials following a standard PBR model.
	}
	\label{fig:material_model}
	\vspace{-2mm}
\end{figure}
}


\newcommand{\figTexSeams}{
\begin{figure}
	\centering
	\setlength{\tabcolsep}{1pt}
	\begin{tabular}{cc}
		\includegraphics[width=0.4\columnwidth]{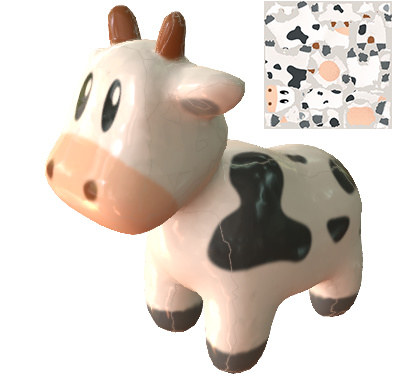} &
		\hspace{5mm}
		\includegraphics[width=0.4\columnwidth]{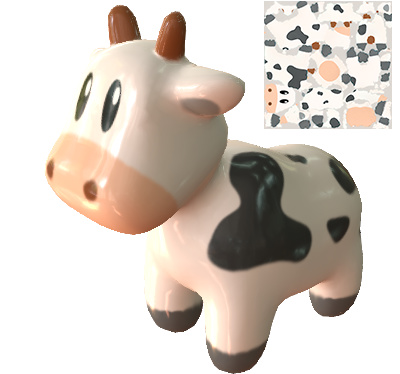}
	\end{tabular}
	\vspace{-4mm}
	\caption{
		Sampling out the volumetric representation to create 2D textures
		results in texture seams (left). However, further optimization (right),
		quickly removes the seams automatically. 
	}
	\vspace{-2mm}
	\label{fig:tex_seams}
\end{figure}
}


\newcommand{\figSG}{
\begin{figure}[t]
	\centering
	\setlength{\tabcolsep}{1pt}
	{\small
	\begin{tabular}{lccc}
		& Reference & SG~128 & Split Sum  \\
		
		\rotatebox[origin=c]{90}{Plastic} &
		\raisebox{-0.5\height}{\includegraphics[width=0.30\columnwidth]{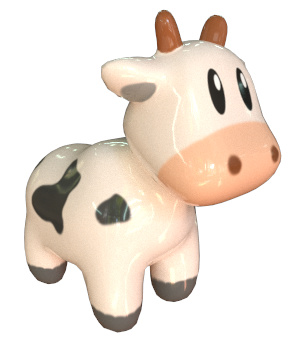}} &
		\raisebox{-0.5\height}{\includegraphics[width=0.30\columnwidth]{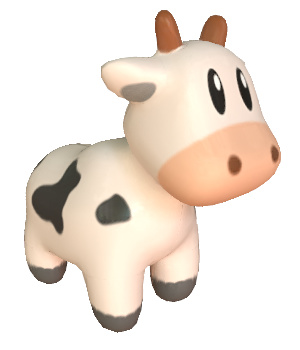}} &
		\raisebox{-0.5\height}{\includegraphics[width=0.30\columnwidth]{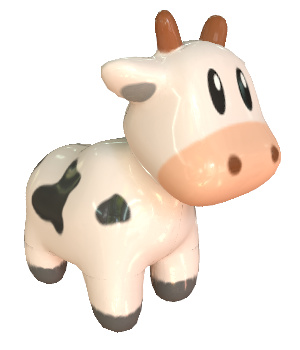}} \\
		& PSNR $|$ SSIM & 33.74 $|$ 0.968 & 36.20 $|$ 0.977 \\
		&
		\includegraphics[width=0.30\columnwidth]{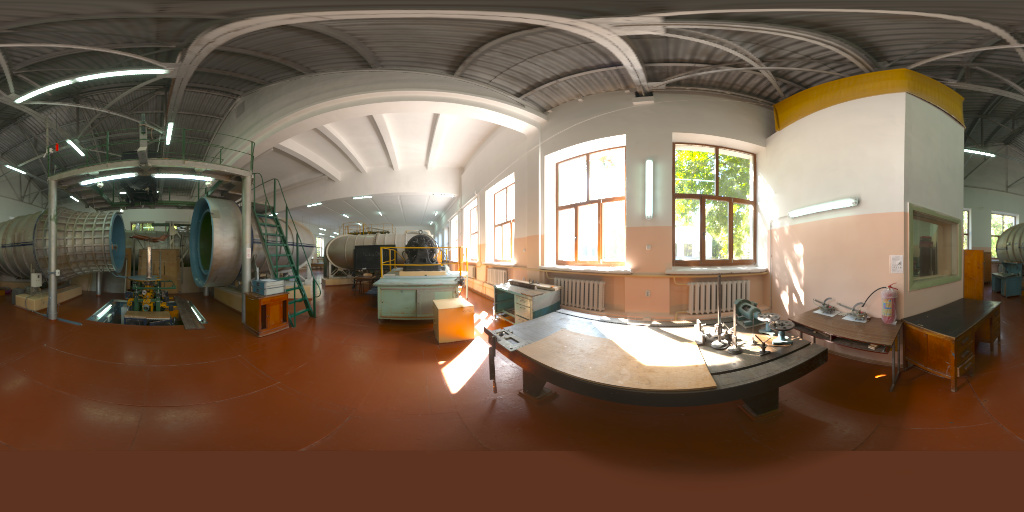} &
		\includegraphics[width=0.30\columnwidth]{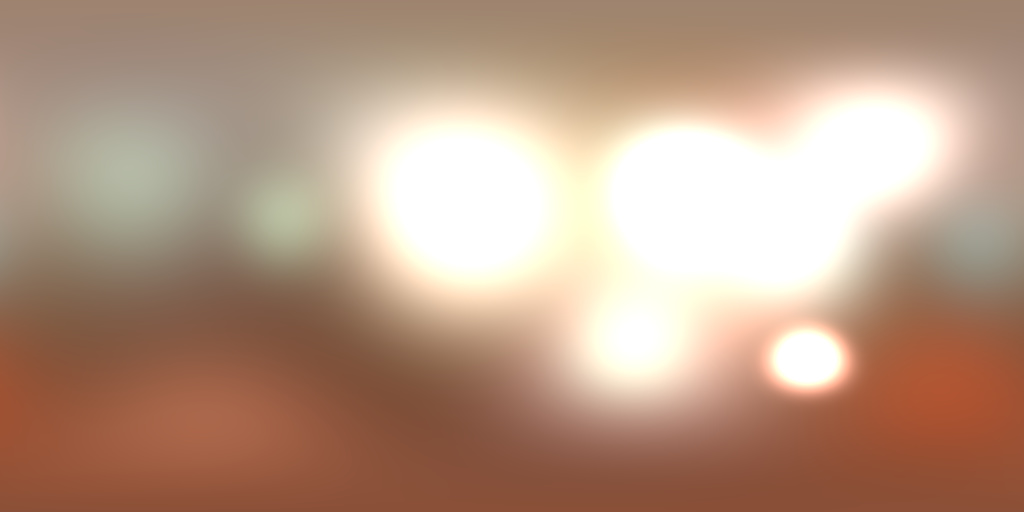} &
		\includegraphics[width=0.30\columnwidth]{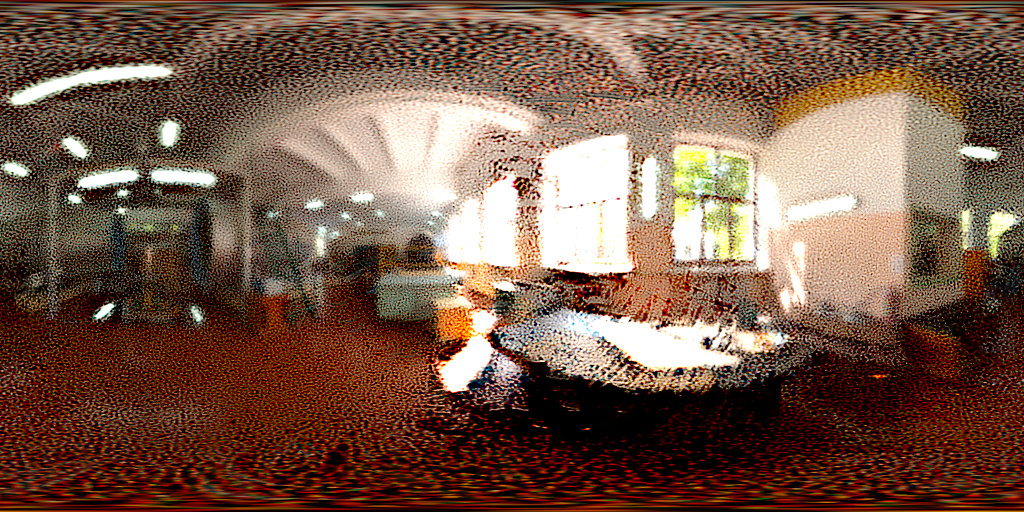} \\
		\rotatebox[origin=c]{90}{Metal} &
		\raisebox{-0.5\height}{\includegraphics[width=0.30\columnwidth]{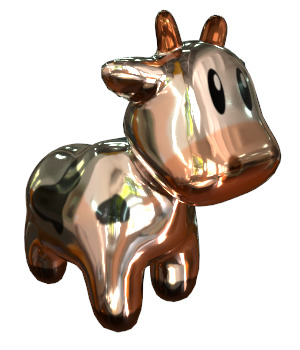}} &
		\raisebox{-0.5\height}{\includegraphics[width=0.30\columnwidth]{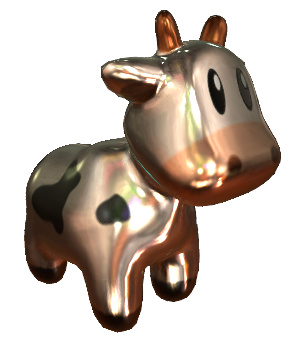}} &
		\raisebox{-0.5\height}{\includegraphics[width=0.30\columnwidth]{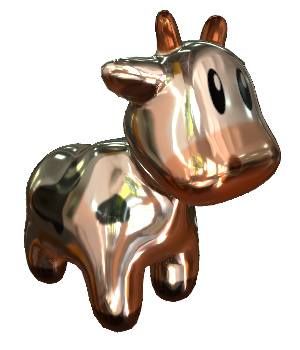}} \\
		& PSNR $|$ SSIM & 26.31 $|$ 0.936 & 30.08 $|$ 0.982 \\
		&
		\includegraphics[width=0.30\columnwidth]{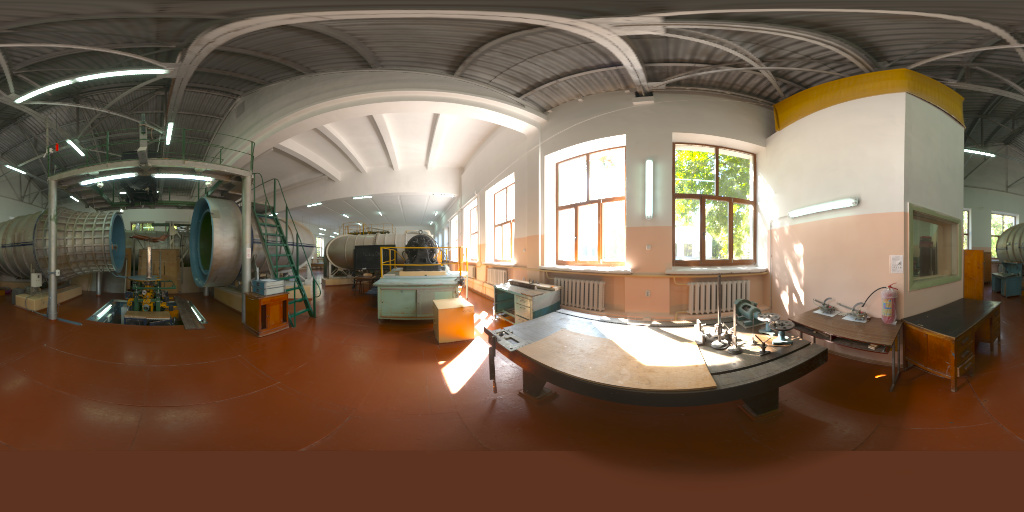} &
		\includegraphics[width=0.30\columnwidth]{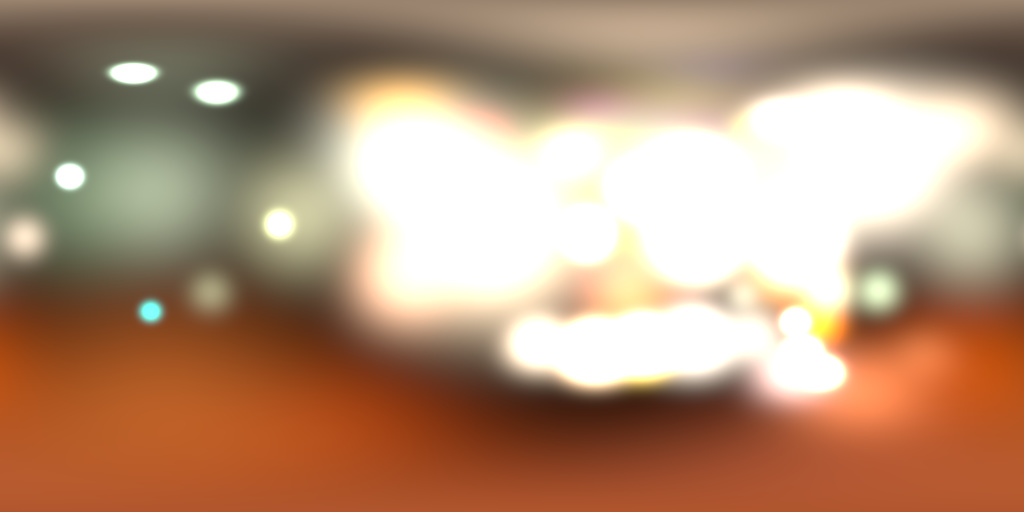} &
		\includegraphics[width=0.30\columnwidth]{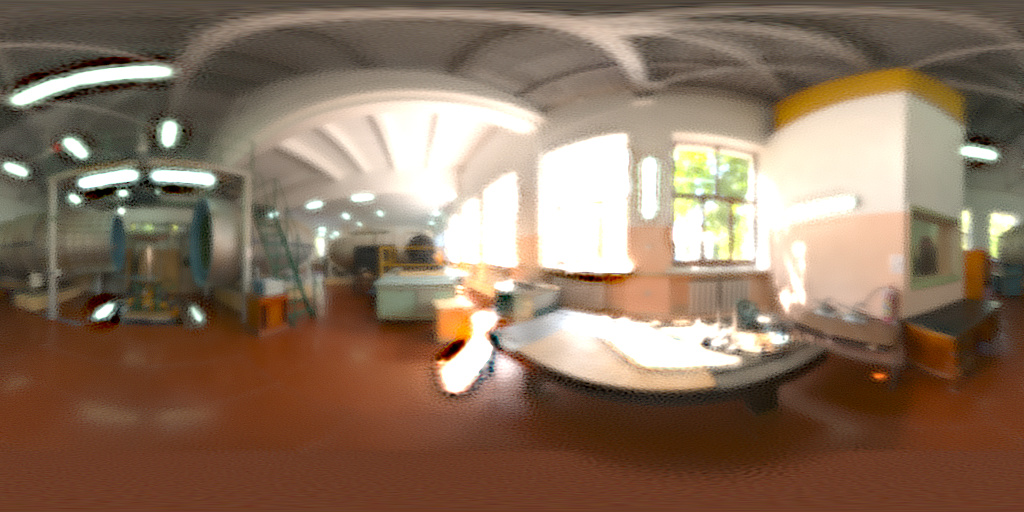} 
	\end{tabular}
	}
	\vspace{-2mm}
	\caption{
		Environment lighting approximated with Spherical Gaussians using 128 lobes 
		vs. Split Sum. The training set consists of 256 path traced images with Monte Carlo sampled 
		environment lighting using a high resolution HDR probe. We assume known	geometry and 
		optimize materials and lighting using identical settings for both methods.
		Reported image metrics are the arithmetic mean of the 16 (novel) views in the test set. 
		Note that the split sum approximation is able to capture high frequency lighting.
		Probe from Poly Haven~\cite{polyhaven}.
	}
	\vspace{-2mm}
	\label{fig:sg}
\end{figure}
}


\newcommand{\figSaxophone}{
\begin{figure}[t]
	\centering
        \setlength{\tabcolsep}{1pt}
        \begin{tabular}{cc}
            \includegraphics[width=0.33\columnwidth]{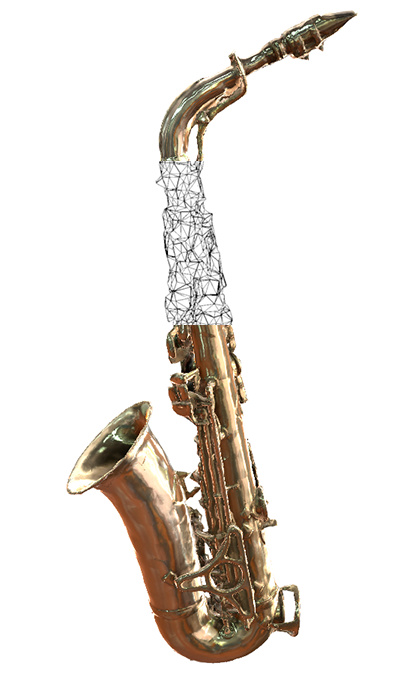} & 
            \includegraphics[width=0.33\columnwidth]{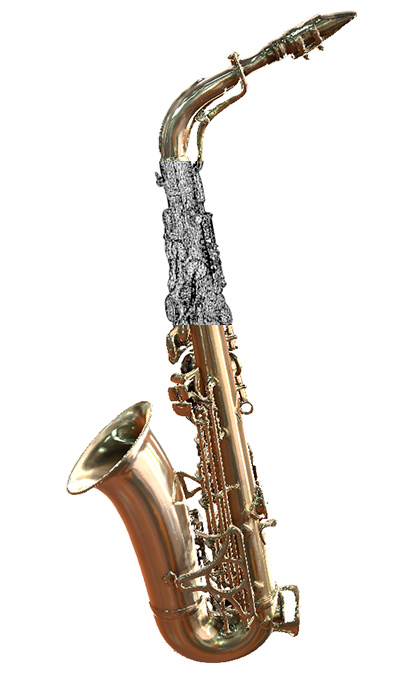} \\
            \includegraphics[width=0.49\columnwidth]{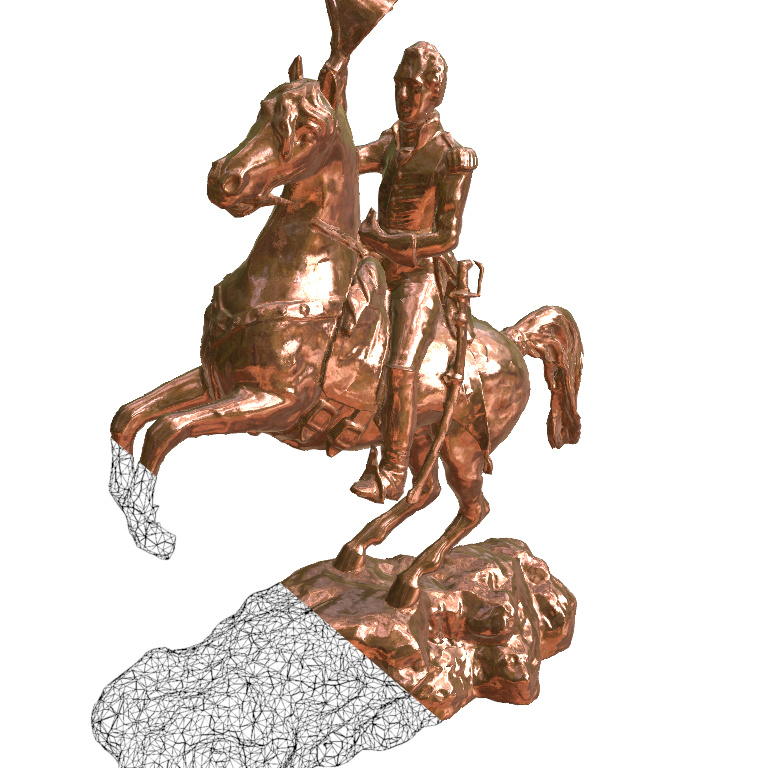} & 
            \includegraphics[width=0.49\columnwidth]{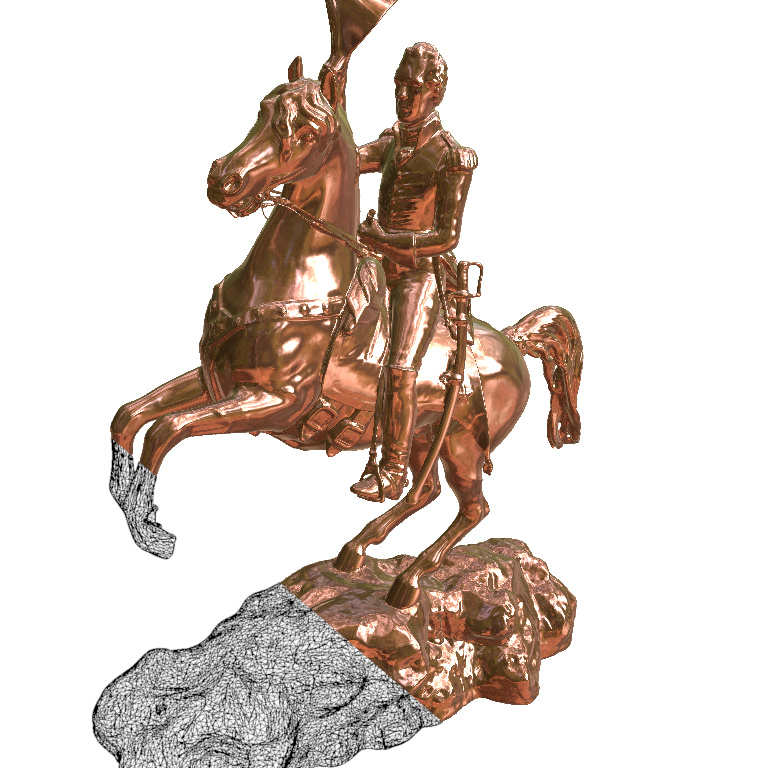} \\
			Our & Reference
        \end{tabular}
	\caption{
		DMTet can accurately capture topology, even in challenging scenarios.
        To illustrate this, we show two examples from the Smithsonian 3D repository~\protect\cite{Smithsonian2020},
        where we jointly learn topology and materials under known environment lighting.
		The left column shows our approximation extracted from multiple 2D observations (5000 views)
		and the right side a rendering of the reference model. In both examples, we start from a tet grid 
        of resolution $128^3$ and optimize the grid SDF values, vertex offsets and material parameters.
	}
	\label{fig:sax}
\end{figure}
}


\newcommand{\nerdComparison}{
	\begin{figure}[tb]
		\centering
		\setlength{\tabcolsep}{1pt}
		{\small
		\begin{tabularx}{\columnwidth}{YYYY}
			Reference & Our & NeRD & NeRF \\
			\includegraphics[width=0.2\columnwidth]{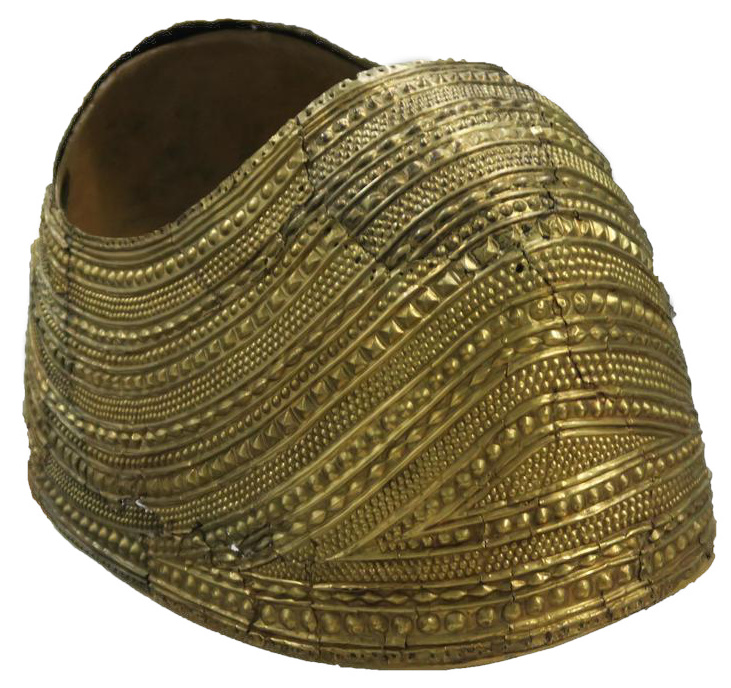} &
			\includegraphics[width=0.2\columnwidth]{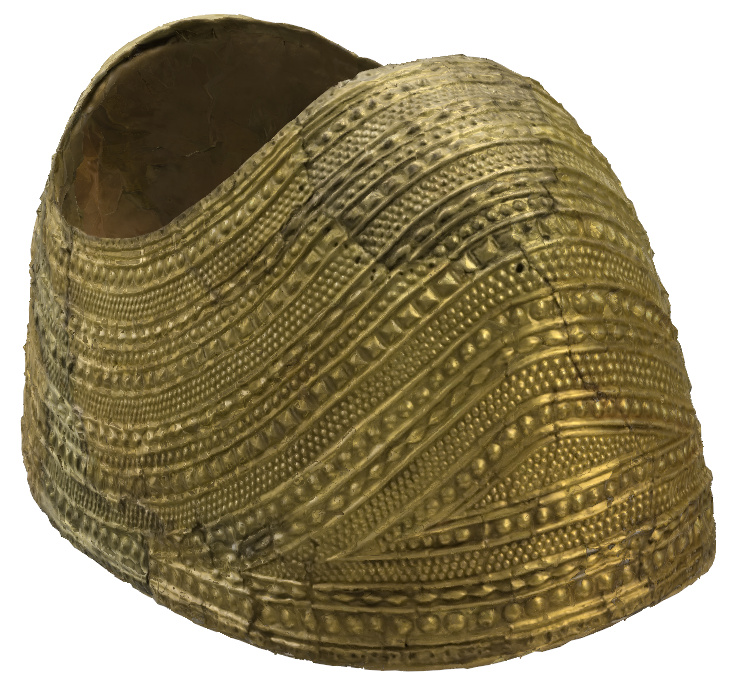} &
			\includegraphics[width=0.2\columnwidth]{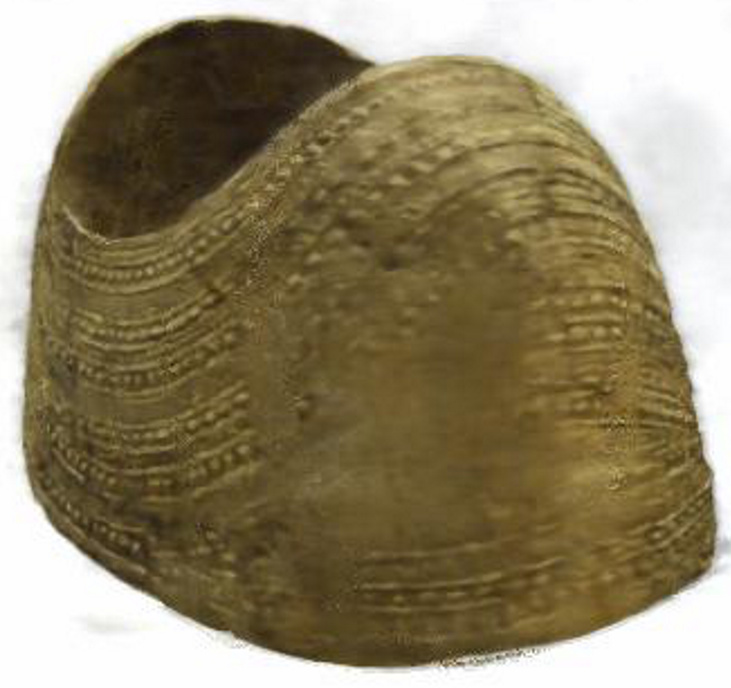} &
			\includegraphics[width=0.2\columnwidth]{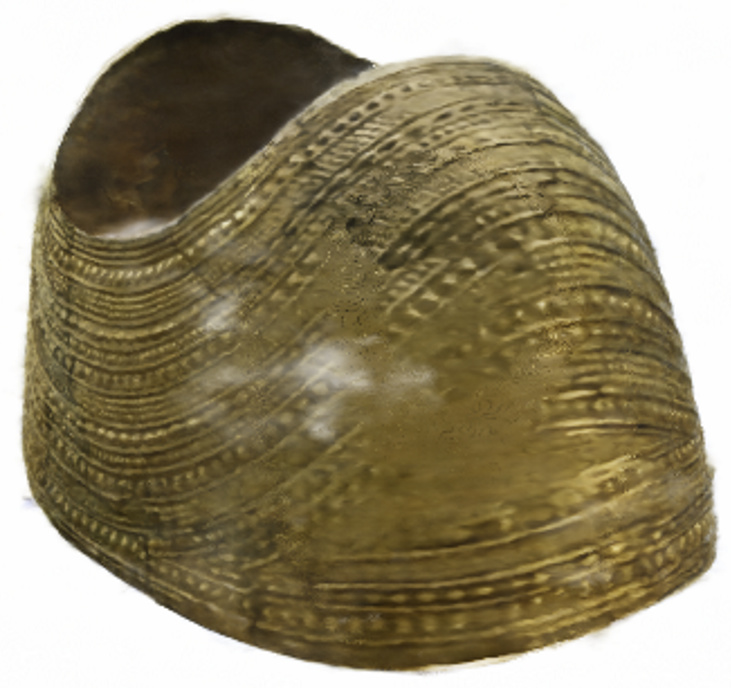} \\
			\small{PSNR} & \small{25.7} & \small{24.3} & \small{24.4} \\
			\includegraphics[width=0.2\columnwidth]{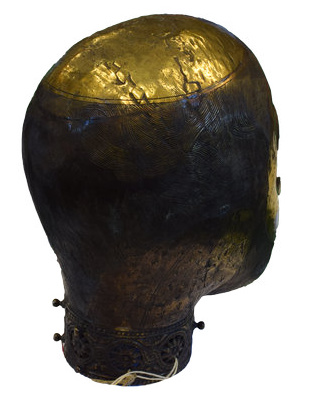} &
			\includegraphics[width=0.2\columnwidth]{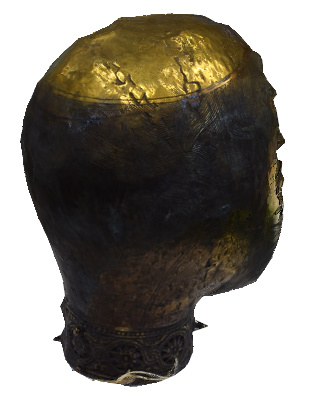} &
			\includegraphics[width=0.2\columnwidth]{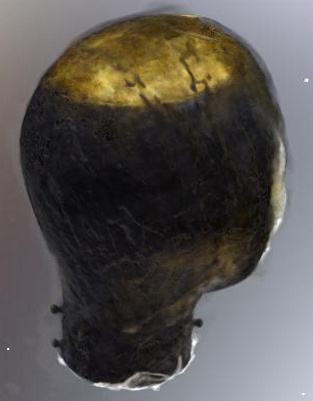} &
			\includegraphics[width=0.2\columnwidth]{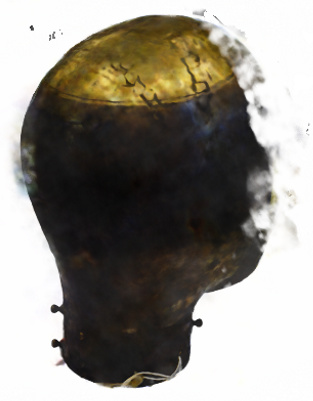} \\
			\small{PSNR} & \small{24.9} & \small{21.8} & \small{19.0} \\
		\end{tabularx}
		}
		\vspace{-2mm}
		\caption{Reconstruction from photographs (datasets from NeRD), comparing our results with NeRD 
			and NeRF. Images in the two rightmost colums were provided by the NeRD authors. 
			We score higher in terms of image metrics, most likely due to our mesh representation enforcing opaque 
			geometry, where competing algorithms rely on volumetric opacity. Despite inconsistencies in camera poses 
			and masks, our results remain sharp while NeRF and NeRD suffer from floating or missing geometry.}
		\vspace{-2mm}
		\label{fig:nerd_cmp}
	\end{figure}
}

\newcommand{\nerdMesh}{
	\begin{figure}[tb]
		\centering
		{
		\begin{tabular}{cc}
			\includegraphics[width=0.42\columnwidth]{figures/photo/nerd_cmp_gold/crop/masked_ref.jpg} &
			\includegraphics[width=0.42\columnwidth]{figures/photo/nerd_cmp_gold/crop/our.jpg} \\
			Reference & Our \\
			\includegraphics[width=0.42\columnwidth]{figures/photo/nerd_cmp_gold/crop/nerd.jpg} &
			\includegraphics[width=0.42\columnwidth]{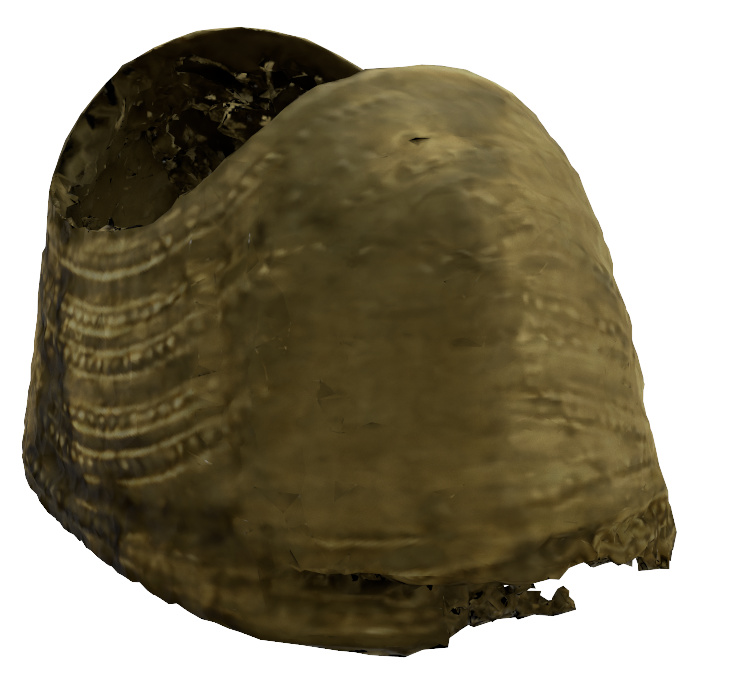} \\
			NeRD (neural)& NeRD (mesh) \\
		\end{tabular}
		}
		\caption{Example of the quality of the neural NeRD representation and their final generated mesh. Note the quality loss in both geometry and appearance (textures).}
		\label{fig:nerd_mesh}
	\end{figure}
}


\newcommand{\figSceneEdit}{
	\begin{figure}[t]
		\setlength{\tabcolsep}{1pt}
		{\small
		\begin{tabular}{ccc}
			\includegraphics[width=0.32\columnwidth]{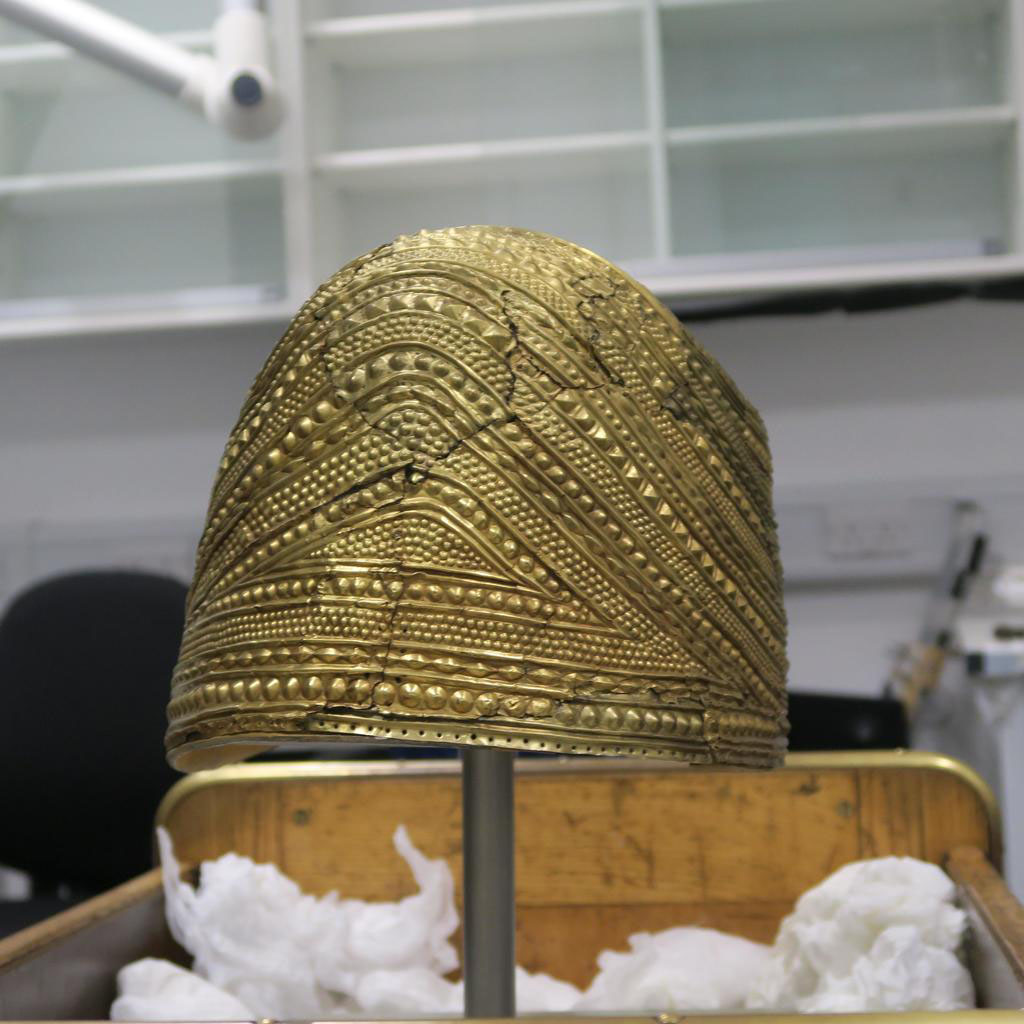} &
			\includegraphics[width=0.32\columnwidth]{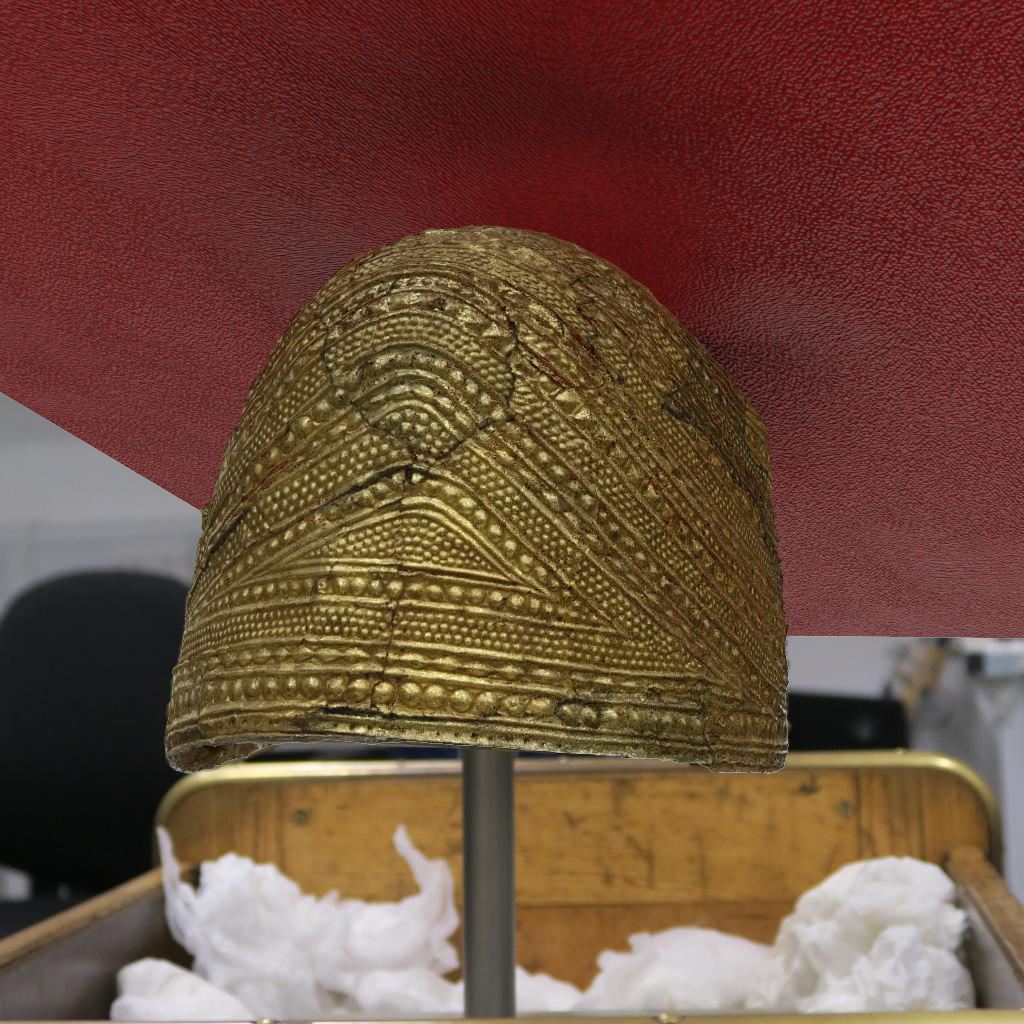} &
			\includegraphics[width=0.32\columnwidth]{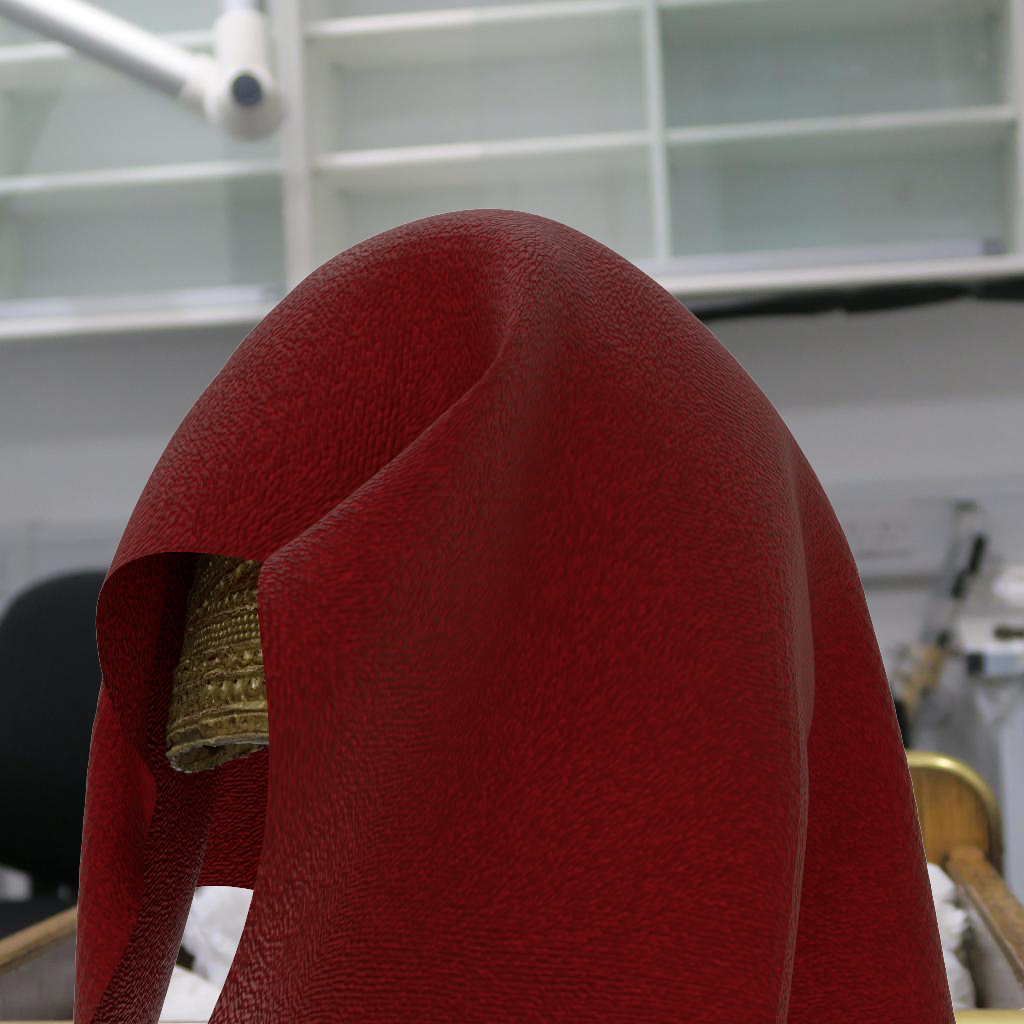} \\
			Reference photo & Scene editing & Cloth simulation \\
			\includegraphics[width=0.32\columnwidth]{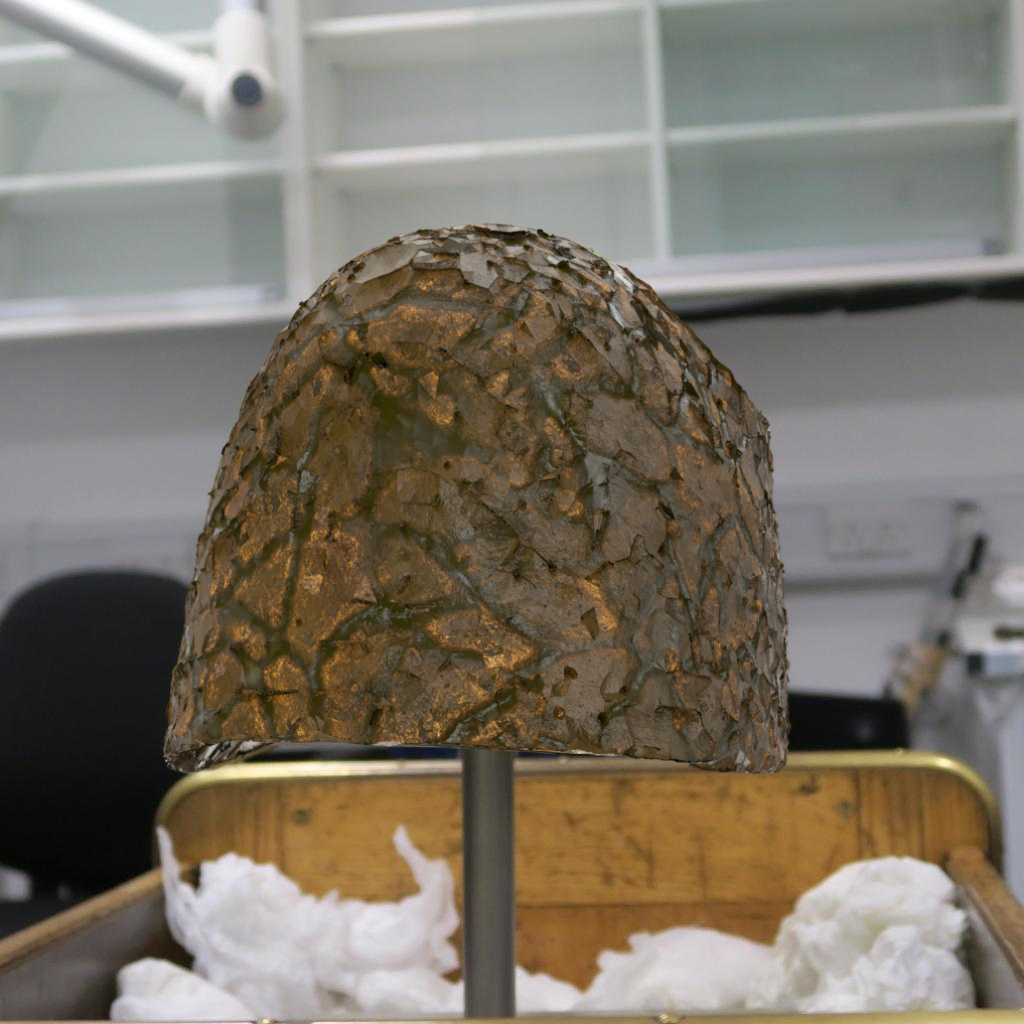} &
			\includegraphics[width=0.32\columnwidth]{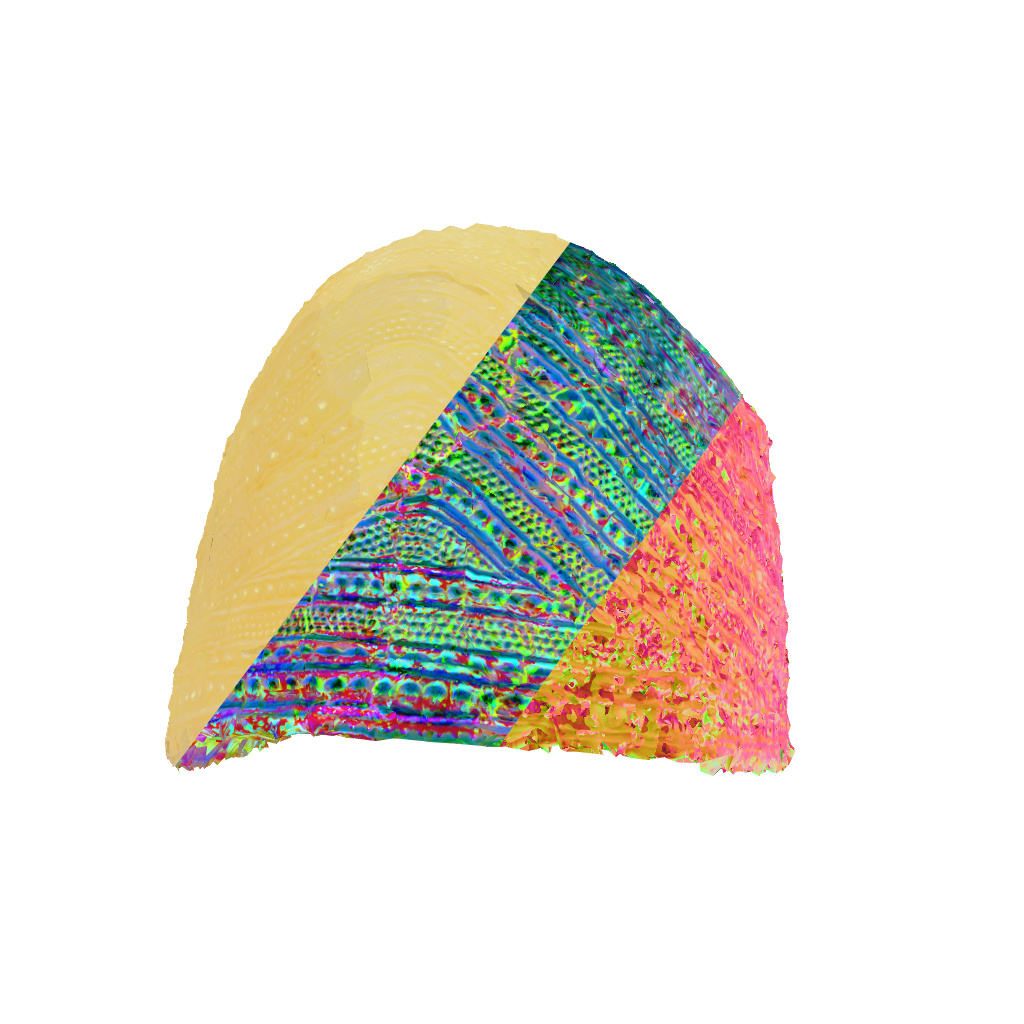} &
			\includegraphics[width=0.32\columnwidth]{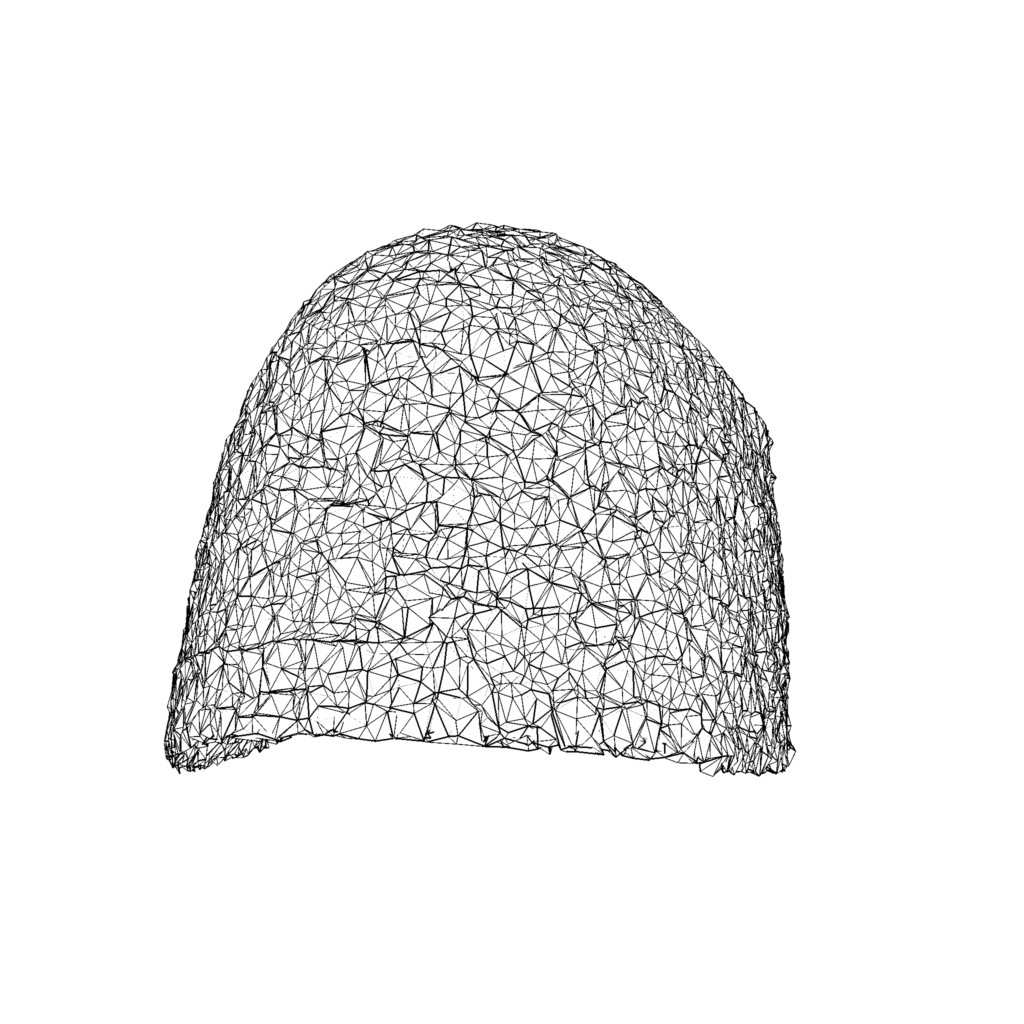} \\
			Material editing & $\kd$/$\korm$/$\mathbf{n}$ & Mesh \\
		\end{tabular}
		}
		\vspace{-3mm}
		\caption{We reconstruct a triangular mesh with unknown topology, 
		spatially-varying materials, and lighting from a set of multi-view images. 
		We show examples of scene manipulation using off-the-shelf modeling tools,  
		enabled by our reconstructed 3D model.}
		\label{fig:scene_editB}
		\vspace{-4mm}
	\end{figure}
}


\newcommand{\figLOD}{
\begin{figure}[t]
    \centering
    \setlength{\tabcolsep}{1pt}
    \begin{tabular}{lccc}
		\rotatebox[origin=l]{90}{\small{LOD 3}} &
        \includegraphics[width=0.32\columnwidth]{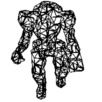} &
        \includegraphics[width=0.32\columnwidth]{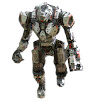} &
        \includegraphics[width=0.32\columnwidth]{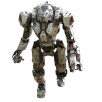} \\
		\rotatebox[origin=l]{90}{\small{LOD 0}} &
		\includegraphics[width=0.32\columnwidth]{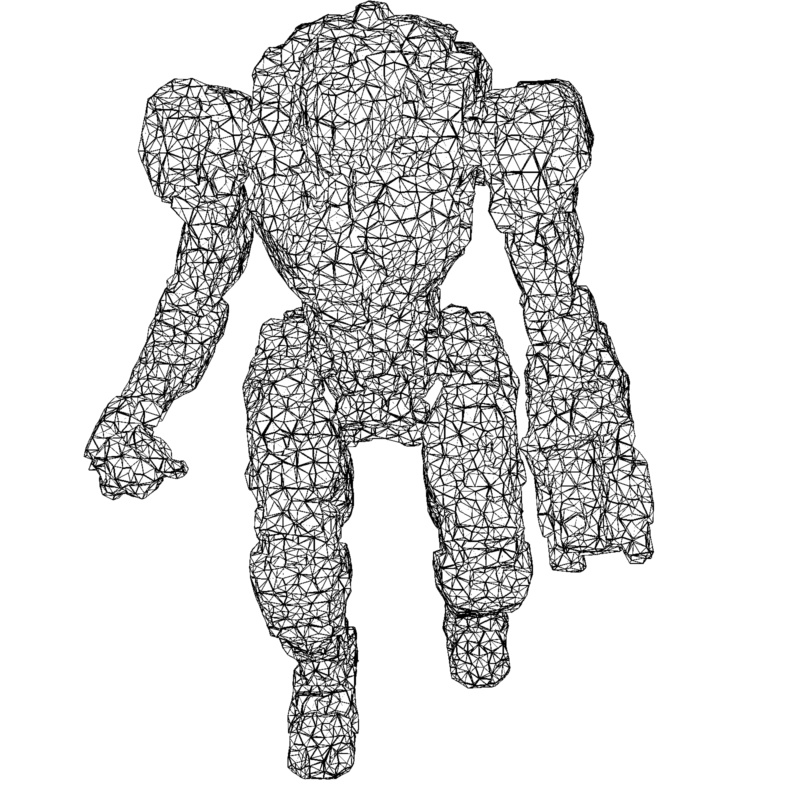} &
        \includegraphics[width=0.32\columnwidth]{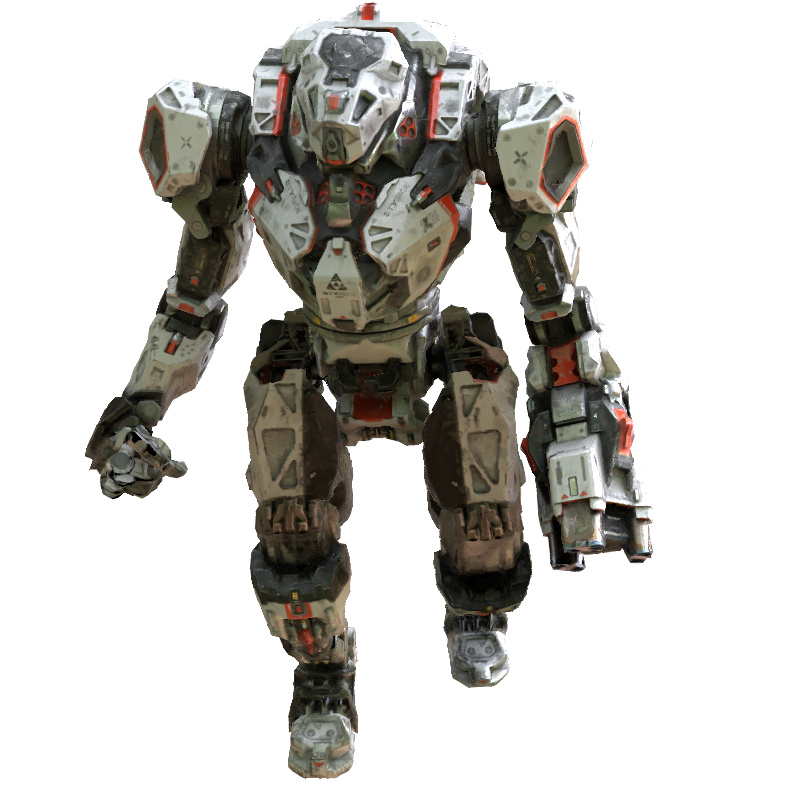} &
        \includegraphics[width=0.32\columnwidth]{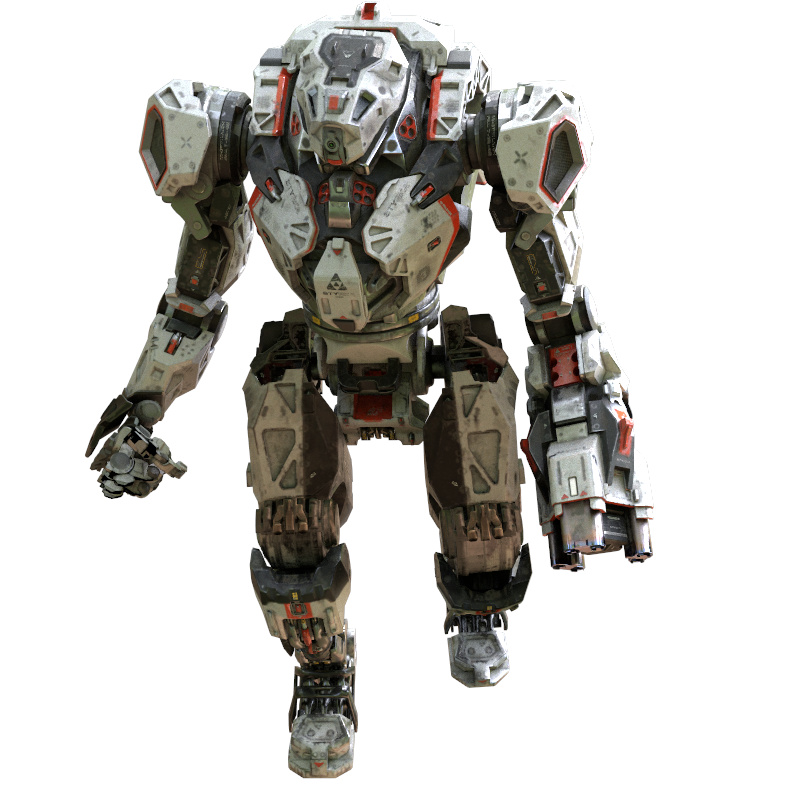} \\
		& Wireframe & Our & Reference \\
     \end{tabular}
    \caption{
        Automatic LOD example: We generated 256 views of an object (with masks \& poses) from a path tracer in two
		resolutions: $1024\times 1024$ and  
		$128\times 128$ pixels, then reconstructed the mesh, materials and lighting to approximate LOD creation.
		\textbf{Top}: LOD level optimized to look good at a resolution of 128$\times$128 pixels with 3k triangles. 
	 	\textbf{Bottom}: LOD level optimized to look good at a resolution of $1024\times 1024$ pixels with 63k triangles.
    }
    \label{fig:lod}
\end{figure}
}


\newcommand{\figNEUS}{
\begin{figure*}
	\centering
    \setlength{\tabcolsep}{1pt}
    {\small
    \begin{tabular}{cccccc}
		\includegraphics[width=0.165\textwidth]{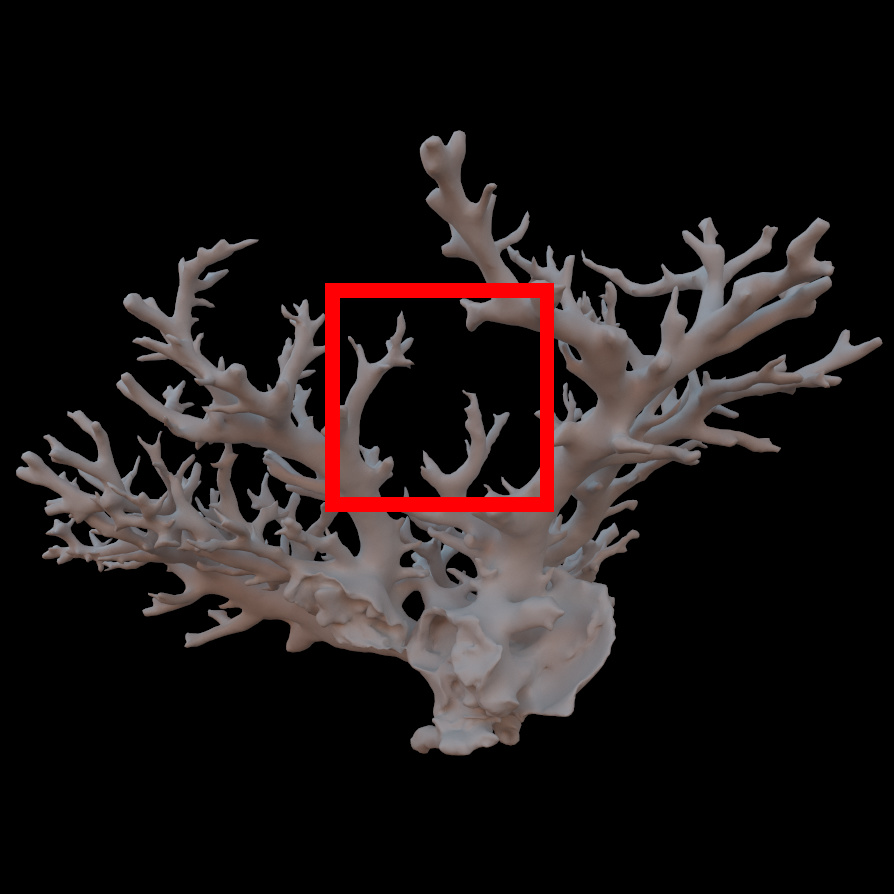} &
        \includegraphics[width=0.165\textwidth]{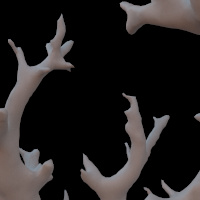} &
        \includegraphics[width=0.165\textwidth]{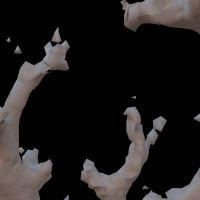} &
        \includegraphics[width=0.165\textwidth]{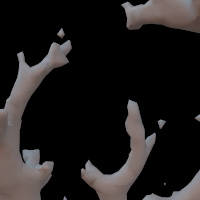} &
        \includegraphics[width=0.165\textwidth]{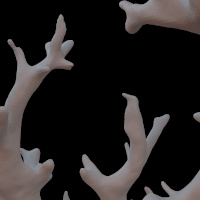} &
        \includegraphics[width=0.165\textwidth]{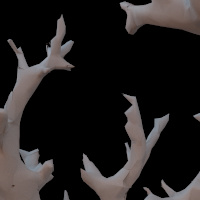} \\
        & Reference & NeRF (57k tris) & NeuS (53k tris) & NeuS (900k tris) & Our (53k tris) \\ 
		& Chamfer~$L_1$ $\times 10^{-4}$ & 33.4 & 9.19 & 5.84 & 4.65
    \end{tabular}
	}
	\vspace{-3mm}
	\caption{
        Triangle mesh extraction from a set of 256 rendered images w/ masks. 
		Damicornis model from the Smithsonian 3D repository~\cite{Smithsonian2020},
        We extracted meshes from NeRF and NeuS using Marching Cubes for a target triangle count of 
        roughly 50k triangles and optimized the example in our pipeline for a similar count. We show 
        renderings of the extracted meshes in a path tracer and report the Chamfer loss. 
        We note that NeuS, which optimizes a surface representation, significantly improves on the 
        volumetric representation used by NeRF for this example. Furthermore, our end-to-end optimization of a 
        triangle mesh improves both the visual quality and the Chamfer loss at a fixed triangle count. 
        When drastically increasing the triangle count in the NeuS mesh extraction (from 53k to 900k 
        triangles), the quality improves significantly, indicating that NeuS has a high quality
        internal surface representation. Still, our mesh with 53k triangles is on par with the high resolution NeuS 
        output, indicating the benefit of directly optimizing the mesh representation. 
	}
	\vspace{-3.5mm}
	\label{fig:neus}
\end{figure*}
}


\newcommand{\figMeshExtract}{
\begin{figure}
	\centering
	\setlength{\tabcolsep}{1pt}
	\begin{tabular}{cc}
		\includegraphics[width=0.49\columnwidth]{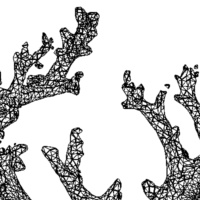} &
		\includegraphics[width=0.49\columnwidth]{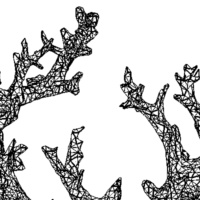} \\
		\small{Marching Cubes} & \small{Our, finetuned} \\
		\includegraphics[width=0.49\columnwidth]{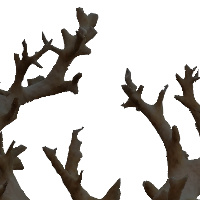} &
		\includegraphics[width=0.49\columnwidth]{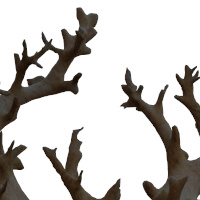} \\
		\small{Reference} & \small{Our, finetuned} 
	\end{tabular}
	\caption{
		Appearance-aware NeRF 3D model extraction. We show insets of the silhouette quality before and after
		our optimization pass, alongside insets of the reference and our rendered result.
	}
	\label{fig:mesh_extract}
\end{figure}
}


\newcommand{\relightbox}[2]{\rotatebox{90}{\makebox[#1\columwidth]{\centering\small #2}}}

\newcommand{\figRelightNerFactorB}{
	\begin{figure}
		\centering
		\setlength{\tabcolsep}{1pt}
		\begin{tabular}{lccccc}
		
			\rotatebox[origin=l]{90}{\tiny{NeRFactor}} &
			\includegraphics[width=0.19\columnwidth]{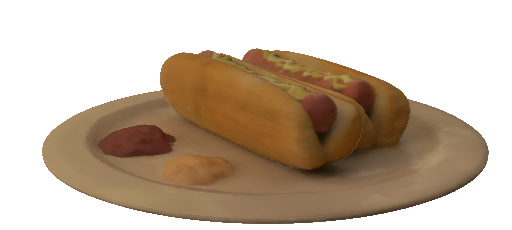} &
			\includegraphics[width=0.19\columnwidth]{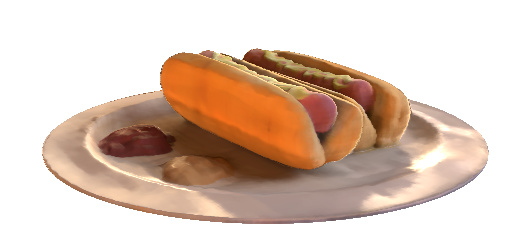} &
			\includegraphics[width=0.19\columnwidth]{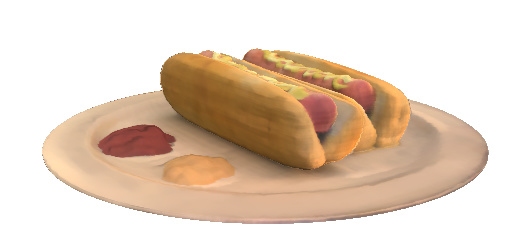} &
			\includegraphics[width=0.19\columnwidth]{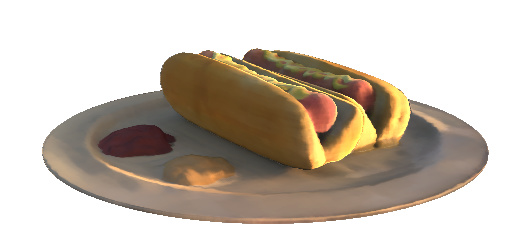} &
			\includegraphics[width=0.19\columnwidth]{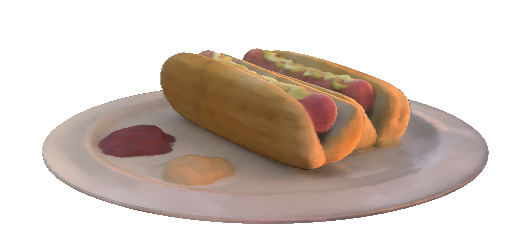} \\
			
			\rotatebox[origin=l]{90}{\tiny{Our}} &
			\includegraphics[width=0.19\columnwidth]{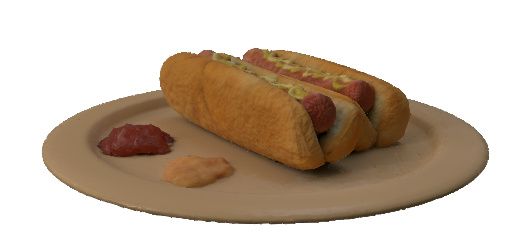} &
			\includegraphics[width=0.19\columnwidth]{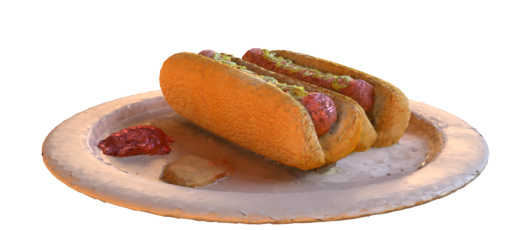} &
			\includegraphics[width=0.19\columnwidth]{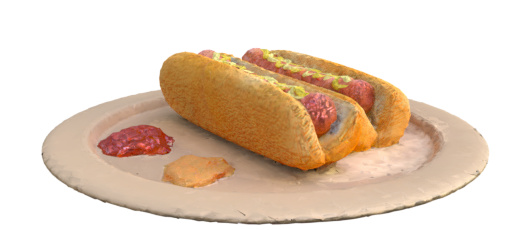} &
			\includegraphics[width=0.19\columnwidth]{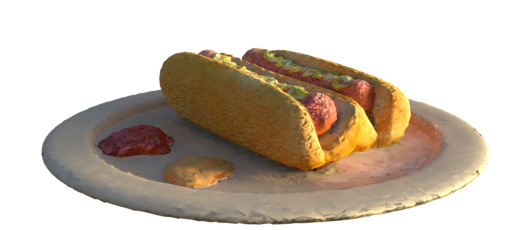} &
			\includegraphics[width=0.19\columnwidth]{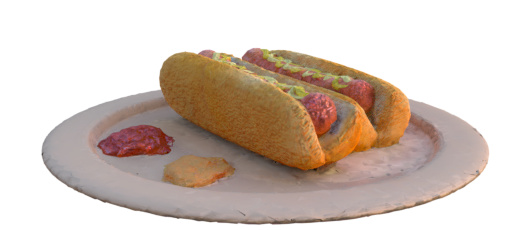} \\
			
			\rotatebox[origin=l]{90}{\tiny{Ref.}} &
			\includegraphics[width=0.19\columnwidth]{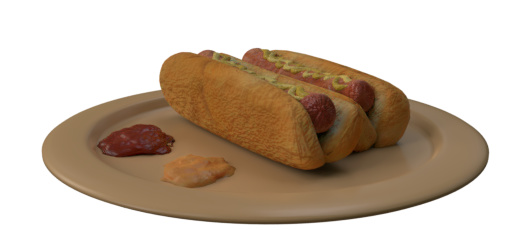} &
			\includegraphics[width=0.19\columnwidth]{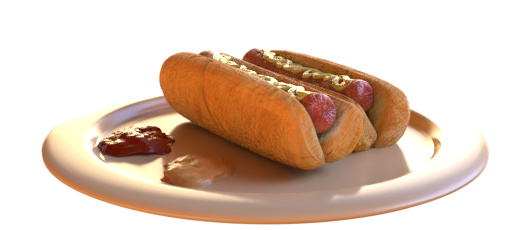} &
			\includegraphics[width=0.19\columnwidth]{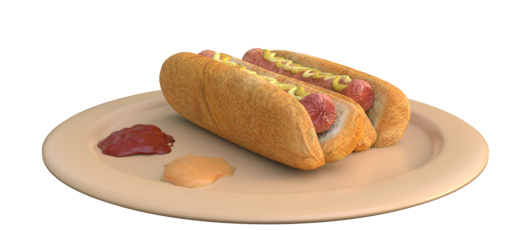} &
			\includegraphics[width=0.19\columnwidth]{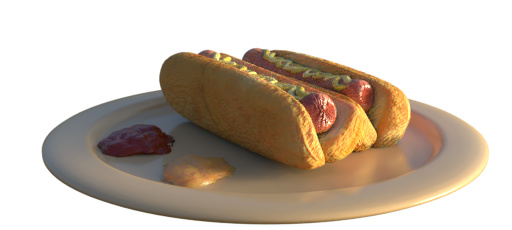} &
			\includegraphics[width=0.19\columnwidth]{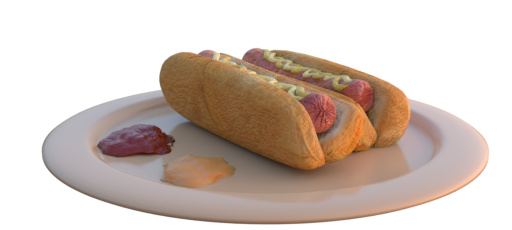} \\

			& \small{Original} & \small{Courtyard} & \small{Forest} & \small{Sunrise} & \small{Sunset} \\
		\end{tabular}
		\vspace{-2mm}
		\caption{
			Relighting quality for a scene from the NeRFactor dataset, with our examples relit using 
			Blender, and NeRFactor results generated using the public code. 
		}
		\vspace{-2mm}
		\label{fig:relight_nerfactor}
	\end{figure}
}

\newcommand{\figRelightNerFactorC}{
	\begin{figure}
		\centering
		\setlength{\tabcolsep}{1pt}
		\begin{tabular}{lccccc}
			\rotatebox[origin=l]{90}{\small{NeRFactor}} &
			\includegraphics[width=0.19\columnwidth]{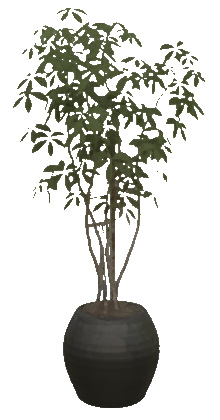} &
			\includegraphics[width=0.19\columnwidth]{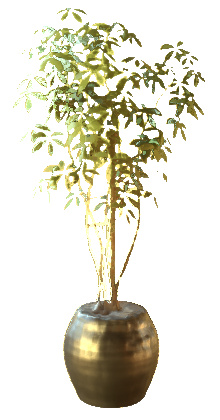} &
			\includegraphics[width=0.19\columnwidth]{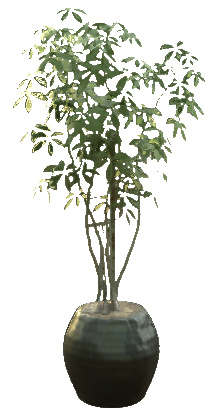} &
			\includegraphics[width=0.19\columnwidth]{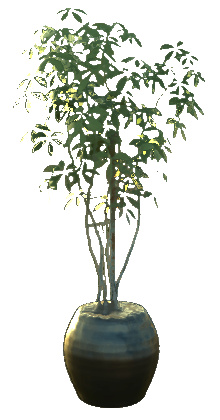} &
			\includegraphics[width=0.19\columnwidth]{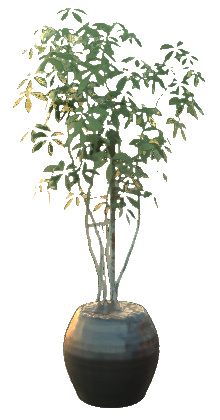} \\
			
			\rotatebox[origin=l]{90}{\small{Our}} &
			\includegraphics[width=0.19\columnwidth]{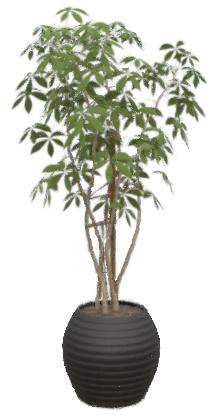} &
			\includegraphics[width=0.19\columnwidth]{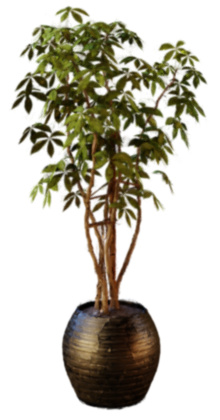} &
			\includegraphics[width=0.19\columnwidth]{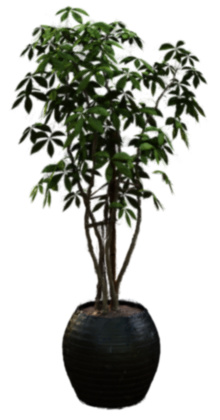} &
			\includegraphics[width=0.19\columnwidth]{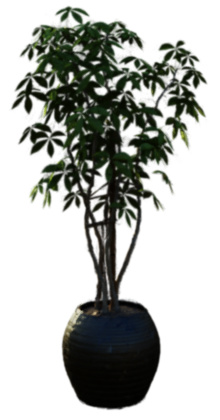} &
			\includegraphics[width=0.19\columnwidth]{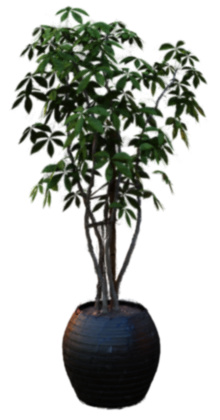} \\
			
			\rotatebox[origin=l]{90}{\small{Reference}} &
			\includegraphics[width=0.19\columnwidth]{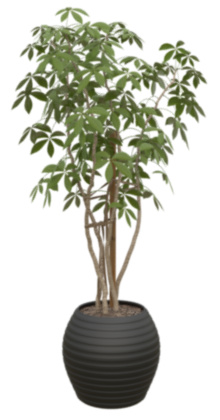} &
			\includegraphics[width=0.19\columnwidth]{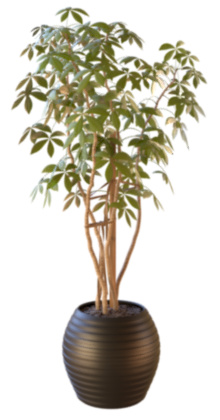} &
			\includegraphics[width=0.19\columnwidth]{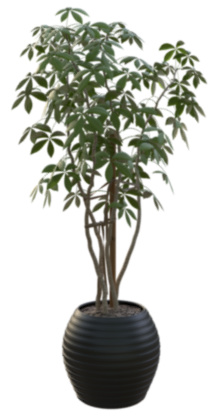} &
			\includegraphics[width=0.19\columnwidth]{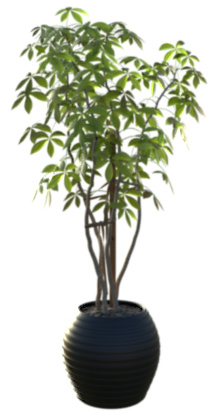} &
			\includegraphics[width=0.19\columnwidth]{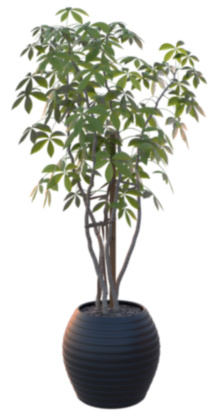} \\
			
			& \small{Original} & \small{Courtyard} & \small{Forest} & \small{Sunrise} & \small{Sunset}
		\end{tabular}
		\caption{
			Relighting quality for a scene from the NeRFactor dataset, with our examples relit using 
			Blender, and NeRFactor results generated using the public code. 
		}
		\label{fig:relight_nerfactor_ficus}
	\end{figure}
}


\newcommand{\figEditNerFactor}{
	\begin{figure}
		\centering
		\setlength{\tabcolsep}{1pt}
		\begin{tabular}{lcc}
			\rotatebox[origin=c]{90}{\small{Scene editing}} &
			\raisebox{-0.5\height}{\includegraphics[width=0.45\columnwidth]{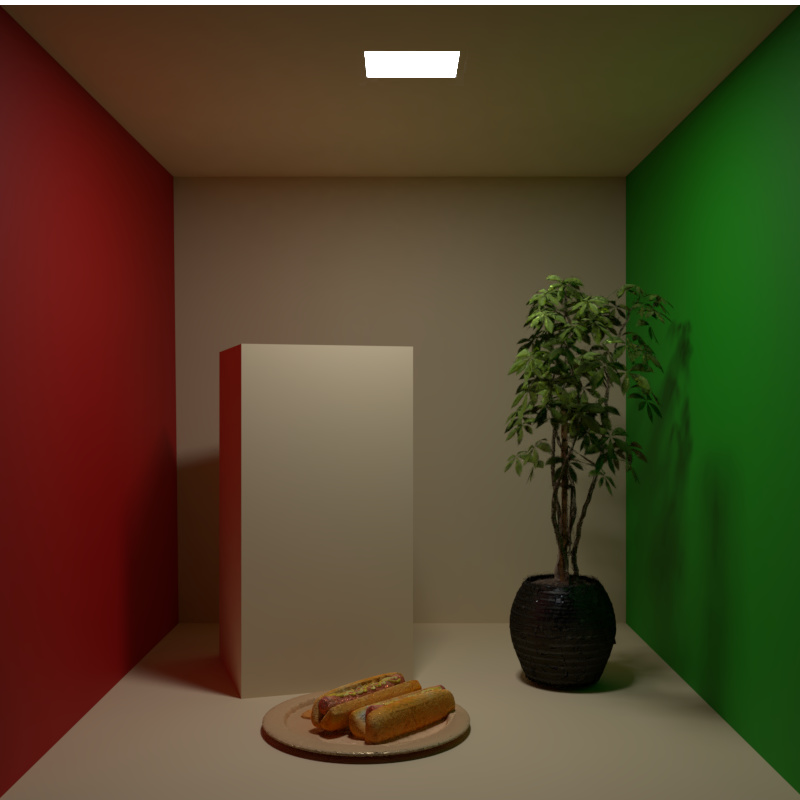}} &
			\raisebox{-0.5\height}{\includegraphics[width=0.45\columnwidth]{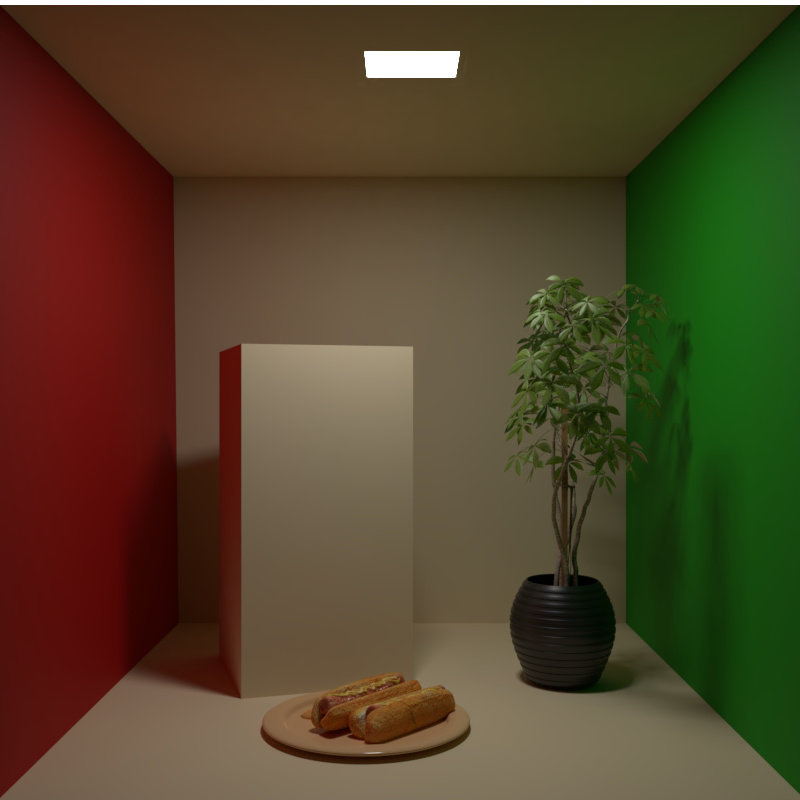}} \\

			\rotatebox[origin=c]{90}{\small{Soft-body sim.}} &
			\raisebox{-0.5\height}{\includegraphics[width=0.45\columnwidth]{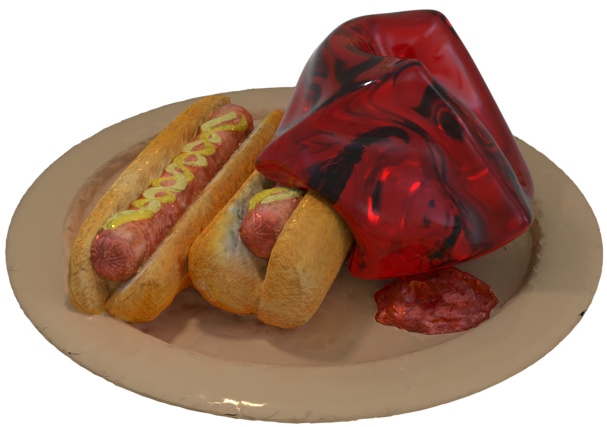}} &
			\raisebox{-0.5\height}{\includegraphics[width=0.45\columnwidth]{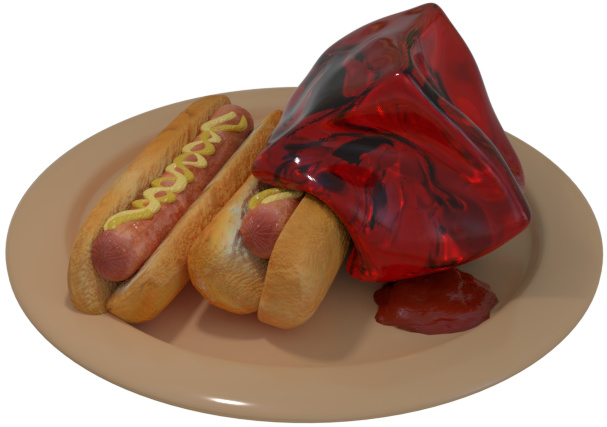}} \\	
			& \small Our result & \small Reference rendering \\
		\end{tabular}
		\vspace{-2mm}
		\caption{To highlight the benefits of our explicit representation, 
			we insert two reconstructed models into the Cornell box. Note that the 
			objects accurately interact with the scene lighting, and cast shadows (e.g., the green wall). 
			Next, we use our reconstructed hotdog model in a soft-body physics 
			simulation, dropping red jelly on the plate. We run the entire simulation (21 frames) 
			on both the reference 3D mesh and our reconstructed mesh, and display the last frame.
			Note that these applications are not currently feasible for neural volumetric representations.
		}
		\vspace{-3mm}
		\label{fig:edit_nerfactor}
	\end{figure}
}


\newcommand{\figAlbedoNerFactorC}{
	\begin{figure}
		\centering
		\setlength{\tabcolsep}{1pt}
		\begin{tabular}{lccc}
			\rotatebox[origin=l]{90}{\small{PhySG}} &
			\includegraphics[width=0.3\columnwidth]{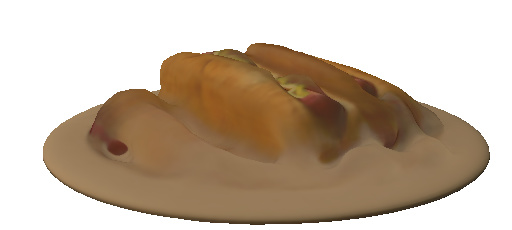} & 
			\includegraphics[width=0.3\columnwidth]{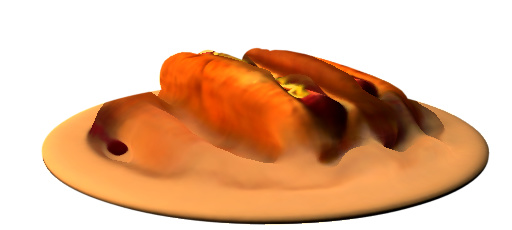} & 
			\includegraphics[width=0.3\columnwidth]{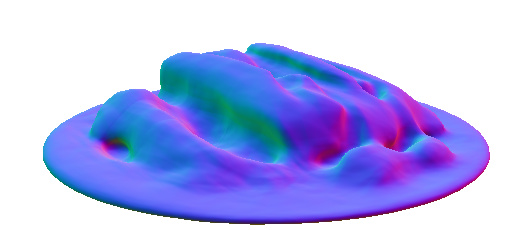} \\

			\rotatebox[origin=l]{90}{\small{NeRFactor}} &
			\includegraphics[width=0.3\columnwidth]{figures/relight2/nerfactor/crop/pred_rgb.jpg} & 
			\includegraphics[width=0.3\columnwidth]{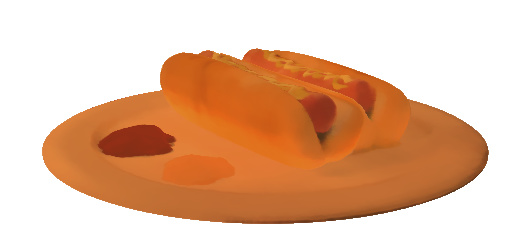} & 
			\includegraphics[width=0.3\columnwidth]{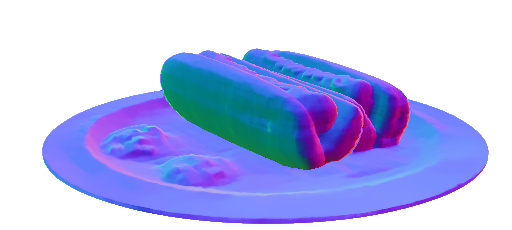} \\

			\rotatebox[origin=l]{90}{\small{Our}} &
			\includegraphics[width=0.3\columnwidth]{figures/relight2/our_neural/crop/val_000003_opt.jpg} &
			\includegraphics[width=0.3\columnwidth]{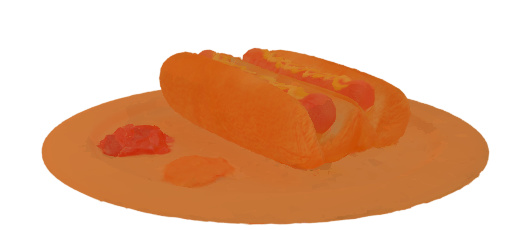} & 
			\includegraphics[width=0.3\columnwidth]{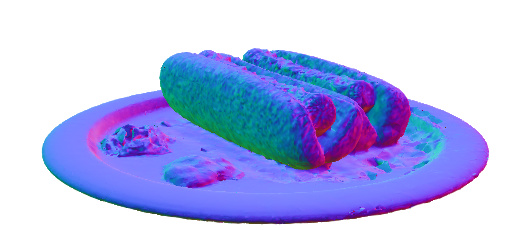} \\
			
			\rotatebox[origin=l]{90}{\small{Reference}} &
			\includegraphics[width=0.3\columnwidth]{figures/relight2/ref/crop/rgba.jpg} &
			\includegraphics[width=0.3\columnwidth]{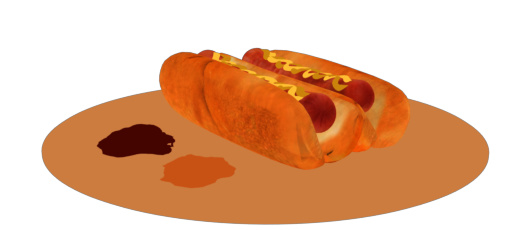} & 
			\includegraphics[width=0.3\columnwidth]{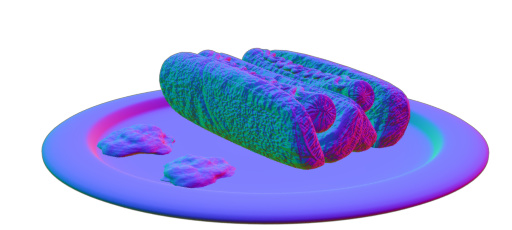} \\
			& \small{Shaded} & \small $\kd$ & \small{normal} \\
		\end{tabular}
		\caption{
			Extracted materials for a scene from the NeRFactor dataset. We directly compare albedo $\kd$ and 
			normals $\mathbf{n}$ to the results of NeRFactor. Specular parameters are omitted as we use  
			different BSDF models. All $\kd$ images have been renormalized using the reference albedo, as
			suggested in NeRFactor.	
		}
		\label{fig:albedo_nerfactor}
	\end{figure}
}

\newcommand{\figChamferSynthetic}{
	\begin{figure}
		\centering
		{\small
		\setlength{\tabcolsep}{1pt}
		\begin{tabular}{ccc}
			\includegraphics[width=0.25\columnwidth]{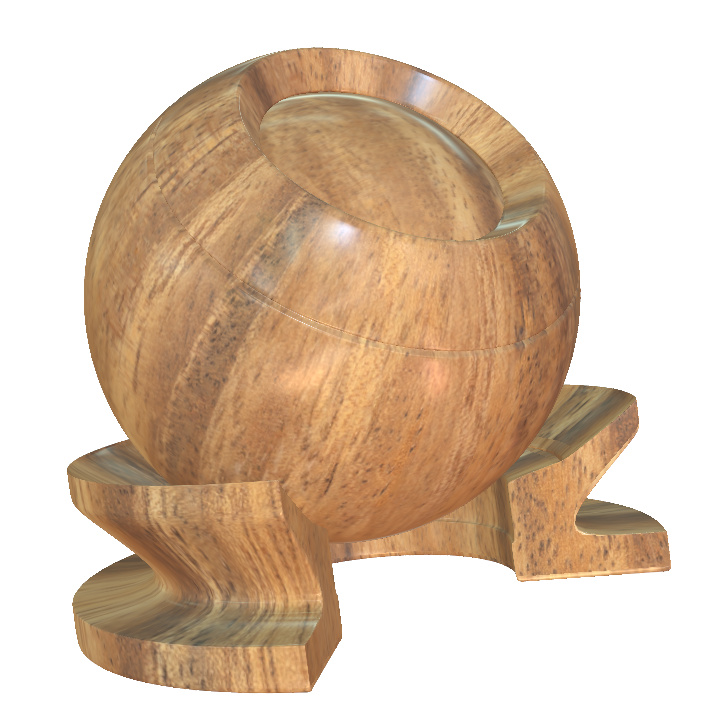} & 
			\includegraphics[width=0.3\columnwidth]{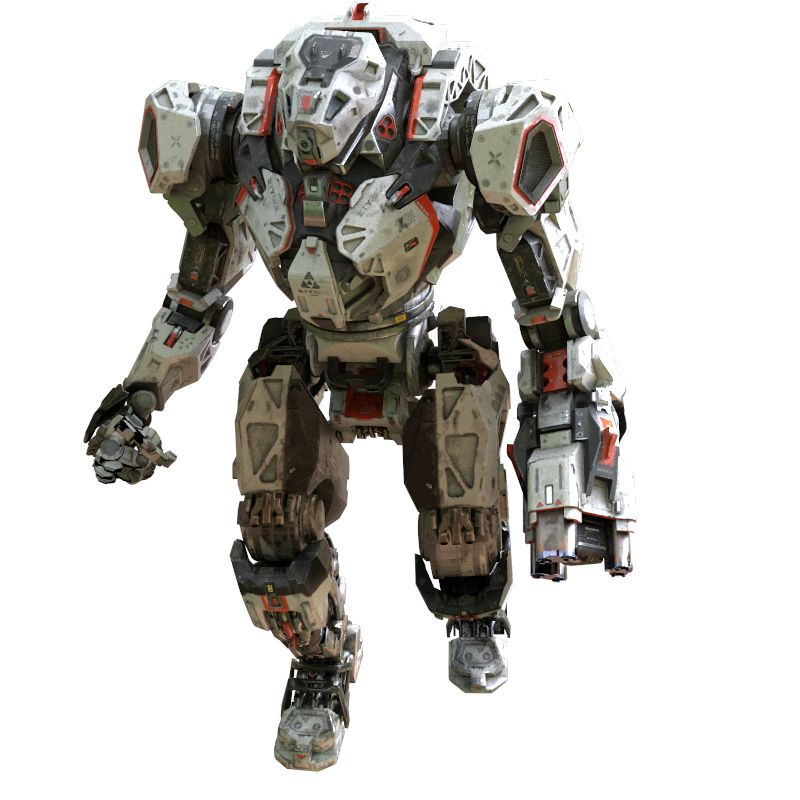} & 
			\includegraphics[width=0.45\columnwidth]{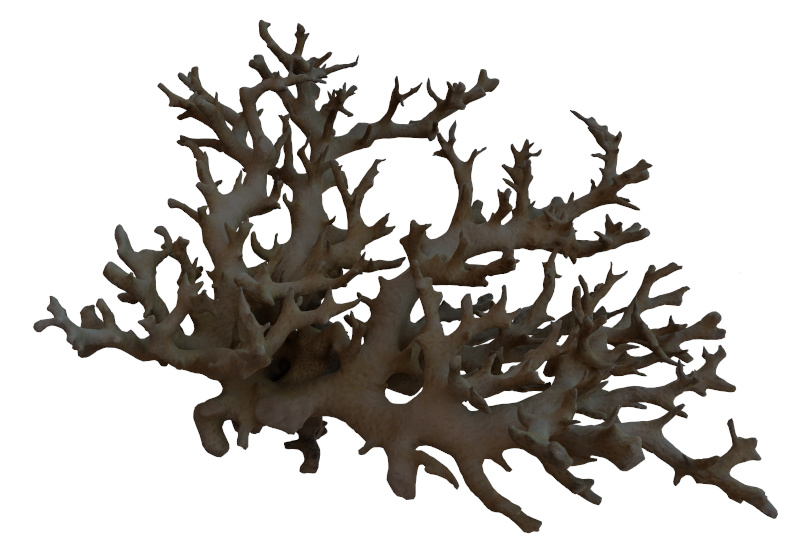} 
		\end{tabular}
		\begin{tabularx}{\columnwidth}{l|YYY}
			& \textsc{Knob} & \textsc{Cerberus} & \textsc{Damicornis} \\
			\hline
			NeRF~\cite{Mildenhall2020} & 2.77e-01 & 9.08e-03 & 3.34e-03 \\
			NeuS~\cite{Wang2021neus} & 2.04e-01 & 2.84e-02 & 5.84e-04 \\
			Our  & 1.87e-01 & 1.03e-02 & 4.66e-04 \\ \hline
		\end{tabularx}
		}
		\caption{
			Synthetic examples with increasing complexity. Each dataset consists of 256 rendered
			images at a resolution of 1024$\times$1024 pixels. 
			We report Chamfer~$L_1$ scores on the extracted meshes for NeRF (neural volume), 
			NeuS (neural implicit), and our explicit approach. Lower score is better.
		}
		\label{fig:chamfer}
	\end{figure}
}

\newcommand{\figKnobComp}{
	\begin{figure}
		\centering
		\setlength{\tabcolsep}{1pt}
		{\small
		\begin{tabularx}{\columnwidth}{cYYYY}
			\multirow{2}{2mm}{\rotatebox[origin=c]{90}{\textsc{Knob}}} &
			\includegraphics[width=0.23\columnwidth]{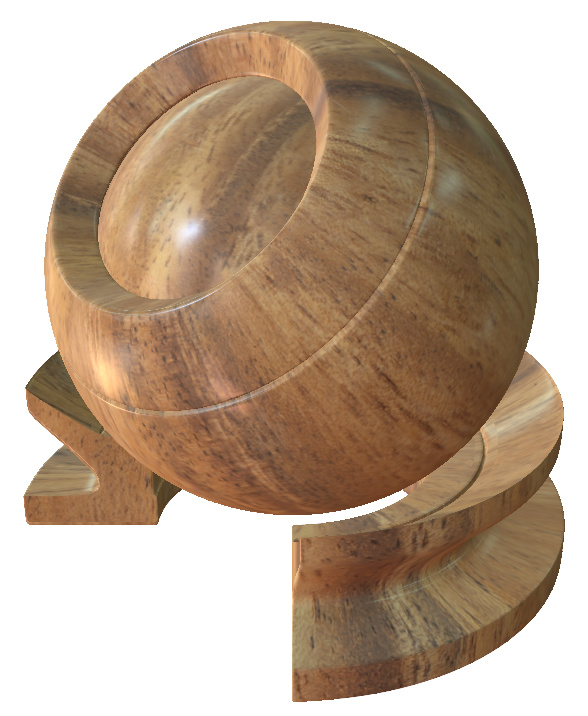} & 
			\includegraphics[width=0.23\columnwidth]{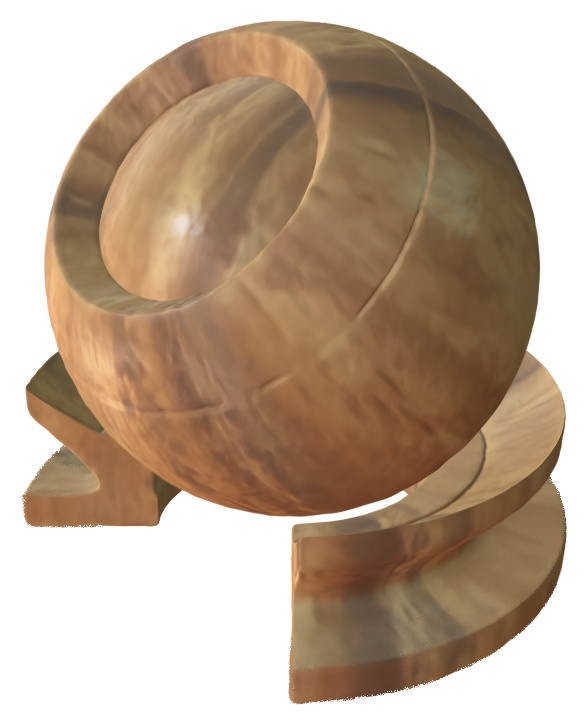} &
			\includegraphics[width=0.23\columnwidth]{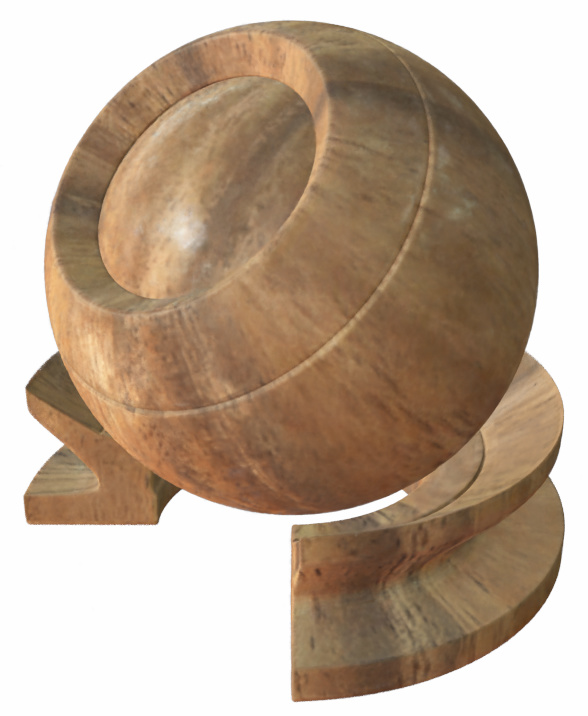} &
			\includegraphics[width=0.23\columnwidth]{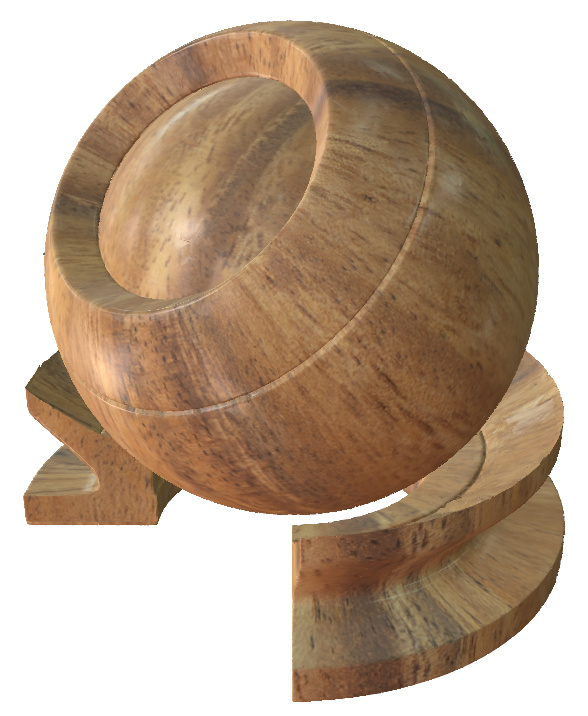} \\
			& \includegraphics[width=0.23\columnwidth]{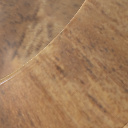} & 
			\includegraphics[width=0.23\columnwidth]{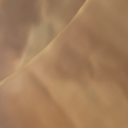} & 
			\includegraphics[width=0.23\columnwidth]{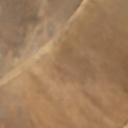} &
			\includegraphics[width=0.23\columnwidth]{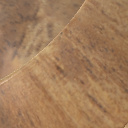} \\
			& PSNR $|$ SSIM & 30.55 $|$ 0.937 & 33.97 $|$ 0.958 & 35.67 $|$ 0.985 \\

			\multirow{2}{2mm}{\rotatebox[origin=c]{90}{\textsc{Cerberus}}} &
			\includegraphics[width=0.23\columnwidth]{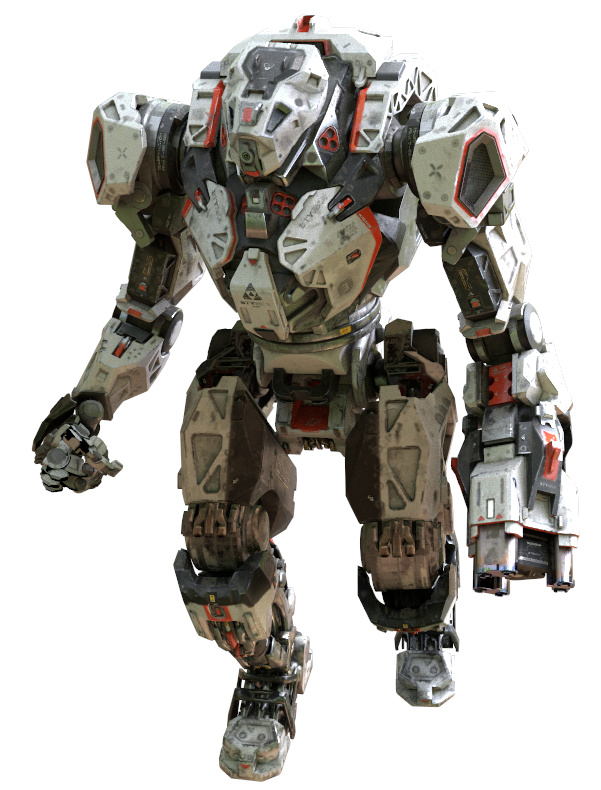} & 
			\includegraphics[width=0.23\columnwidth]{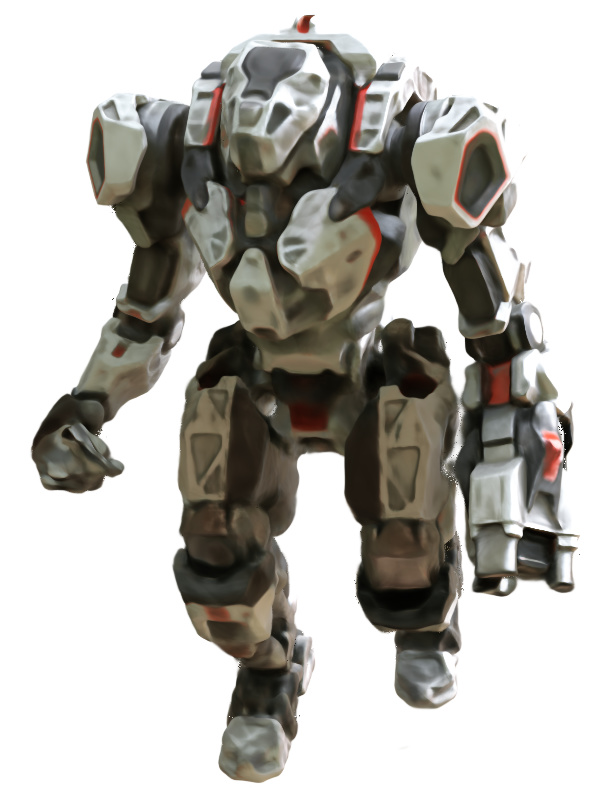} &
			\includegraphics[width=0.23\columnwidth]{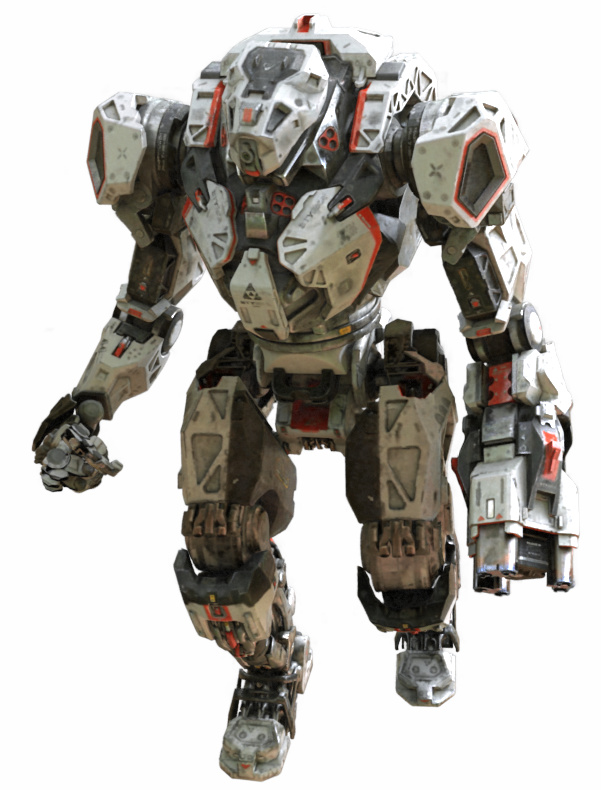} &
			\includegraphics[width=0.23\columnwidth]{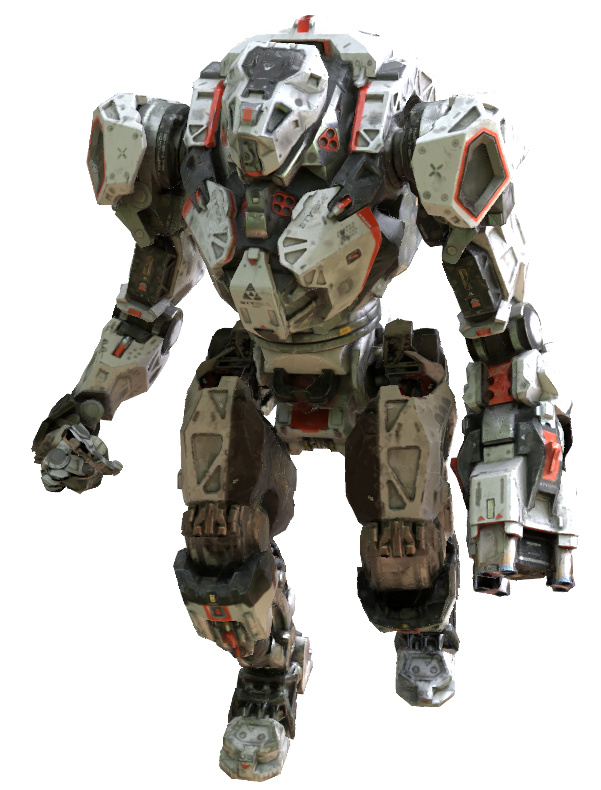} \\
			& \includegraphics[width=0.23\columnwidth]{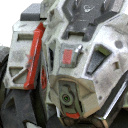} & 
			\includegraphics[width=0.23\columnwidth]{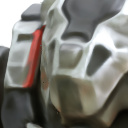} & 
			\includegraphics[width=0.23\columnwidth]{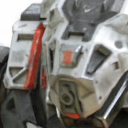} &
			\includegraphics[width=0.23\columnwidth]{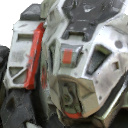} \\
			& PSNR $|$ SSIM & 26.68 $|$ 0.927 & 34.25 $|$ 0.982 & 30.49 $|$ 0.968 \\
			& Reference & NeuS & NeRF & Our \\
		\end{tabularx}
	}
	\vspace*{-3mm}
	\caption{
		Visual quality examples on the synthetic \textsc{Knob} and \textsc{Cerberus} datasets. 
		We observe slightly blurry results from NeuS. 
	}
	\vspace*{-2mm}
	\label{fig:knob_comp}
	\end{figure}
}

\newcommand{\figPhySGComp}{
\begin{figure}
	\centering
	\setlength{\tabcolsep}{1pt}
	{\small
		\begin{tabularx}{\columnwidth}{cYYYY}
			\rotatebox[origin=c]{90}{\textsc{Chair}} &
			\raisebox{-0.5\height}{\includegraphics[width=0.24\columnwidth]{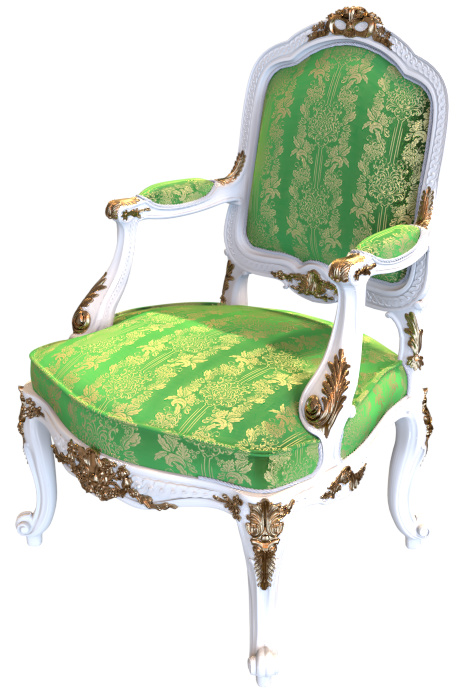}} & 
			\raisebox{-0.5\height}{\includegraphics[width=0.24\columnwidth]{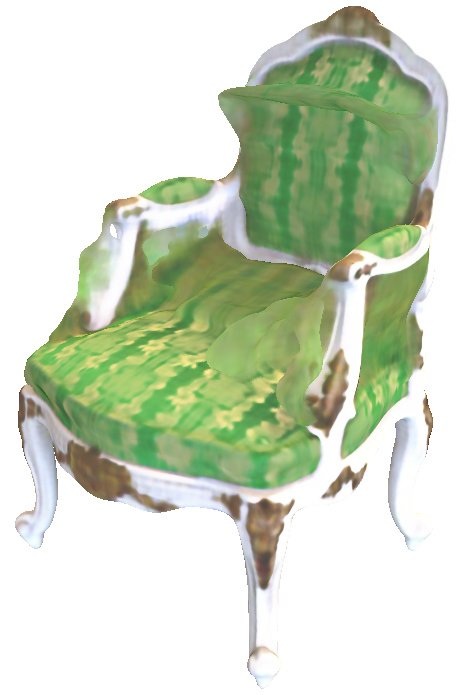}} &
			\raisebox{-0.5\height}{\includegraphics[width=0.24\columnwidth]{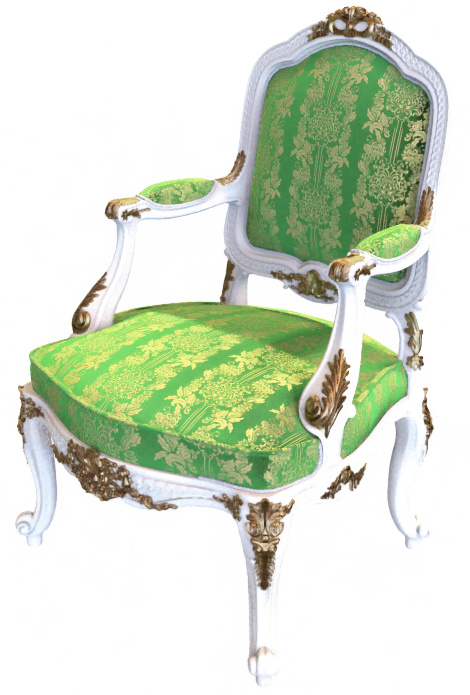}} &
			\raisebox{-0.5\height}{\includegraphics[width=0.24\columnwidth]{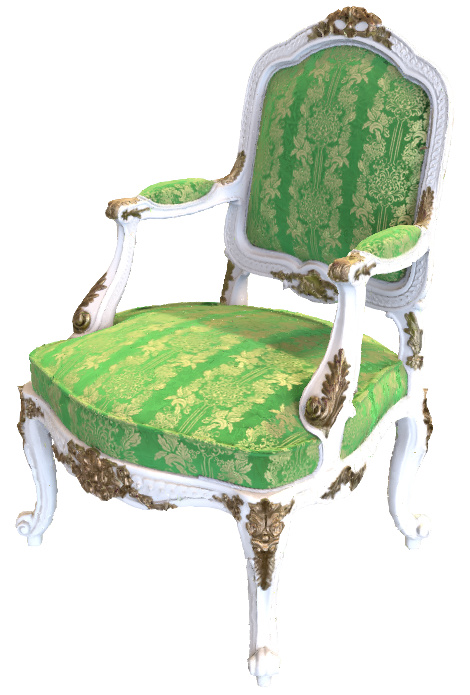}} \\
			\rotatebox[origin=c]{90}{\textsc{Mic}} &
			\raisebox{-0.5\height}{\includegraphics[width=0.24\columnwidth]{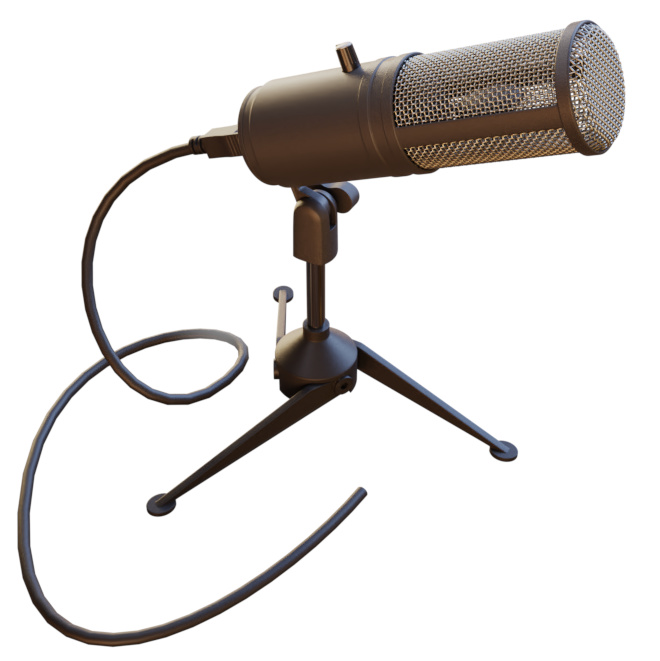}} & 
			\raisebox{-0.5\height}{\includegraphics[width=0.24\columnwidth]{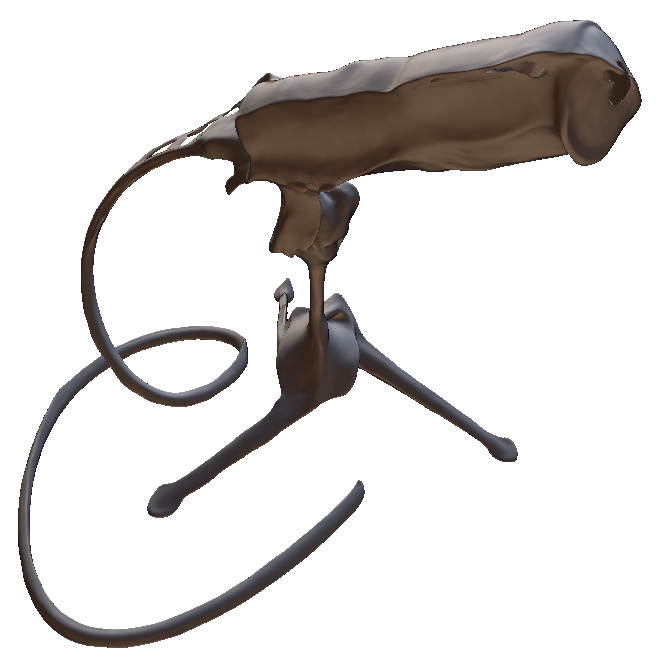}} &
			\raisebox{-0.5\height}{\includegraphics[width=0.24\columnwidth]{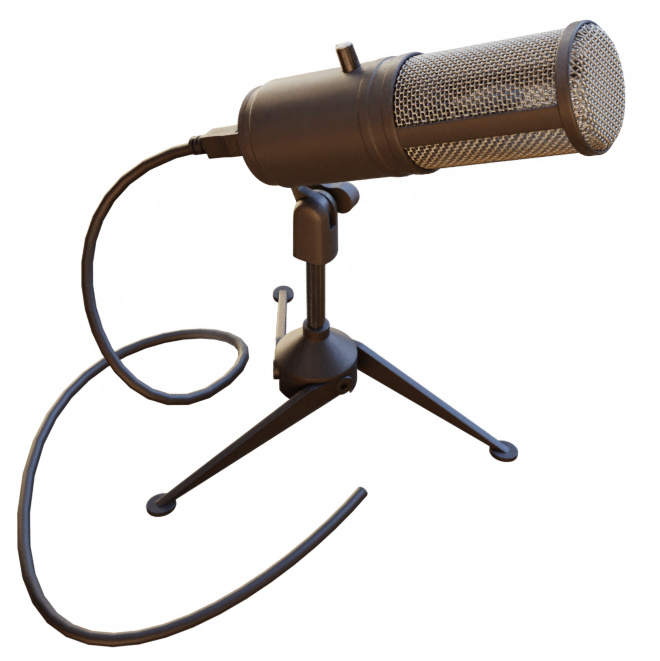}} &
			\raisebox{-0.5\height}{\includegraphics[width=0.24\columnwidth]{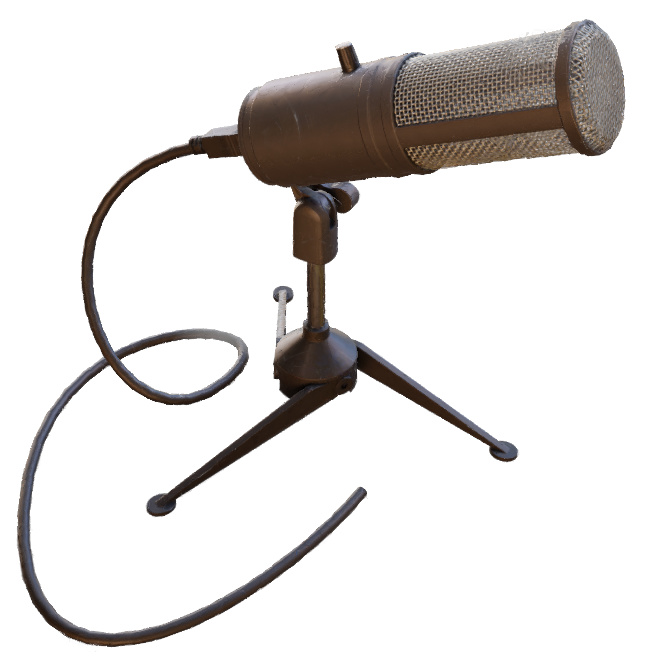}} \\
			\rotatebox[origin=c]{90}{\textsc{Ship}} &
			\raisebox{-0.5\height}{\includegraphics[width=0.24\columnwidth]{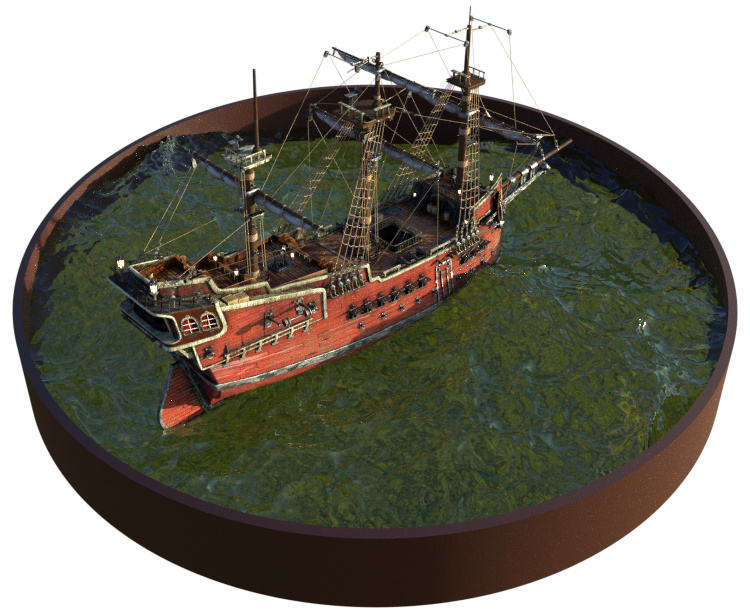}} & 
			\raisebox{-0.5\height}{\includegraphics[width=0.24\columnwidth]{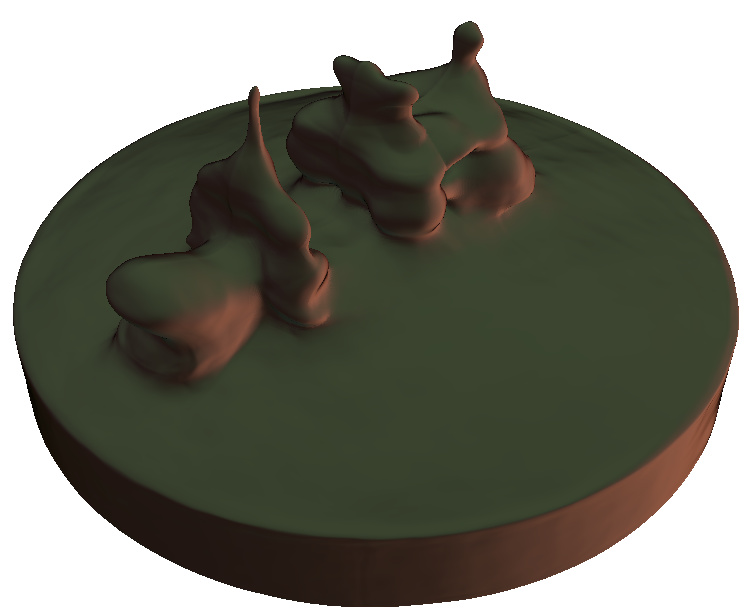}} &
			\raisebox{-0.5\height}{\includegraphics[width=0.24\columnwidth]{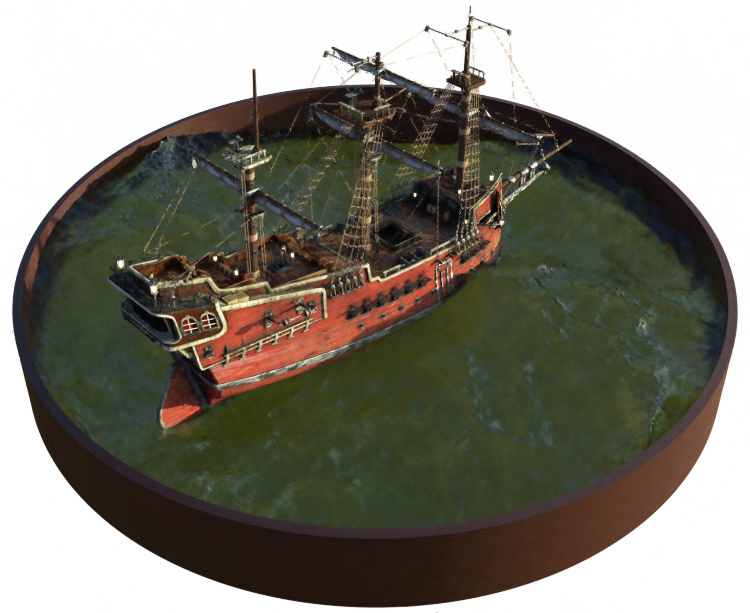}} &
			\raisebox{-0.5\height}{\includegraphics[width=0.24\columnwidth]{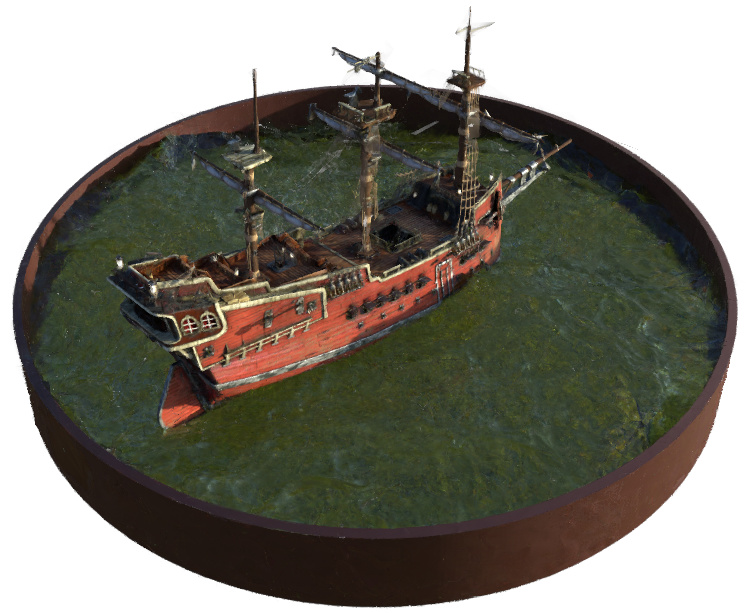}} \\
			& Reference & PhySG & MipNeRF & Our \\
		\end{tabularx}
	}
	\vspace*{-3mm}
	\caption{Visual quality examples from the NeRF realistic synthetic dataset comparing our method to PhySG and MipNeRF. PhySG struggles to 
		accurately capture the complex geometry and spatially varying materials of the dataset.}
	\label{fig:physg_comp}
\end{figure}
}

\newcommand{\figKnobLambert}{
	\begin{figure}
		\centering
		\setlength{\tabcolsep}{1pt}
		\begin{tabular}{cccc}
			\includegraphics[width=0.24\columnwidth]{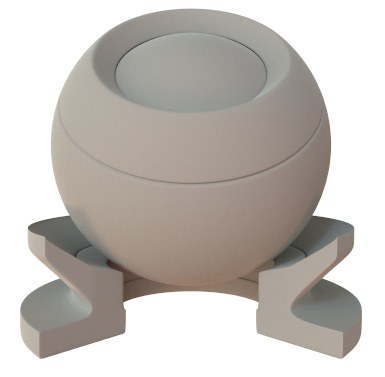} & 
			\includegraphics[width=0.24\columnwidth]{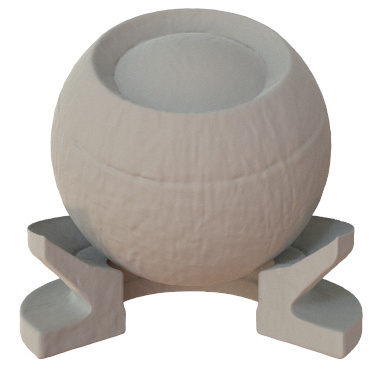} &
			\includegraphics[width=0.24\columnwidth]{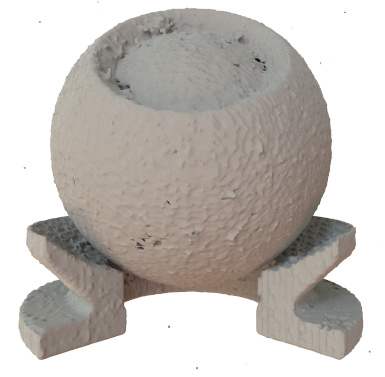} &
			\includegraphics[width=0.24\columnwidth]{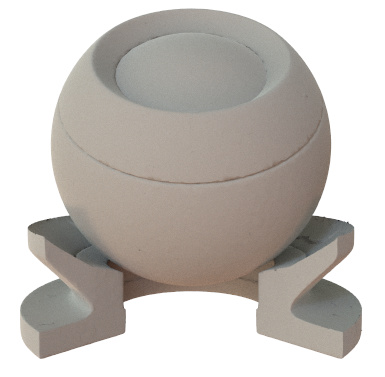} \\
			82k tris & 2.5M tris & 766k tris & 119k tris \\
			\includegraphics[width=0.24\columnwidth]{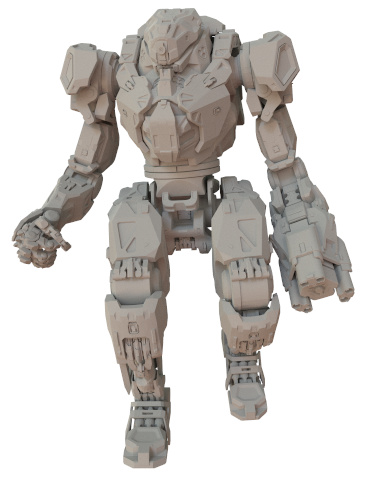} & 
			\includegraphics[width=0.24\columnwidth]{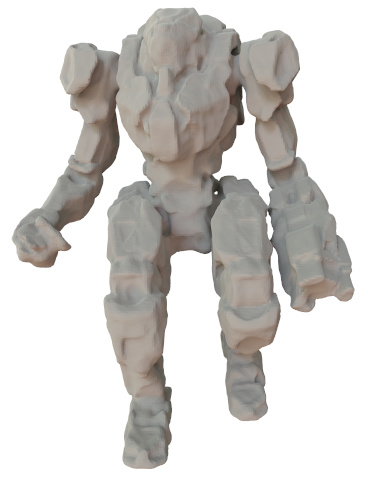} &
			\includegraphics[width=0.24\columnwidth]{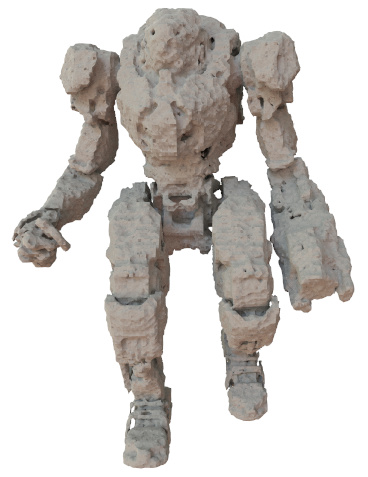} &
			\includegraphics[width=0.24\columnwidth]{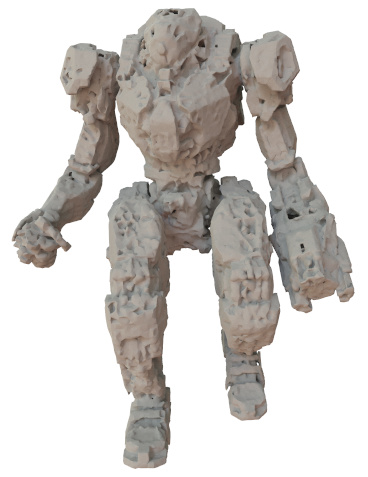} \\
			192k tris & 323k tris & 314k tris & 311k tris \\
			Reference & NeuS & NeRF & Our
		\end{tabular}
		\caption{
			Extracted mesh quality visualization  examples on the synthetic \textsc{Knob} and \textsc{Cerberus}
			datasets. 
		}
		\label{fig:knob_lambert}
	\end{figure}
}

\newlength{\myfigsize}
\setlength{\myfigsize}{0.14\textwidth}

\newcommand{\figMegaNerf}{
	\begin{figure*}
		\centering
		\setlength{\tabcolsep}{1pt}

		\begin{tabular}{lcccccc}
			& \small{Reference} & \small{Our} & \small{$\kd$} & \small{$\korm$} & \small{normals} & \small{HDR probe} \\

			\rotatebox[origin=c]{90}{\small{Chair}} &
			\raisebox{-0.5\height}{\includegraphics[width=\myfigsize]{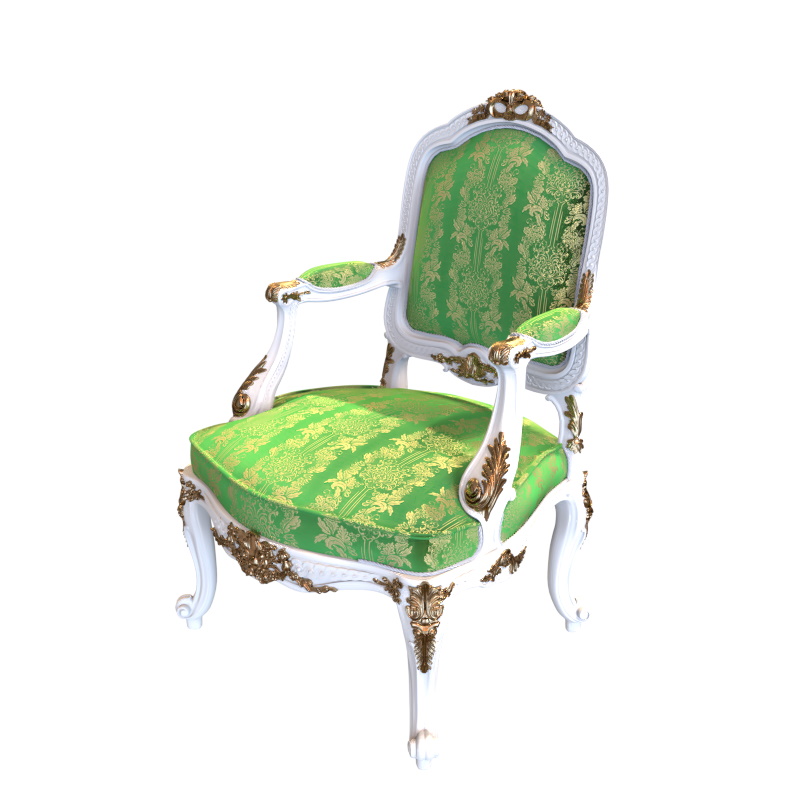}} &
			\raisebox{-0.5\height}{\includegraphics[width=\myfigsize]{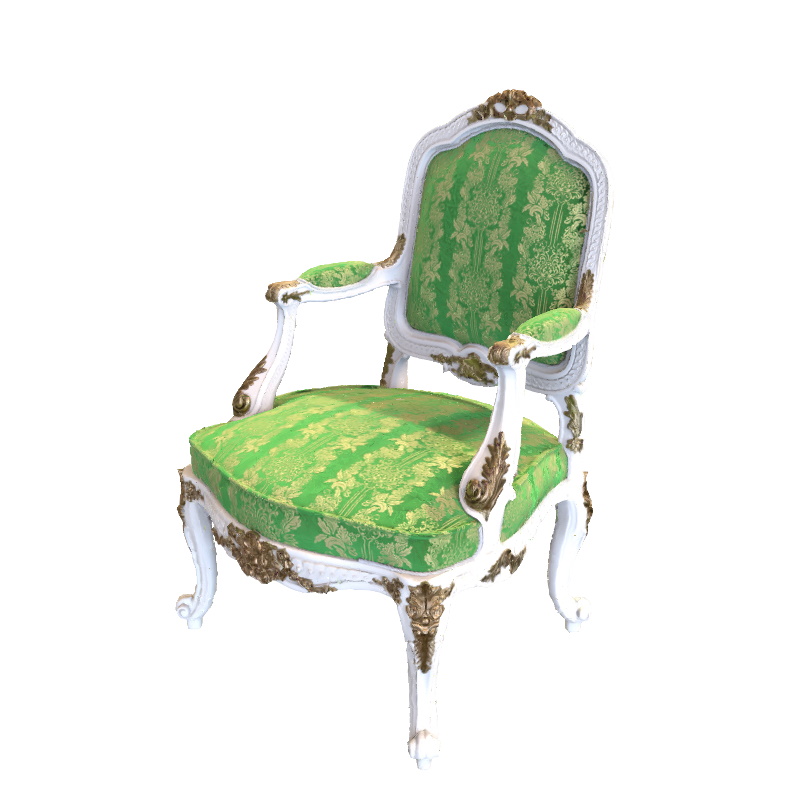}} &
			\raisebox{-0.5\height}{\includegraphics[width=\myfigsize]{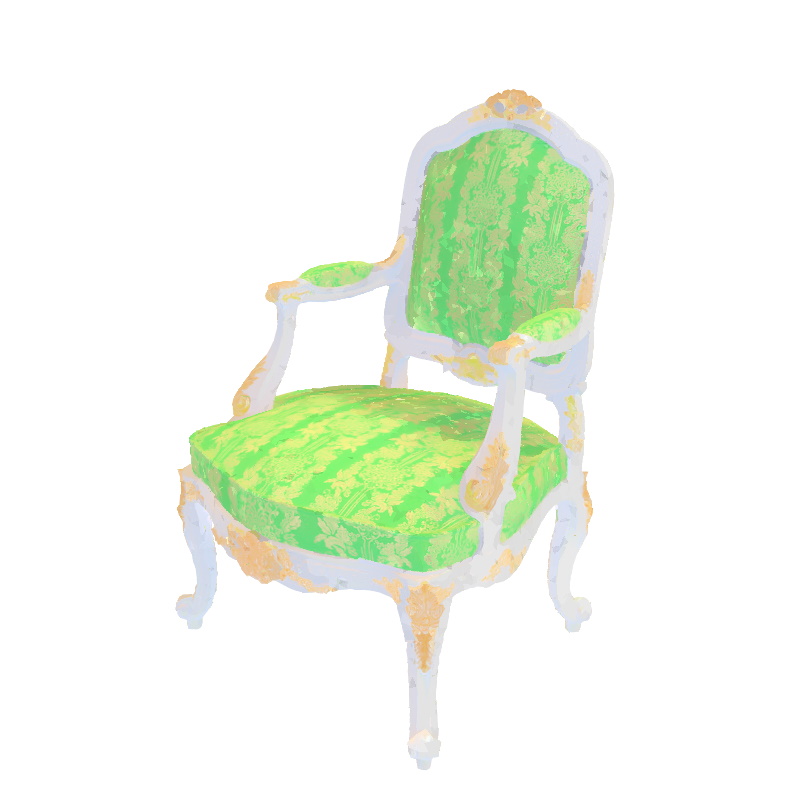}} &
			\raisebox{-0.5\height}{\includegraphics[width=\myfigsize]{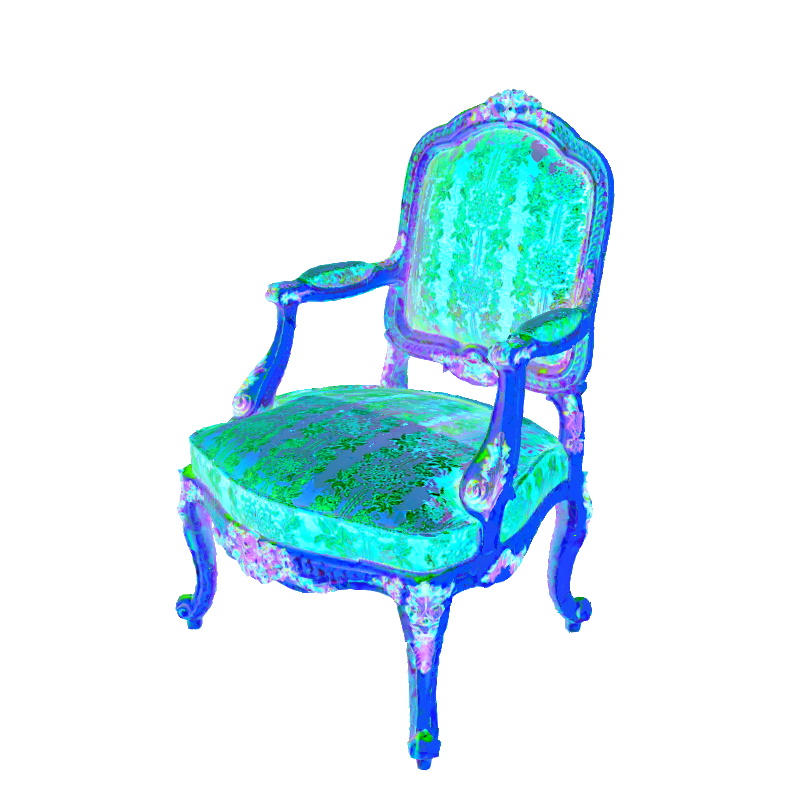}} &
			\raisebox{-0.5\height}{\includegraphics[width=\myfigsize]{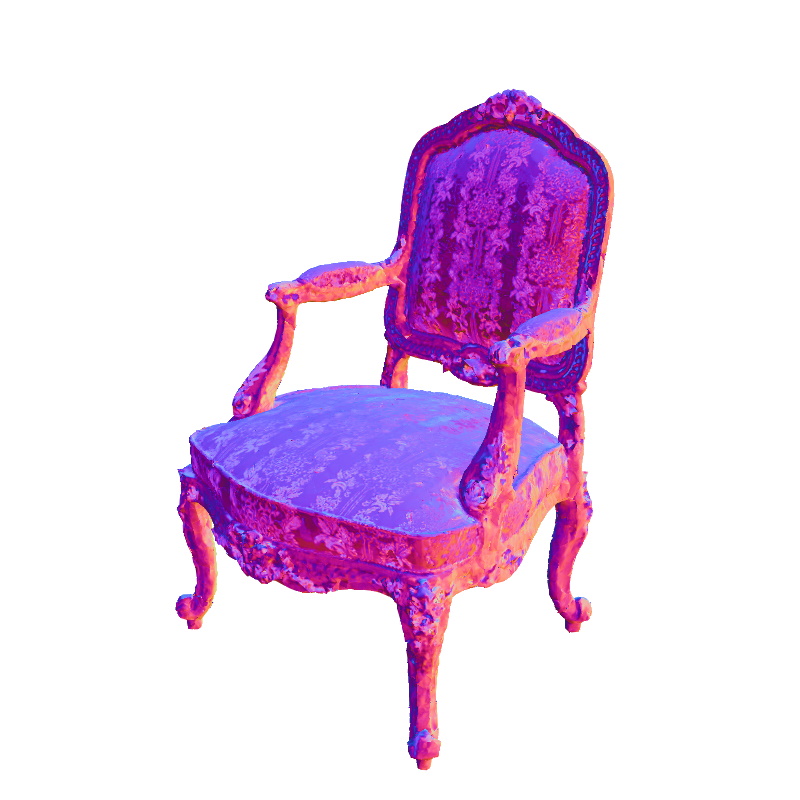}} &
			\raisebox{-0.5\height}{\includegraphics[width=\myfigsize]{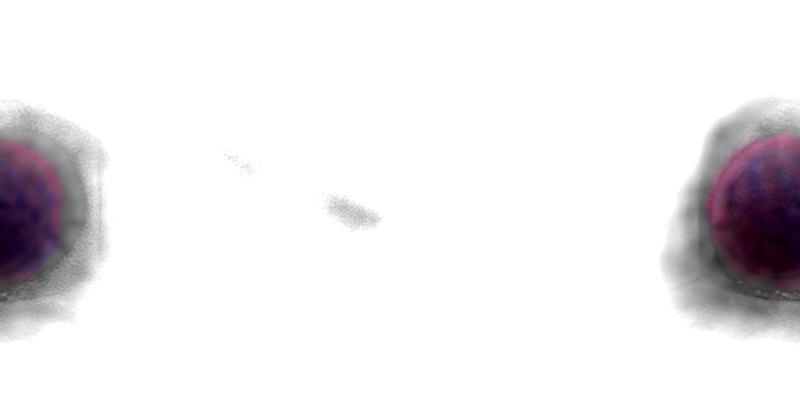}} \\

			\rotatebox[origin=c]{90}{\small{Lego}} &
			\raisebox{-0.5\height}{\includegraphics[trim={0cm 3cm 0cm 2cm},clip,width=\myfigsize]{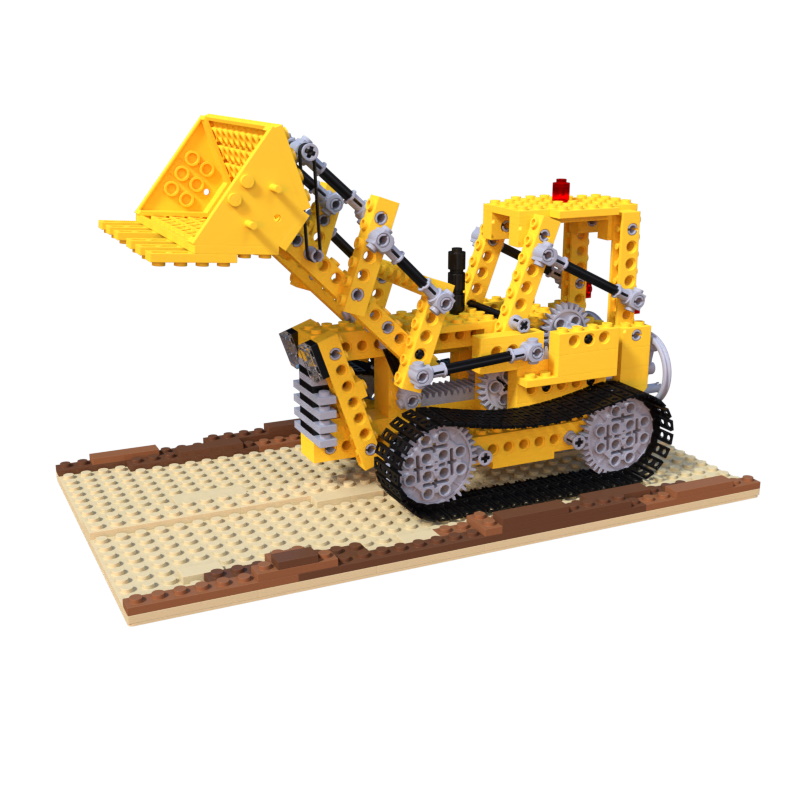}} &
			\raisebox{-0.5\height}{\includegraphics[trim={0cm 3cm 0cm 2cm},clip,width=\myfigsize]{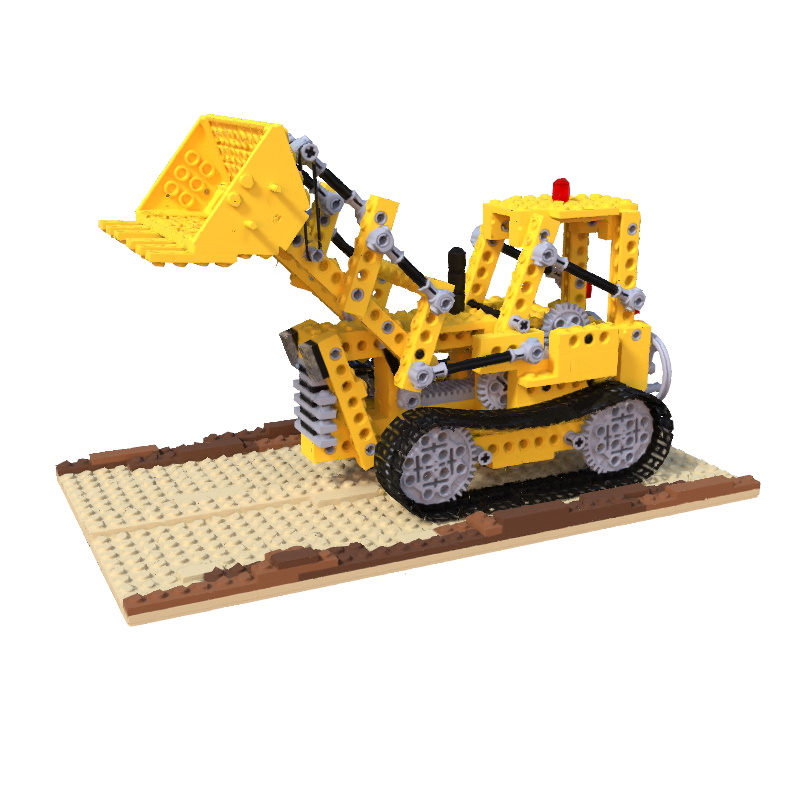}} &
			\raisebox{-0.5\height}{\includegraphics[trim={0cm 3cm 0cm 2cm},clip,width=\myfigsize]{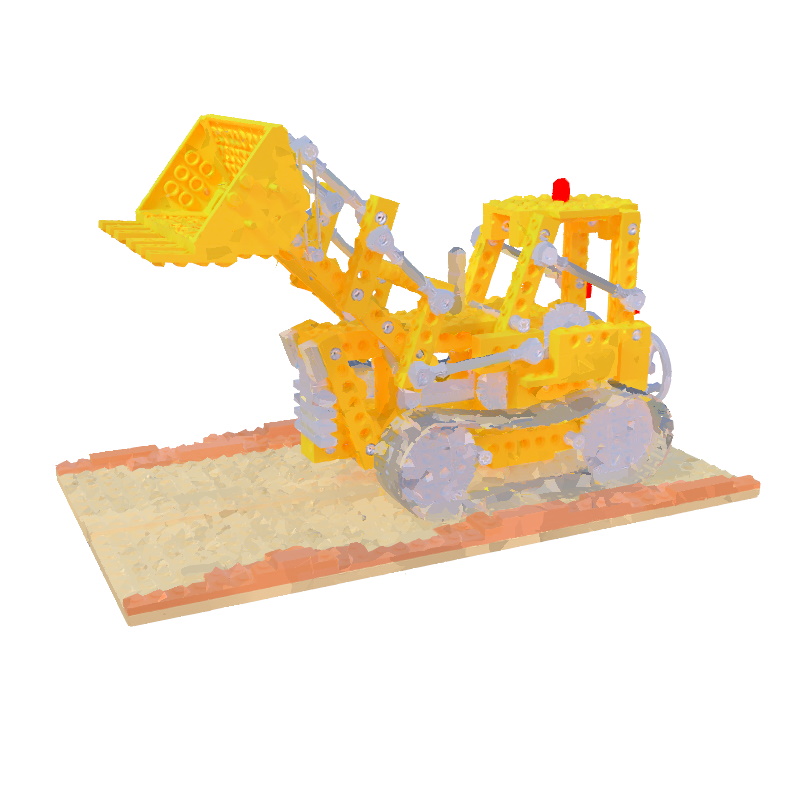}} &
			\raisebox{-0.5\height}{\includegraphics[trim={0cm 3cm 0cm 2cm},clip,width=\myfigsize]{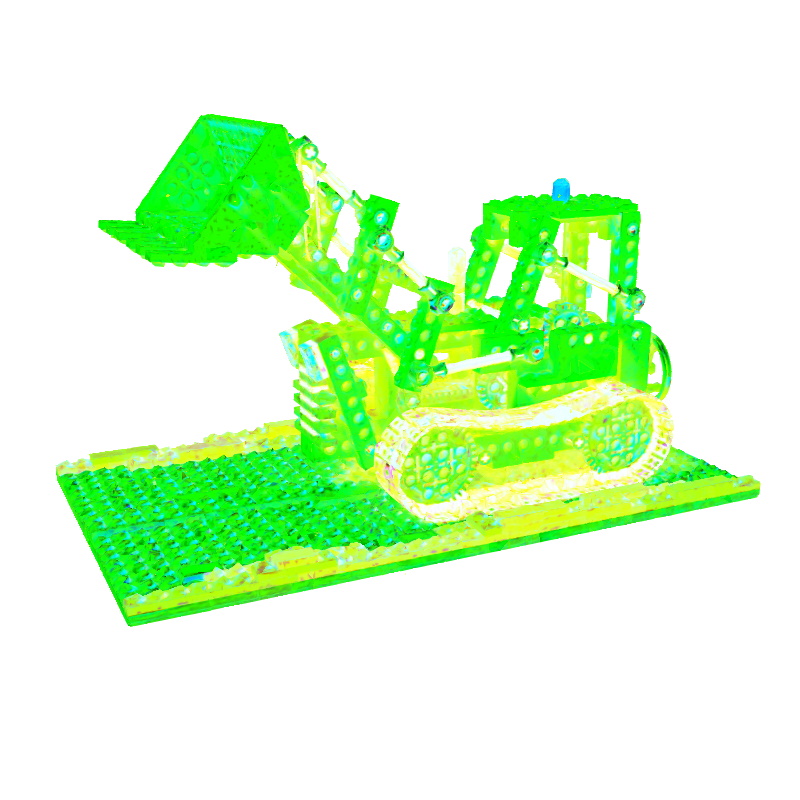}} &
			\raisebox{-0.5\height}{\includegraphics[trim={0cm 3cm 0cm 2cm},clip,width=\myfigsize]{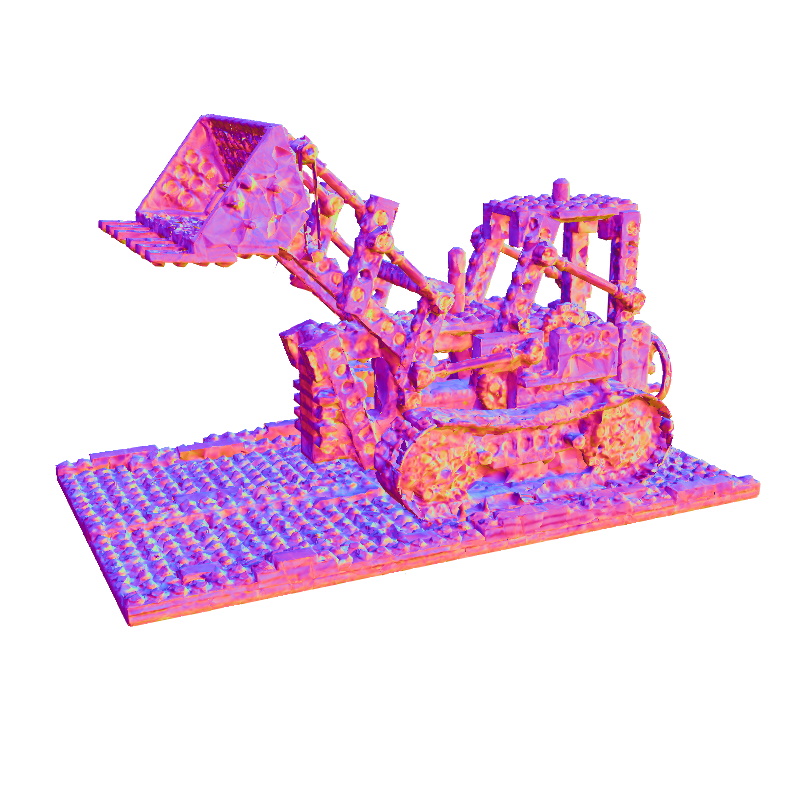}} &
			\raisebox{-0.5\height}{\includegraphics[width=\myfigsize]{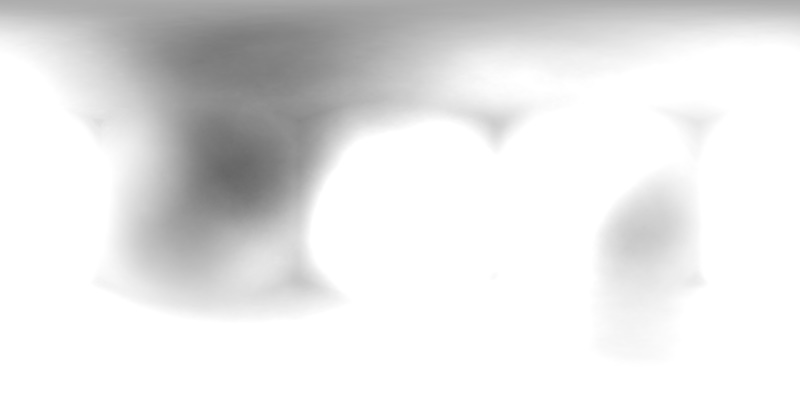}} \\

			\rotatebox[origin=c]{90}{\small{Drums}} &
			\raisebox{-0.5\height}{\includegraphics[width=\myfigsize]{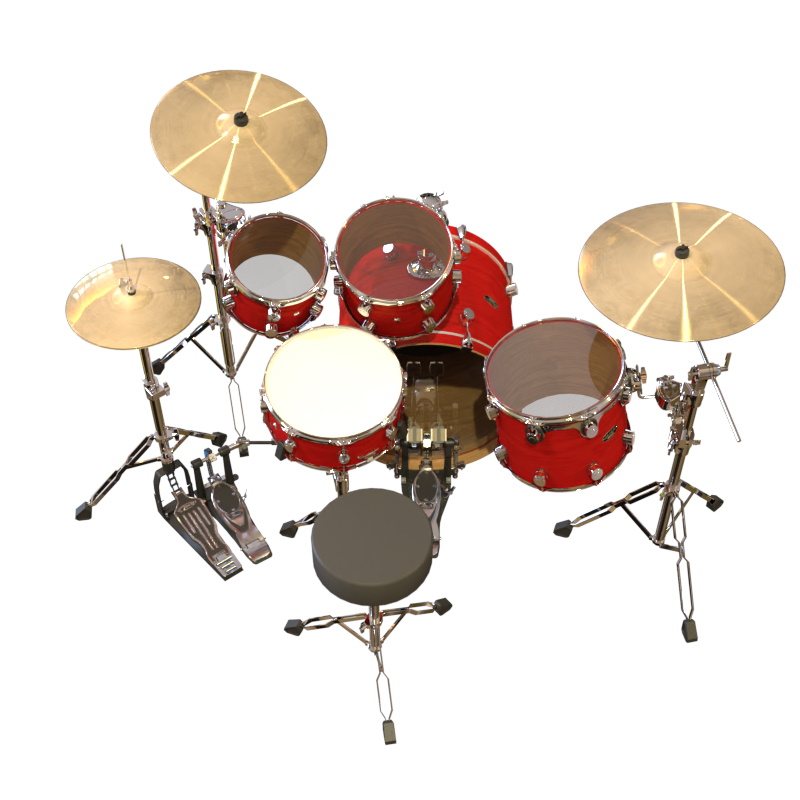}} &
			\raisebox{-0.5\height}{\includegraphics[width=\myfigsize]{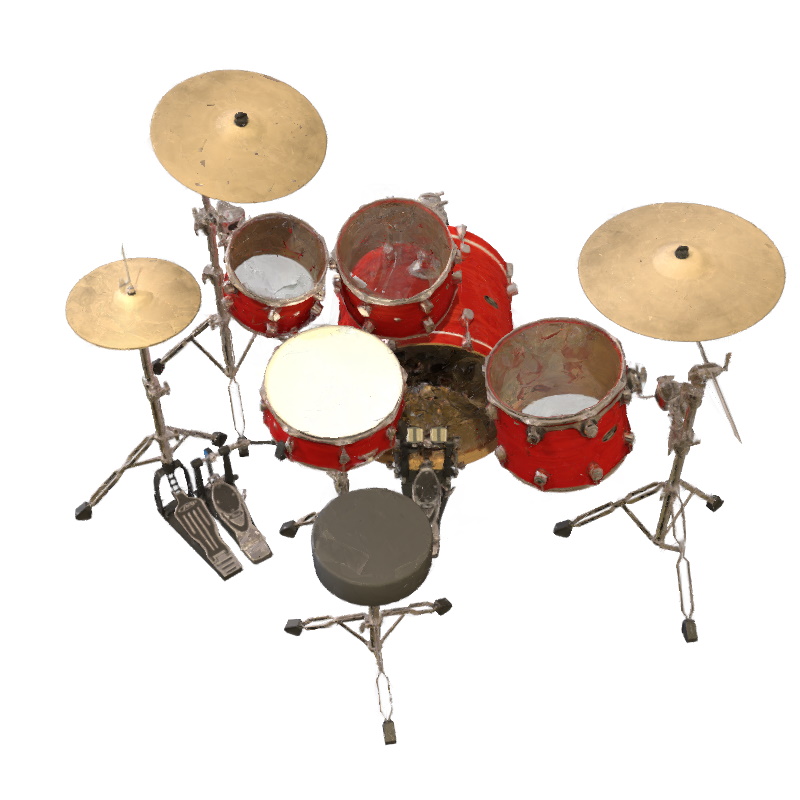}} &
			\raisebox{-0.5\height}{\includegraphics[width=\myfigsize]{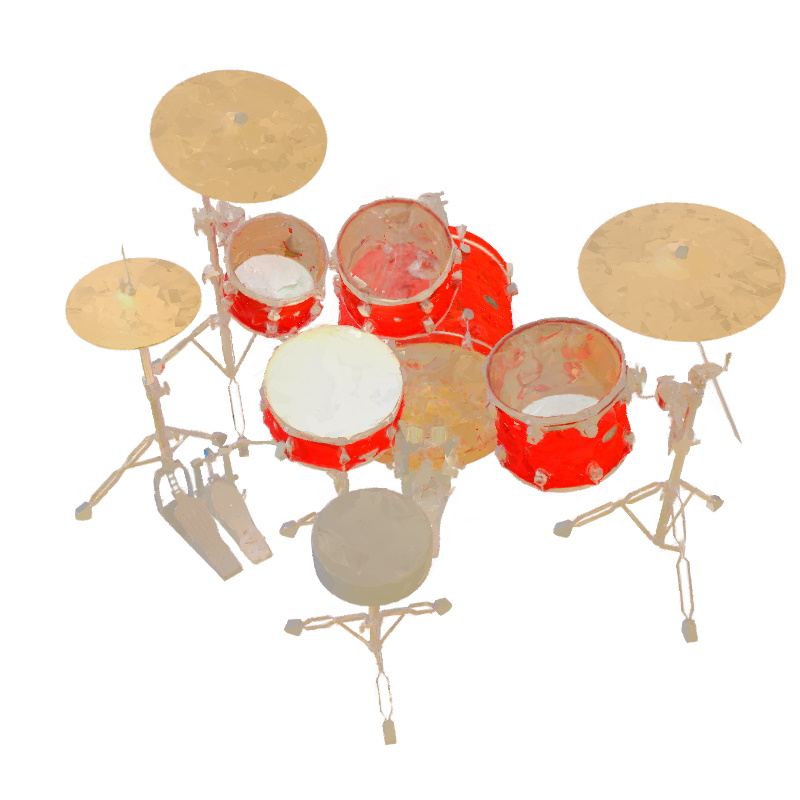}} &
			\raisebox{-0.5\height}{\includegraphics[width=\myfigsize]{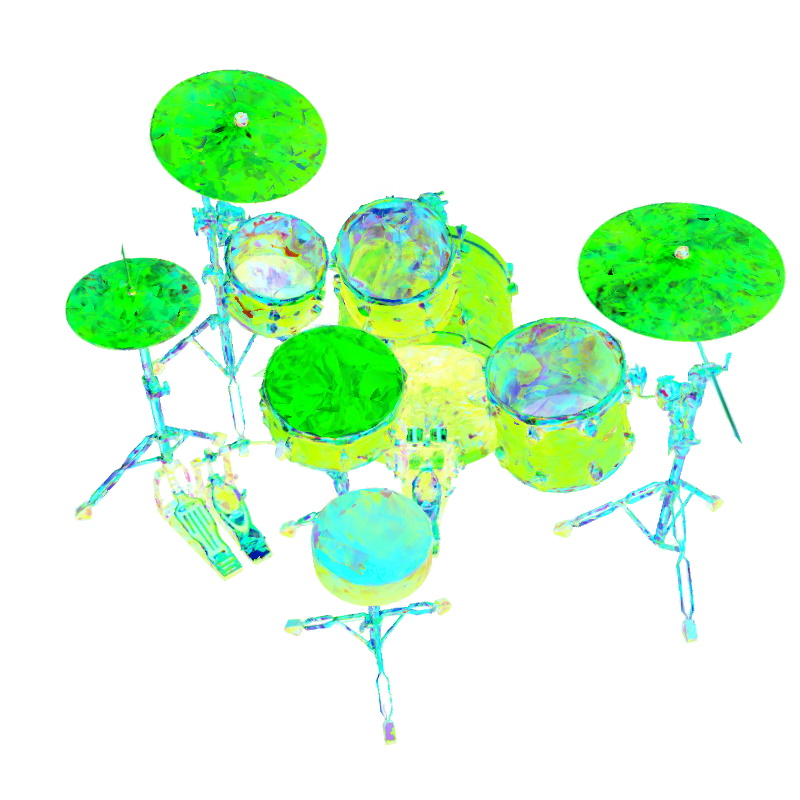}} &
			\raisebox{-0.5\height}{\includegraphics[width=\myfigsize]{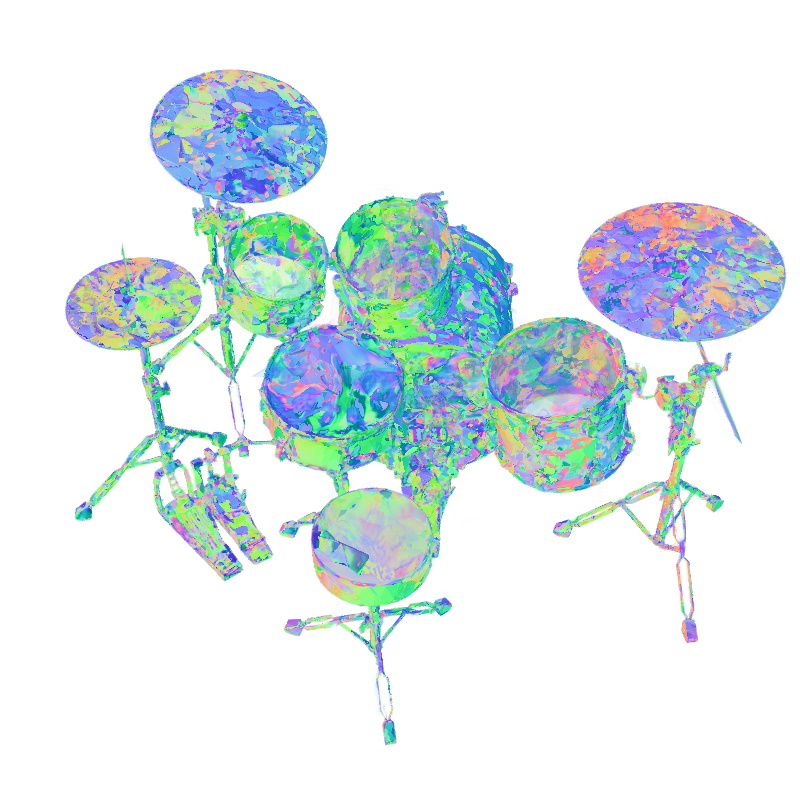}} &
			\raisebox{-0.5\height}{\includegraphics[width=\myfigsize]{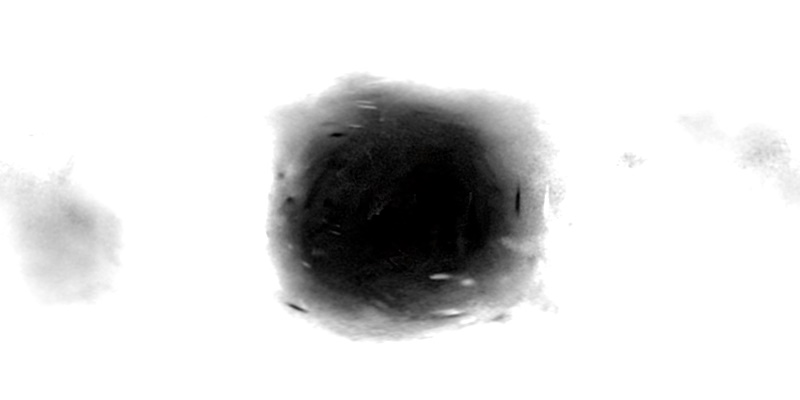}} \\

			\rotatebox[origin=c]{90}{\small{Ficus}} &
			\raisebox{-0.5\height}{\includegraphics[trim={0cm 5cm 0cm 2cm},clip,width=\myfigsize]{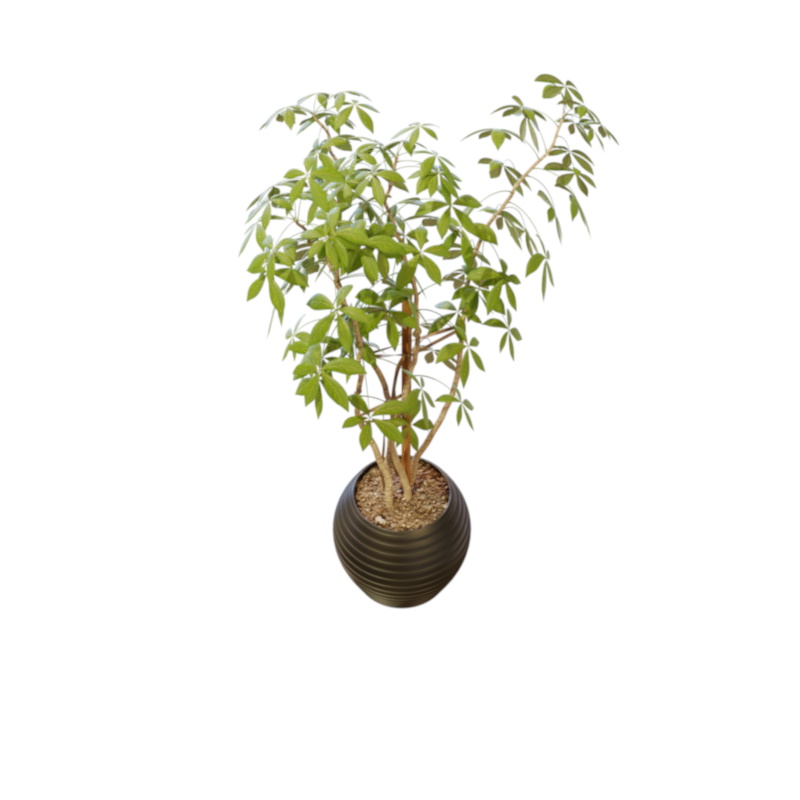}} &
			\raisebox{-0.5\height}{\includegraphics[trim={0cm 5cm 0cm 2cm},clip,width=\myfigsize]{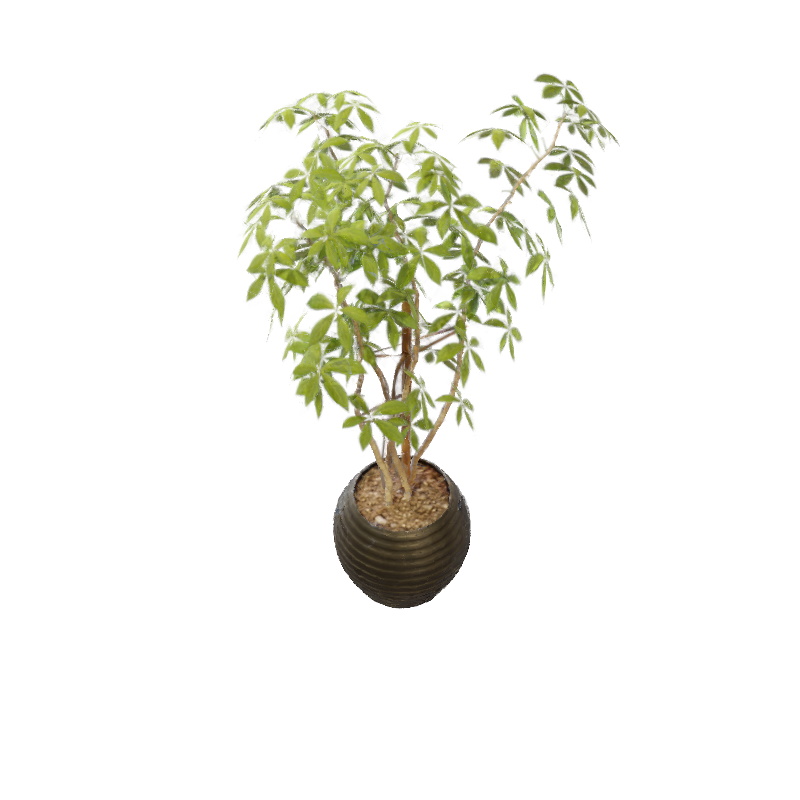}} &
			\raisebox{-0.5\height}{\includegraphics[trim={0cm 5cm 0cm 2cm},clip,width=\myfigsize]{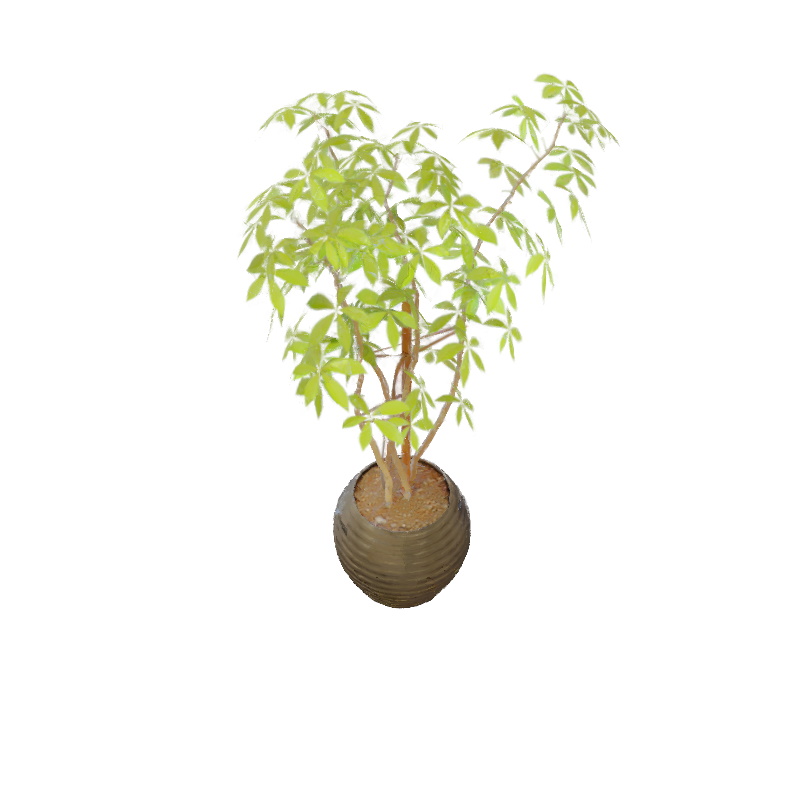}} &
			\raisebox{-0.5\height}{\includegraphics[trim={0cm 5cm 0cm 2cm},clip,width=\myfigsize]{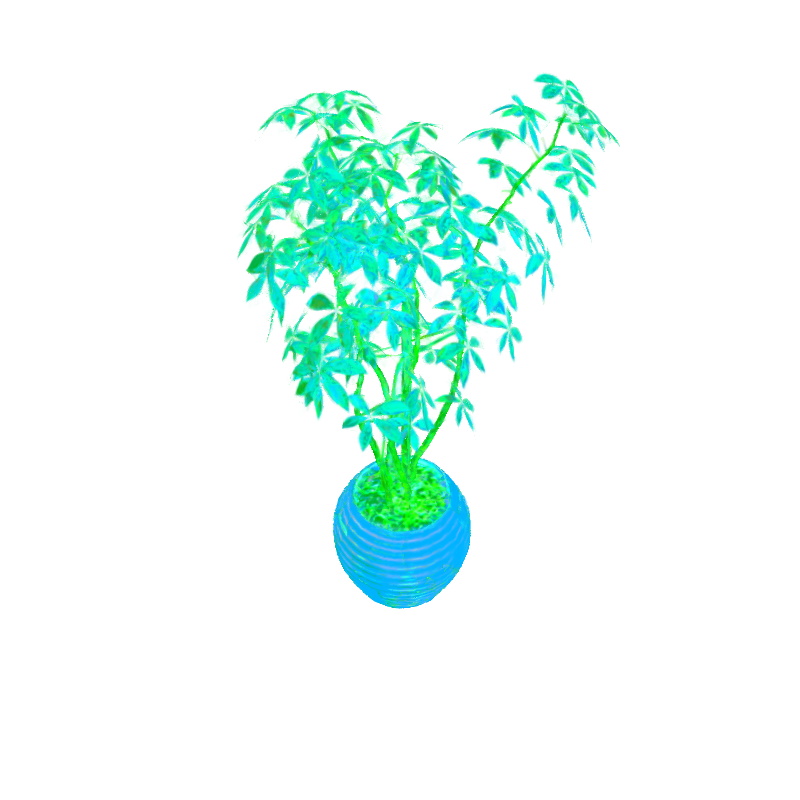}} &
			\raisebox{-0.5\height}{\includegraphics[trim={0cm 5cm 0cm 2cm},clip,width=\myfigsize]{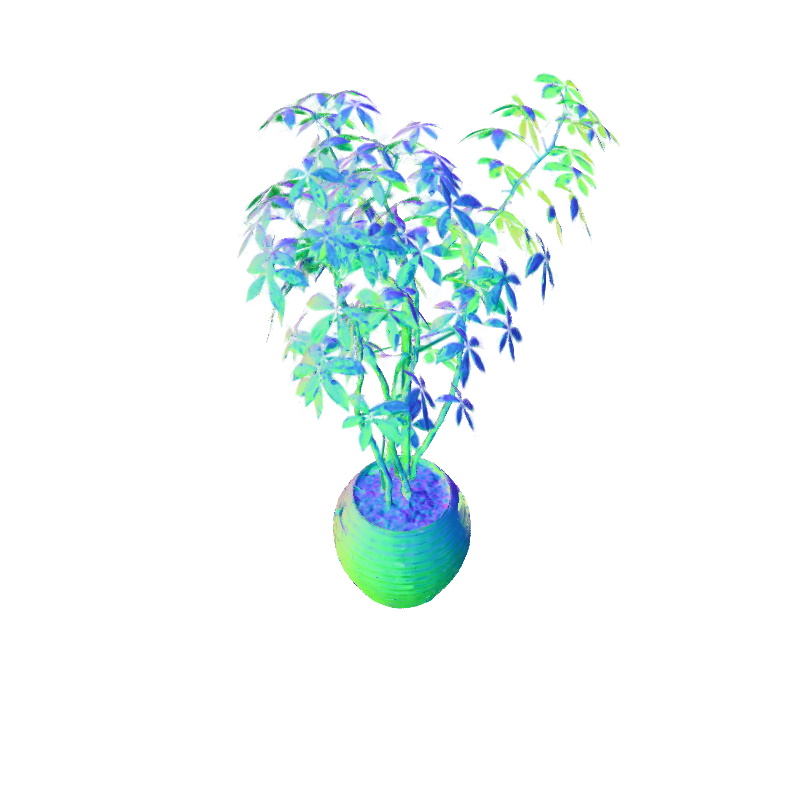}} &
			\raisebox{-0.5\height}{\includegraphics[width=\myfigsize]{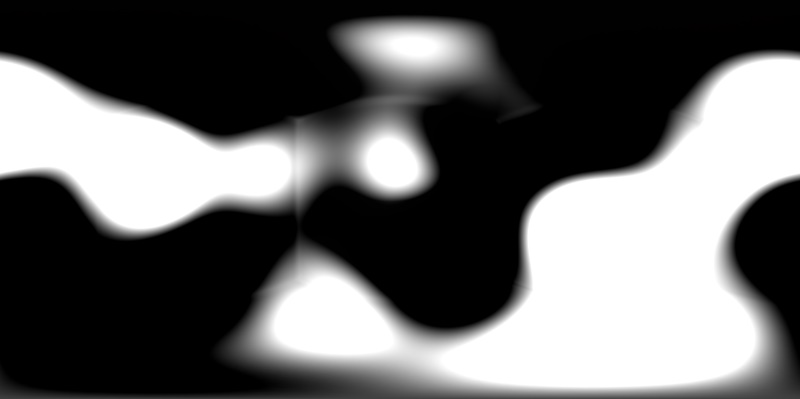}} \\

			\rotatebox[origin=c]{90}{\small{Hotdog}} &
			\raisebox{-0.5\height}{\includegraphics[trim={0cm 2cm 0cm 2cm},clip,width=\myfigsize]{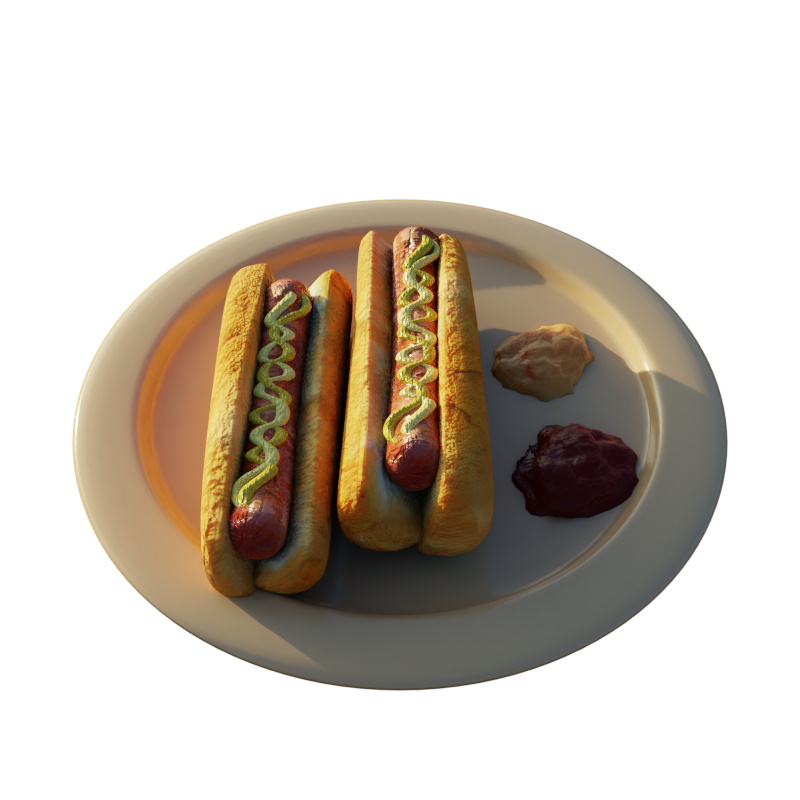}} &
			\raisebox{-0.5\height}{\includegraphics[trim={0cm 2cm 0cm 2cm},clip,width=\myfigsize]{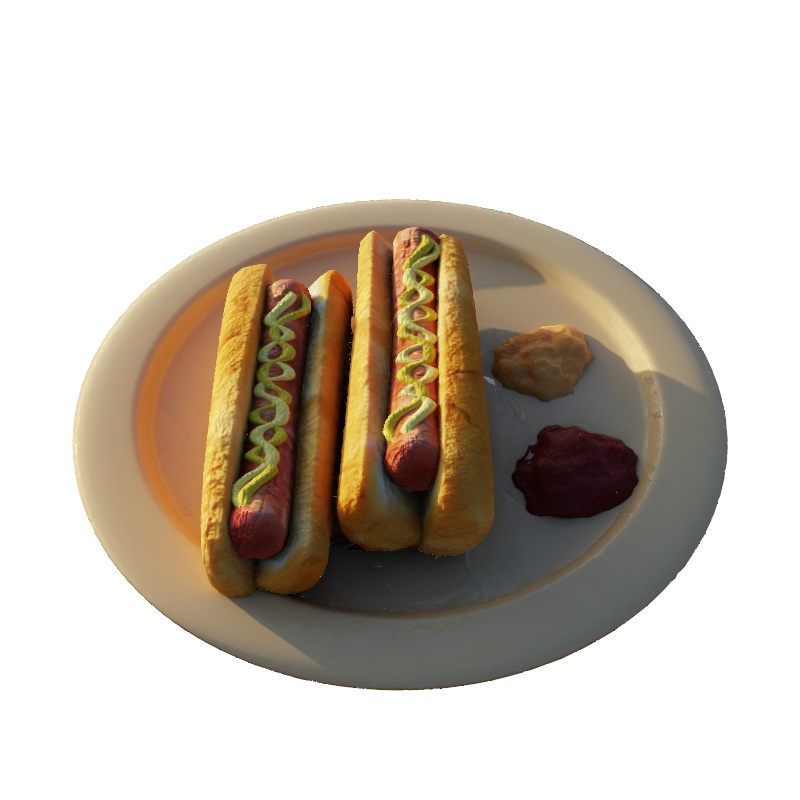}} &
			\raisebox{-0.5\height}{\includegraphics[trim={0cm 2cm 0cm 2cm},clip,width=\myfigsize]{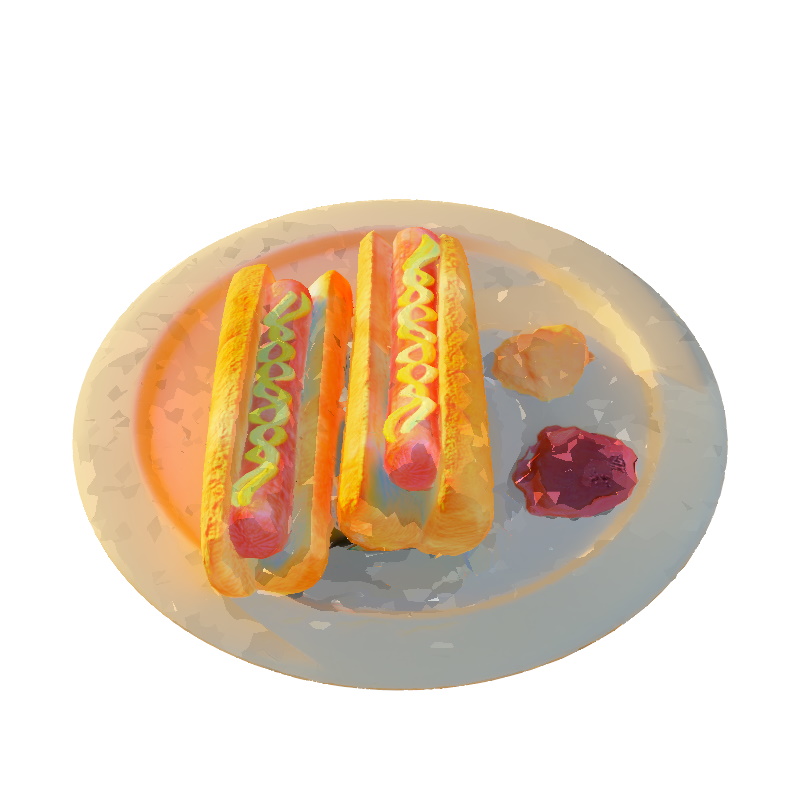}} &
			\raisebox{-0.5\height}{\includegraphics[trim={0cm 2cm 0cm 2cm},clip,width=\myfigsize]{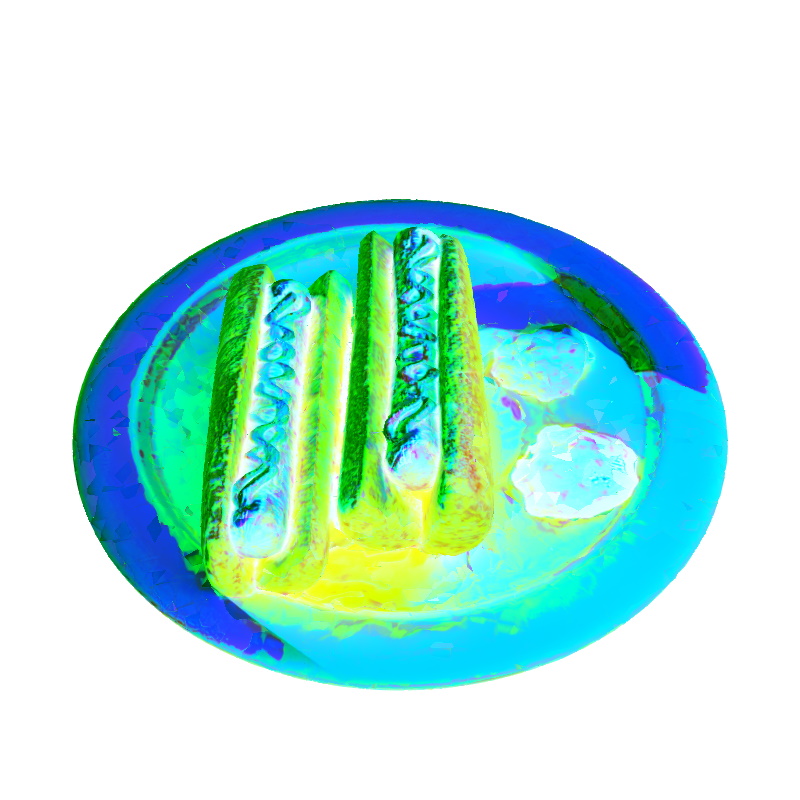}} &
			\raisebox{-0.5\height}{\includegraphics[trim={0cm 2cm 0cm 2cm},clip,width=\myfigsize]{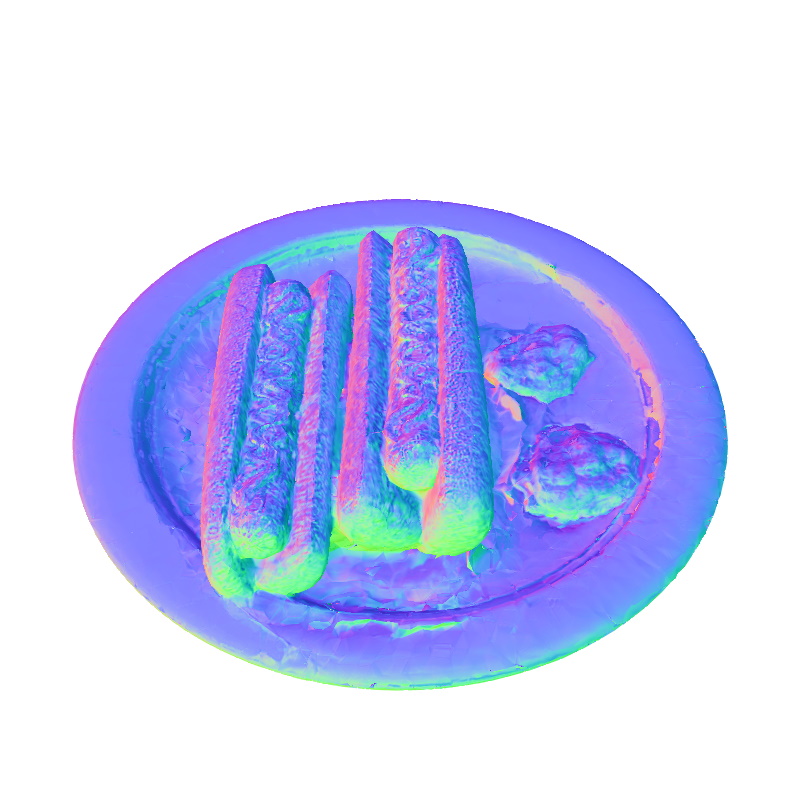}} &
			\raisebox{-0.5\height}{\includegraphics[width=\myfigsize]{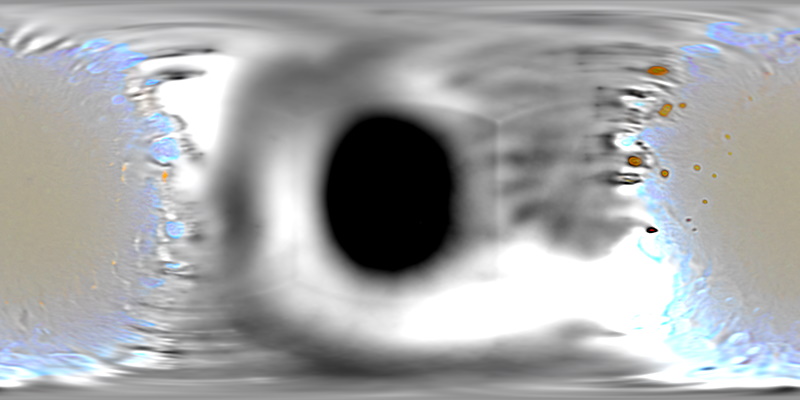}} \\

			\rotatebox[origin=c]{90}{\small{Materials}} &
			\raisebox{-0.5\height}{\includegraphics[trim={0cm 2cm 0cm 2cm},clip,width=\myfigsize]{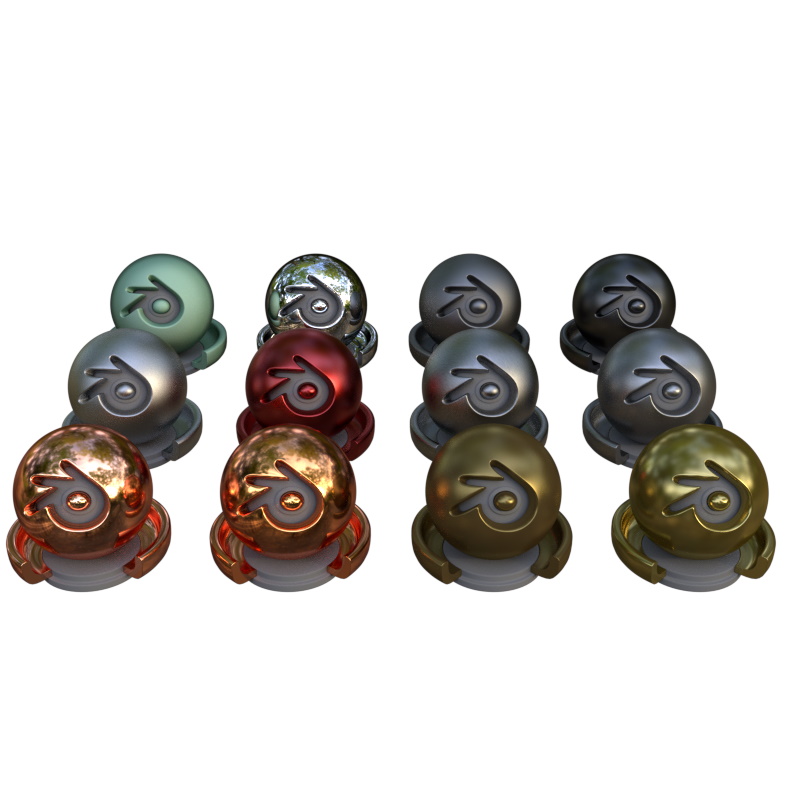}} &
			\raisebox{-0.5\height}{\includegraphics[trim={0cm 2cm 0cm 2cm},clip,width=\myfigsize]{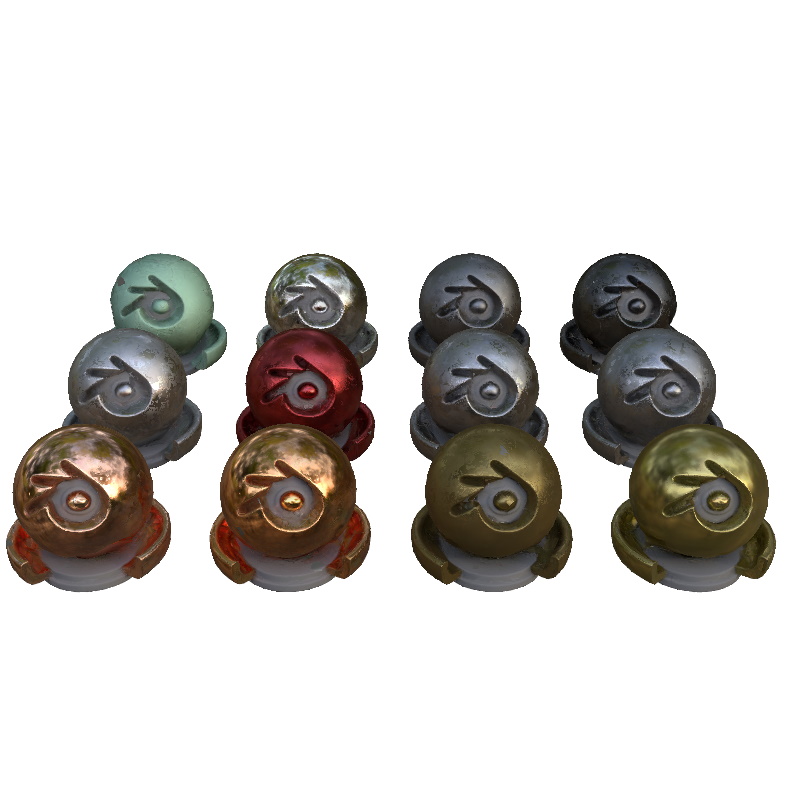}} &
			\raisebox{-0.5\height}{\includegraphics[trim={0cm 2cm 0cm 2cm},clip,width=\myfigsize]{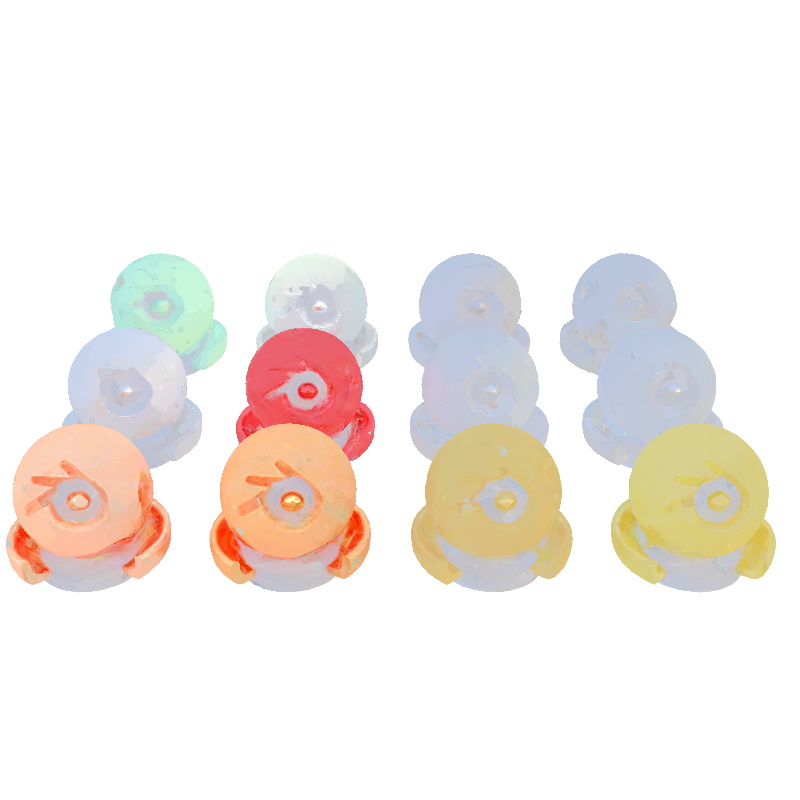}} &
			\raisebox{-0.5\height}{\includegraphics[trim={0cm 2cm 0cm 2cm},clip,width=\myfigsize]{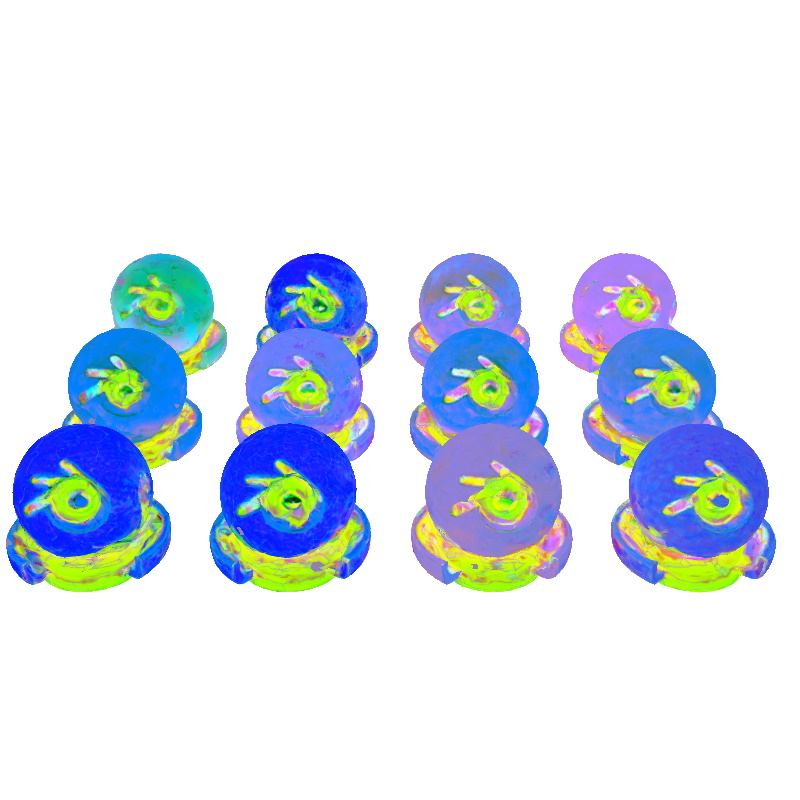}} &
			\raisebox{-0.5\height}{\includegraphics[trim={0cm 2cm 0cm 2cm},clip,width=\myfigsize]{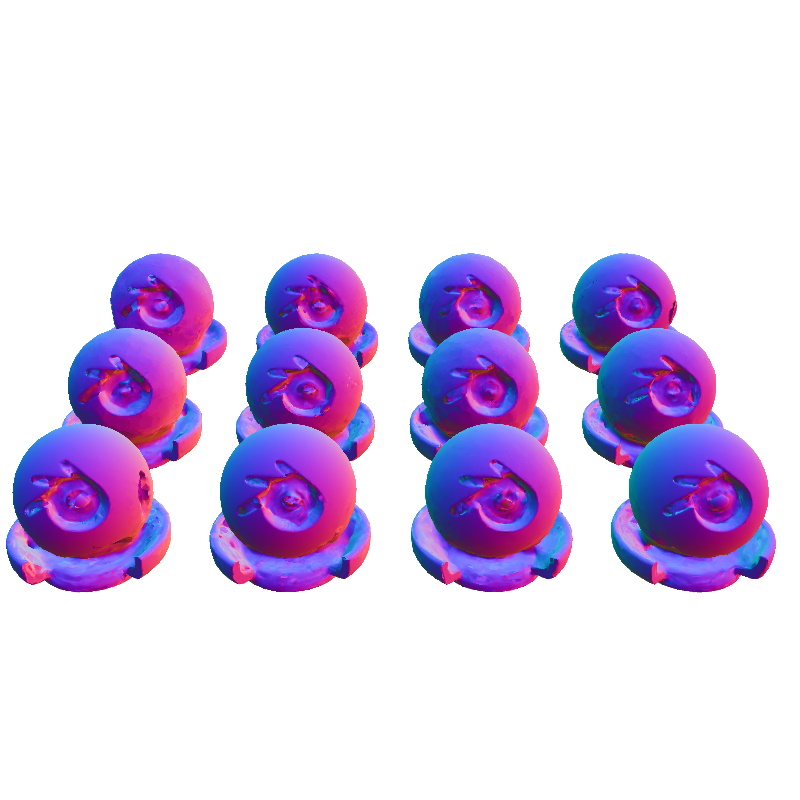}} &
			\raisebox{-0.5\height}{\includegraphics[width=\myfigsize]{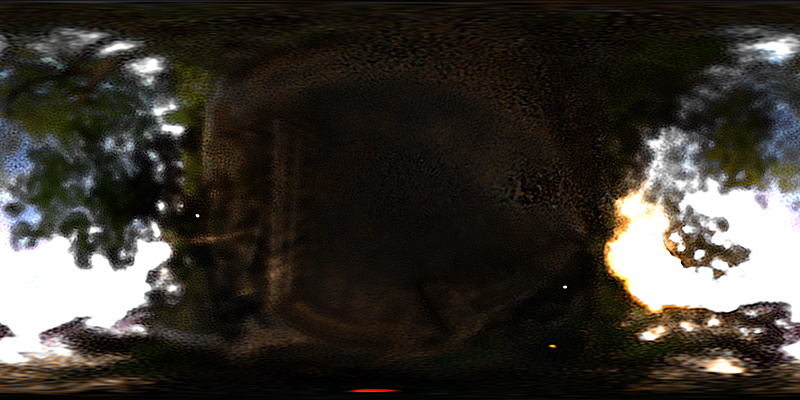}} \\

			\rotatebox[origin=c]{90}{\small{Mic}} &
			\raisebox{-0.5\height}{\includegraphics[trim={0cm 1cm 0cm 2cm},clip,width=\myfigsize]{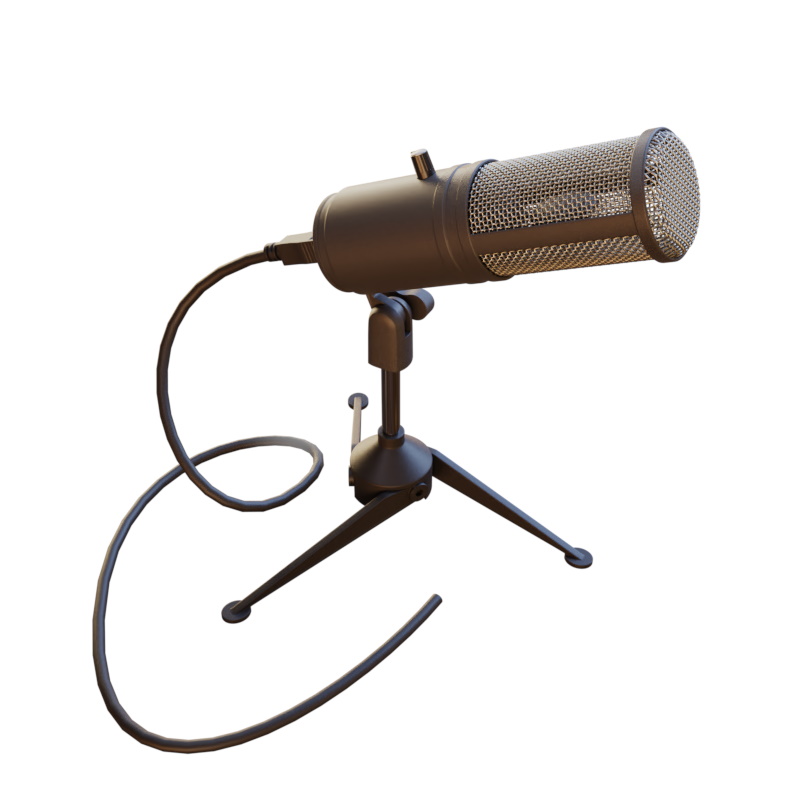}} &
			\raisebox{-0.5\height}{\includegraphics[trim={0cm 1cm 0cm 2cm},clip,width=\myfigsize]{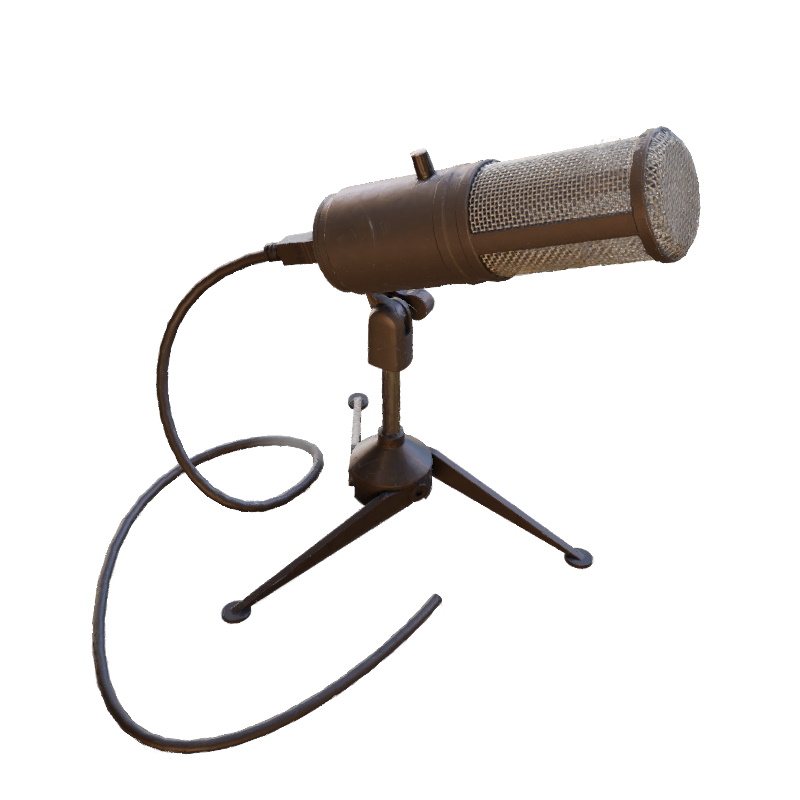}} &
			\raisebox{-0.5\height}{\includegraphics[trim={0cm 1cm 0cm 2cm},clip,width=\myfigsize]{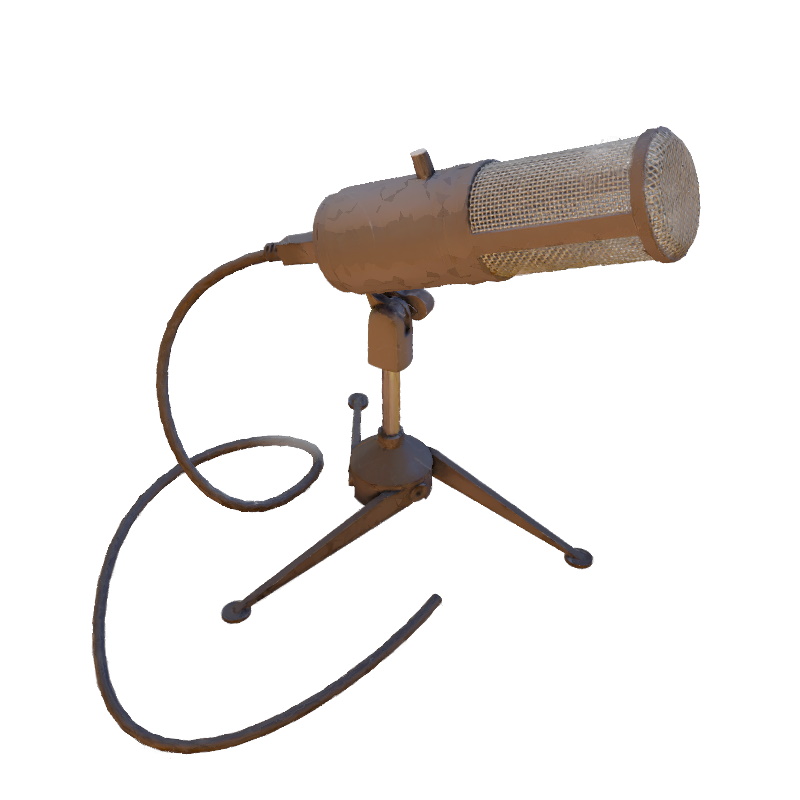}} &
			\raisebox{-0.5\height}{\includegraphics[trim={0cm 1cm 0cm 2cm},clip,width=\myfigsize]{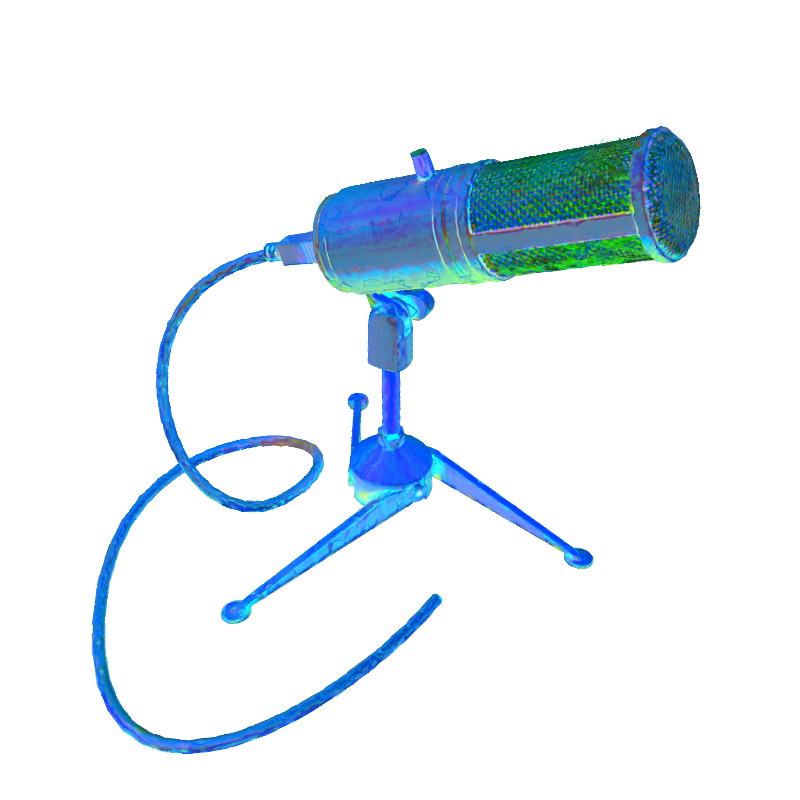}} &
			\raisebox{-0.5\height}{\includegraphics[trim={0cm 1cm 0cm 2cm},clip,width=\myfigsize]{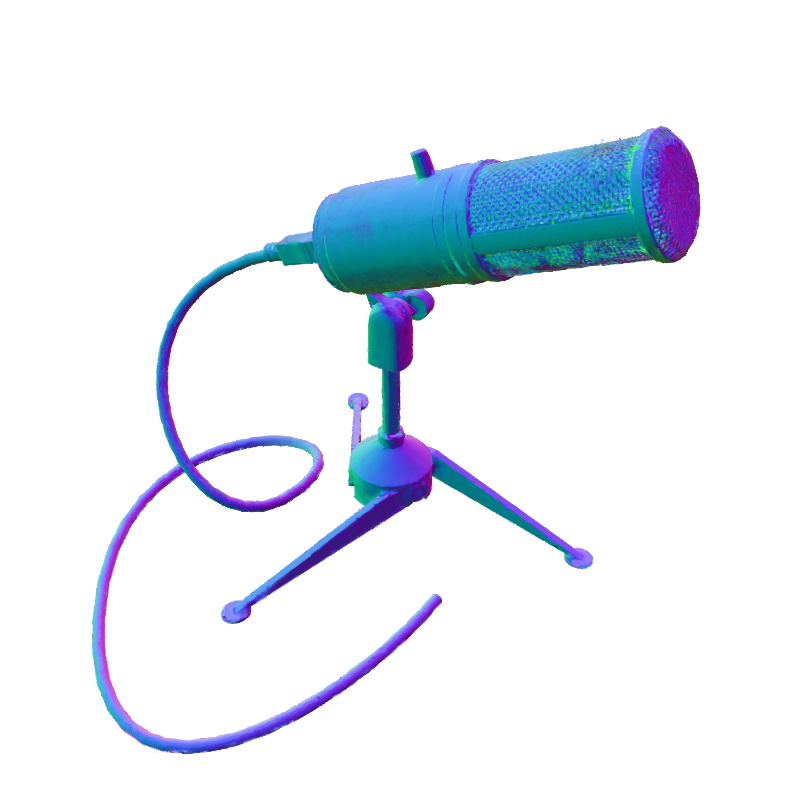}} &
			\raisebox{-0.5\height}{\includegraphics[width=\myfigsize]{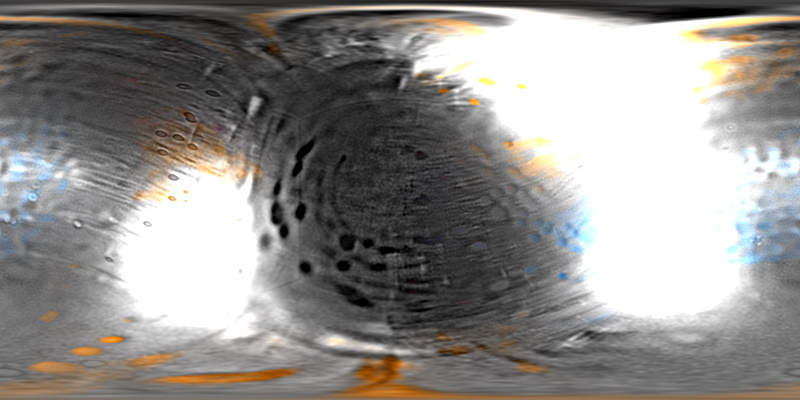}} \\

			\rotatebox[origin=c]{90}{\small{Ship}} &
			\raisebox{-0.5\height}{\includegraphics[trim={0cm 0cm 0cm 4cm},clip,width=\myfigsize]{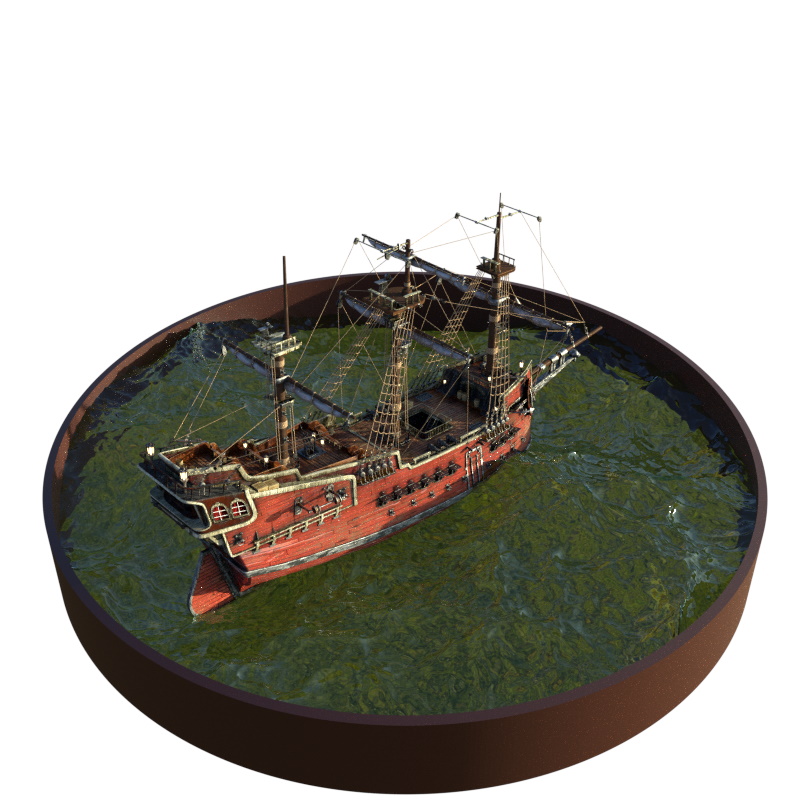}} &
			\raisebox{-0.5\height}{\includegraphics[trim={0cm 0cm 0cm 4cm},clip,width=\myfigsize]{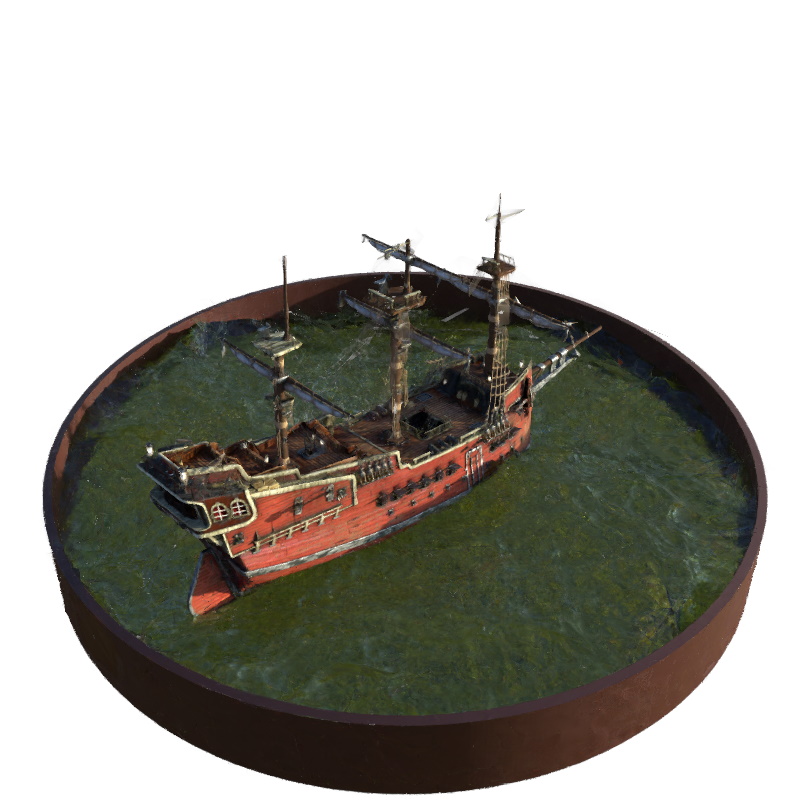}} &
			\raisebox{-0.5\height}{\includegraphics[trim={0cm 0cm 0cm 4cm},clip,width=\myfigsize]{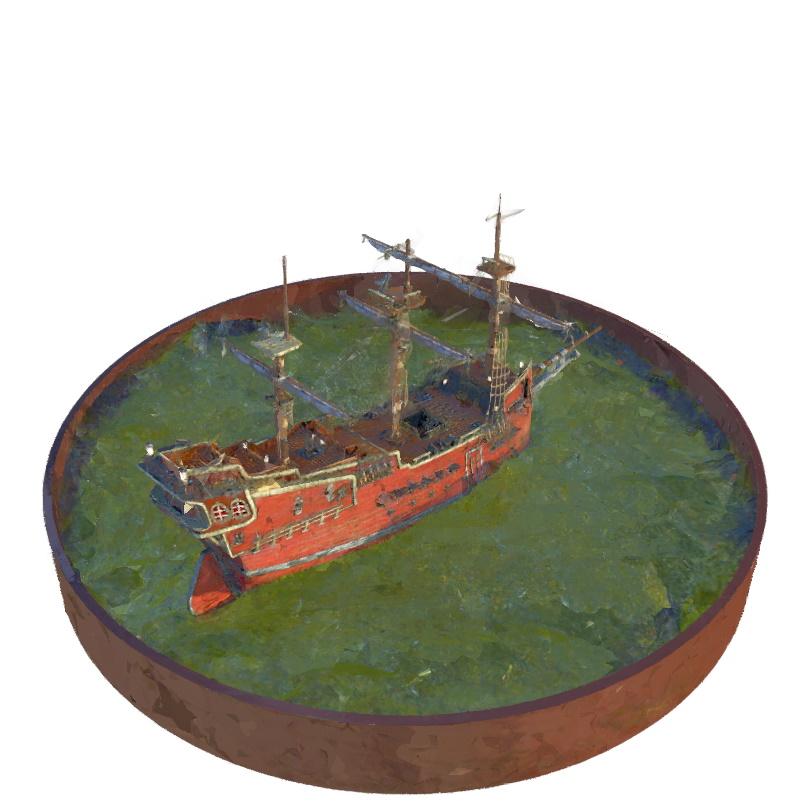}} &
			\raisebox{-0.5\height}{\includegraphics[trim={0cm 0cm 0cm 4cm},clip,width=\myfigsize]{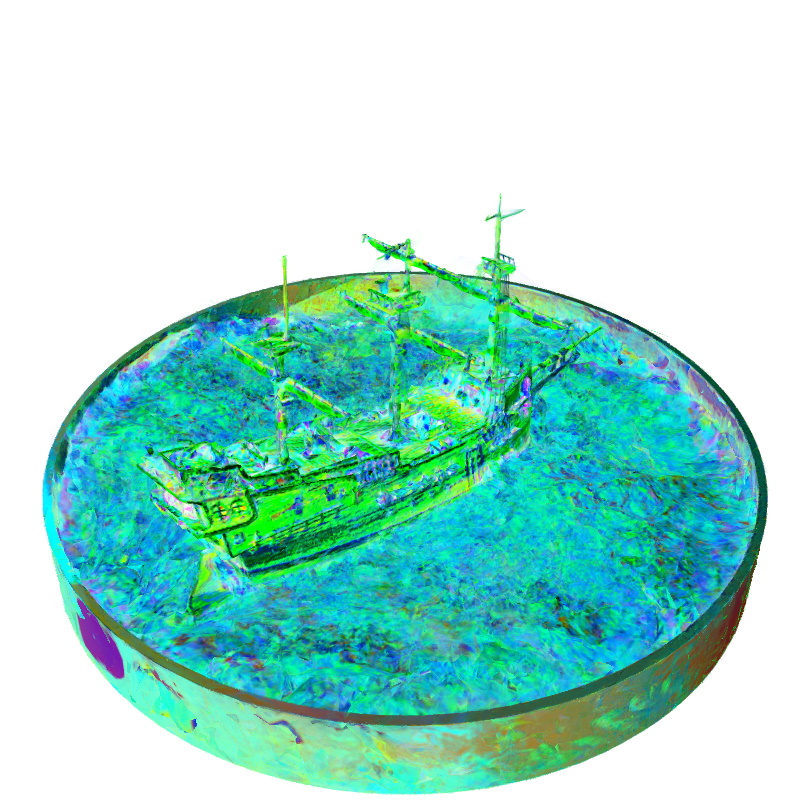}} &
			\raisebox{-0.5\height}{\includegraphics[trim={0cm 0cm 0cm 4cm},clip,width=\myfigsize]{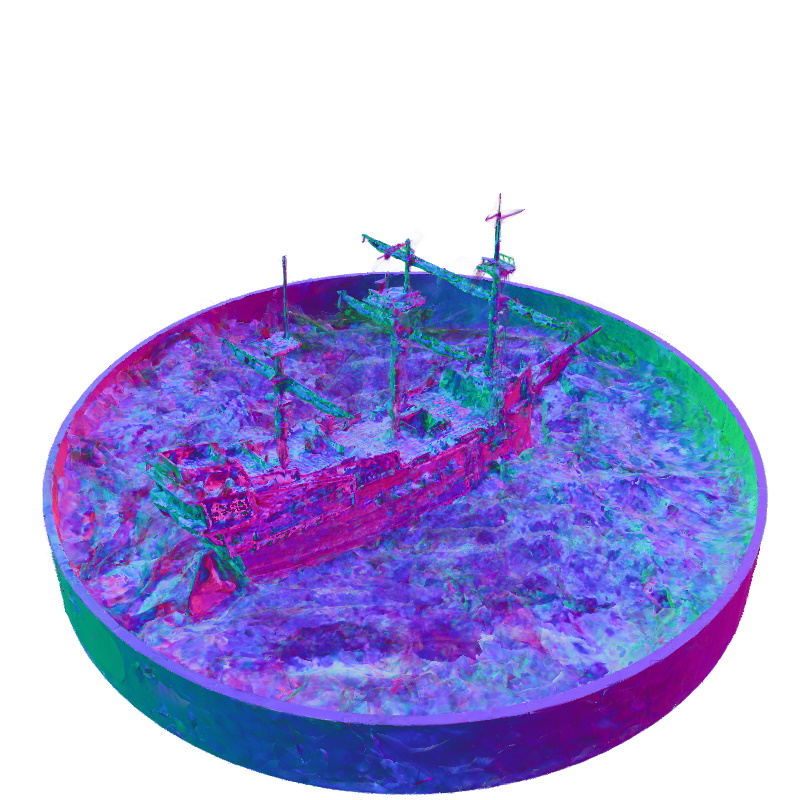}} &
			\raisebox{-0.5\height}{\includegraphics[width=\myfigsize]{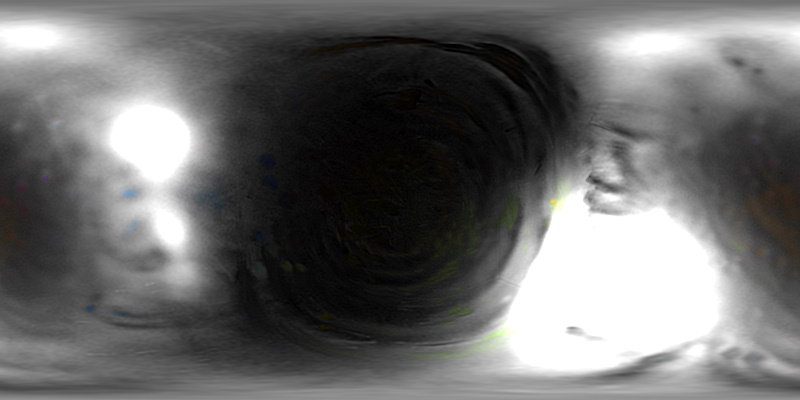}} \\

			& \small{Reference} & \small{Our} & \small{$\kd$} & \small{$\korm$} & \small{normals} & \small{HDR probe}
		\end{tabular}
		\caption{
			Our decomposition results on the NeRF Synthetic dataset. We show our rendered models alongside the material textures: 
			diffuse ($\kd$), roughness/metalness ($\korm$), the normals, and the extracted lighting. 
		}
		\label{fig:nerf_decomposition}
	\end{figure*}
}


\newcommand{\tabRelightNerfactor}{
\begin{table}[tb]
	\centering
	{\small
		\setlength{\tabcolsep}{2pt}
		\begin{tabularx}{\columnwidth}{l|YYY|YYY}
			& \multicolumn{3}{c|}{\textbf{Relighting}} & \multicolumn{3}{c}{$\kd$}  \\ 
			& PSNR$\uparrow$ & SSIM$\uparrow$ & LPIPS$\downarrow$ & PSNR$\uparrow$ & SSIM$\uparrow$ & LPIPS$\downarrow$  \\ \hline
			NeRFactor   & 23.78          & 0.907          & 0.112  & 23.11          & 0.917          & 0.094             \\
			Our         & 24.53          & 0.914          & 0.085  & 24.75          & 0.924          & 0.092             \\
			\hline
		\end{tabularx}
	}
	\vspace{-2mm}
	\caption{
		Relighting quality for NeRFactor's synthetic dataset. The reported image metrics are the 
		arithmetic mean over 8 validation views and 8 light probes for all 4 test scenes. We also show 
		metrics for the $\kd$ (albedo) textures. 
		Following NeRFactor, we note that the scale factor between material and 
		lighting is indeterminable and therefore normalize albedo images by the average intensity of the 
		reference before measuring errors. 
	}
	\vspace{-2mm}
	\label{tab:nerfactor_relight_avg}
\end{table}
}

\newcommand{\tabRelightNerfactorBreakdown}{
	\begin{table}[tb]
		\centering
		{\small
		\begin{tabularx}{\columnwidth}{l|YYYYY}
			\multicolumn{6}{c}{PSNR$\uparrow$} \\
			Scene       & Drums       & Ficus       & Hotdog      & Lego        & Avg    \\ \hline
			NeRFactor   & 21.94       & 22.35       & 25.59       & 25.25       & 23.82  \\
			Our         & 22.63       & 25.71       & 28.77       & 21.03       & 25.24  \\
			\hline
			\multicolumn{6}{c}{SSIM$\uparrow$} \\
			Scene       & Drums       & Ficus       & Hotdog      & Lego         & Avg   \\ \hline
			NerFactor   & 0.912       & 0.930       & 0.917       & 0.870        & 0.907 \\
			Our         & 0.915       & 0.961       & 0.932       & 0.846        & 0.911 \\
			\hline 
			\multicolumn{6}{c}{LPIPS$\downarrow$} \\
			Scene       & Drums       & Ficus       & Hotdog      & Lego         & Avg   \\ \hline
			NeRFactor   & 0.092       & 0.095       & 0.129       & 0.132        & 0.112 \\
			Our         & 0.083       & 0.046       & 0.092       & 0.119        & 0.085 \\
			\hline
			\multicolumn{6}{c}{FLIP$\downarrow$} \\
			Scene       & Drums       & Ficus       & Hotdog      & Lego         & Avg   \\ \hline
			NeRFactor   & 0.083       & 0.081       & 0.110       & 0.095        & 0.092 \\
			Our         & 0.087       & 0.063       & 0.074       & 0.163        & 0.097 \\
			\hline
		\end{tabularx}
		}
		\caption{Relighting quality results for the four scenes in NeRFactor's synthetic 
			dataset. The reported metrics are the arithmetic mean over eight validation views relit with
			eight different light probes.}
		\label{tab:nerfactor_relight}
	\end{table}
}

\newcommand{\tabAlbedoNerfactorBreakdown}{
	\begin{table}[tb]
		\centering
		{\small
		\begin{tabularx}{\columnwidth}{l|YYYYY}
			\multicolumn{6}{c}{PSNR$\uparrow$} \\
			Scene       & Drums       & Ficus       & Hotdog      & Lego        & Avg    \\ \hline
			NeRFactor   & 21.70       & 20.26       & 26.90       & 23.58       & 23.11  \\
			Our         & 20.22       & 30.33       & 27.26       & 21.17       & 24.75  \\
			\hline
			\multicolumn{6}{c}{} \\	\multicolumn{6}{c}{SSIM$\uparrow$} \\
			Scene       & Drums       & Ficus       & Hotdog      & Lego         & Avg   \\ \hline
			NeRFactor   & 0.898       & 0.930       & 0.932       & 0.906        & 0.917 \\
			Our         & 0.899       & 0.975       & 0.945       & 0.877        & 0.924 \\
			\hline 
			\multicolumn{6}{c}{} \\	\multicolumn{6}{c}{LPIPS$\downarrow$} \\
			Scene       & Drums       & Ficus       & Hotdog      & Lego         & Avg   \\ \hline
			NeRFactor   & 0.088       & 0.086       & 0.094       & 0.106        & 0.094 \\
			Our         & 0.095       & 0.037       & 0.088       & 0.149        & 0.092 \\
			\hline
		\end{tabularx}
		}
		\caption{
		}
		\label{tab:nerfactor_relight}
	\end{table}
}


\newcommand{\tabViewInterp}{
\begin{table}[tb]
	\centering
	{\small
		\begin{tabularx}{\columnwidth}{lYYY}
			Method      & PSNR$\uparrow$ & SSIM$\uparrow$ & LPIPS$\downarrow$ \\ \hline
			PhySG       & 18.91          & 0.847          & 0.182             \\
			NeRF        & 31.00          & 0.947          & 0.081             \\
			Mip-NeRF    & 33.05          & 0.961          & 0.067             \\
			Our         & 29.05          & 0.939          & 0.081             \\
			\hline
		\end{tabularx}
	}
	\vspace{-2mm}
	\caption{
		Average results for the eight scenes in the NeRF realistic synthetic dataset. Each scene consists of 100 
		training images, and 200 test images, with masks and known camera poses. 
		Results from NeRF are taken from Table~4 of the NeRF paper~{\protect\cite{Mildenhall2020}}. 
		PhySG and Mip-NeRF were retrained using public source code. 
	}
	\vspace{-2mm}
	\label{tab:view_interpolation_avg}
\end{table}
}

\newcommand{\tabViewInterpBreakdown}{
	\begin{table}[tb]
		\centering
		{\scriptsize
		\setlength{\tabcolsep}{2pt}		
		\begin{tabularx}{\columnwidth}{l|YYYYYYYYY}
			\multicolumn{10}{c}{PSNR$\uparrow$} \\
			Scene       & Chair       & Drums       & Ficus       & Hotdog      & Lego        & Mats.       & Mic         & Ship        & Avg         \\ \hline
			PhySG       & 21.87       & 16.45       & 17.40       & 21.57       & 18.81       & 18.02       & 19.16       & 18.06       & 18.91       \\
			NeRF        & 33.00       & 25.01       & 30.13       & 36.18       & 32.54       & 29.62       & 32.91       & 28.65       & 31.00       \\
			MipNeRF     & 35.08       & 25.56       & 33.44       & 37.38       & 35.46       & 30.63       & 36.38       & 30.46       & 33.05       \\
			Our         & 31.60       & 24.10       & 30.88       & 33.04       & 29.14       & 26.74       & 30.78       & 26.12       & 29.05       \\
			\hline
			\\
			\multicolumn{10}{c}{SSIM$\uparrow$} \\
			Scene       & Chair       & Drums       & Ficus       & Hotdog      & Lego        & Mats.       & Mic         & Ship        & Avg         \\ \hline
			PhySG       & 0.890       & 0.823       & 0.861       & 0.894       & 0.812       & 0.837       & 0.904       & 0.756       & 0.847       \\
			NeRF        & 0.967       & 0.925       & 0.964       & 0.974       & 0.961       & 0.949       & 0.980       & 0.856       & 0.947       \\
			MipNeRF     & 0.980       & 0.934       & 0.981       & 0.982       & 0.978       & 0.959       & 0.991       & 0.885       & 0.961       \\
			Our         & 0.969       & 0.916       & 0.970       & 0.973       & 0.949       & 0.923       & 0.977       & 0.833       & 0.939       \\
			\hline
			\\
			\multicolumn{10}{c}{LPIPS$\downarrow$} \\
			Scene       & Chair       & Drums       & Ficus       & Hotdog      & Lego        & Mats.       & Mic         & Ship        & Avg         \\ \hline
			PhySG       & 0.122       & 0.188       & 0.144       & 0.163       & 0.208       & 0.182       & 0.108       & 0.343       & 0.182       \\
			NeRF        & 0.046       & 0.091       & 0.044       & 0.121       & 0.050       & 0.063       & 0.028       & 0.206       & 0.081       \\
			MipNeRF     & 0.041       & 0.104       & 0.045       & 0.038       & 0.053       & 0.054       & 0.024       & 0.177       & 0.067       \\
			Our         & 0.045       & 0.101       & 0.048       & 0.060       & 0.061       & 0.100       & 0.040       & 0.191       & 0.081       \\
			\hline
			\\
			\multicolumn{10}{c}{FLIP$\downarrow$} \\
			Scene       & Chair       & Drums       & Ficus       & Hotdog      & Lego        & Mats.       & Mic         & Ship        & Avg         \\ \hline
			PhySG       & 0.088       & 0.140       & 0.115       & 0.109       & 0.139       & 0.139       & 0.069       & 0.159       & 0.119       \\
			MipNeRF     & 0.028       & 0.073       & 0.035       & 0.026       & 0.036       & 0.043       & 0.016       & 0.061       & 0.040       \\
			Our         & 0.034       & 0.065       & 0.041       & 0.033       & 0.042       & 0.060       & 0.024       & 0.080       & 0.047       \\
			\hline
		\end{tabularx}
		}
		\caption{
			Image quality metrics for the NeRF realistic synthetic dataset. Each training set consists of 100 images with masks and known camera poses,
			and the reported image metrics are the arithmetic mean over the 200 images in the test set. Results for NeRF are based on Table~4 of the 
			original paper~{\protect\cite{Mildenhall2020}}, with new measurements for PhySG and MipNeRF using their respective publicly available source 
			code. We additionally report FLIP mean scores~\cite{Andersson2020}.
			Note that the Hotdog outlier LPIPS score for NeRF is consistent with the original paper, but probably a bug.
		}
		\label{tab:view_interpolation}
	\end{table}
}


\newcommand{\tabViewInterpNerfactor}{
\begin{table}[tb]
	\centering
	{\small
		\begin{tabularx}{\columnwidth}{lYYY}
			Method      & PSNR$\uparrow$ & SSIM$\uparrow$ & LPIPS$\downarrow$ \\ \hline
			NeRF        & 31.08          & 0.956          & 0.064             \\
			NeRFactor   & 26.87          & 0.930          & 0.099             \\
			Our         & 31.65          & 0.967          & 0.054             \\
			\hline
		\end{tabularx}
	}
	\vspace{-2mm}
	\caption{
		View interpolation error metrics on NeRFactor's variant of the NeRF synthetic dataset. 
		The reported image metrics are the arithmetic mean over the eight validation 
		images of all four scenes.}
	\vspace{-2mm}
	
	\label{tab:nerfactor_avg}
\end{table}
}

\newcommand{\tabViewInterpNerfactorBreakdown}{
\begin{table}[tb]
	\centering
	{\small
	\begin{tabularx}{\columnwidth}{l|YYYYY}
		\multicolumn{6}{c}{PSNR$\uparrow$} \\
		Scene       & Drums       & Ficus       & Hotdog      & Lego        & Avg    \\ \hline
		PhySG       & 14.35       & 15.25       & 24.49       & 17.10       & 17.80  \\
		NeRF        & 27.67       & 28.05       & 36.71       & 31.89       & 31.08  \\
		NeRFactor   & 24.63       & 23.14       & 31.60       & 28.12       & 26.87  \\
		Our         & 28.45       & 31.20       & 36.26       & 30.70       & 31.65  \\
		\hline
		\\
		\multicolumn{6}{c}{SSIM$\uparrow$} \\
		Scene       & Drums       & Ficus       & Hotdog      & Lego         & Avg   \\ \hline
		PhySG       & 0.807       & 0.838       & 0.909       & 0.771        & 0.831 \\
		NeRF        & 0.951       & 0.957       & 0.971       & 0.944        & 0.956 \\
		NeRFactor   & 0.933       & 0.937       & 0.948       & 0.900        & 0.930 \\
		Our         & 0.959       & 0.978       & 0.981       & 0.951        & 0.967 \\
		\hline
		\\
		\multicolumn{6}{c}{LPIPS$\downarrow$} \\
		Scene       & Drums       & Ficus       & Hotdog      & Lego         & Avg   \\ \hline
		PhySG       & 0.215       & 0.176       & 0.153       & 0.278        & 0.206 \\
		NeRF        & 0.069       & 0.055       & 0.058       & 0.075        & 0.064 \\
		NeRFactor   & 0.082       & 0.087       & 0.101       & 0.124        & 0.099 \\
		Our         & 0.063       & 0.047       & 0.048       & 0.057        & 0.054 \\
		\hline
		\\
		\multicolumn{6}{c}{FLIP$\downarrow$} \\
		Scene       & Drums       & Ficus       & Hotdog      & Lego         & Avg   \\ \hline
		PhySG       & 0.163       & 0.133       & 0.076       & 0.168        & 0.135 \\
		NeRF        & 0.045       & 0.045       & 0.030       & 0.037        & 0.039 \\
		NeRFactor   & 0.058       & 0.071       & 0.050       & 0.058        & 0.059 \\
		Our         & 0.037       & 0.037       & 0.023       & 0.030        & 0.032 \\
		\hline
	\end{tabularx}
	}
	\caption{
		View interpolation results for the four scenes of NeRFactor's synthetic dataset. 
		The NeRF column shows the baseline NeRF trained as part of NeRFactor's setup, 
		and is different from the NeRF in our other view interpolation results. 
		Each training set consists of 100 images with masks and known camera poses,	and the reported image metrics 
		are the arithmetic mean over the eight images in the test set. 
	}
	\label{tab:nerfactor}
\end{table}
}

\newcommand{\tabViewInterpChamfer}{
\begin{table}[tb]
	\centering
	{\scriptsize
	\setlength{\tabcolsep}{2pt}		
	\begin{tabularx}{\columnwidth}{l|YYYYYYYYY}
		\multicolumn{9}{c}{Chamfer Loss$\downarrow$} \\
		Scene           & Chair       & Drums       & Ficus       & Hotdog      & Lego        & Mats.       & Mic         & Ship    \\ \hline
		PhySG           & 0.1341      & 0.4236      & 0.0937      & 0.2420      & 0.2592      & -           & 0.2712      & 0.7118  \\
		NeRF (w/o mask) & 0.0185      & 0.0536      & 0.0115      & 4.6010      & 0.0184      & 0.0057      & 0.0124      & 2.0111  \\
		NeRF (w/  mask) & 0.0435      & 0.0326      & 0.0145      & 0.0436      & 0.0201      & 0.0082      & 0.0122      & 0.2931  \\
		Our             & 0.0574      & 0.0325      & 0.0154      & 0.0272      & 0.0267      & 0.0180      & 0.0098      & 0.3930  \\
		\hline
		\\
		\multicolumn{9}{c}{Triangles$\downarrow$ (Thousands)} \\
		Scene           & Chair       & Drums       & Ficus       & Hotdog      & Lego        & Mats.       & Mic         & Ship  \\ \hline
		PhySG           & 353         & 439         & 489         & 725         & 498         & -           & 386         & 557   \\
		NeRF (w/o mask) & 192         & 261         & 585         & 869         & 2259        & 2411        & 261         & 1087  \\
		NeRF (w/  mask) & 494         & 548         & 440         & 694         & 1106        & 594         & 307         & 3500  \\
		Our             & 102         &  65         & 39          & 57          & 111         & 58          & 22          & 190   \\
		\hline
	\end{tabularx}
	}
	\caption{Chamfer loss and triangle counts for reconstructed meshes for the NeRF realistic synthetic dataset. We compare to the meshes produced
		by PhySG, and also generate meshes from the NeRF volume using density thresholding and marching cubes. Note that we primarily focus on 
		opaque geometry, so the \textsc{Drums}, \textsc{Ship}, and \textsc{Ficus} scenes with transparency are challenging cases.}
	\label{tab:view_interpolation_chamfer}
\end{table}
}

\newcommand{\tabDTUChamfer}{
\begin{table}[tb]
	\centering
	{\small
	\begin{tabularx}{\columnwidth}{l|YYYYY}
		\multicolumn{6}{c}{Chamfer loss$\downarrow$} \\
		Scene   & NeRF & DVR  & Our  & IDR  & NeuS \\ \hline
		scan65  & 1.44 & 1.06 & 1.03 & 0.79 & 0.72 \\
		scan106 & 1.44 & 0.95 & 1.07 & 0.67 & 0.66 \\
		scan118 & 1.13 & 0.71 & 0.69 & 0.51 & 0.51 \\ 
		\hline     
	\end{tabularx}
	}
	\caption{
		Quantitative evaluation on the DTU dataset w/ mask. Chamfer distances are measured
		in the same way as NeuS~\cite{Wang2021neus}, IDR~\cite{yariv2020multiview}, and DVR~\cite{Niemeyer2020CVPR}. 
		Results for NeRF, IDR and NeuS are taken from Table~1 in the NeuS paper~\cite{Wang2021neus}, 
		and the DVR results are taken from Table 8, 9 and 10 in the DVR supplemental material. We also reevaluated the 
		DVR scores using the DTU MVS dataset evaluation scripts~\cite{Jensen2014} to verify the evaluation pipeline. 
		Our Chamfer distances are lower than NeRF, roughly on par with DVR, but higher than the current state-of-the-art 
		(IDR/NeuS). Still, we find these results encouraging, considering that we provide an explicit mesh with factorized
		materials.
	}
	\label{tab:dtu_chamfer}
\end{table}
}


\newcommand{\figGoldMaskErrors}{
\begin{figure}[tb]
	\centering
	\setlength{\tabcolsep}{1pt}
	\begin{tabular}{cccc}
		\includegraphics[width=0.23\columnwidth]{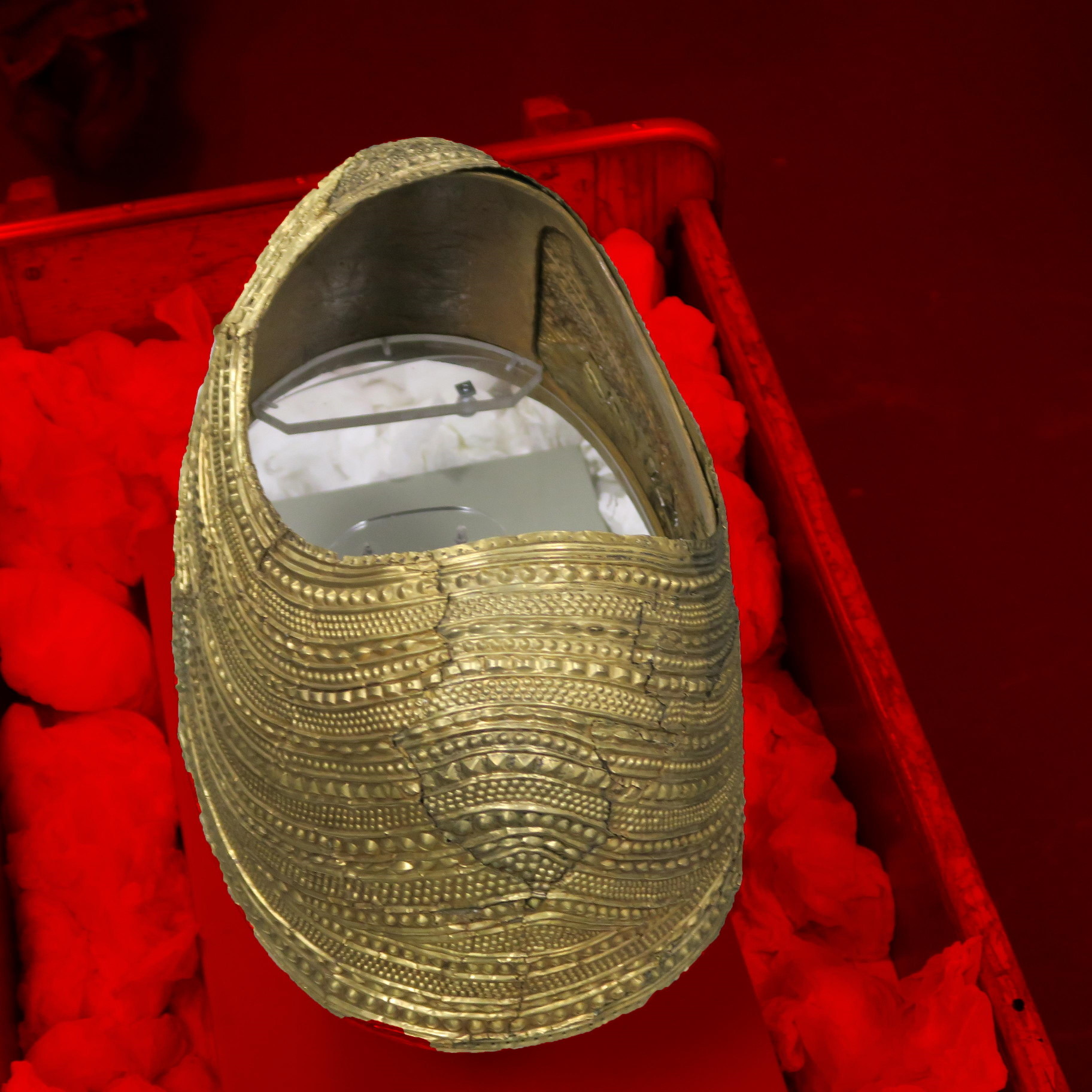} & 
		\includegraphics[width=0.23\columnwidth]{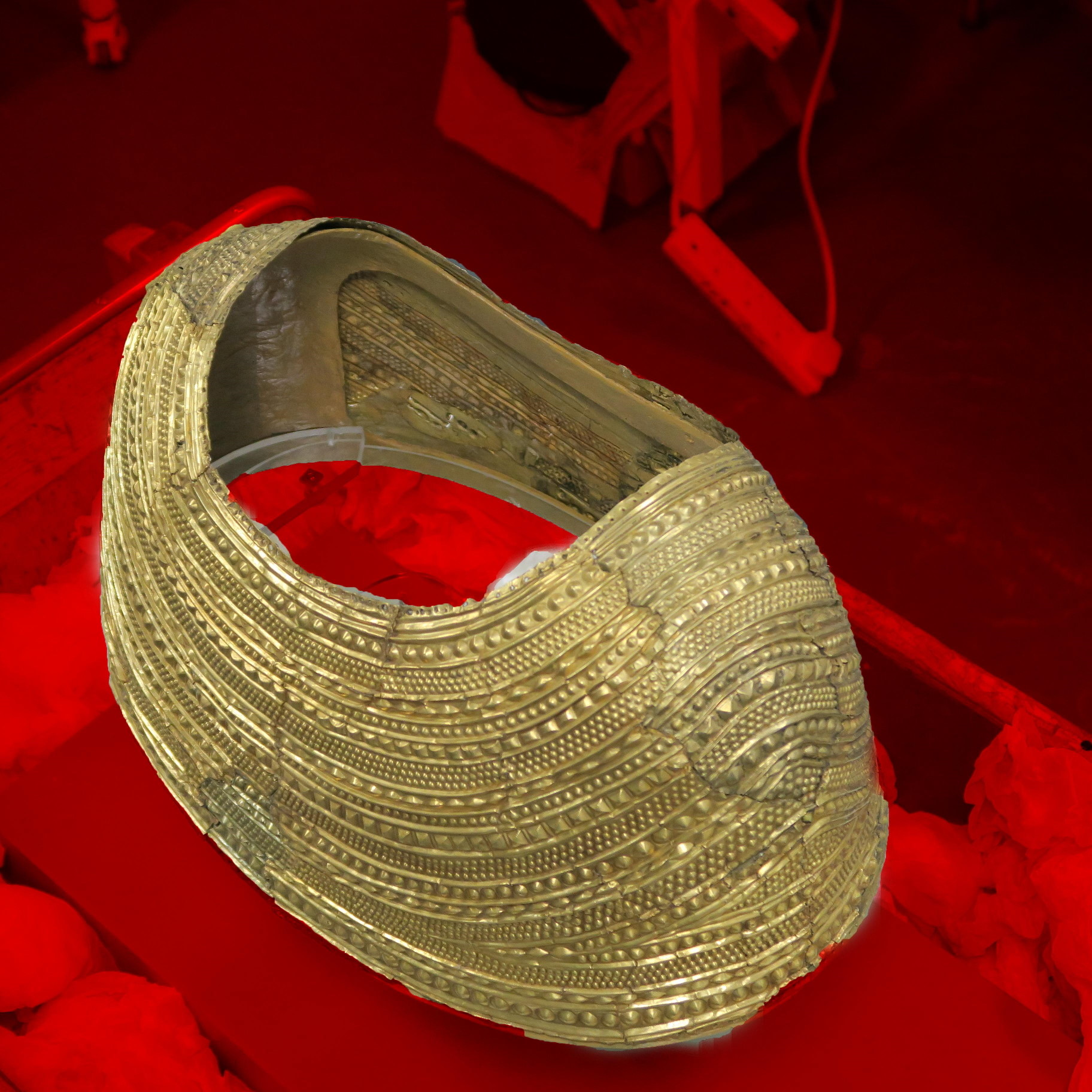} &
		\includegraphics[width=0.23\columnwidth]{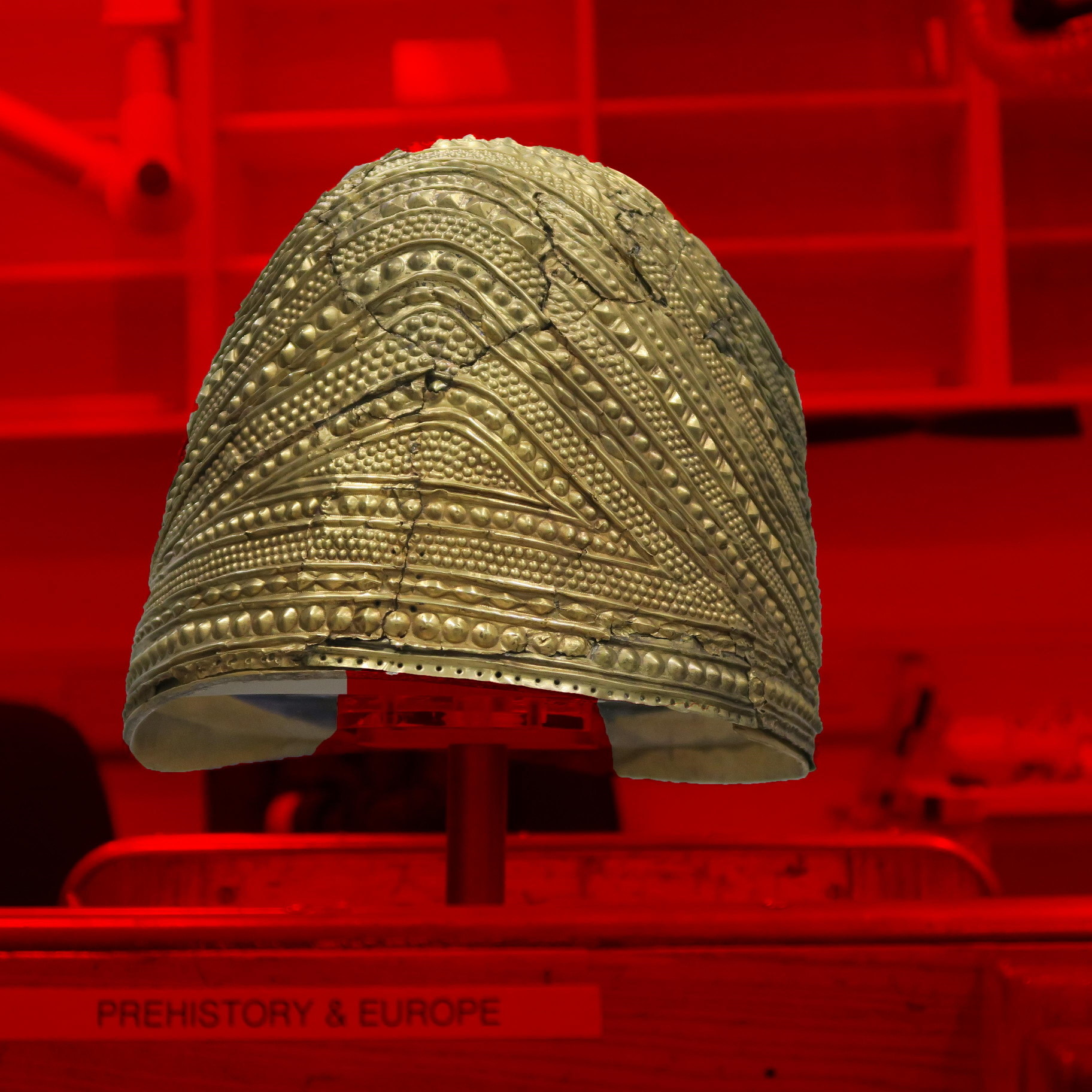} &
		\includegraphics[width=0.23\columnwidth]{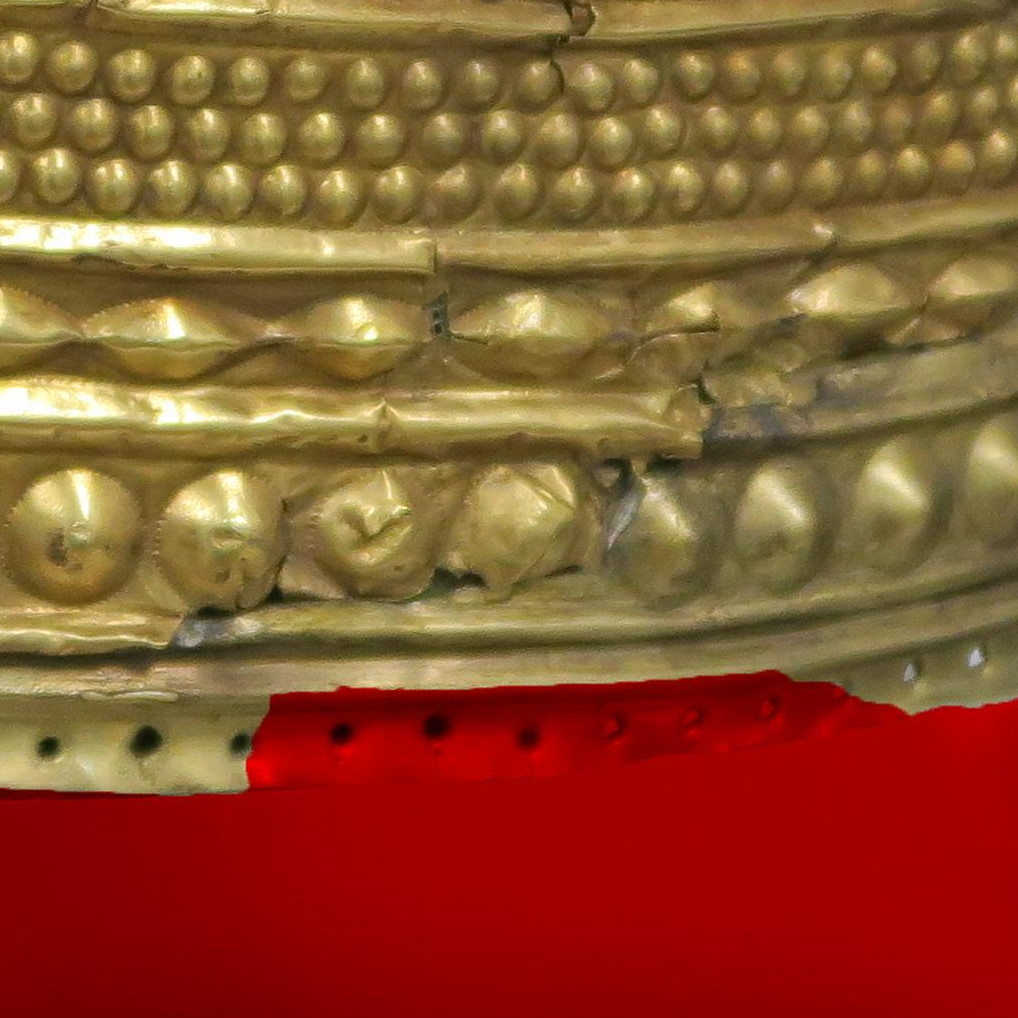}
	\end{tabular}
	\vspace*{-3mm}
	\caption{Examples of masking errors for the \emph{Mold Gold Cape} dataset. Note the inconsistencies in 
		classifying the plastic mount as both part of the object and background. 
	}
	\label{fig:gold_mask_errors}
\end{figure}
}

\newcommand{\figMaskTechniques}{
\begin{figure}[tb]
	\centering
	\setlength{\tabcolsep}{1pt}
	{\small
		\begin{tabular}{cccc}
			\includegraphics[width=0.24\columnwidth]{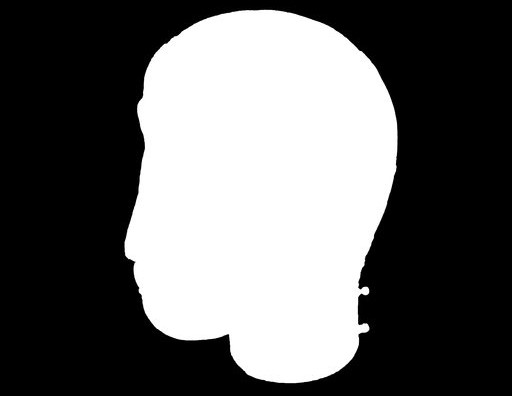} &
			\includegraphics[width=0.24\columnwidth]{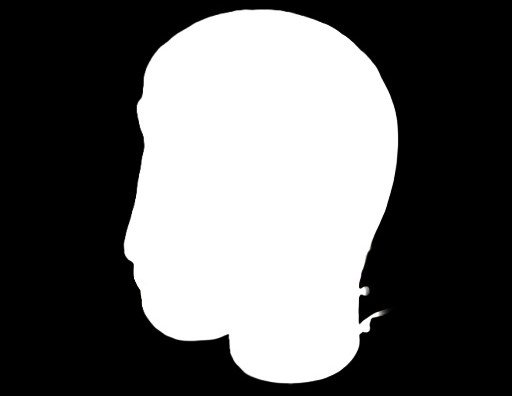} &
			\includegraphics[width=0.24\columnwidth]{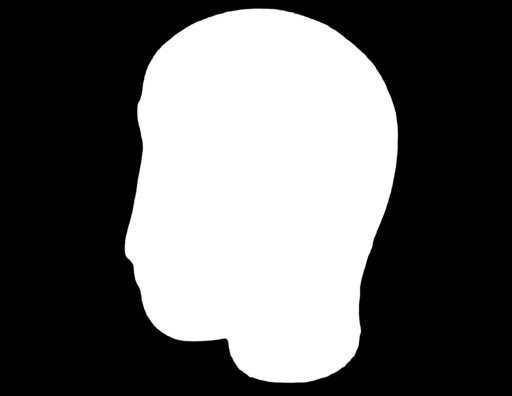} &
			\includegraphics[width=0.24\columnwidth]{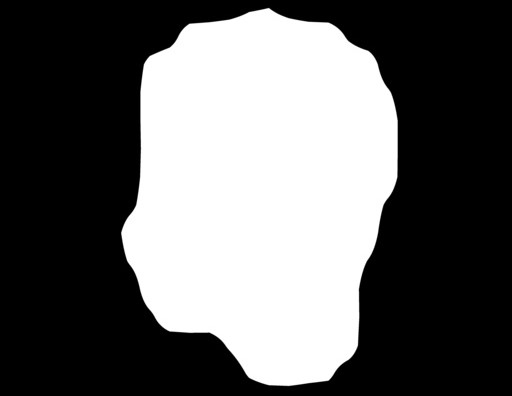} \\
			\includegraphics[width=0.24\columnwidth]{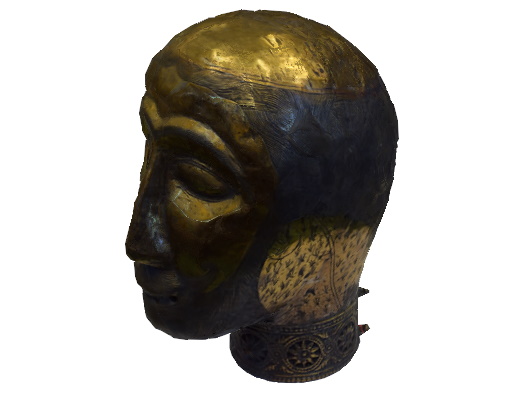} &
			\includegraphics[width=0.24\columnwidth]{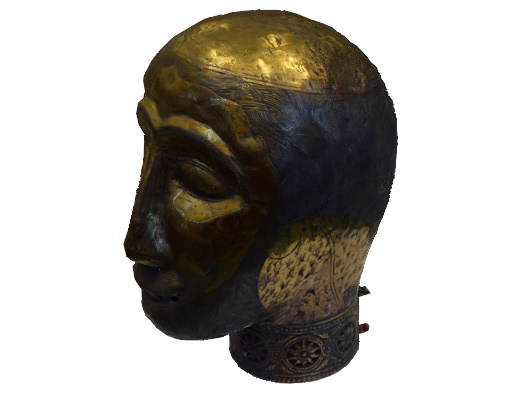} &
			\includegraphics[width=0.24\columnwidth]{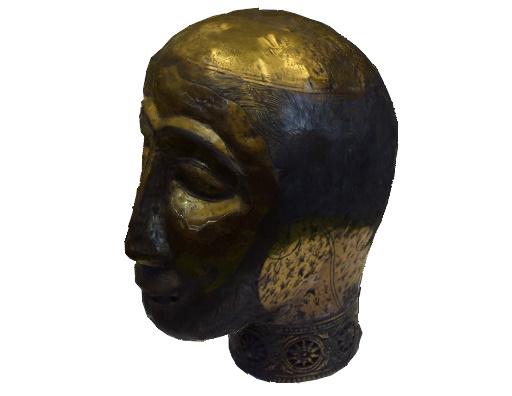} &
			\includegraphics[width=0.24\columnwidth]{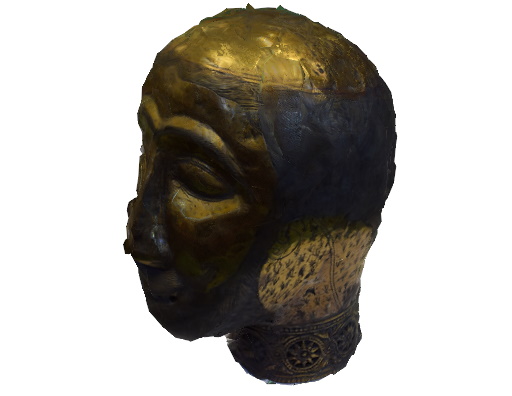} \\
			Manual & Photoshop & D2: PointRend & D2: M-RCNN \\
			26.55~dB & 25.94~dB & 25.31~dB & 22.83~dB
		\end{tabular}
	}
	\vspace*{-2mm}
	\caption{We compare four segmentation techniques for the \emph{Ethiopian Head} dataset:
		The original (crowdsourced) masks, automatic segmentation in Photoshop (using the object selection tool), 
		and two versions of Detectron2.	For Detectron2, we run two pretrained instance segmentation models which 
		predict accurate and coarse segmentations respectively. PSNR$\uparrow$ scores are the arithmetic mean of 
		all reconstructed frames.}
	\label{fig:mask_techniques}
\end{figure}
}

\newcommand{\figSyntheticMaskCorruption}{
\begin{figure}[t]
	\centering
	\setlength{\tabcolsep}{1pt}
	{\small
		\begin{tabular}{lcccc}
			&
			\includegraphics[width=0.23\columnwidth]{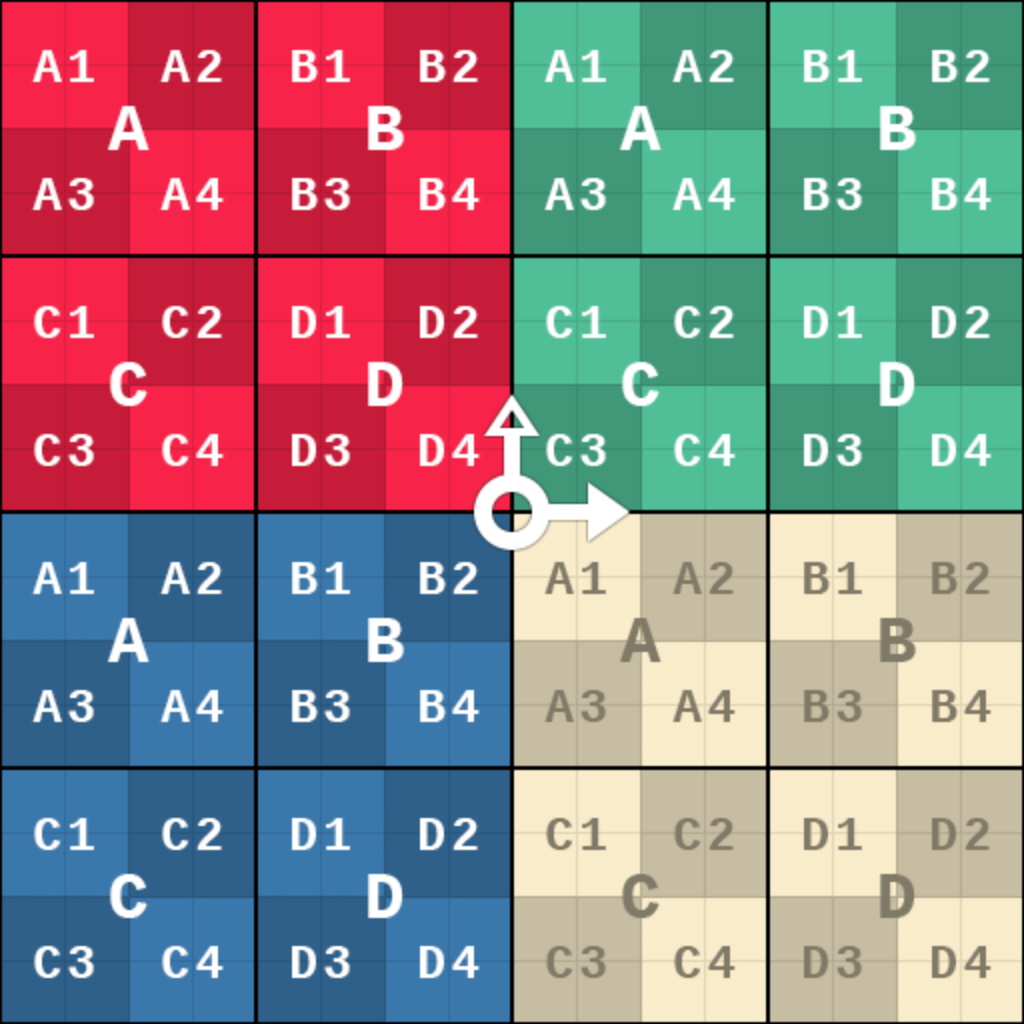} &
			\includegraphics[width=0.23\columnwidth]{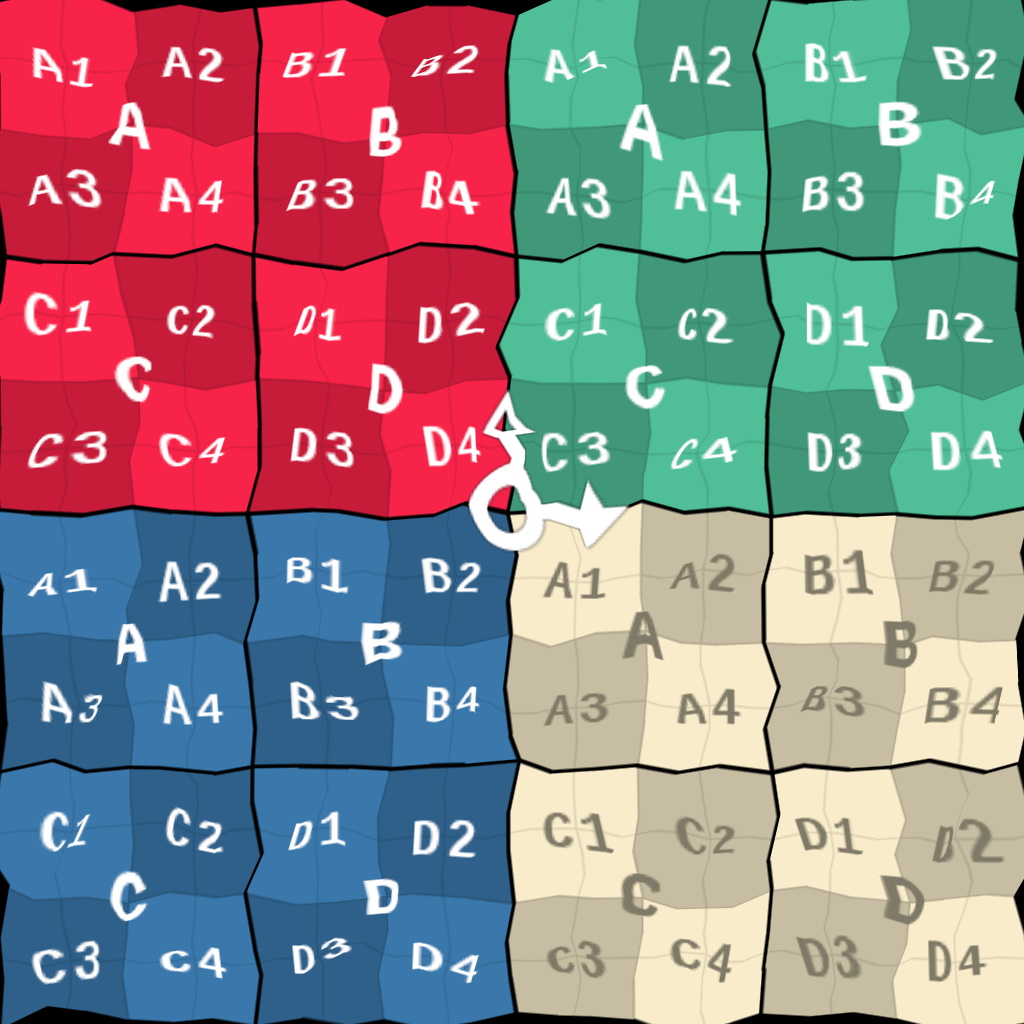} &
			\includegraphics[width=0.23\columnwidth]{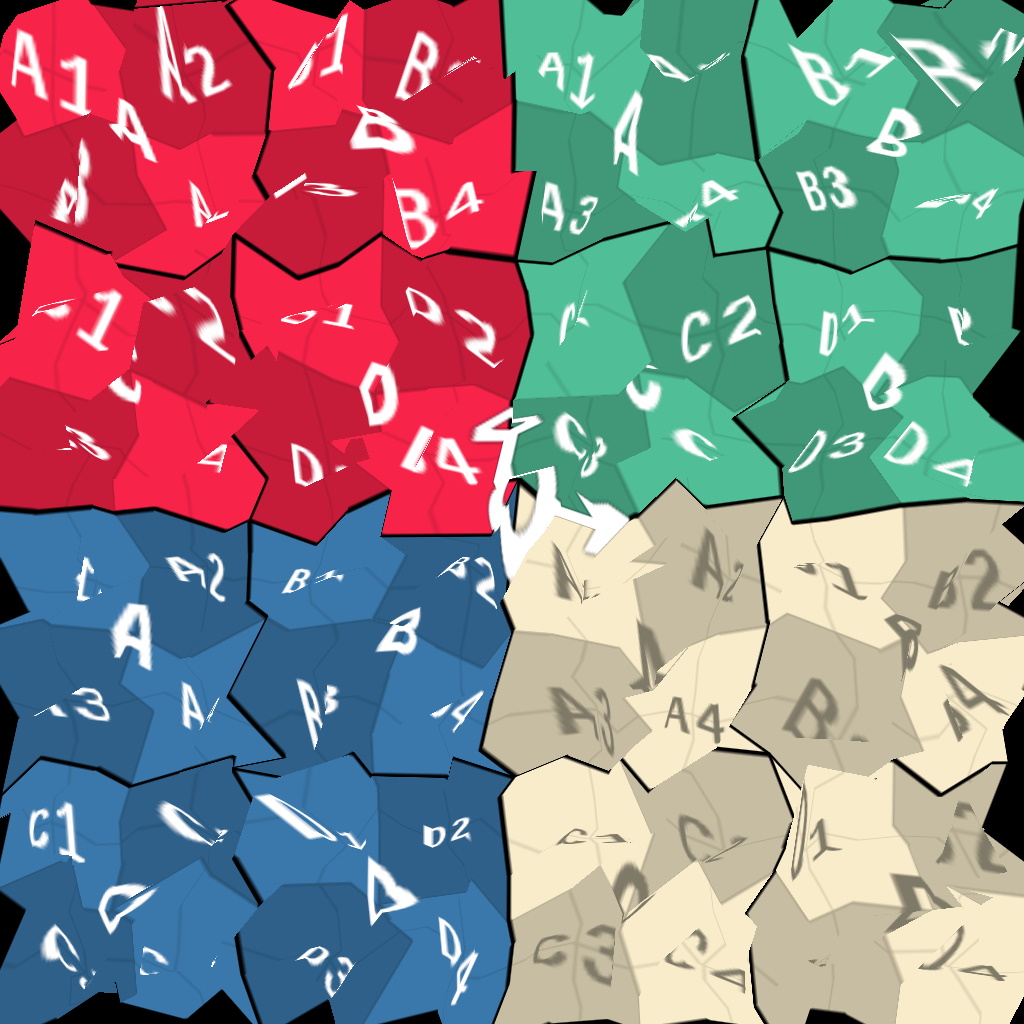} &
			\includegraphics[width=0.23\columnwidth]{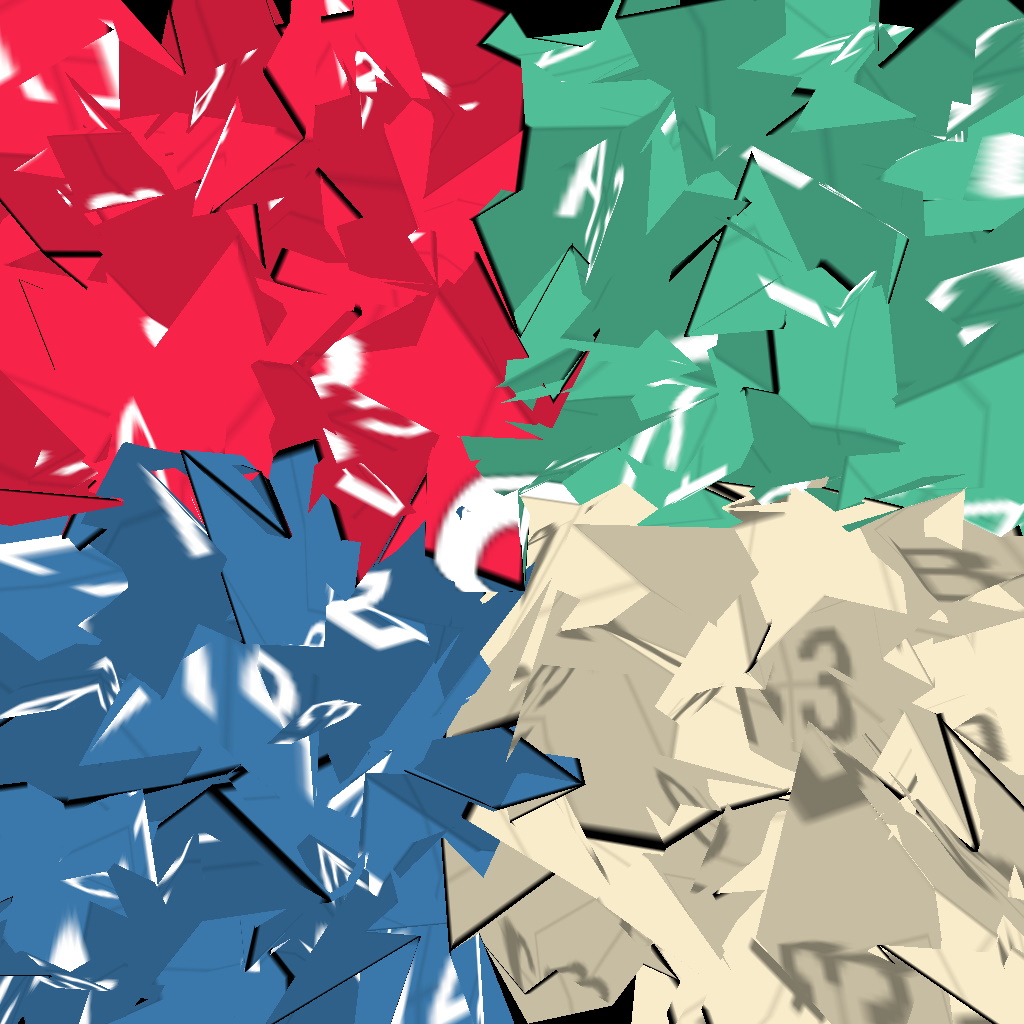} \\
			&
			\includegraphics[width=0.23\columnwidth]{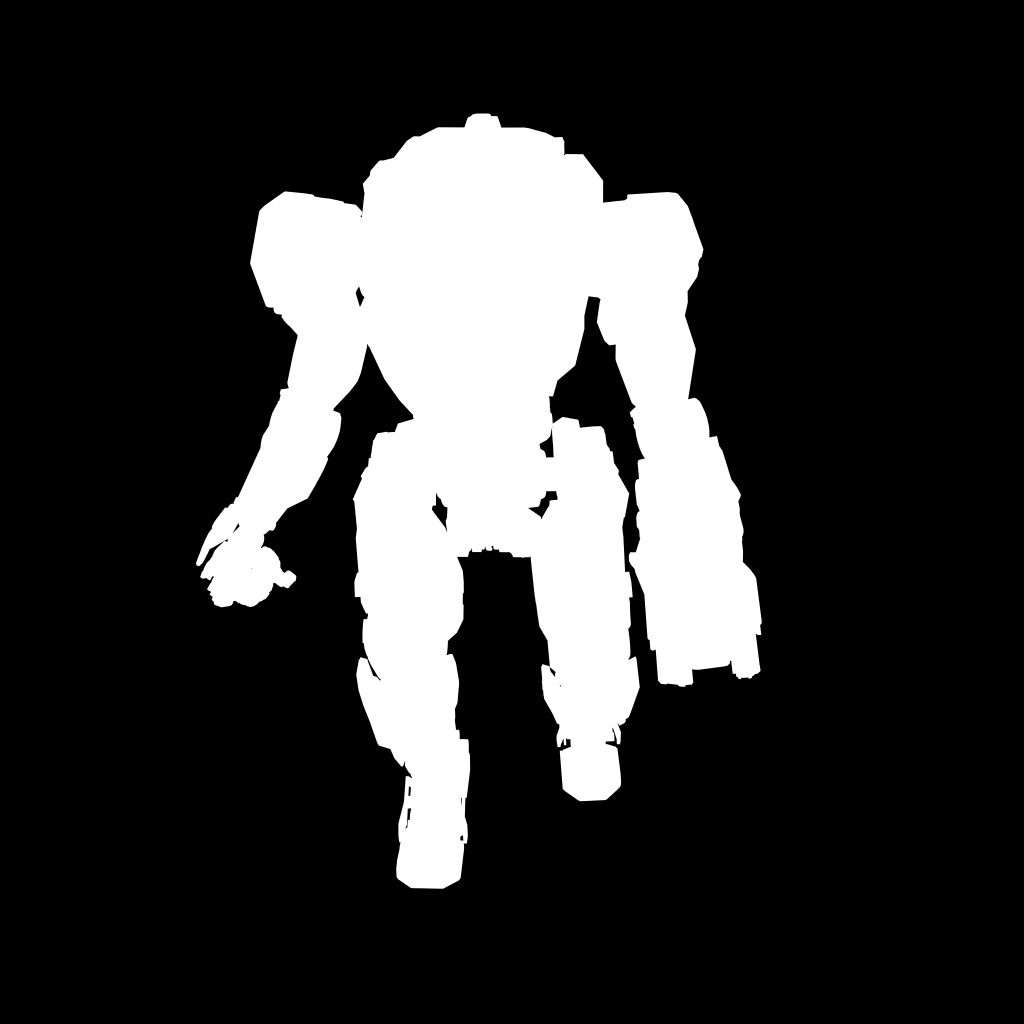} &
			\includegraphics[width=0.23\columnwidth]{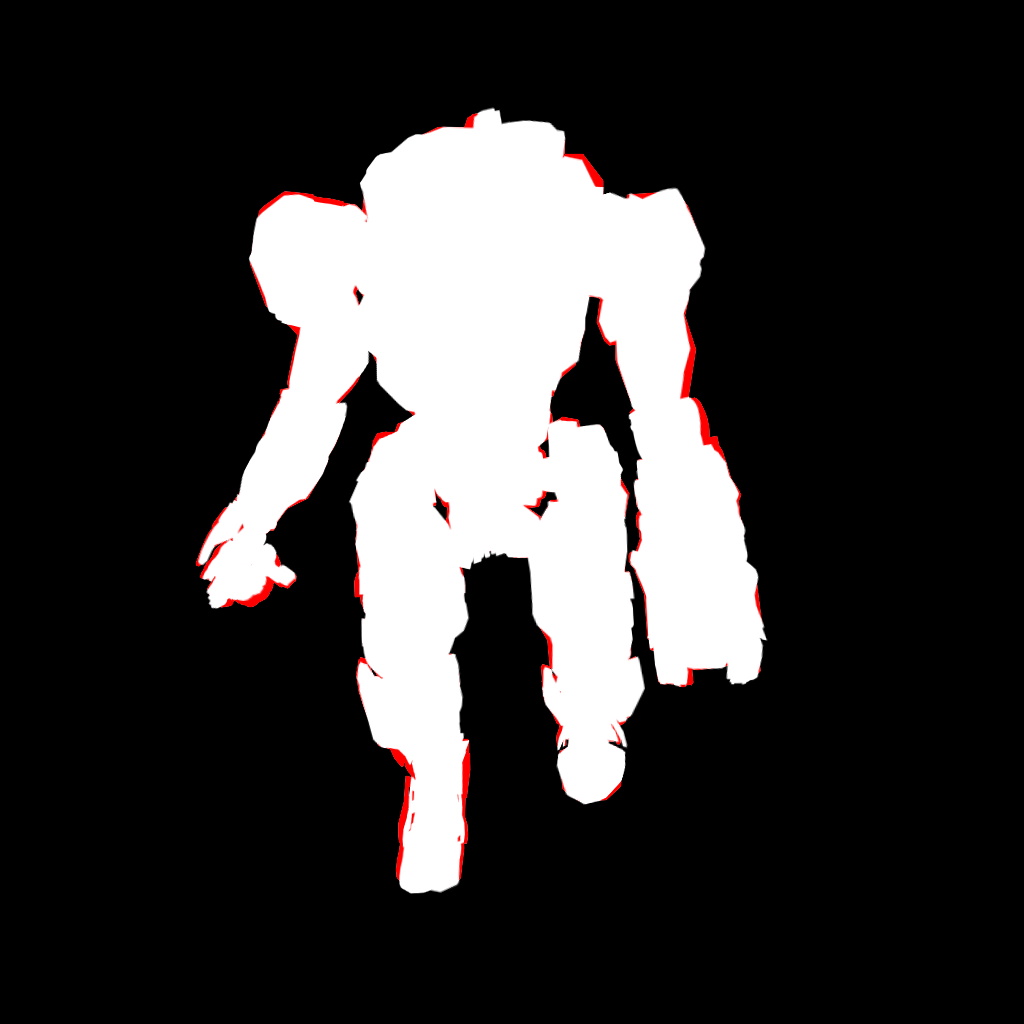} &
			\includegraphics[width=0.23\columnwidth]{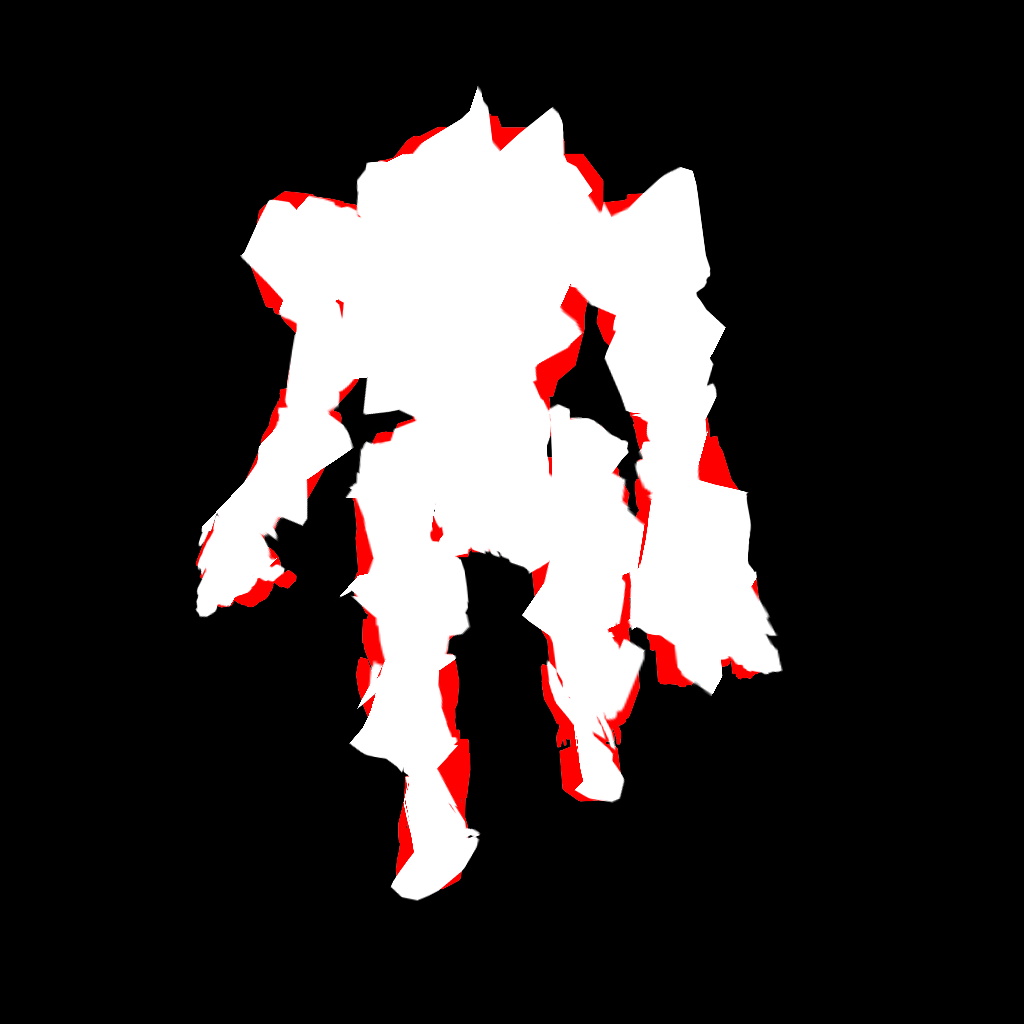} &
			\includegraphics[width=0.23\columnwidth]{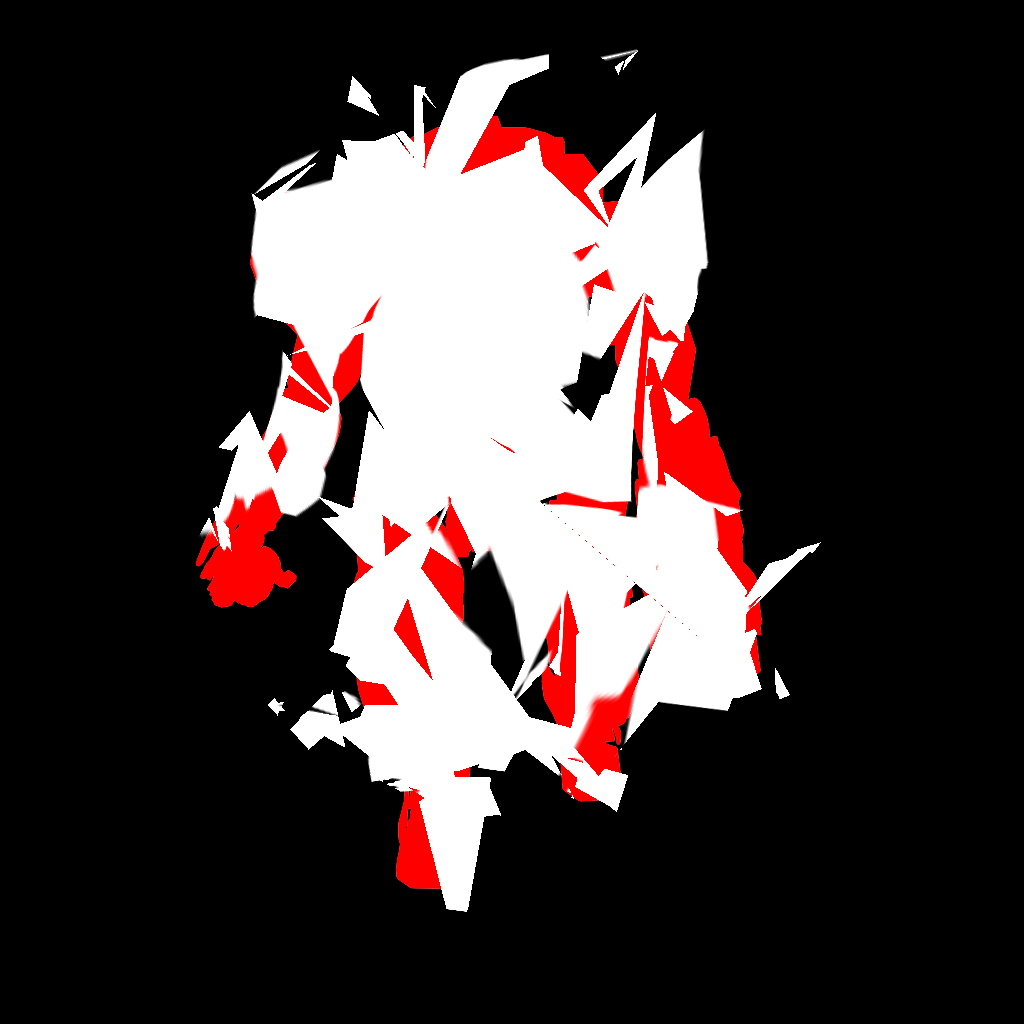} \\
			&
			\includegraphics[width=0.24\columnwidth]{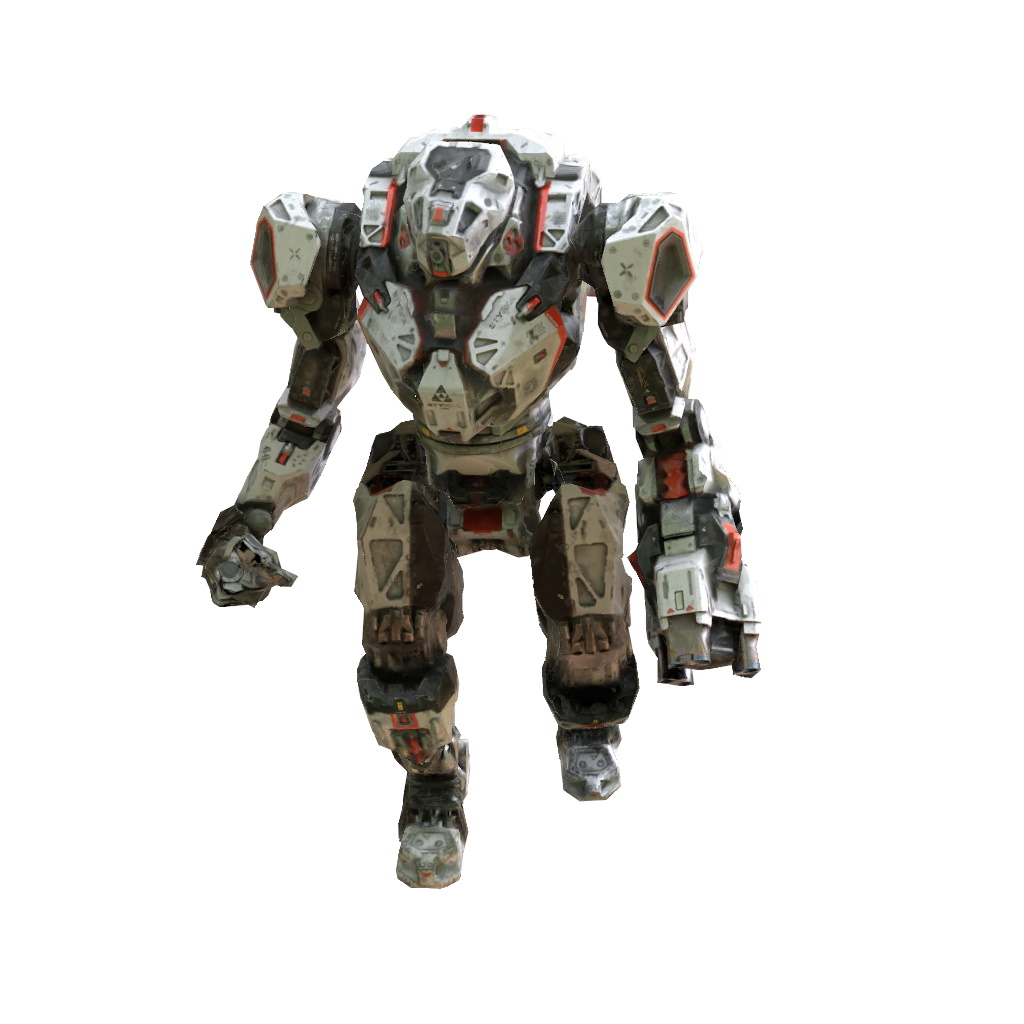} &
			\includegraphics[width=0.24\columnwidth]{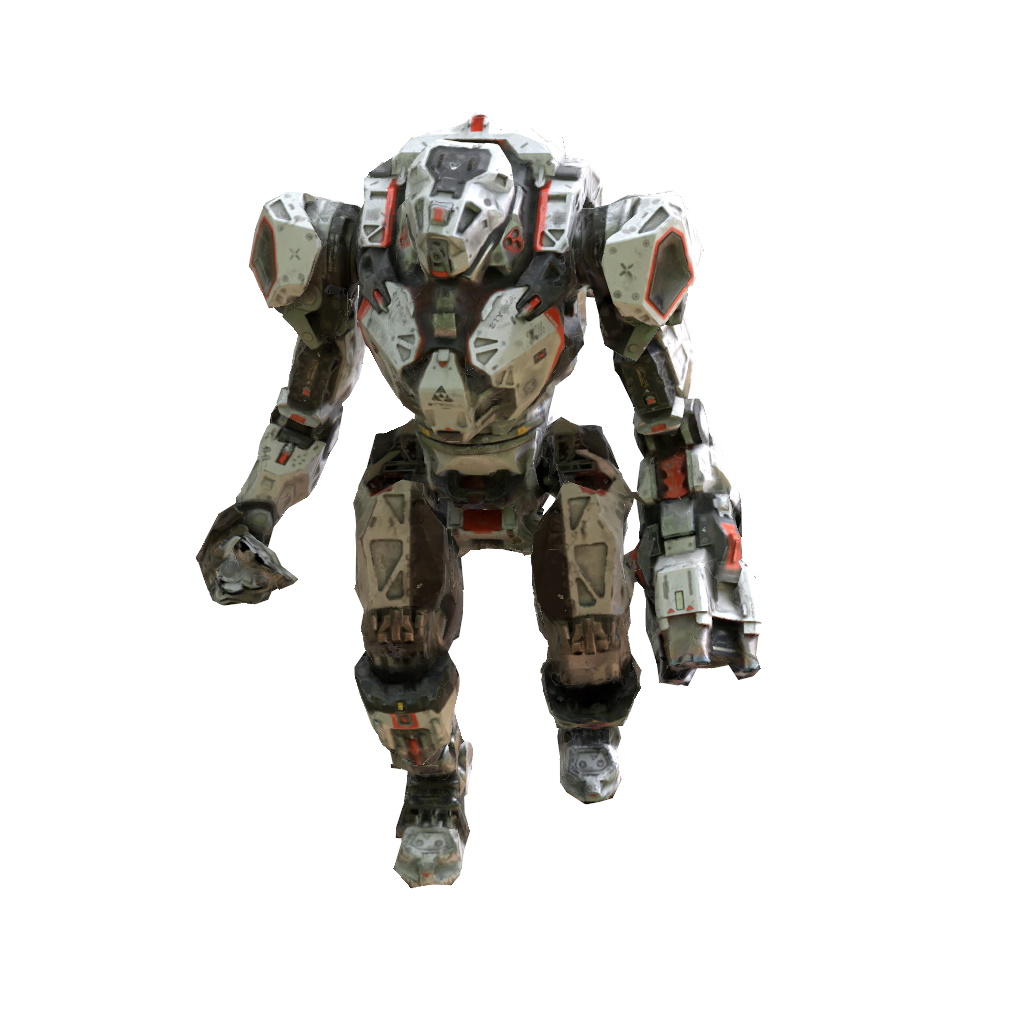} &
			\includegraphics[width=0.24\columnwidth]{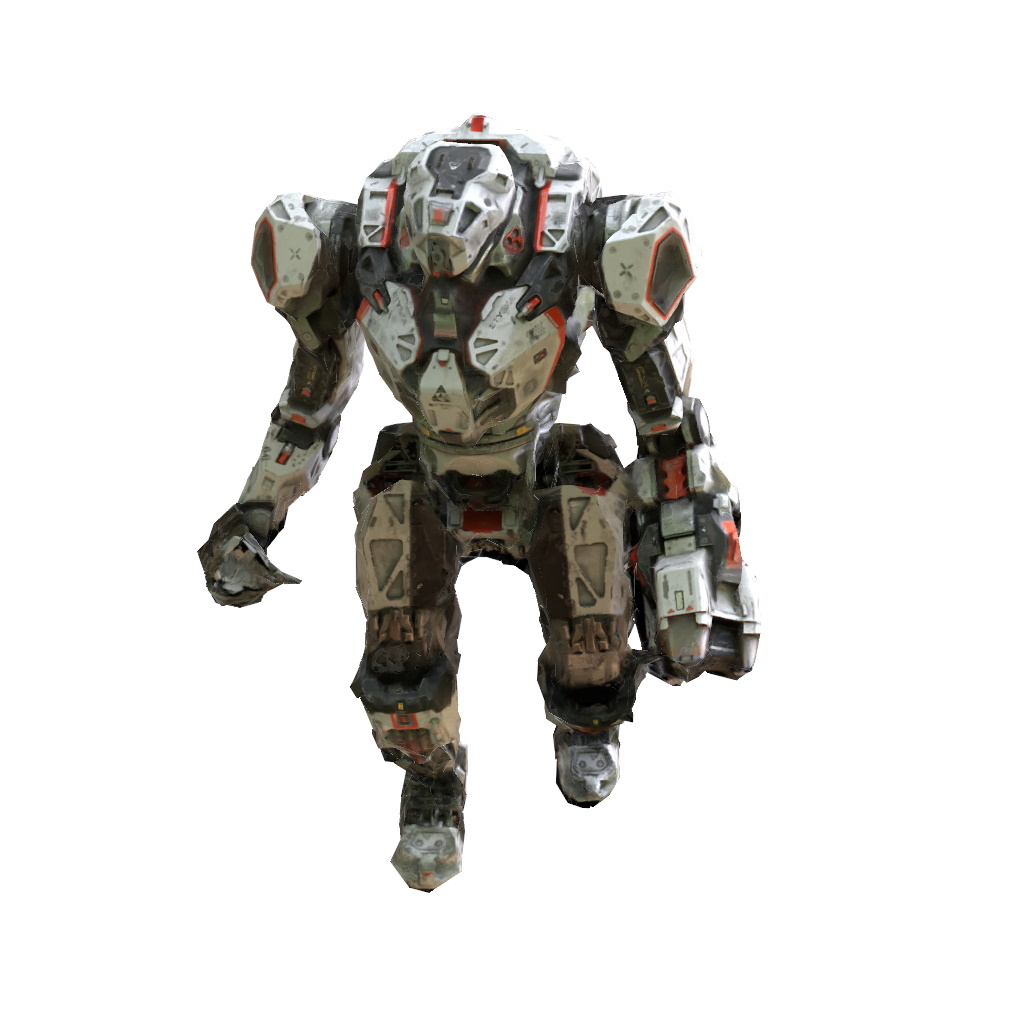} &
			\includegraphics[width=0.24\columnwidth]{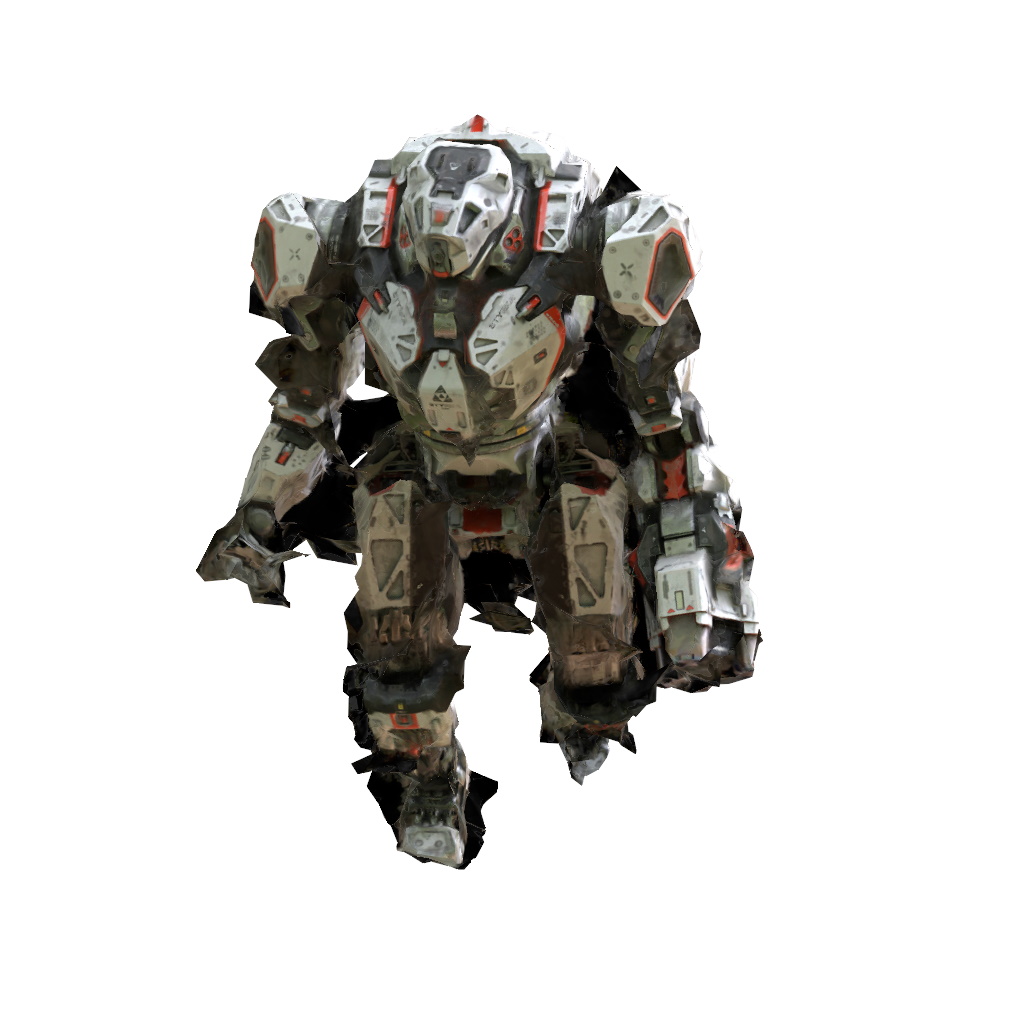} \\
			& $\sigma = 0$ & $\sigma = 5~\mathrm{px}$ & $\sigma = 16~\mathrm{px}$ & $\sigma = 51~\mathrm{px}$ \\
			\textbf{U} & 26.94~dB & 25.60~dB & 22.93~dB & 17.85~dB \\ 
			\textbf{C} & 26.94~dB & 24.36~dB & 20.09~dB & 16.08~dB  
		\end{tabular}
	}
	\vspace*{-2mm}
	\caption{To evaluate the impact of corrupted masks, we warp	perfect masks by texture-mapping them on a 
		grid, displacing each of the $25\times 25$ vertices by zero-mean Gaussian noise with increasing 
		standard deviation, $\sigma$. 
		From top to bottom, we show a warped texture (to give a sense of the magnitude of 
		corruption), the corrupted masks with the reference mask shown in red, and our 
		reconstruction.
		The training set consists of 200 images, and PSNR$\uparrow$ scores are computed as the arithmetic 
		mean of 50 validation images.
		The `uncorrelated` series, \textbf{U}, are generated with unique random numbers
		for each frame, while in the ``correlated'' scores, \textbf{C}, we corrupt all masks using the \emph{same} 
		random seed, simulating a segmentation with systematic bias.}
	\label{fig:synthetic_mask_corruption}
\end{figure}
}

\maketitle

\begin{abstract}
	\vspace{-2mm}

  We present an efficient method for joint optimization of topology, materials and lighting 
  from multi-view image observations. Unlike recent multi-view reconstruction approaches, 
  which typically produce entangled 3D representations encoded in neural networks, we output 
  triangle meshes with spatially-varying materials and environment 
  lighting that can be deployed in any traditional graphics engine unmodified.
  We leverage recent work in differentiable rendering, 
  coordinate-based networks to compactly represent volumetric texturing, 
  alongside differentiable marching tetrahedrons to enable gradient-based optimization directly 
  on the surface mesh. Finally, we introduce a differentiable formulation of the split sum 
  approximation of environment lighting to efficiently recover all-frequency lighting. 
  Experiments show our extracted models used in advanced scene editing, material decomposition, 
  and high quality view interpolation, all running at interactive rates in triangle-based 
  renderers (rasterizers and path tracers). 
  \blfootnote{Project page: \url{https://nvlabs.github.io/nvdiffrec/}}
\end{abstract}


\vspace{-3mm}
\section{Introduction}
\vspace{-0.5mm}

3D content creation is a challenging, mostly manual task which requires 
both artistic modeling skills and technical knowledge. Efforts to automate
3D modeling can save substantial production costs or allow for 
faster and more diverse content creation. Photogrammetry~\cite{Pollefeys2002,Snavely2006} is a popular technique to assist in 
this process, where multiple photos of an object
are converted into a 3D model. 
Game studios leverage photogrammetry to quickly build
highly detailed virtual landscapes~\cite{Embark2021}.
\figSceneEdit
However, this is a multi-stage process, including 
multi-view stereo~\cite{schoenberger2016mvs} to align cameras and find 
correspondences, geometric simplification, texture parameterization, 
material baking and delighting. 
This complex pipeline has many steps with conflicting optimization 
goals and errors that propagate between stages. Artists  
rely on a plethora of software tools and significant manual adjustments
to reach the desired quality of the final 3D model.

Our goal is to frame this process as an \emph{inverse rendering} task, and optimize
as many steps as possible jointly, driven by the quality of the rendered images
of the reconstructed model, compared to the captured input imagery. 
Recent work approaches 3D reconstruction with neural rendering, and provides 
high quality novel view synthesis~\cite{Mildenhall2020}. However, these methods typically produce representations 
that entangle geometry, materials and lighting into neural networks, and thus cannot easily support 
scene editing operations. Furthermore, to use them in traditional graphics engines, one needs to extract geometry 
from the network using methods like Marching Cubes which may lead to poor surface quality, 
particularly at low triangle counts. Recent neural methods can disentangle shape, materials, 
and lighting\cite{Boss2021,Zhang2020physg,Zhang2021nerfactor}, but sacrifice reconstruction quality.
Also, the materials encoded in neural networks cannot easily be edited or extracted in a form compatible with
traditional game engines. 
In contrast, we reconstruct 3D content compatible with traditional graphics engines, 
supporting re-lighting and scene editing. 

In this paper, we present a highly efficient inverse rendering method capable of extracting triangular 
meshes of unknown topology, with spatially-varying materials and lighting from multi-view images. 
We assume that the object is illuminated under one unknown environment lighting condition, 
and that we have corresponding camera poses and masks indicating the object in these images, 
as in past work~\cite{Boss2021}. 
Our approach learns topology and vertex positions for a surface mesh without
requiring any initial guess for the 3D geometry. 
The heart of our method is a differentiable surface model based on a deformable tetrahedral mesh~\cite{Shen2021},
which we extend to support spatially-varying materials and high dynamic range (HDR) environment lighting,
through a novel differentiable split sum approximation. 
We optimize geometry, materials and lighting (50M+ parameters) jointly using a highly optimized 
differentiable rasterizer with deferred shading~\cite{Laine2020,Hasselgren2021}. 
The resulting 3D model can be deployed without conversion on any device supporting triangle rendering, 
including phones and web browsers, and renders at interactive rates. 

Experiments show our extracted models used in scene editing (e.g., Figure~\ref{fig:scene_editB}), 
material decomposition, and high quality view interpolation, all running at interactive rates in triangle-based 
renderers (rasterizers and path tracers).

\vspace{-0.5mm}
\section{Related Work}
\vspace{-0.5mm}

\figSystem

\subsection{Multi-view 3D Reconstruction}

\paragraph{Classical methods} for multi-view 3D reconstruction either exploit inter-image 
correspondences~\cite{schoenberger2016mvs,furukawa2009accurate,agarwal2011building,galliani2016gipuma} to estimate 
depth maps or use voxel grids to represent shapes~\cite{de1999poxels,seitz1999photorealistic}. The former methods 
typically fuse depth maps into point clouds, optionally generating meshes~\cite{kazhdan2013screened}. 
They rely heavily on the quality of matching, and errors are hard to rectify during post-processing. 
The latter methods estimate occupancy and color for each voxel and are often limited by the cubic memory requirement.

\vspace{-9pt}
\paragraph{Neural implicit representations} leverage differentiable rendering to reconstruct 3D geometry with appearance
from image collections~\cite{Jiang2020sdfdiff,Mildenhall2020,Niemeyer2020CVPR}. 
NeRF~\cite{Mildenhall2020} and follow-ups~\cite{martin2021nerf,niemeyer2021giraffe,pumarola2021d,kaizhang2020,wang2021nerfmm,garbin2021fastnerf,mueller2022instant,Reiser2021,yu2021plenoctrees,Wizadwongsa2021NeX}, 
use volumetric representations and compute radiance by ray marching through a neurally encoded 5D light field.
While achieving impressive results on novel view synthesis, geometric quality suffers from the ambiguity of volume 
rendering~\cite{kaizhang2020}.
Surface-based rendering methods~\cite{Niemeyer2020CVPR, yariv2020multiview} use implicit 
differentiation to obtain gradients, optimizing the underlying surface directly. 
Unisurf~\cite{Oechsle2021} is a hybrid approach that gradually reduces the sampling region, encouraging 
a volumetric representation to converge to a surface, and NeuS~\cite{Wang2021neus} provides an unbiased conversion 
from SDF into density for volume rendering.
Common for all methods is that they rely on ray marching for rendering, which is computationally expensive both during 
training and inference. While implicit surfaces can be converted to meshes for fast inference, this introduces additional 
error not accounted for during optimization~\cite{Shen2021}.
We optimize the end-to-end image loss of an explicit mesh representation, supporting intrinsic decomposition of 
shape, materials and lighting by design, and utilizing efficient differentiable rasterization~\cite{Laine2020}.

\vspace{-9pt}
\paragraph{Explicit surface representations} are proposed to estimate explicit 3D mesh from 
images~\cite{Gao2020,Liao2018,Shen2021,softras,Chen2019dibrender,kaolin,chen2021dibrpp}. Most approaches 
assume a given, fixed mesh topology~\cite{softras,Chen2019dibrender,kaolin,chen2021dibrpp}, but 
this is improved in recent work~\cite{Liao2018,Gao2020,Shen2021}.
In particular,
DMTet~\cite{Shen2021} directly optimizes the surface mesh using a differentiable marching tetrahedral layer. 
However, it focuses on training with 3D supervision. In this work, we extend DMTet to 2D supervision, 
using differentiable rendering to jointly optimize topology, materials, and lighting. 

\vspace{-0.5mm}
\subsection{BRDF and Lighting Estimation}
\vspace{-0.5mm}

Beyond geometry, several techniques estimate surface radiometric properties from images. 
Previous work on BTF and SVBRDF estimation rely on special viewing configurations, lighting patterns or complex capturing 
setups~\cite{hendrik2003planned,Gardner2003,Ghosh2009,Guarnera2016,Weinmann2015,bi2020neural,boss2020two,Schmitt2020CVPR,haindl_filip13visual}.
Recent methods exploit neural networks to predict BRDF from images~\cite{Gao2019,Guo2020,li2020inverse,li2018learning,Merlin2021,Luan2021}.
Differentiable rendering methods~\cite{softras,Chen2019dibrender,StyleGAN3D,chen2021dibrpp,Hasselgren2021} learn to predict geometry, 
SVBRDF and, in some cases, lighting via 2D image loss. Still, their shape is generally deformed from a sphere and cannot represent arbitrary topology.

\begin{table}[b]
	\centering
	{\small
		\begin{tabularx}{\columnwidth}{l|XcXX}
			Method                              &  Geometry & Factorize    & Training  & Inference \\ \hline
			NeRF~\cite{Mildenhall2020}          &  NV       &              & day       & seconds   \\
			NGP-NeRF~\cite{mueller2022instant}  &  NV       &              & minutes   & ms        \\
			NeRD~\cite{Boss2021}                &  NV       & \checkmark   & days      & seconds   \\
			NerFactor~\cite{Zhang2021nerfactor} &  NV       & \checkmark   & days      & seconds   \\
			PhySG~\cite{Zhang2020physg}         &  NS       & \checkmark   & day       & seconds   \\
			NeuS~\cite{Wang2021neus}            &  NS       &              & day       & seconds   \\
			Our                                 &  Mesh     & \checkmark   & hour      & ms        \\
			\hline
		\end{tabularx}
	}
	\vspace{-2mm}
	\caption{
		Taxonomy of methods. NV: Neural volume, NS: Neural surface. Factorize indicates if the method 
		supports some decomposition of geometry, materials, and lighting. 
	}
	\vspace{-2mm}
	\label{tab:taxonomy}
\end{table}

Neural implicit representations 
successfully estimate lighting and BRDF from image collections. 
Bi et al.~\cite{bi2020neural} and NeRV~\cite{nerv2021} model light transport to support advanced lighting effects, 
e.g., shadows, but have very high computational cost.
Most related to our work are neural 3D reconstruction methods for jointly estimating shape, BRDFs and lighting from 
images~\cite{Boss2021,Zhang2020physg,Zhang2021nerfactor,boss2021neuralpil},
while providing an \emph{intrinsic decomposition} of these terms.
Illumination is represented using mixtures of spherical Gaussians (NeRD~\cite{Boss2021}, PhySG~\cite{Zhang2020physg}), 
or low resolution envmaps (NeRFactor~\cite{Zhang2021nerfactor}), in both cases
limited to low frequency illumination. 
In contrast, we propose a differentiable split sum lighting model, also 
adopted by the concurrent work Neural-PIL~\cite{boss2021neuralpil}. 
These neural implicit methods use multiple MLPs to factorize the representation, 
resulting in long training and inference times. 
Furthermore, these methods forgo the vast ecosystem of available 3D modeling and 
rendering tools, ``reinventing the wheel'' for tasks such as rendering, 
scene editing~\cite{yang2021objectnerf} and simulation.
In contrast, our output is directly compatible with existing renderers and modeling tools.
We optimize an explicit surface mesh, BRDF parameters, and lighting
stored in an HDR probe, achieving faster training speed and better decomposition results. 
Table~\ref{tab:taxonomy} shows a high level comparison of the methods.

\vspace{-0.5mm}
\section{Our Approach}
\vspace{-0.5mm}

\figNEUS

We present a method for 3D reconstruction supervised by multi-view images of an 
object illuminated under one unknown environment lighting condition, together with known camera poses and background 
segmentation masks.
The target representation consists of triangle meshes, spatially-varying materials (stored in 2D textures) and 
lighting (a high dynamic range environment probe). 
We carefully design the optimization task to \emph{explicitly} render triangle meshes, while
robustly handling arbitrary topology. Hence, unlike most recent work using neural implicit surface or volumetric
representations, we directly optimize the target shape representation.

Concretely, we adapt Deep Marching Tetrahedra~\cite{Shen2021} (DMTet) to work in the setting of 2D supervision,
and jointly optimize shape, materials, and lighting. 
At each optimization step, the shape representation -- parameters of a signed distance field (SDF) defined on a grid
with corresponding per-vertex offsets  --
is converted to a triangular surface mesh using a marching tetrahedra layer. Next, we render the extracted surface mesh in a differentiable 
rasterizer with deferred shading, and compute loss in image space on the rendered image compared to a reference image. 
Finally, the loss gradients are back-propagated to update the shape, textures and lighting parameters. 
Our approach is summarized in Figure~\ref{fig:system} and each step is described in detail below;
Section~\ref{sec:topology} outlines our topology optimization, Section~\ref{sec:shadingmodel}
presents the spatially-varying shading model, and our approach for reconstructing all-frequency environment
lighting is described in Section~\ref{sec:splitsum}.

\vspace{-9pt}
\paragraph{Optimization task}
Let $\phi$ denote our optimization parameters (i.e., SDF values and vertex offsets representing the shape, 
spatially varying material and light probe parameters).
For a given camera pose, $c$, the differentiable renderer produces an 
image $I_{\phi}(c)$. The reference image $I_{\mathrm{ref}}(c)$ is a view from the same camera.
Given a loss function $L$, we minimize the empirical risk
\begin{equation}
\underset{\phi}{\mathrm{argmin}}\ \mathbb{E}_{c}\big[L\big(I_{\phi}(c), I_{\mathrm{ref}}(c)\big)\big]
\end{equation}
using Adam~\cite{Kingma2014} based on gradients w.r.t.~the optimization parameters,
$\partial L/\partial\phi$, which are obtained through differentiable rendering.
Our renderer uses physically based shading and produces images with high dynamic range.
Therefore, the objective function must be robust to the full range of floating-point values.
Following recent work in differentiable rendering~\cite{Hasselgren2021},
our loss function is $L = L_\mathrm{image} + L_\mathrm{mask} + \lambda L_{\mathrm{reg}}$, an image space loss, 
$L_\mathrm{image}$ ($L_1$ norm on tone mapped colors), a mask loss, $L_\mathrm{mask}$ (squared $L_2$) 
and a regularizer $L_{\mathrm{reg}}$ (Equation~\ref{eq:reg}) to improve geometry.
Please refer to the supplemental material for details. 

\vspace{-9pt}
\paragraph*{Assumptions} For performance reasons we 
use a differentiable rasterizer with deferred shading~\cite{Hasselgren2021}, 
hence reflections, refractions (e.g., glass), and translucency are not supported.
During optimization, we only renderer direct lighting without shadows. 
Our shading model uses a diffuse Lambertian lobe and a specular, isotropic microfacet GGX lobe, which 
is commonly used in modern game engines~\cite{Karis2013,Lagarde2014}. Both dielectrics and metal materials are supported.
We note that our approach directly generalizes to a differentiable path tracer~\cite{Nimier2019,Nimier2020}, but at a significantly
increased computational cost.

\vspace{-0.5mm}
\subsection{Learning Topology}
\vspace{-0.5mm}
\label{sec:topology}
\figDMTetFormulation
Volumetric and implicit shape representations (e.g., SDFs) can be converted to meshes through Marching Cubes~\cite{Lorensen1997} (MC) 
in a post-processing step. However, MC inevitably imposes discretization errors. As a result, the output mesh quality, 
particularly at the moderate triangle counts typically used in real-time rendering, is often not sufficient. 
Similarly, simplifying dense extracted meshes using decimation tools may introduce errors in rendered appearance. 
To avoid these issues, we explicitly render triangle meshes during optimization. 
We leverage Deep Marching Tetrahedra~\cite{Shen2021} (DMTet) in a 2D supervision
setting through differentiable rendering. DMTet is a hybrid 3D representation that represents a shape with 
a discrete SDF defined on vertices of a deformable tetrahedral grid. The SDF is converted to triangular 
mesh using a differentiable marching tetrahedra layer (MT), as illustrated in Figure~\ref{fig:DMTetFormulation}. 
The loss, in our case computed on renderings of the 3D model, is back-propagated to the implicit field
to update the surface topology. This allows us to directly optimize the surface mesh and rendered appearance \emph{end-to-end}.

We illustrate the advantage of end-to-end learning in Figure~\ref{fig:neus}, where we compare our 
meshes to those generated by competing methods. While NeRF~\cite{Mildenhall2020} (volumetric rep.) 
and NeuS~\cite{Wang2021neus} (implicit surface rep.) provide high quality view interpolation, 
the quality loss introduced in the MC step is significant at low triangle counts.

Given a tetrahedral grid with vertex positions $v$, DMTet learns SDF values, $s$, 
and deformation vectors $\Delta v$. The SDF values and deformations can either be stored 
explicitly as values per grid vertex, or implicitly~\cite{Niemeyer2020CVPR,Oechsle2021} 
by a neural network. At each optimization step, the SDF is first
converted to a triangular surface mesh using MT, which is shown to be differentiable w.r.t. 
SDF and can change surface topology in DMTet~\cite{Shen2021}. Next, the extracted mesh is rendered using a 
differentiable rasterizer to produce an output image, and image-space loss gradients
are back-propagated to the SDF values and offsets (or network weights).
A neural SDF representation can act as a smoothness prior, which can be beneficial in producing 
well-formed shapes. Directly optimizing per-vertex attributes, on the other hand, can capture
higher frequency detail and is faster to train. In practice, the optimal choice of parametrization depends 
on the ambiguity of geometry in multi-view images. We provide detailed analysis in the supplemental materials.

To reduce floaters and internal geometry, we regularize the SDF values of DMTet similar to 
Liao~et~al.~\cite{Liao2018}. Given the binary cross-entropy $H$, sigmoid function $\sigma$, 
and the sign function $\sign(x)$, we define the regularizer as
\begin{equation}
\small
L_{\mathrm{reg}} = \sum_{{i,j} \in \mathbb{S}_e}
H\left(\sigma(s_i), \sign\left(s_j\right)\right) + H\left(\sigma(s_j), \sign\left(s_i\right)\right),
\label{eq:reg}
\end{equation}
where we sum over the set of unique edges, $\mathbb{S}_e$, in the tetrahedral grid, for which
$\sign(s_i) \neq \sign(s_j)$. 
Intuitively, this reduces the number of sign flips and simplifies the 
surface, penalizing internal geometry or floaters.
We ablate the choice of regularization loss in the supplemental material.

\vspace{-0.5mm}
\subsection{Shading Model}
\label{sec:shadingmodel}
\vspace{-0.5mm}

\paragraph*{Material Model}
We follow previous work in differentiable rendering~\cite{Hasselgren2021} 
and use the physically-based (PBR) material model from Disney~\cite{Burley12}.
This lets us easily import game assets and render our optimized models 
directly in existing engines without modifications. This material model 
combines a diffuse term with an isotropic, specular GGX lobe~\cite{Walter2007}. 
Referring to Figure~\ref{fig:material_model}, the
diffuse lobe parameters $\kd$ are provided as a four-component texture, 
where the optional fourth channel $\alpha$ represents transparency. 
The specular lobe is described by a roughness value, $r$, for the GGX normal 
distribution function and a metalness factor, $m$, which interpolates between plastic and metallic
appearance by computing a specular highlight color according to 
$\ks = (1-m) \cdot 0.04 + m \cdot \kd$~\cite{Karis2013}. Following 
a standard convention, we store these values in a texture $\korm = (o,r,m)$, where $o$ is left unused.
Finally, we include a tangent space normal map, $\mathbf{n}$, to capture
high frequency shading detail. We regularize material parameters using a 
smoothness loss~\cite{Zhang2021nerfactor}, please refer to our supplemental material 
for details.

\figMaterialModel

\vspace{-9pt}
\paragraph*{Texturing}
Automatic texture parametrization for surface meshes is an active research area  
in computer graphics. We optimize topology, which requires continually updating the 
parametrization, potentially introducing discontinuities into the training process.
To robustly handle texturing during topology optimization, we leverage volumetric texturing, 
and use world space position to index into our texture. This ensures that the 
mapping varies smoothly with both vertex translations and changing topology. 

The memory footprint of volumetric textures grows cubically, which is unmanageable 
for our target resolution. We therefore extend the approach of PhySG~\cite{Zhang2020physg}, using a 
multilayer perceptron (MLP) to encode all material parameters in a 
compact representation. This representation can adaptively allocate detail near the 2D manifold
representing the surface mesh, which is a small subset of the dense 3D volume.  
More formally, we let a positional encoding + MLP represent a mapping $\mathbf{x} \to (\kd, \korm, \mathbf{n})$, 
e.g., given a world space position $\mathbf{x}$, compute the base color, $\kd$, specular parameters, 
$\korm$ (roughness, metalness), and a tangent space normal perturbation, $\mathbf{n}$. 
We leverage the tiny-cuda-nn framework~\cite{tiny-cuda-nn},
which provides efficient kernels for hash-grid positional encoding~\cite{mueller2022instant} and MLP evaluations.

Once the topology and MLP texture representation have converged, we \emph{re-parametrize} the model:
we generate unique texture coordinates using xatlas~\cite{xatlas} and sample 
the MLP on the surface mesh to initialize 2D textures, then continue the optimization with fixed 
topology. Referring to Figure~\ref{fig:tex_seams}, this effectively removes texture seams 
introduced by the $(u,v)$-parametrization, and may also increase texture detail as we can use 
high resolution 2D textures for each of $\kd$, $\korm$, and $\mathbf{n}$. 
This results in 2D textures compatible with standard 3D tools and game engines.

\figTexSeams

\vspace{-0.5mm}
\subsection{Image Based Lighting}
\vspace{-0.5mm}
\label{sec:splitsum}

We adopt an image based lighting model, where the scene environment light is given by a high-resolution 
cube map. Following the rendering equation~\cite{Kajiya1986}, we compute the outgoing radiance $L(\omega_o)$ 
in direction $\omega_o$ by:
\begin{equation}
\vspace{-1mm}
	L(\omega_o) = \int_\Omega L_i(\omega_i)f(\omega_i,\omega_o) (\omega_i \cdot \mathbf{n}) d\omega_i.
    \label{eq:ibl}
\vspace{-1mm}
\end{equation}
This is an integral of the product of the incident radiance, $L_i(\omega_i)$ from 
direction $\omega_i$ and the BSDF $ f(\omega_i, \omega_o)$. The integration domain is 
the hemisphere $\Omega$ around the surface intersection normal, $\mathbf{n}$. 

Below, we focus on the specular part of the outgoing radiance, where, in our case, 
the BSDF is a Cook-Torrance microfacet specular shading model~\cite{Cook1982} according to:
\begin{equation}
\vspace{-1mm}
f(\omega_i,\omega_o) = \frac{D \ G \ F }{4 (\omega_o \cdot \mathbf{n}) (\omega_i \cdot \mathbf{n})},
\vspace{-1mm}
\end{equation}
where $D$, $G$ and $F$ are functions representing the GGX~\cite{Walter2007} normal distribution (NDF), 
geometric attenuation and Fresnel term, respectively.

High quality estimates of image based lighting can be obtained by Monte Carlo integration. 
For low noise levels, large sample counts are required,
which is typically too expensive for interactive applications. 
Thus, spherical Gaussians (SG) and spherical harmonics (SH) are common approximations for image based 
lighting~\cite{Chen2019dibrender,Zhang2020physg,Boss2021}.
They allow for control over the lighting frequency through varying the number of SG lobes (or SH coefficients), and are
efficient representations for low to medium frequency lighting. However, representing high frequency and highly specular 
materials requires many SG lobes, which comes at a high runtime cost and hurts training stability.

We instead draw inspiration from real-time rendering, where the \emph{split sum} approximation~\cite{Karis2013} 
is a popular, efficient method for all-frequency image based lighting. 
Here, the lighting integral from Equation~\ref{eq:ibl} is approximated as: 
{\small
\begin{equation}
\vspace{-1mm}
    L(\omega_o) \approx\int_\Omega f(\omega_i,\omega_o) (\omega_i \cdot \mathbf{n}) d\omega_i
                 \int_\Omega L_i(\omega_i)D(\omega_i,\omega_o) (\omega_i \cdot \mathbf{n}) d\omega_i.
    \label{eqn:splitsum}
\end{equation}
}
The first term represents the integral of the specular BSDF with a solid white environment light. It only depends on the parameters 
$\cos\theta = \omega_i \cdot \mathbf{n}$ and the roughness $r$ of the BSDF, and can be precomputed and stored in a 2D lookup 
texture. The second term represents the integral of the incoming radiance with the specular NDF, $D$. 
Following Karis~\cite{Karis2013}, this term is also pre-integrated and represented by a filtered cubemap. In each mip-level, 
the environment map is integrated against $D$ for a fixed roughness value (increased roughness at coarser mips).

The split sum approach is popular for its modest runtime cost, using only two 
texture lookups: query the 2D lookup texture representing the first term based on $(r, \cos\theta)$
and the mip pyramid at level $r$, in direction $\omega_o$. This should be compared to 
evaluating SG products with hundreds of lobes for each shading point. 
Furthermore, it uses the standard GGX parametrization, which means that we can relight our extracted models with 
different kinds of light sources (point, area lights etc.) and use our reconstructed materials with no modifications 
in most game engines and modeling tools. 

We introduce a \emph{differentiable} version of the split sum shading model to learn  
environment lighting from image observations through differentiable rendering. 
We let the texels of a cube map (typical resolution $6\times512\times512$) be the trainable parameters. 
The base level represents the pre-integrated lighting for the lowest supported roughness value, 
and each smaller mip-level is constructed from the base level using the pre-filtering 
approach from Karis~\cite{Karis2013}. 

\figRelightNerFactorB

To obtain texel gradients, we express the lighting computations using PyTorch's auto-differentiation. 
However, the pre-filtering of the second term in Equation~\ref{eqn:splitsum} must be updated
in each training iteration, and therefore warrant a specialized CUDA implementation to reduce the training cost. 
This term can either be estimated through Monte-Carlo integration (BSDF importance sampling), 
or by pre-filtering the environment map in a solid-angle footprint derived from the NDF. 
To reduce noise, at the cost of introducing some bias, we chose the latter approach.
Please refer to our supplemental material for implementation details. 

We additionally create a single filtered low-resolution ($6\times16\times16$) cube map representing the \emph{diffuse} lighting. 
The process is identical to the filtered specular probe, sharing the same trainable parameters, average-pooled 
to the mip level with roughness $r=1$. The pre-filtering of the diffuse term only uses the cosine term, 
$\omega_i \cdot \mathbf{n}$. The two filtering steps are fully differentiable and are performed at 
each training step.

\vspace{-0.5mm}
\section{Experiments}
\vspace{-0.5mm}

\figEditNerFactor

In the following, we evaluate our system for a variety of applications. 
To emphasize our approach's explicit decomposition into a triangle mesh and materials, 
we show re-lighting, editing, and simulation using off-the shelf tools. 
We also compare to recent neural methods supporting factorization: NeRD~\cite{Boss2021} 
and NeRFactor~\cite{Zhang2021nerfactor}. 
While not our main focus, we include view interpolation results to establish a 
baseline comparison to state-of-the-art methods. 
Finally, we compare our split-sum approximation against spherical Gaussians for image-based lighting.

\vspace{-0.5mm}
\subsection{Scene Editing and Simulation}
\label{sec:scene_edit}
\vspace{-0.5mm}

\tabRelightNerfactor

Our factorized scene representation enables advanced scene editing.
Previous work using density-based neural representations only supports 
rudimentary relighting and simple forms of scene edits~\cite{Boss2021,Zhang2020physg,Zhang2021nerfactor}. 

In Figure~\ref{fig:relight_nerfactor} we compare the relighting quality of our reconstructed model, rendered using 
the Blender Cycles path tracer, with the results of NeRFactor (rendered by evaluating a neural network). 
A quantitative summary is provided in Table~\ref{tab:nerfactor_relight_avg}, where we also measure the quality of the 
reconstructed albedo textures. We note that our method produces more detailed results and outperforms NeRFactor in all metrics. 
Our artifacts come mainly from the mismatch between training (using a rasterizer), and inference (using full global illumination). 
In areas with strong shadowing or color bleeding, our material and geometry quality suffer. A 
differentiable path tracer~\cite{Nimier2019} can likely improve the material separation in our pipeline, 
but would require significantly more computations. 

Our representation can be directly deployed in the vast collection of
3D content generation tools available for triangle meshes. This greatly facilitates scene editing, 
which is still very challenging for neural volumetric representations~\cite{Zhang2021nerfactor}.
We show advanced scene editing examples in Figure~\ref{fig:edit_nerfactor}, where we add our reconstructed models 
from the NeRFactor dataset to the Cornell box and use them in a soft-body simulation. 
Note that our models receive scene illumination, cast accurate shadows, and robustly act as colliders for virtual objects.
In Figure~\ref{fig:scene_editB}, and the supplemental video, we show another example,
where an object is reconstructed from real-world photographs and then used as a collider for a virtual cloth object. 
The combined scene is then rendered using our extracted environment light.
Note that shading of the virtual object looks plausible given the reference photo.
We also show material editing on the same example. 

\vspace{-0.5mm}
\subsection{View interpolation}
\label{sec:view_interpolation}
\vspace{-0.5mm}

\paragraph{Synthetic datasets} We show results for the NeRF \emph{realistic synthetic image} dataset in 
Table~\ref{tab:view_interpolation_avg} and a visual example of the \textsc{Materials} scene in 
Figure~\ref{fig:teaser}.
Per-scene results and visual examples are included in our supplemental material,
where we also include Chamfer loss on extracted meshes. 
Our method consistently performs on par with NeRF, with better quality in some scenes. 
The margins are smaller for perceptually based image metrics (SSIM 
and LPIPS). We speculate that density-based volume approaches can more efficiently minimize PSNR than our  
opaque meshes. However, the effect of slightly moving a silhouette edge will not be as detrimental to a perceptual metric.

\tabViewInterp
\figTeaser

The \textsc{Drums} and \textsc{Ship} scenes are failure cases for our method. We assume mostly direct lighting, with no significant 
global illumination effects, and these scenes contain both significant intra-scene reflections, refractions, and caustics. 
Interestingly, while material reconstruction suffers, we still note high quality results for view interpolation. 

\tabViewInterpNerfactor

Given that we factorize into explicit shape, materials and lighting, we have slightly lower quality on novel view synthesis
than methods specialized for view-interpolation. To put this in context, in Table~\ref{tab:nerfactor_avg} we compare
our aproach against NeRFactor, which performs a similar factorization, and our approach. We observe a $4.21$ dB PSNR image 
quality reduction for NeRFactor compared to the NeRF baseline. This is consistent with NeRD~\cite{Boss2021} which do not 
provide source code but report a $4.17$~dB quality drop for their factorized representation on a subset of the NeRF synthetic 
dataset. In contrast, our quality is significantly higher, while still providing the flexibility of a factorized representation.

\vspace{-9pt}
\paragraph{Real-world datasets}
NeRD~\cite{Boss2021} provides a small dataset of real-world photo scans with auto-generated (inaccurate) coverage masks 
and diverse camera placement. 

\nerdComparison

Visual and quantitative results are shown in Figure~\ref{fig:nerd_cmp}, where we have masked out the background for the
reference object. Due to inconsistencies in the dataset, both NeRF and NeRD struggle to find sharp geometry borders with
transparent ``floaters'' and holes. In contrast, we get sharp silhouettes and a significant boost in image quality. 
The results reported for NeRD are for their volumetric representation. Note that NeRD can generate output meshes as a 
post-processing step, but at a significant quality loss (visual comparison included in our supplemental material).

\vspace{-0.5mm}
\subsection{Comparing Spherical Gaussians and Split Sum}
\vspace{-0.5mm}

\figSG

In Figure~\ref{fig:sg}, we compare our differentiable split sum environment
lighting approximation, from Section~\ref{sec:splitsum} against the commonly used spherical Gaussian (SG) 
model. 
Split sum captures  the lighting much more faithfully across all frequencies, while still having a lower runtime cost. 
In our implementation, we observe a $~5\times$ reduction of optimization time compared to SG with 128 lobes. 
At inference, evaluating the split sum approximation is extremely fast, 
requiring just two texture lookups.

\vspace{-0.5mm}
\section{Limitations and Conclusions}
\vspace{-0.5mm}

Our main limitation is the simplified shading model, not accounting for global illumination or shadows. 
This choice is intentional to accelerate optimization, but is a limiting factor for material extraction and 
relighting. With the current progress in differentiable path tracing, we look forward to this limitation being 
lifted in future work. We additionally rely on alpha masks to separate foreground from background. While 
our method seems quite robust to corrupted masks, it would be beneficial to further incorporate this step into 
the system. Other limitations include the static lighting assumption, not optimizing camera poses, and high compute 
resource and memory consumption during training.
Apart from deepfakes, which are common to all scene reconstruction methods, 
we are not aware of and do not foresee nefarious use cases of our method.

In summary, we show results on par with state-of-the-art for view synthesis and 
material factorization, while directly optimizing an \emph{explicit} representation: triangle meshes with 
materials and environment lighting.
Our representation is, by design, directly compatible with modern 3D engines and modeling tools, 
which enables a vast array of applications and simplifies artist workflows.
We perform end-to-end optimization driven by appearance of the rendered model, 
while previous work often sidestep the error from mesh extraction through Marching Cubes. 
Our method can be applied as an appearance-aware
converter from a (neural) volumetric or SDF representation to triangular 3D models with materials,
complementing many recent techniques.

{\small
\bibliographystyle{ieee_fullname}
\bibliography{paper}
}

\cleardoublepage 
\section{Supplemental Material}

In the following, we supplement the paper with additional results, ablations and implementation details. 
In Section~\ref{sec:novel} we present novel use cases for our method: automatic level of detail creation from images,
and appearance aware model extraction.
In Section~\ref{sec:results}, we present additional results, including an evaluation of geometric quality, 
additional per-scene statistics and visual examples. 
Finally, in Section~\ref{sec:impl} we provide implementation details, including efficient split-sum pre-integration,
regularizer terms and losses.

\section{Novel applications}
\label{sec:novel}
\subsection{Level-of-detail From Images}

Inspired by a recent work in appearance-driven automatic 3D model simplification~\cite{Hasselgren2021}, 
we demonstrate level-of-detail (LOD) creation directly from rendered images of an object. 
The previous technique requires an initial 
guess with fixed topology and known lighting. We generalize this approach and showcase LOD creation
directly from a set of images, i.e., we additionally learn both topology and lighting. 
To illustrate this, we generated 256 views (with masks \& poses) from a 
path tracer, rendered in two resolutions: 1024$\times$1024 pixels and  
128$\times$128 pixels, then reconstructed the mesh, materials and lighting in our pipeline
to create two LOD levels (geometry and spatially-varying materials). We show visual results in 
Figure~\ref{fig:lod} and in the supplemental video. 
\figLOD

\subsection{Appearance-Aware NeRF 3D Model Extractor}
\figMeshExtract

We devise a way to extract 3D models from neural radiance fields~\cite{Mildenhall2020} (NeRF) in a format 
compatible with traditional 3D engines. 
Our pipeline for this task has three steps: 
\begin{equation*}
\mathrm{NeRF} \rightarrow \mathrm{Marching\ Cubes} \rightarrow \mathrm{Differentiable\ renderer}. 
\end{equation*}
The dataset consists of 256 images of the Damicornis model~\cite{Smithsonian2020} (with masks and poses),
rendered in a path tracer. We first train a NeRF model 
and extract the mesh with Marching Cubes. Next, we finetune the extracted mesh and learn 
materials parameters (2D textures) using our differentiable renderer (with DMTet topology optimization disabled), 
still only supervised by the images in the dataset.
The output is a triangle mesh with textured PBR materials compatible with traditional engines.
As a bonus, the silhouette quality improves over the Marching Cubes extraction, which is illustrated in 
Figure~\ref{fig:mesh_extract}. 

\subsection{3D Model Extraction with Known Lighting}
\figSaxophone

We observe that the DMTet representation successfully learns challenging topology and materials jointly, 
even for highly specular models and when lit using high frequency lighting. We illustrate this in a joint 
shape and material optimization task with known environment lighting, optimized using a
large number of views. In Figure~\ref{fig:sax} we show two examples from the 
Smithsonian 3D repository~\cite{Smithsonian2020}. Note the quality
in both the extracted materials and geometric detail.

\tabRelightNerfactorBreakdown
\figRelightNerFactorC
\figAlbedoNerFactorC

\section{Results}
\label{sec:results}
\subsection{Scene Editing and Simulation}

This section supplements Section~\ref{sec:scene_edit} in the main paper. In Table~\ref{tab:nerfactor_relight} we present per-scene breakdowns
of relighting results corresponding to Table~\ref{tab:nerfactor_relight_avg} in the main paper. An additional visual relighting example is shown in 
Figure~\ref{fig:relight_nerfactor_ficus}, where we relight the Ficus scene with four different light probes, comparing 
to the results of NeRFactor~\cite{Zhang2021nerfactor}. Figure~\ref{fig:albedo_nerfactor} shows a visual example of material 
separation with albedo, $\kd$, and normals, $\mathbf{n}$. In Figure~\ref{fig:nerf_decomposition} we show our lighting, 
material and shape separation for all scenes in the NeRF synthetic dataset.
We note that we achieve significantly more detailed normals 
(thanks to the tangent space normal map included in our shading model) and albedo mostly decorrelated from lighting. 
Our remaining challenges are areas with strong shadows or global illumination effects, which are currently 
not rendered in our simplified shading model used during optimization.

\subsection{View interpolation}

\tabViewInterpBreakdown
\figPhySGComp
\tabViewInterpNerfactorBreakdown
\figKnobComp

This section supplements Section~\ref{sec:view_interpolation} in the main paper. In Table~\ref{tab:view_interpolation} we show per-scene breakdowns 
of the view-interpolation results corresponding to Table~\ref{tab:view_interpolation_avg} in the main paper, evaluated on the 
NeRF Synthetic Dataset. Figure~\ref{fig:physg_comp} shows a visual comparison to PhySG and MipNeRF for the \textsc{Chair}, 
\textsc{Microphone} and \textsc{Ship} scenes. We note that PhySG struggles to capture the complex geometry of the NeRF dataset. 

To study view interpolation quality for techniques which support material decomposition, 
we report per-scene breakdowns of view-interpolation result in Table~\ref{tab:nerfactor}. 
This corresponds to Table~\ref{tab:nerfactor_avg} in the main paper. We use the NeRFactor dataset (which is 
a subset of the NeRF dataset with simplified lighting conditions) and 
compare with NeRFactor and PhySG.

In Figure~\ref{fig:knob_comp} we additionally compare view interpolation quality on a small synthetic dataset containing three 
scenes with increasing geometric complexity: \textsc{Knob}, \textsc{Damicornis} and \textsc{Cerberus}, 
each dataset consists of 256 views with masks and known camera poses, and is validated on 200 novel views.
We compare against NeRF (neural volumetric representation) and NeuS~\cite{Wang2021neus} (neural implicit representation).
We provided masks at training for both approaches.  We note that on this dataset, our method performs on par with NeRF, 
and consistently produces results with greater detail and sharpness than NeuS.

\subsection{Geometry}

\tabViewInterpChamfer
\figChamferSynthetic
\figKnobLambert
\nerdMesh

Our primary targets are appearance-aware 3D reconstructions which render efficiently in real-time (e.g. for a game or interactive 
path tracer). As part of that goal, our shading model includes tangent space normal maps, which is a commonly 
used technique to capture the appearance of high frequency detail at modest triangle counts. 
For these reasons, we consider image quality our main evaluation metric, but additionally report Chamfer scores in 
Table~\ref{tab:view_interpolation_chamfer} for completeness. 
When comparing with NeRF~\cite{Mildenhall2020}, we use pretrained checkpoints provided by
JaxNeRF\footnote{https://github.com/google-research/google-research/tree/master/jaxnerf}~\cite{jaxnerf2020github},
which we denote NeRF w/o mask.
We note that the pretrained models suffer greatly from floater geometry in some scenes. To that 
end, we additionally show results for NeRF (w/ mask) which further utilizes coverage masks and 
regularizes density, trading some image quality for better geometric accuracy.
We use the NGP-NeRF~\cite{mueller2022instant} code base to generate the NeRF (w/ mask) results. 
To calculate the Chamfer scores, we sample 2.5M points on both
predicted mesh and ground mesh respectively, and calculate the Chamfer distance between the two point clouds.

While our meshes have considerably lower triangle count that the MC extractions, we are still competitive in terms of Chamfer loss. 
Note that we primarily focus on opaque geometry, hence, the \textsc{Drums}, \textsc{Ship}, and \textsc{Ficus} scenes with transparency 
are challenging cases.

In Figure~\ref{fig:chamfer}, we report Chamfer loss on three synthetic datasets of increasing
geometric complexity. Interestingly, the neural implicit version performs very well 
on the organic shapes, but struggles on the \textsc{Cerberus} robot model,
where NeRF provide the lowest Chamfer loss. 
Visual comparisons of rendered reconstruction quality
are included in Figure~\ref{fig:knob_comp} and a visualization of the Lambertian shaded mesh is included in Figure~\ref{fig:knob_lambert}.

We additionally present an example of an output mesh generated by NeRD~\cite{Boss2021} in Figure~\ref{fig:nerd_mesh}. 
The impact of the mesh extraction step is notable, both to geometry and material quality. As we only have this 
single data point, with no means of accurately aligning the meshes for measuring geometric loss (NeRD does not provide source code), 
we will not provide metrics.

\subsection{Quality of Segmentation Masks}

Like many related works (e.g. NeRD~\cite{Boss2021}, DVR~\cite{Niemeyer2020CVPR}, 
and IDR~\cite{yariv2020multiview}) our method relies on foreground segmentation masks. While 
this is a limitation we hope to see lifted in future work, we note that our method is robust to moderate 
levels of mask corruption, as can be expected from automated methods or crowdsourced annotation.

\figGoldMaskErrors
\figMaskTechniques

Both the DTU MVS dataset (Figure~\ref{fig:DTUNeural}) and the NeRD dataset (Figure~\ref{fig:nerd_mesh}) rely 
on manually annotated masks, with some frames containing large errors and inconsistencies, as shown in 
Figure~\ref{fig:gold_mask_errors}.
In Figure~\ref{fig:mask_techniques} we automatically generate segmentation masks of varying quality
for a real-world dataset. We generate masks using two versions of Detectron2~\cite{wu2019detectron2}, 
namely PointRend~\cite{kirillov2019pointrend} and Mask R-CNN~\cite{matterport_maskrcnn_2017}. Additionally, 
we generated another version of masks using the ``object finder'' tool in Adobe Photoshop 2022. 
As expected, reconstruction quality decreases gracefully with lower mask 
quality. Subjectively, the silhouette looks best using the automatic mask generated in Photoshop. 

\figSyntheticMaskCorruption

In Figure~\ref{fig:synthetic_mask_corruption}, we show a synthetic experiment where we corrupt perfect masks
with increasing levels of noise. Reconstruction quality decreases gracefully as a function of corruption level, 
and while the quality reduction is significant, our system is stable even for large corruptions. 
Here, all masks in the dataset are corrupted, while segmentation algorithms typically produce 
good results with a few localized errors, so this experiment is a stress-test even for low noise levels. 
We speculate that \emph{surface-based} representations more robustly handle mask corruptions, as inaccuracies 
in silhouettes are less objectionable than the ``floater'' geometry generated by density-based approaches.

\subsection{Multi-View Stereo Datasets}

\figDTUNeural
\figDTUAblation
\figDamicornisAblation
Our experiments with scans from a limited view angle, low number of views, and/or varying 
illumination, e.g., the DTU MVS datasets~\cite{Jensen2014}, shows that our approach work less 
well than the recent neural implicit versions, such as NeuS~\cite{Wang2021neus}, Unisurf~\cite{Oechsle2021}, 
and IDR~\cite{yariv2020multiview}, which we attribute to a more regularized, smoother shape representation 
for the neural implicit approaches, and our physically-bases shading model which assumes constant lighting. 
We provide quantitative results for three scans from DTU in Table~\ref{tab:dtu_chamfer}, and visual examples of 
our results on three scans in Figure~\ref{fig:DTUNeural}.

\tabDTUChamfer

The sparse viewpoints and varying illumination (which breaks our shading model assumption of constant lighting) 
in the DTU datasets lead to strong ambiguity in the reconstructed geometry. In this case, we noticed that 
directly optimizing the per-vertex SDF values results in high-frequency noise in the surface mesh. 
Instead, we follow the approach of the neural implicit approaches and use an MLP to parametrize the SDF values, 
which implicitly regularize the SDF, and, as a consequence, the resulting surface geometry produced by DMTet. 
The smoothness of the reconstructed shape can be controlled by the frequency 
of the positional encoding applied to the inputs of the MLP, as shown in Figure~\ref{fig:DTUAblation}. 
On the contrary, in case of densely sampled 
viewpoints and constant illumination, we observed that directly optimizing per-vertex attributes better captures 
high-frequency details, as shown in Figure~\ref{fig:DamicornisAblation}, and is faster to train. 
We use direct optimization of per-vertex SDF values in all results presented in the paper, except for the DTU 
scans, and the NeRF hotdog example, where we obtained better geometry reconstruction using the MLP parameterization.

We use the same MLP as in DVR~\cite{Niemeyer2020CVPR}, which consists of five fully connected residual layers with 
256 hidden features. In addition, we adopt the positional encoding in NeRF~\cite{Mildenhall2020} and progressively
fit the frequencies similar to SAPE~\cite{hertz2021sape}. More specifically, for an input position $p$ and a set 
of encoding functions $e_1, e_2, \ldots, e_n$ with increasing frequencies, we multiply each encoding $e_n(p)$ with a soft 
mask $\alpha_n (t)$ at training iteration $t$. The first $n_{base}$ encodings are always exposed to the network, 
and we linearly reveal the rest during training such that:
\begin{eqnarray}
\alpha_n (t) &=&  
\begin{cases}
1 & n \leq n_{base} \\
\mathrm{min}(1,\frac{t}{t_f}) & n >  n_{base}
\end{cases}
\end{eqnarray}
where $t_f$ is the iteration when all encodings are fully revealed. In practice, we find that progressively fitting the frequencies produces less high-frequency artifacts 
on the reconstructed surface than the non-progressive scheme.

\section{Implementation}
\label{sec:impl}

\subsection{Optimization}

Unless otherwise noted, we start from a tetrahedral grid of resolution 128 (using 1.53M tetrahedra and 277k vertices). 
As part of the Marching Tetrahedral step, each tetrahedron can generate up to two triangles. 

We initialize the per-vertex SDF values to random values in the range $[-0.1, 0.9]$, such that a random selection 
of approximately 10\% of the SDF values will  report ``inside' status at the beginning of optimization. 
The per-vertex offsets are initialized to zero.

Textures are initialized to random values within the valid range. We also provide min/max values per texture channels, 
which are useful when optimizing from photographs, where we follow NeRFactor~\cite{Zhang2021nerfactor} and us a range on the albedo texture 
of $\kd \in [0.03, 0.8]$. Similarly, we limit the minimal roughness value (green channel of the $\korm$ texture) to
0.08 (linearized roughness). The tangent space normal map is initialized to $(0,0,1)$, i.e.,
following the surface normal with no normal perturbation.
The environment light texels are initialized to random values in the range $[0.25, 0.75]$, which we empirically found to be 
a reasonable starting point in our tests. 

We use the Adam~\cite{Kingma2014} optimizer with default settings combined with a learning rate scheduler 
with an exponential falloff from 1.0 to 0.1 over 5000 iterations.
We typically train for 5000 iteration using a mini-batch of eight images, rendered at the native resolution of 
the images in the datasets (typically in the range from 512$\times$512 pixels to 1024$\times$1024 pixels).
Next, after texture reparametrization, we finetune geometry and 2D textures with locked topology 
for another 5000 iterations. The entire process takes approximately an hour on a single NVIDIA V100 GPU, with indicative 
results after a few minutes. We include a training visualization in the supplemental video.

In DTU experiments, we set $n=6$, $n_{base}=4$ and $t_f=2500$ for the progressive positional encoding. We disable the normal perturbation and second-stage optimization to get the best geometric quality, and train DMTet for 10k iterations.

\subsection{Losses and Regularizers}
\label{sec:loss}
\paragraph{Image Loss}
Our renderer uses physically based shading and produces images with high dynamic range.
Therefore, the objective function must be robust to the full range of floating-point values.
Following recent work in differentiable rendering~\cite{Hasselgren2021}, 
our image space loss, $L_\mathrm{image}$, computes the $L_1$ norm on tone mapped colors. As tone map operator, 
we transform linear radiance values, $x$, according to $x' = \Gamma(\log(x + 1))$,
where $\Gamma(x)$ is the sRGB transfer function~\cite{srgb96}:
\begin{eqnarray}
\Gamma(x) &=&  
\begin{cases}
12.92x & x \leq 0.0031308 \\
(1+a)x^{1/2.4} -a & x > 0.0031308
\end{cases} \\ 
a &=& 0.055.  \nonumber
\end{eqnarray}

\newcommand{\di}{\boldsymbol{\delta}_i}
\newcommand{\dip}{{\di}{\!\!'\hspace*{0.2mm}}}
\newcommand{\vj}{\boldsymbol{v}_{\hspace*{-0.2mm}j}}
\newcommand{\vi}{\boldsymbol{v}_{\hspace*{-0.2mm}i}}

\paragraph{Light Regularizer} Real world datasets contain primarily neutral, white lighting. To 
that end, we use a regularizer for the environment light that penalizes color shifts. Given the 
per-channel average intensities $\overline{c_i}$, we define the regularizer as:
\begin{equation}
L_\text{light} = \frac{1}{3}\sum^3_{i=0} \left|\overline{c_i} - \frac{1}{3}\sum^3_{i=0}\overline{c_i} \right|.
\end{equation}

\paragraph{Material Regularizer}
As mentioned in the paper, we regularize material parameters using a smoothness loss similar to 
NeRFactor~\cite{Zhang2021nerfactor}. Assuming that $\kd\left(\mathbf{x}\right)$ denotes the $\kd$ 
parameter at world space position $\mathbf{x}$ and $\mathbf{\epsilon}$ is a random displacement 
vector, we define the regularizer as:
\begin{equation}
L_\text{mat} = \sum_{\mathbf{x}_\text{surf}} \left|\kd\left(\mathbf{x}_\text{surf}\right) - \kd\left(\mathbf{x}_\text{surf} + \mathbf{\epsilon}\right)\right|.
\end{equation}
To account for the lack of global illumination and shadowing in our differentiable 
renderer, we use an additional, trainable visibility term which can be considered a regularizer. We store this term 
in the otherwise unused $o$-channel of the $\boldsymbol{k}_{\hspace*{-0.1mm}\mathrm{orm}}$ specular 
lobe parameter texture and use it to directly modulate the radiance estimated by our split sum shading model. 
Thus, it is similar to a simple ambient occlusion term and does not account for directional visibility.

\paragraph{Laplacian Regularizer}
In the second pass, when topology is locked, we use a Laplacian regularizer~\cite{Sorkine2005} 
on the triangle mesh to regularize the vertex movements. 
The uniformly-weighted differential $\di$ of vertex $\vi$ is given by 
$\di = \vi - \frac{1}{\left|N_i\right|}\sum_{j \in N_i} \vj$, 
where $N_i$ is the one-ring neighborhood of vertex $\vi$.
We follow Laine et~al.~\cite{Laine2020} and use a Laplacian regularizer term given by
\begin{equation}
L_{\boldsymbol{\delta}} = \frac{1}{n}\sum_{i=1}^n\left\lVert\di - \dip\right\rVert^2,
\end{equation}
where $\dip$ is the uniformly-weighted differential of the input mesh (i.e., the output mesh
from the first pass).

\figRegAblation
\paragraph{SDF Regularizer}
If we only optimize for image loss, internal faces which are not visible from any viewpoint do not receive any gradient 
signal. This leads to random geometry inside the object, as shown in Fig.~\ref{fig:RegAblation}, which is undesirable for 
extracting compact 2D textures. To remove the internal faces, we regularize the SDF values of DMTet similar to 
Liao et al.~\cite{Liao2018} as described in the main paper (Eqn. 2). The $L_1$ smoothness loss 
proposed by Liao et al., adapted from occupancy to SDF values, can be written as:
\begin{equation}
L_{\mathrm{smooth}} = \sum_{{i,j} \in \mathbb{S}_e} |s_i - s_j|,
\label{eq:reg}
\end{equation}
where $\mathbb{S}_e$ is the set of unique edges, and $s_i$ represents the per-vertex SDF values.
In contrast, our regularization loss explicitly penalizes the sign change of SDF values over edges in the tetrahedral grid. 
Empirically, our loss more efficiently removes internal structures, as shown in Fig.~\ref{fig:RegAblation}.

\figRegVisibility
In our DTU experiments, we use an additional regularization loss to removes the floaters behind the visible surface, as illustrated in Fig~\ref{fig:RegVisibility}. Specifically, for a triangular face $f$ extracted from tetrahedron $T$, if $f$ is not visible in current training views, we encourage the SDFs at vertices of $T$ to be positive with BCE loss.

\subsection{Split Sum Implementation Details}

We represent the trainable parameters for incoming lighting as texels of a cube map 
(typical resolution $6\times512\times512$). The base level represents the pre-integrated lighting for the lowest
supported roughness value, which then linearly increases per mip-level. 
Each filtered mip-map is computed by average-pooling the base level texels to the current resolution 
(for performance reasons, the quantization this process introduces is an acceptable approximation for our use case).
Then, each level is convolved with the GGX normal distribution function. 
We pre-compute accurate filter bounds per mip-level (the filter bound is a function 
of the roughness, which is constant per mip level).

\newlength{\commentWidth}
\setlength{\commentWidth}{1.5cm}
\newcommand{\atcp}[1]{\tcp*[r]{\makebox[\commentWidth]{#1\hfill}}}

\begin{algorithm}[t]
	\SetAlgoNoEnd
	\SetAlgoNoLine
	\KwIn{output gradient: $\frac{\partial L}{\partial Y}$, weight tensor: $W$ }
	$\frac{\partial L}{\partial \mathbf{X}} = \mathbf{0}$ \;
	\For{$i,j \in \mathrm{pixels}$}
	{
		\For{$k,l \in \mathrm{footprint}$}
		{
			$\frac{\partial L}{\partial x_{i,j}}$ += $W^T_{k,l} \cdot \frac{\partial L}{\partial y_{i + k, j+l}}$ \atcp{gather}
			$\frac{\partial L}{\partial x_{i + k, j+l}}$ += $W_{k,l} \cdot \frac{\partial L}{\partial y_{i,j}}$ \atcp{scatter}
		}
	}
	\caption{Computation of the loss gradient w.r.t, inputs, $\frac{\partial L}{\partial X}$, for a 2D convolution, 
	expressed as a gather or scatter operation. 
	We use the notation $x_{i,j}$ to denote element $(i,j)$ of the tensor $X$.}
	\label{alg:scatter_gather}
\end{algorithm}

The loss gradients w.r.t. the inputs, $\frac{\partial L}{\partial X}$, for 
a convolution operation can be computed as a gather operation using products of the transposed weight tensor, $W^T$,
and the output gradient, $\frac{\partial L}{\partial Y}$, within the filter footprint.  
However, in cube maps, the filter footprint may extend across cube edges or corners, which 
makes a gather operation non-trivial. 
We therefore express the gradient computation as a \emph{scatter} operation, 
which can be efficiently implemented on the GPU using non-blocking \texttt{atomicadd} instructions.
We illustrate the two approaches in Algorithm~\ref{alg:scatter_gather}.

\section{Scene Credits}
Mori Knob from Yasotoshi Mori (CC BY 3.0). Cerberus model used with permission from NVIDIA.
Damicornis, Saxophone, and Jackson models courtesy of the Smithsonian 3D repository~\cite{Smithsonian2020}, 
(CC0). Spot model (public domain) by Keenan Crane.
NeRD datasets (moldGoldCape, ethiopianHead) (Creative Commons Attribution-NonCommercial-ShareAlike 4.0 International).
The NeRF and NeRFactor datasets contain renders from modified blender models located on blendswap.com:
chair by 1DInc (CC-0), drums by bryanajones (CC-BY),  ficus by Herberhold (CC-0), hotdog by erickfree (CC-0), 
lego by Heinzelnisse (CC-BY-NC), materials by elbrujodelatribu (CC-0), mic by up3d.de (CC-0), 
ship by gregzaal (CC-BY-SA).
Probes from Poly Haven~\cite{polyhaven} (CC0) and the probes provided in the NeRFactor dataset
which are modified from the probes (CC0) shipped with Blender.
DTU scans from the DTU MVS dataset~\cite{Jensen2014}.

\figMegaNerf

\end{document}